\PassOptionsToPackage{hyperfootnotes=false}{hyperref}
\documentclass{article}

\usepackage[preprint,nonatbib]{neurips_2026}

\usepackage{subcaption}
\DeclareCaptionFormat{myformat}{\fontsize{8}{9}\selectfont#1#2#3}
\usepackage{etoolbox}

\usepackage[utf8]{inputenc}
\usepackage[T1]{fontenc}    \usepackage{hyperref}     \hypersetup{
  colorlinks=true,
  linkcolor=blue!50!black,
  citecolor=blue!50!black,
  urlcolor=blue!50!black
}

\usepackage{url}            \usepackage{booktabs}       \usepackage{amsfonts}       \usepackage{nicefrac}       \usepackage{microtype}      \usepackage{xcolor}         \usepackage{algorithm}
\usepackage{algpseudocode}
\usepackage{setspace}
\usepackage[export]{adjustbox}
\usepackage{graphicx}
\usepackage{tikz}

\usepackage{amsmath, amssymb, amsthm}
\usepackage{mathrsfs}
\usepackage{mathtools}
\DeclarePairedDelimiterXPP{\trace}[1]{\operatorname{tr}}[]{}{#1}
\DeclarePairedDelimiterXPP{\ptrace}[2]{\operatorname{tr}_{#1}}[]{}{#2}

\usepackage{thmtools, thm-restate}

\makeatletter
\newcommand{\thmfirsttimeonly}[1]{\ifthmt@thisistheone
    \textcolor{red}{\emph{#1}}\fi
}
\makeatother

\declaretheoremstyle[
spaceabove=6pt,
spacebelow=-.5em,
headfont=\normalfont\bfseries,
notefont=\mdseries,
notebraces={(}{)},
bodyfont=\normalfont,
]{mystyle}

\newcommand{\mtrans}[1]{{#1}^{\top}}
\newcommand{\ctrans}[1]{{#1}^{\dagger}}

\usepackage{enumitem}
\usepackage{fancyvrb}
\usepackage{color}
\usepackage{listings}
\usepackage{xcolor}
\newcommand{\R}{\ensuremath \mathbb R}
\newcommand{\C}{\ensuremath \mathbb C}
\newcommand{\tr}[1]{\ensuremath \mathrm{Tr}(#1)}

\usepackage{cleveref}
\Crefname{rthm}{Theorem}{Theorems}
\Crefname{rdefn}{Definition}{Definitions}

\usepackage{xstring}

\newcommand{\loss}{\ensuremath{\mathscr L}}
\newcommand{\chinacr}{\ensuremath{\mathscr A}}

\newcommand{\setfmt}[1]{\ensuremath{\mathcal {#1}}}
\newcommand{\data}{\setfmt{D}}
\newcommand{\datasize}{\ensuremath{d}}

\newcommand{\modelsize}{\ensuremath{m}}
\newcommand{\computeamount}{\ensuremath{\mathfrak{c}}}

\newcommand{\weakly}{\ensuremath{\zeta}}
\newcommand{\spectralradius}{\ensuremath{\varrho}}

\newcommand{\mfnuma}{\ensuremath{1}}
\newcommand{\mfnumb}{\ensuremath{2}}
\newcommand{\mfnumc}{\ensuremath{3}}
\newcommand{\mfa}{\ensuremath{\phi_\mfnuma}}
\newcommand{\mfb}{\ensuremath{\phi_\mfnumb}}
\newcommand{\mfc}{\ensuremath{\phi_\mfnumc}}
\newcommand{\mfna}{\ensuremath{1-(x+1)^{-1}}}
\newcommand{\mfunca}{\ensuremath{1-(x+\beta)^{-\alpha}}} \newcommand{\gradmfunca}{\ensuremath{\alpha(x+\beta)^{-\alpha-1}}} 
\newcommand{\neggradmfunca}{\ensuremath{-\gradmfunca}} \newcommand{\gradmfuncaarg}[1]{\ensuremath{\alpha(#1+\beta)^{-\alpha-1}}}

\newcommand{\parama}{\ensuremath{x \geq 0, \alpha > 0, \beta > 0}}

\newcommand{\mfnb}{\ensuremath{1-e^{-x}}}
\newcommand{\mfuncb}{\ensuremath{1-e^{-x}}} \newcommand{\gradmfuncb}{\ensuremath{e^{-x}}} \newcommand{\paramb}{\ensuremath{x \geq 0, \alpha > 0}}

\newcommand{\mfnc}{\ensuremath{\frac{x}{x+1}}}
\newcommand{\mfuncc}{\ensuremath{\frac{x}{(1+x^\alpha)^{1/\alpha}}}} \newcommand{\gradmfuncc}{\ensuremath{(1+x^\alpha)^{-\frac{1}{\alpha}-1}}} \newcommand{\gradmfunccarg}[1]{\ensuremath{(1+{#1}^\alpha)^{-\frac{1}{\alpha}-1}}} \newcommand{\paramc}{\ensuremath{x \geq 0, \alpha > 0}}

\DeclareMathOperator{\rank}{rank}
\DeclareMathOperator*{\argmax}{argmax}

\usepackage[
style=numeric-comp, sorting=none,          natbib=true,
  backend=bibtex,        maxnames=6,            minnames=1,            firstinits=true,
  backref=true,
  hyperref=true, url=true,
  isbn=false,
  arxiv=abs
  ]{biblatex}
\AtEveryBibitem{\clearfield{abstract}} \AtEveryBibitem{\clearfield{note}} \AtEveryBibitem{\clearfield{eprint}} \DeclareNameAlias{default}{last-first}   

\usepackage[most]{tcolorbox}
\tcbset{
  colback=gray!5,   colframe=gray!60, boxrule=0.4pt,
  arc=2mm,          left=6pt,right=6pt,top=6pt,bottom=6pt,
}
\newtcolorbox{intuition}[1][]{
enhanced,
  colback=gray!4,
  colframe=gray!60,
  boxrule=0.4pt,
  arc=2mm,
  left=6pt,right=6pt,top=6pt,bottom=6pt,
  fonttitle=\bfseries,
  title=#1
}

\newtcolorbox[auto counter,number within=section,]{explanation}[2][]{
float, floatplacement=htbp, enhanced,
  colback=gray!4,
  colframe=gray!60,
  boxrule=0.4pt,
  arc=2mm,
  left=6pt,right=6pt,top=6pt,bottom=6pt,
  fonttitle=\bfseries,
  title=Explanation~\thetcbcounter: #2,
label type=explanation, fontupper=\footnotesize,
  #1
}

\title{How Much Is a Dataset Worth? Scaling Laws, the Vendi Score, and Matrix Spectral Functions}

\author{Jeff A.\ Bilmes\textsuperscript{1,2} \\
  \texttt{bilmes@uw.edu}
  \And
  Gantavya Bhatt\textsuperscript{1} \\
  \texttt{gbhatt2@uw.edu}
  \And
  Arnav M. Das\textsuperscript{1} \\
  \texttt{arnavmd2@uw.edu}
  \\[0.8em]
  \textsuperscript{1}Department of Electrical \& Computer Engineering\\  
  \textsuperscript{2}Paul G. Allen School of Computer Science \& Engineering\\
  University of Washington, Seattle\\
  Seattle, WA 98195
}

\date{2026}

\newcommand{\randomspeedup}{\ensuremath{35{,}000\times}}

\definecolor{FigGreen}{HTML}{00A551}
\definecolor{FigRed}{HTML}{FF3333}
\definecolor{FigBlue}{HTML}{333399}

\newcommand{\myxshape}{\scalebox{1}[0.8]{\textbf{\textsf{X}}}}
\newcommand{\mysquareshape}{\ensuremath{\blacksquare}}
\newcommand{\mycircshape}{\raisebox{-0.25ex}{\scalebox{1.7}{$\bullet$}}}

\makeatletter
\begin{document}
\maketitle

\newcommand{\FigureOneCaptionA}{ImageNet-1K test performance versus compute budget,
measured via number of training samples seen. All training sets
are perfectly class-balanced over the 1000 classes matching the
test distribution. Colors indicate dataset value:
\textcolor{FigGreen}{high (green)}, \textcolor{FigRed}{random
  (red)}, and \textcolor{FigBlue}{poor (blue)}; markers indicate
dataset size: small (\myxshape),
medium (\mycircshape), and
large (\mysquareshape). Random shows 10 independent 
balanced class-stratified random
sampled subsets at each size.
\par
}

\newcommand{\FigureOneCaptionB}{Best held-out ImageNet-1K test performance at the largest (32M) compute budget
versus a fast surrogate data-valuation score. The $y$-axis requires
full model training and evaluation, while the $x$-axis requires only an
inexpensive surrogate dataset appraisal set-function evaluation.
The random (\textcolor{FigRed}{red}) points plot the performance
of ten independent class-stratified random subsets,
demonstrating random's strong concentration of statistical properties
at all sizes along both axes.
\par
}

\begin{figure}[htbp]
  \centering

  \begin{subfigure}[t]{0.54\textwidth}
    \centering
    \includegraphics[
      width=\linewidth,
      valign=c
    ]{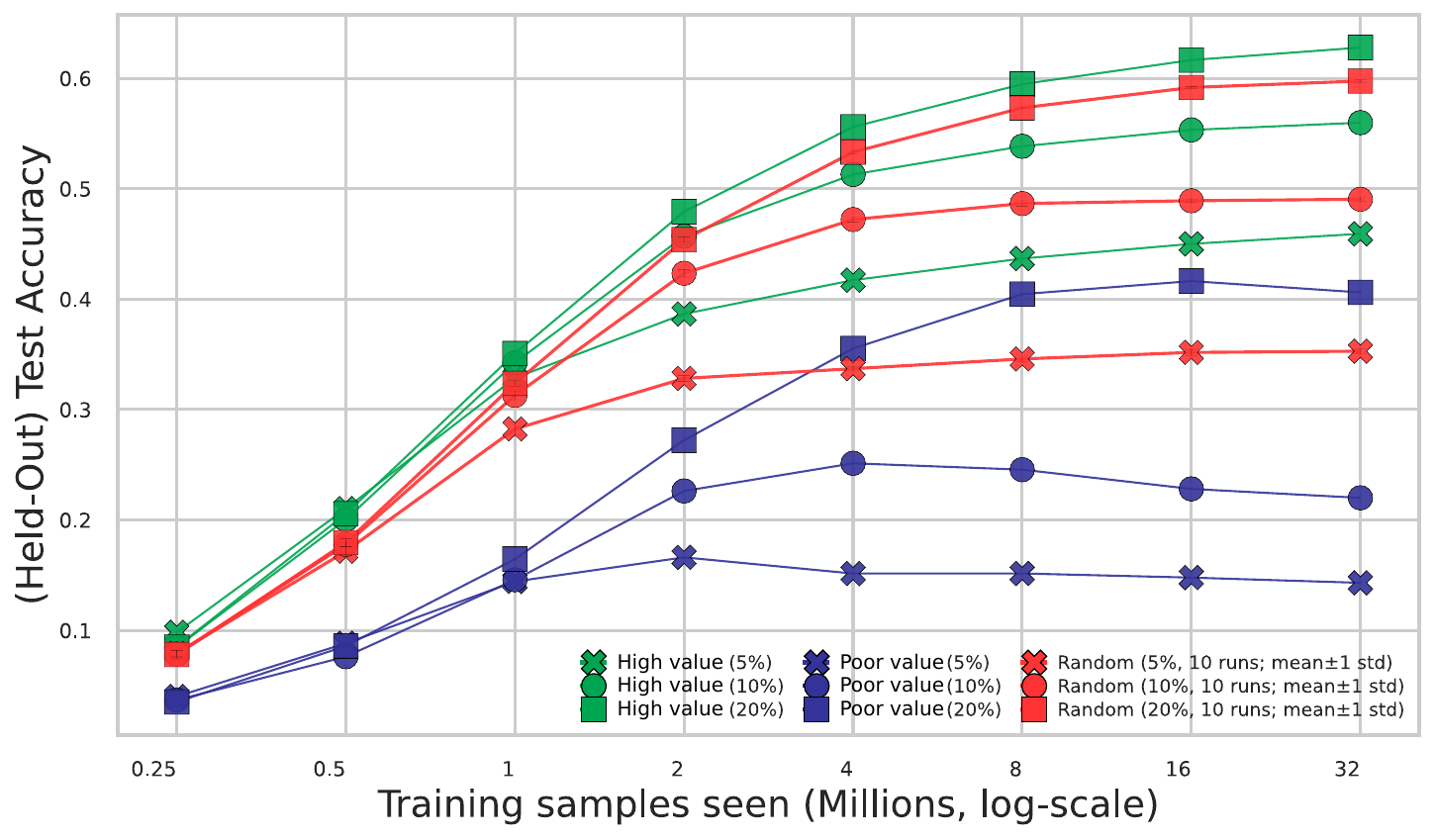}
    \caption{
      \FigureOneCaptionA \looseness-1}
    \label{fig:class-balanced-all-merged-teaser}
  \end{subfigure}
  \hfill
  \begin{subfigure}[t]{0.44\textwidth}
    \centering
    \includegraphics[
      width=\linewidth,
      valign=c
    ]{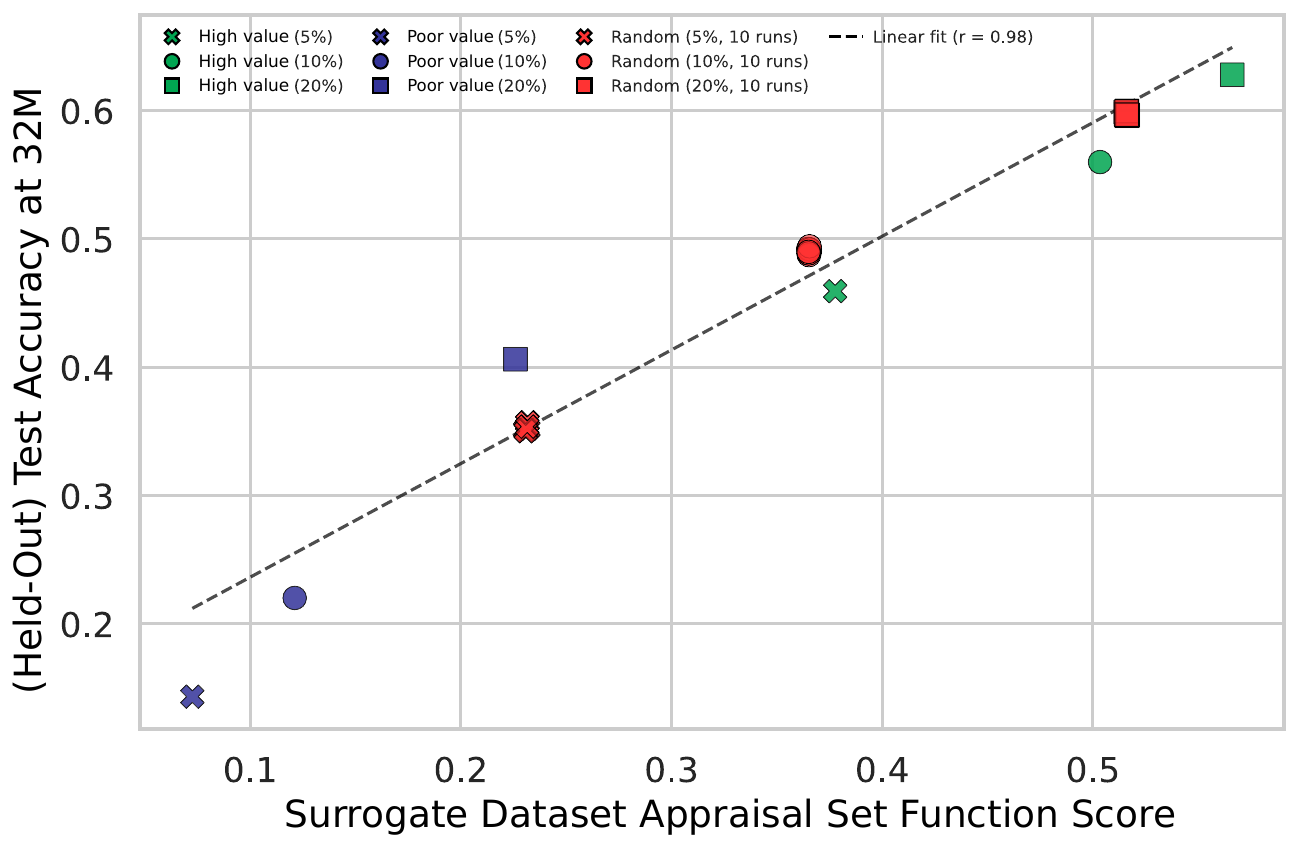}
    \caption{
      \FigureOneCaptionB }
    \label{fig:class-balanced-accuracy-vs-fl-score-at-32M-teaser}
  \end{subfigure}

  \caption{ \textbf{Dataset value is not determined by size or compute alone, even when
    all datasets are perfectly class balanced.} In (a), the ranking induced by
    dataset value is largely preserved across more than two orders of
    magnitude of compute budget, suggesting that offline data
    appraisal is meaningful beyond a single training budget. Size
    (given as percent of the full dataset) generally helps within a fixed value category, but it does not
    dominate: small balanced high-valued datasets (\textcolor{FigGreen}{green \myxshape}) outperform large
    balanced poor-valued datasets (\textcolor{FigBlue}{blue
      \mysquareshape}) across all compute budgets. Poor datasets
    at all sizes eventually overfit, declining at large compute budgets, consistent with
    early stopping acting as regularization; good datasets at all sizes continue
    improving even at the largest compute budget (32M), while random datasets appear
    closer to saturation.
In (b), the surrogate dataset valuation score captures this without relying 
    only on size or class balance. The score is a non-additive set function
    that evaluates every sample in the context of the other samples, thereby
    estimating the information in the dataset. It provides a cheap
    proxy for the more expensive final test performance obtained after
    full training. Details of the data-valuation function and the construction
    of the good, random, and poor datasets are given in the paper
    and further results in~\Cref{app:additional_compute_budgets}.
    We visualize high and poor value data subsets for different classes of ImageNet-1K in~\Cref{app:viz_good_bad_sets}.
    \looseness-1
}
    \label{fig:main-teaser}
\end{figure}

\begin{abstract}

  Training data is central to machine learning, yet it remains
  challenging to efficiently appraise dataset value for a learning
  process. Neural scaling laws appraise data through dataset size,
  while the recently proposed Vendi Score uses Von Neumann (quantum)
  entropy to measure dataset value. In this paper, we show both that
  common neural-scaling-law objectives and the Vendi Score are
  submodular. We further show that the Vendi Score is a special case
  of a broader class of submodular objectives that we call matrix
  spectral functions. This class also includes determinantal (DPP)
  objectives, which are known to be log-submodular, as well as many
  others, including neural-scaling-law-inspired matrix spectral
  hybrids. We also introduce weakly matrix monotone functions and show
  how they lead to weakly submodular matrix spectral functions,
  yielding a broad family of practical objectives for data selection
  and appraisal. A key challenge is scale. We thus develop
  secular-equation-based updates that avoid repeated
  eigendecompositions during greedy optimization, reducing
  marginal-gain evaluation for $m$-dimensional embeddings by an $O(m)$
  factor relative to oracle queries. In our implementation, this
  yields an average empirical speedup of $\approx \randomspeedup$,
  making direct optimization of the Vendi Score feasible on
  ImageNet-1K-scale datasets.
Thus
  enabled, we compare how well several objectives predict the value of
  training subsets for held-out test performance under fixed-size,
  class-balanced, and fixed training-budget regimes, including the
  Vendi Score, DPPs, facility location, and three new matrix spectral
  variants. Across multiple datasets, facility location 
  performs the best.
Direct optimization also reveals that, while
  the Vendi Score is predictive over moderate score ranges, pushing
  the objective to higher values can make it a poor downstream performance proxy.
  Other matrix spectral variants, including
  weakly submodular ones, perform better, though still below facility
  location. We also find that uniformly at random fixed-size subsets,
  both unconstrained and class-balanced, are remarkably concentrated in
  both appraisal scores and held-out performance, indicating that
  random sampling fails to expose 
  much variety.
Finally, we show that size, class balance, and training
  budget do not alone determine data value: even when controlling for these
  factors, performance ranges smoothly from good to
  bad. 
\looseness-1

\end{abstract}

\section{Introduction, Background, and Data Appraisal}
\label{sec:intro}

In this paper, we focus on two specific but common data appraisal methods
and show how they are related in that they both have the property of
submodularity~\citep{fujishige2005submodular,bilmes2022submodularity,tohidi2020submodularity}.
Specifically,
neural scaling laws, although non-additive, appraise data based on a simple
concave composed with cardinality objective.  On the other hand, the Vendi
score~\cite{friedmanvenditmlr,pasarkar2024cousins,pasarkar2026vendi,jung2025prismatic,nguyen2025vendi,sirigiri2026diversity,jalali2026conditional}
(and its G-Vendi variant~\cite{pasarkar2026vendi}) valuates a dataset based on
the entropy of the eigenvalues of an underlying positive semi-definite matrix,
similar to the Von Neumann (i.e., quantum) entropy~\citep{bennett2002quantum,petz2007quantum,nielsen2010quantum}. 
We show that common neural scaling laws (when seen as set functions)
are submodular, and also that the log Vendi score\footnote{We focus exclusively on the $q=1$ order Vendi score where it relates to the
Shannon entropy, although one can define other $q \neq 1$ orders that correspond
to the Renyi entropy of order $q$ that generalizes Shannon entropy. 
Our fast secular equation-based 
computation (Section~\ref{sec:fast-matrix-spectral-functions}) applies not only to any matrix
spectral function but also to any order Vendi score.} is, along with log
determinantal point processes, a special case of a more general class we call
the ``Matrix Spectral'' submodular functions. This means that one can
immediately construct conditional Vendi scores and Vendi-based conditional
mutual information constructs as one can do with any submodular
function~\cite{iyer2021generalized}. Matrix Spectral functions are defined by
applying a matrix function to a positive semi-definite matrix. If the matrix
function has a certain property, namely its negative derivative is (weakly)
matrix monotone (all defined below), the resulting matrix spectral function is
(weakly) submodular. We introduce weakly matrix monotone functions,
and their properties may be of independent theoretical interest.
In order to
scale to larger dataset sizes,
we develop a 
provably fast 
secular-function 
strategy to repeatedly compute queries like $f(S + a)$
for a different $a$ and fixed $S$, as required by the greedy (and many other)
submodular optimization algorithms. 
Under our implementation of this,
timing numbers for directly maximizing the Vendi score
show an average speedup of
about $\randomspeedup$ 
(Figure~\ref{fig:greedy-maximization-speedup-random-data})
over traditional purely oracle function calls.
Thus, the Vendi score (and any matrix
spectral function) can be 
feasibly
directly 
maximized
while having the
classic $1-1/e$ approximation guarantee.

It has been stated that the Vendi score is a good way to judge the diversity,
and hence the value, of a training data subset. For example,
recently~\cite{jung2025prismatic} showed good correlation between 
a gradient-based Vendi
score and test set accuracy. However, the evaluated subsets were not
produced by directly optimizing the Vendi score but rather utilized
indirect proxy methods, begging the question if the full range (both low and
high) of Vendi scores were considered. Armed with the aforementioned ability to
directly optimize the Vendi score, we compare the abilities of six 
functions to
predict the value of a training data subset.
This includes the Vendi score, determinantal
point processes (DPPs), facility location (a classic submodular
function), and three new matrix spectral variants (one inspired
by neural scaling laws). We apply this to several datasets (including
ImageNet-1K). A number of our findings are surprising. First, we find Vendi
correlation much worse than anticipated; this discovery was possible only
due to the aforementioned secular-function speedups
that enabled direct objective optimization. We find moreover that the facility
location function significantly beats the correlations of all of the other
functions. We also observe a remarkable concentration of statistical
properties of fixed-size, both balanced and otherwise,
uniformly at random sets; random subsets all seem similar to each
other in terms both of appraisal value and downstream held-out
performance. Without set function direct optimization however, random's
concentrated nature would not have been made apparent. We also find that data subsets
of the same size, both unconstrained and with class balance enforced, can have
widely varying appraisal score and test performance, showing that size alone is
a poor proxy for data value. Similarly, in a fixed compute-budget scenario, we
show that less computation can be much better if you have the right data. We
also find that more computation is not always better, as there are small
valuable balanced datasets that continue to improve with more computation, and
large balanced datasets that not only do worse than the small datasets but also
eventually overfit. In other words, not only computation but the data itself is
crucial. Also, finding these datasets utilized various forms of submodular
optimization on the right submodular function.

\Cref{sec:neural-scaling-laws} shows that common neural scaling laws are
submodular.
\Cref{sec:vendi} shows that the Vendi score is a
special case of the more general matrix
spectral function class (\Cref{sec:matrix_spectral_functions}). 
We introduce the notion of weakly matrix monotone
functions (\Cref{sec:weakly-matrix-monotone}) and
show how they lead to $\weakly$-weakly DR submodular matrix spectral functions
(which still have approximation guarantees). Section~\ref{sec:fast-matrix-spectral-functions} 
describes the secular-equation speedups.
\Cref{sec:experiments} applies these 
methods to data appraisal by
directly optimizing the Vendi, and other functions to
compare their ability to predict downstream performance.
\Cref{sec:conclusions} concludes and discusses future
work. Many additional background discussions, details, 
explanations, results, all proofs, and a table
of contents, are given in the
appendices starting in~\Cref{sec:paper-toc}.

\section{Neural Scaling Laws are Submodular}
\label{sec:neural-scaling-laws}

Recently, our community has expressed a desire to
explain how the test loss (or accuracy) of a large neural model
depends, in practice, on the amount of training data, the size of the model, and the
amount of compute used to train the model. The neural scaling law
literature  studies this, empirically quantifying how loss
decreases with more data and/or parameters, often
modeled using a parameterized power law function~\cite{HestnessEtAl2017,KaplanEtAl2020,hoffmann2022training,pearce2024reconcilingkaplanchinchillascaling,porian2025resolvingdiscrepanciescomputeoptimalscaling}. 
This resulted in the Chinchilla laws, still a key reference
point for compute-optimal scaling of large models, which has the
following form:
\begin{align}
\loss(\modelsize,\datasize) \propto \loss_\infty + a \modelsize^{-\alpha} + b \datasize^{-\beta}
\end{align}
where $\loss(\modelsize,\datasize)$ measures the test loss (e.g.,
cross entropy or negative log-likelihood on a held-out test set), $\loss_\infty$ is
the irreducible loss (analogous to Bayes error), $\modelsize$ is the number
of model parameters, $\datasize$ measures the amount
of data (e.g., number of training tokens, or number of samples), $a \geq 0$ and $b \geq 0$ are two fitted
linear coefficients and $\alpha >0$ and $\beta > 0$ are two fitted
exponential coefficients. In the above equation, the quantity of
computation $\computeamount$ is implicit and assumed to 
have the relation $\computeamount \propto \modelsize \datasize$.  This
equation is used to guide what is known as ``compute-optimal''
training; i.e., for a given fixed amount of computational resources $\computeamount$ (and thus
financial cost),
what is the best balance between the amount of training data $\datasize$ and the
number of parameters $\modelsize$? 
This leads to rules of thumb such as the 20:1 tokens:parameter ratio~\cite{dey2023cerebrasgptopencomputeoptimallanguage} 
resulting from the Chinchilla approach~\cite{hoffmann2022training} obtained by
first fitting $\loss(\modelsize,\datasize)$ to data,
and then minimizing it subject to
$\modelsize \datasize = \computeamount$.
Major government policy and business investment
decisions have been
made based on 
this formulation~\cite{villalobos2024run,digitalstrategy_ec_eu_ai_act_gpaicompute_thresholds,BusinessInsider_Goldman_Data_2025,usco_part3_ai_training_2025}.

Notably, the Chinchilla measure of data $\datasize$ is defined purely
in terms of the count of the number of tokens/samples used. If we
keep $\modelsize$ fixed, this yields the data-loss relationship
$\loss(\datasize) = c + b \datasize^{-\beta}$ where $c$ is a
constant. This function expresses
a diminishing returns property, data becomes less valuable
as we obtain more of it.
Suppose $\data$ is a dataset of size $\datasize=|\data|$,
then this can be seen as a set function
$\loss(\data) = c + b |\data|^{-\beta}$. 
As loss is a measure of error or cost, it
can be converted into a
mathematically equivalent accuracy-like or data-value measure 
via $\chinacr(\data) = c' - b |\data|^{-\beta}$, where $c'$ is a different
constant and $\chinacr(\data)$ measures value. Any
loss or cost measure $\loss(\data)$ (measuring
the cost of using $\data$)
can thus be
converted into a mathematically equivalent value measure $\chinacr(\data) = c' - \loss(\data)$ 
(measuring the value of using $\data$), and vice versa.
One example of this is to measure value or quality using negative log perplexity~\cite{kaiser2017one,kaiser2017depthwise}
which is the same thing as average conditional log-likelihood and
is also the same as the negative cross-entropy loss (where
we would set $c'=0$ and $\loss(\data)$ is a model of cross-entropy loss).
We therefore immediately have the following:
\begin{restatable}[Chinchilla is Submodular]{rlem}{chinchillaissubmodular}
\label{lem:chinchilla-submodular}
The Chinchilla scaling laws measure data value using a simple submodular function.
\thmfirsttimeonly{(Proven in Section~\ref{sec:proofs-neural-scaling-laws}, page~\pageref{proof:chinchilla-submodular}.)}
\end{restatable}

Thus, such neural scaling law papers are implicitly
measuring the value of data using not only a submodular function (with
a diminishing returns property) but one that is particularly simple in
that it measures value only via size, and one that assumes all
training samples are created equal. In these laws, the amount of training data, the size of the model, and
the amount of compute are all proxies for more fundamental quantities.
For example, the amount of training data is a proxy for the
information in the training data relevant to a model to solve a task. The
justification for these simple proxies is scalability, but this 
is at the expense of accuracy.  It is highly unlikely that all samples of a given size are 
equally valuable. It is also unlikely that every sample's value is influenced by all other
samples identically (i.e., $\chinacr(\data+v) = \chinacr(\data+u)$ for 
samples $u,v \notin \data$).

More recent cluster based scaling laws, such as~\cite{kang2025autoscalescaleawaredatamixing}, 
go a step further, with relations such as $\loss(\datasize_1, \datasize_2, \dots, \datasize_k) = c +
\sum_{k=1}^K (c_i + \datasize_i)^{-\beta_i}$ over $K$ separate data
domains, where again $c$ is constant (with respect to data), and $\datasize_i$ is
the amount of data in the $i$-th domain. This is more flexible than the original 
Chinchilla law, and mimics stratified sampling,
but is also still submodular. This 
form can again be converted to a mathematically equivalent accuracy set function:
$\chinacr(\data) = c' - \sum_{k=1}^K g_i(\data \cap \data_i)$
where $g_i(\setfmt{A}) = (c_i + |\setfmt{A}|)^{-\beta_i}$
and $\data_i$ is one of $K$ ``domains'' of $\data$.  Within each 
domain, data is again valued by only its size and all samples
within a domain are considered equally valuable. We have the following: 
\begin{restatable}[Cluster-based Scaling Laws are Submodular]{rcor}{clusterscalingissubmodular}
\label{cor:cluster-scaling-submodular}
Cluster-based generalizations of neural scaling laws measure
data value using a submodular function.
\thmfirsttimeonly{(Proven in Section~\ref{sec:proofs-neural-scaling-laws}, page~\pageref{proof:cluster-scaling-submodular}.)}
\end{restatable}

A third, still more recent, scaling law we consider captures the
notion that the value of a sample depends on how often the sample is repeatedly
used during training over multiple epochs~\citep{goyal2024scaling} and
this depends on the compute budget $\computeamount$. Here,
a smaller dataset is repeated a larger number of times while
a larger dataset is repeated a smaller number of times in order
to consume a given fixed compute budget $\computeamount$, measured in units of total samples used for training. 
Such scaling laws
observe that each time a sample is used over multiple epochs, 
its value decreases relative to the previous time it was used. This
has led to hypotheses~\cite{goyal2024scaling,mohri2026bitter} that with larger compute budgets $\computeamount$,
it is better to use more data since every sample is used less frequently. One
such scaling law~\citep{goyal2024scaling} is the following:
\begin{align}
\loss_\computeamount (\datasize_1, \datasize_2, \dots, \datasize_k)
= c + b \datasize_1^{-\beta_1}\prod_{j=2}^k \left(\frac{\datasize_j}{\datasize_{j-1}}\right)^{-\beta_j}
\end{align}
where here $\datasize_j$ is number of samples seen at the end of $j^\text{th}$
epoch, $b > 0$, $c > 0$, and where $\beta_{j} = \beta
\left(\frac{1}{2}\right)^{(j-1)/\tau}$ for $j \geq 1$ where $\beta > 0$ is an initial scaling exponent and
$\tau > 0$ is a half-life parameter. Here, $\loss_\computeamount ( \{
\datasize_j \}_j )$ implicitly assumes that $k = \computeamount / \datasize_1$
epochs are performed over a given amount of data and thus can be seen as a
function of the base datasize $\loss_\computeamount (\datasize)$ by setting
$\datasize_j = j \datasize$. Viewed as a value-based set function
$\chinacr_\computeamount (\data) = c' - \loss_\computeamount (|\data|)$, this is
also a submodular function.
\begin{restatable}[Epoch-based Scaling Laws are Submodular]{rthm}{epochscalingissubmodular}
\label{thm:epoch-scaling-submodular}
Epoch-based generalizations of neural scaling laws measure
data value using a submodular function.
\thmfirsttimeonly{(Proven in Section~\ref{sec:proofs-neural-scaling-laws}, 
page~\pageref{proof:epoch-scaling-submodular}.)
}
\end{restatable}
The proof of this third theorem is much less obvious than the proofs of the
above first two. Moreover, the fact that a scaling-law data valuation approach that
considers a sample's value over multiple epochs is still submodular is not
surprising when considering that a different member of the submodular 
function family is good at valuing data across a wide range of compute budgets, as we saw
in~\Cref{fig:main-teaser}.

While the above neural scaling laws are submodular, their weakness is that they all value
data only by size rather than collective utility of the unique
properties of each data sample. On the other hand, the above power-law style
concave function $\mfa(x) = \mfunca$, $\parama$,
and the concave function
$\mfb(x) = \mfuncb, \paramb$ (which, as we argue below
after the proof of \Cref{thm:epoch-scaling-submodular}, is similar
to that used for the epoch-based
scaling law of~\cite{goyal2024scaling}),
can be useful building blocks for more complex 
data appraisal functions,
which
we turn to next.

\section{Matrix Spectral Functions}
\label{sec:matrix_spectral_functions}

Given a scalar function $\phi : \R_+ \to \R$, and
a symmetric positive semidefinite matrix $B \in \R^{m \times m}$,
there are many ways~\cite{loewner1934monotone,lancaster1985theory,bhatia1997matrix,higham2008functions,bhatia2009positive} to define the application of scalar function $\phi$
to matrix $B$. One particularly useful way is via the eigen-decomposition of $B$. Let $B = Q \Lambda \mtrans{Q}$ be the
eigen-decomposition of $B$ where $Q$ is an orthonormal matrix of eigenvectors and $\Lambda$ is a diagonal 
matrix of eigenvalues. Since the matrix is positive semi-definite, all eigenvalues are non-negative.
Then we can define $\phi(B) = Q \phi(\Lambda) \mtrans{Q}$ where $\phi(\Lambda)$ is the diagonal matrix 
obtained by applying $\phi$ to each diagonal entry of $\Lambda$. 
This is known as the spectral function definition of $\phi(B)$. Then one can map this back down to
a scalar via $\trace{\phi(B)} = \sum_{i=1}^m \phi(\lambda_i(B))$ which is the 
sum of the diagonal entries of $\phi(B)$ and which, therefore, is also equal to the
sum of $\phi$ applied to the eigenvalues of $B$, where $\lambda_i(B)$ is the $i^\text{th}$ eigenvalue of $B$.
\looseness-1

Now, given a function of a matrix $B$, one can also define set functions where the set determines
a sub-matrix on which to operate. Let $B$ be an $n \times n$ matrix and $X \subseteq V$ where
$V = [n]$. Then for any $X \subseteq V$, let $B_X$ be the $|X| \times |X|$ principal
submatrix of $B$ consisting of the rows and columns indexed by $X$.
We can define a set function $f(X) = \trace{\phi(B_X)}$ which is
the sum of the diagonal entries of the matrix $\phi(B_X)$ and which, therefore,
is the same as applying $\phi$ to each of the eigenvalues of $B_X$ and summing the results.
We call such a set function a \emph{matrix spectral function}.
The cost of evaluating $f(X)$ is that of an eigenvalue
problem, $O(n^3)$ for an arbitrary subset $X \subseteq V$.

The matrix $B$, being positive semidefinite matrix, could easily arrive from
a positive semidefinite kernel function $k: \mathcal{X} \times \mathcal{X} \to \R_+$ 
applied to a dataset $\data$ of $n$ samples, where $B_{ij} = k(x_i, x_j)$ for $x_i, x_j \in \data$
and $x_i \in \R^m$ is the $i$-th sample of $\data$. Therefore, $B_X$ is the kernel 
matrix of the subset of data samples indexed by $X$. In the case of a linear kernel,
then $B_X$ has a particularly simple form, namely $B_X = \data[X] \mtrans{\data[X]}$ where we view
$\data$ is an $n \times m$ design 
matrix 
and $\data[X]$ is the $|X| \times m$ row sub-matrix of $\data$ corresponding to the subset $X$.
Since $\phi(B_X)$ applies $\phi$ to the eigenvalues of $B_X$, and since,
by a form of eigenvalue duality~\cite{bhatia1997matrix},
the eigenvalues of $\data[X] \mtrans{\data[X]}$ are the same as the eigenvalues of $\mtrans{\data[X]} \data[X]$, 
$\phi(B_X)$ can be seen as applying $\phi$ to the eigenvalues of $\mtrans{\data[X]} \data[X]$
which is always an $m \times m$ matrix, and thus computing the eigenvalues is an
$O(m^3)$ operation rather than $O(|X|^3) \in O(n^3)$, which is a significant speedup when $m \ll n$.

The question next becomes, is it ever the case that $f(X)$ has useful properties, such
as submodularity. The reason this is useful is that if $f$ is submodular (and monotone non-decreasing),
this opens the doors to many provably good approximation algorithms for both maximization and
minimization (as summarized in~\Cref{sec:background-submodularity}). A classic result is that if $f$ is submodular monotone non-decreasing then
the simple and fast greedy algorithm for solving $\text{OPT} = \max_{X \subseteq V : |X| \leq k} f(X)$
has an approximation $f(X_\text{greedy}) \geq (1 - 1/e) \text{OPT}$ where $X_\text{greedy}$ is the set 
returned by the greedy algorithm. 
In order to answer this question, we need to define a few additional concepts.

A function $\psi: \R_+ \to \R$ is said to be \emph{matrix monotone} if for any
two positive semidefinite matrices $A$ and $B$ of the same size, if $A \preceq
B$ (i.e., $B - A$ is positive semidefinite), then $\psi(A) \preceq \psi(B)$.
This means that $\psi$ preserves the order of positive semidefinite matrices.
The function $\psi$ is said to be \emph{matrix antitone} if the opposite
holds, i.e., if $A \preceq B$ then $\psi(A) \succeq \psi(B)$. Now,
in~\citep{friedland2011submodular,audenaert2010strongly}, it was first shown
that the function $f(A) = \trace{\phi(A)}$ is submodular if $-\phi'$ (the
negative of the derivative of $\phi$) is matrix monotone. This 
was strengthened to be also necessary in~\citep{lewin2014family}. We have:
\begin{restatable}[Matrix Spectral Submodular Functions (\cite{friedland2011submodular,audenaert2010strongly,lewin2014family})]
{rthm}{matrixspectralissubmodular}
\label{thm:matrix-spectral-submodular}
Let $\phi: \R_+ \to \R$ be a matrix function whose negative derivative is a matrix monotone 
function, i.e., for all matrices $0 \preceq A \preceq B$, we 
have $0 \preceq -\phi'(A) \preceq -\phi'(B)$. 
Then the resulting matrix spectral function $f(X) = \trace{\phi(B_X)}$ is submodular for
any symmetric positive semidefinite matrix $B$.
Conversely, if $f(X) = \trace{\phi(B_X)}$ is submodular for all positive semidefinite matrices $B$,
then $-\phi'$ is a matrix monotone function.
\end{restatable}

When are functions matrix monotone? A famous result by
Loewner~\cite{loewner1934monotone,bendat1955monotone,bhatia2009positive} states
that any matrix monotone function $\psi : \R_+ \to \R$ can be represented as an
integral taken relative to an arbitrary positive measure $\mu$ on $[0, \infty)$
(see~\Cref{eq:integral-representation-matrix-monotone-function}). This means
that there is an infinite family of matrix monotone functions to choose from,
but the important practical question is if there are such functions that are
simple, useful, and analytical. We have the following:
\begin{restatable}[Functions with Matrix Monotone Negative Derivatives]{rthm}{matrixmonotonefunctions}
\label{thm:simple-analytical-matrix-monotone-functions}
The following functions all have matrix monotone negative
derivatives:
\begin{gather}
\xi_1(x) = -(t+x)\log(t+x) \text{ for any } t \geq 0,\\
\xi_2(x) = \log (t+x) \text{ for any } t \geq 0,\\
\xi_3(x) = x^\eta  \text{ for any } 0 < \eta \leq 1 \text{ (which includes $\sqrt{x}$ and $\sqrt[3]{x}$)},\\
\xi_4(x) = -x^\eta \text{ for any } 1 \leq \eta \leq 2.
\end{gather}
The proof of the above may be found in a number of
different sources, including~\cite{bhatia1997matrix,bhatia2009loewner,simon2019loewner,friedland2011submodular,audenaert2010strongly,
lewin2014family}.
Furthermore, any non-negative weighted affine combination of the
above four,
$\xi(x) = w_0 + \sum_{i=1}^4 w_i \xi_i(x)$,
where $w_i \geq 0$ for $i \in \{0,1,2,3,4\}$, has
matrix monotone negative derivatives.
\end{restatable}

This immediately recovers some well known results. For example, it is known
that the log determinantal point process~\citep{kulesza2012determinantal} function
$f(A) = \log \det (tI + B_A)$ is submodular, and this follows immediately from the above by noting that
the function $\phi(x) = \log (t+x)$ has a matrix monotone negative derivative, 
and thus the resulting matrix spectral function $f(X) = \trace{\log(tI + B_X)}$ is submodular
for any $t \geq 0$ whenever the underlying matrix $B$ is positive semidefinite.

Any real-valued function $\phi : \R_+ \to \R$ with a matrix monotone
negative derivative will be concave. The reason for this is the application of
this function to $1 \times 1$ matrices, which are just scalars. If $-\phi'$ is
matrix monotone, then for any two scalars $0 \leq a \leq b$, we have $0
\leq -\phi'(a) \leq -\phi'(b)$, which means that $\phi'$ is non-increasing and
thus $\phi$ is concave. On the other hand, just because $\phi$ is concave
does not mean that $-\phi'$ is matrix monotone (we give some examples
in Section~\ref{sec:weakly-matrix-monotone}).

As a result, the above theorems may at first seem like an immediate consequence of 
summing the application of a monotone non-decreasing concave function to the eigenvalues
of $B$ and then noting that for any rank-1 update with vector $u \in \R^m$,
the eigenvalues of $B + u \mtrans{u}$ are greater than or equal to the eigenvalues of $B$
(this is proven in Theorem~\ref{thm:eigenvalue_updates}, page~\pageref{thm:eigenvalue_updates}).
However, we note that each such rank-1 update does more than increase the eigenvalues, it
also affects the eigenspace (i.e., eigenvectors). So while the eigenvalues
grow for $B + u \mtrans{u}$, they correspond to a different eigenspace than
the eigenvectors of $B$, and one cannot simply associate the two sets of
eigenvalues with each other in order to apply the diminishing returns
property of the concave function. For this reason, the proof of 
Theorem~\ref{thm:matrix-spectral-submodular} is much
more involved (see also~\Cref{thm:weakly-matrix-monotone-functions-lead-to-weakly-submodular-outer-product-matrix-spectral-functions}).
The key idea is that scalar concavity of $\phi$ (i.e., scalar monotone non-decreasingness of
$-\phi'$), and matrix monotonicity of $-\phi'$ are two quite different
things.

\subsection{The log Vendi Score is Submodular}
\label{sec:vendi}

The Vendi score~\cite{friedmanvenditmlr,pasarkar2024cousins,pasarkar2026vendi,jung2025prismatic,nguyen2025vendi,sirigiri2026diversity,jalali2026conditional}
was recently proposed~\cite{friedman2022vendi} as a strategy to measure
the diversity (and hence value) of a dataset.
The log Vendi score can be defined via the Shannon entropy formulation as follows:
\begin{align}
\log \text{Vendi}(X) = \!\bigg(-\sum_{i=1}^m \lambda_i(B_X) \log \lambda_i(B_X)\bigg)
\end{align}
where $\lambda_i(B_X)$ is the $i^\text{th}$ eigenvalue of $B_X$ and $B_X$ is the
principal submatrix,
corresponding to the subset $X \subseteq V$,
of a positive semidefinite kernel matrix $B$. Therefore, the Vendi score\footnote{Note that the
Vendi score can be generalized to use Renyi entropy rather than
Shannon entropy, as further discussed in Section~\ref{sec:general-vendi-score}.} 
can be seen as the exponentiated Shannon spectral entropy
of the eigenvalues of $B_X$, and can be expressed as $\exp f$ where
$f(X) = \trace{-B_X \log(B_X)} = \trace{ \phi(B_X)}$ where $\phi(x) = - x \log x$. Therefore,
the log Vendi score is a matrix spectral function with this particular $\phi$ function.
Since we have established above that $\phi$ has a matrix monotone negative derivative, 
we get the following:
\begin{restatable}{rthm}{logvendisubmodular}
\label{cor:log-vendi-submodular}
The log Vendi score is a matrix spectral function 
that uses a matrix function $\phi$
with a matrix monotone
negative derivative. Therefore, the log Vendi score is submodular.
\end{restatable}

This strongly relates the Vendi score with determinantal point
process; the difference is a change of eigenvalue evaluation function
between $\phi(x) = \log(t+x)$ and $\phi(x) = - x \log x$.  We can view
the factorization $B = Q \Lambda \mtrans{Q}$ as a change of basis,
where in the eigenbasis, $\Lambda$ (the diagonal vector of
eigenvalues) measures the side-lengths of a hyper-rectangle
in $m$-dimensions. The determinantal point process measures the
log volume of this hyper-rectangle, while the Vendi score measures the
exponentiated Shannon entropy of the hyper-rectangle's side-lengths.
Other functions $\phi$ measure the side-lengths 
in other ways, but all matrix spectral functions
are fundamentally a function of the spectrum of a positive 
semidefinite (PSD) matrix, i.e.,
the hyper-rectangle's transformed side-lengths.

The Von Neumann (quantum) entropy is defined as
$S(\rho) = -\trace{\rho \log \rho} = -\sum_{i=1}^m \lambda_i(\rho) \log \lambda_i(\rho)$
where $\rho$ is a positive semidefinite Hermitian matrix with trace 1 (i.e., a
density matrix in quantum mechanics) and $\lambda_i(\rho)$ is the $i^\text{th}$
eigenvalue of $\rho$. It was established
in 1973~\cite{lieb1973proof} that Von Neumann entropy satisfies
the strong subadditivity (SSA) property, which can
be stated as follows. Given 
a density matrix $\rho_{ABC}$ 
over three subsystems $A$, $B$, and $C$
(meaning $\rho_{ABC} \in \mathcal B(\mathcal H_A \otimes \mathcal H_B \otimes \mathcal H_C)$,
where $\mathcal H_A$, $\mathcal H_B$, and $\mathcal H_C$ are the Hilbert spaces of the three subsystems
and $\otimes$ corresponds to the tensor, or Kronecker, product), then the
SSA property states that
$S(\rho_{AB}) + S(\rho_{BC}) \geq S(\rho_{ABC}) + S(\rho_B)$
where $\rho_{AB}$, $\rho_{BC}$, and $\rho_B$ are the density matrices obtained by 
taking partial traces over the appropriate subsystems. This 
is indeed {\bf a} submodularity property, but it is
a {\bf different submodularity property } associated
specifically with the $\phi(x) = - x \log x$ function and
properties of the partial trace operation. It is not
the case that $-\phi'$ being matrix monotone is, alone, sufficient
to establish SSA, and thus the SSA property is not a 
consequence of the general matrix spectral submodularity result.
We offer complete details, 
including a counterexample for the lack of SSA given only $-\phi'$ being matrix monotone, 
in Appendix~\ref{sec:von-neumann-entropy}.

\subsection{Weakly Matrix Monotone Functions Lead to Weak Submodularity}
\label{sec:weakly-matrix-monotone}

Section~\ref{sec:matrix_spectral_functions} mentions
that not all monotone non-decreasing concave
functions have matrix monotone negative derivatives.
While Theorem~\ref{thm:simple-analytical-matrix-monotone-functions} lists
some useful ones, this unfortunately rules out some concave functions that 
have been found to be useful for 
the class of submodular functions known as ``feature based'' functions~\cite{bilmes2022submodularity}: these
are functions that take the form $f(A) = \sum_u \phi_u(m_u(A))$ where $m_u(A)$
is a non-negative additive function and $\phi_u$ is concave. We consider
three such concave functions here all on domain $x \geq 0$. First, 
$\mfa(x) = \mfna$,
is a power-law style function inspired from the Chinchilla scaling laws 
from Section~\ref{sec:neural-scaling-laws}. The second
is $\mfb(x) = \mfnb$, which 
is a form of saturating exponential function. A third
function family\footnote{We here in this section
present only a simplification of these functions
in which case $\mfa = \mfc$.
These are generalized, however, to
parameterized forms
in~\Cref{eq:gen_param_mfa}
and~\Cref{eq:gen_param_mfc}
in~\Cref{sec:proofs-weakly-submodular-matrix-spectral-functions}
in which case they are no longer identical for all
values of their parameters.}
we consider is $\mfc(x) = \mfnc$. We have:
\begin{restatable}[Some Useful Concave Functions Do Not Have Matrix Monotone Negative Derivatives]{rlem}{concavebutnotneggradmatrixmonotone}
\label{lem:concave-but-not-matrix-monotone}
The functions $\mfa$, $\mfb$, and $\mfc$ defined
above are not guaranteed
to have matrix monotone negative derivatives and hence
do not guarantee submodularity when applied as matrix spectral functions.
\end{restatable}
The proof is in Appendix~\ref{sec:proofs-weakly-submodular-matrix-spectral-functions}, page~\pageref{proof:concave-but-not-matrix-monotone}.
Hence, we can not expect to apply any concave function that works with a feature-based function
to a matrix spectral function and expect submodularity.

On the other hand, it is well established that the function does not need to be completely
submodular for algorithms like the greedy algorithm (Algorithm~\ref{alg:greedy-max})
to have an approximation guarantee. For example,
consider the following:
\begin{restatable}[$\weakly$-weakly DR submodularity~\citep{el2020optimal,lehmann2001combinatorial}]{rdefn}{weaklyDRsubmodularity}
\label{defn:weakly-DR-submodularity}
A set function $f: 2^V \to \R$ is said to be $\weakly$-weakly DR submodular for
some $\weakly \in [0,1]$ if for all subsets $X \subseteq Y \subseteq V$ and for
all elements $s \in V \setminus Y$, we have 
that $f(X \cup \{s\}) - f(X) \geq \weakly (f(Y \cup \{s\}) - f(Y))$.
\end{restatable}
This means that the marginal gain of adding an element $s$ to a smaller set $X$
is at least $\weakly$ times the marginal gain of adding the same element $s$ to a
larger set $Y$. When $\weakly = 1$, this reduces to the standard definition of
submodularity. When $\weakly < 1$, this allows for some violation of the
diminishing returns property, but still provides some structure that can be
exploited by algorithms. Any $\weakly$-weakly DR submodular function has a greedy
approximation guarantee of $f(X_\text{greedy}) \geq (1 - e^{-\weakly})
\text{OPT}$ where $X_\text{greedy}$ is the set returned by the greedy algorithm
and $\text{OPT}$ is the optimal value of the maximization problem (this is
shown in Section~\ref{sec:background-submodularity}).
We
next introduce a weak generalization of these ideas.
\begin{restatable}[$\weakly$-weakly matrix monotone]{rdefn}{weaklyMatrixMonotone}
\label{defn:weakly-matrix-monotone}
Given function $\psi : \R_+ \to \R$, we say that $\psi$ is $\weakly$-weakly
matrix monotone for some $\weakly \in [0,1]$ if for all $0 \preceq A \preceq B$,
we have $\weakly \psi(A) \preceq \psi(B)$. We say that $\psi$ is
$\weakly$-weakly matrix antitone if for all $0 \preceq A \preceq B$, we have
$\psi(A) \succeq \weakly \psi(B)$.
Moreover, suppose that $g=-\psi$ is a non-positive function (i.e., $\psi \geq 0$).
For $A \preceq B$, we say that $g$ is $\weakly$-weakly matrix 
monotone if $(1/\weakly) g(A) \preceq g(B)$ which is the
same as $-1/\weakly g(A) \succeq - g(B)$, or
that $\psi$ is $\weakly$-antitone.
\end{restatable}

Suppose we are given a function $\phi: \R_+ \to \R$ such that $g = -\phi'$ is not matrix monotone
but instead is only $\weakly$-weakly matrix monotone for some $\weakly \in [0,1]$. The
following shows that this results in a weakly DR submodular function.
\begin{restatable}[Weakly Matrix Monotone Negative Gradients yield Weakly DR Submodular Outer Product Matrix Spectral Functions]
{rthm}{weaklyMatrixMonotoneFunctionsLeadToWeaklySubmodularOuterProductMatrixSpectralFunctions}
\label{thm:weakly-matrix-monotone-functions-lead-to-weakly-submodular-outer-product-matrix-spectral-functions}
Let $\phi: \R_+ \to \R$ be a 
monotone non-decreasing normalized ($\phi(0) = 0$)
function such that $g = -\phi'$ is $\weakly$-weakly
matrix monotone for some $\weakly \in [0,1]$. 
Let $B$ be any symmetric
positive semidefinite $m \times m$ matrix 
expressible as $B = \mtrans{\data} \data$
for some $\data \in \R^{n \times m}$,
and, for $X \subseteq V$, define
$B_X = \mtrans{\data[X]} \data[X]$ as the 
corresponding matrix for subset $X \subseteq V$.
Then the matrix spectral
function $f(X) = \trace{\phi(B_X)}$ is
monotone non-decreasing $\weakly$-weakly DR submodular.\looseness-1
\end{restatable}
The proof of this theorem 
(Appendix~\ref{sec:proofs-weakly-submodular-matrix-spectral-functions},
page~\pageref{proof:weakly-matrix-monotone-functions-lead-to-weakly-submodular-outer-product-matrix-spectral-functions})
is a main theoretical novel contribution of the paper.
We also have the following corollary:
\begin{restatable}[Weakly Matrix Monotone Negative Gradients yield Weakly DR Submodular Matrix Spectral Functions]
{rcor}{weaklyMatrixMonotoneFunctionsLeadToWeaklySubmodularMatrixSpectralFunctions}
\label{cor:weakly-matrix-monotone-functions-lead-to-weakly-submodular-matrix-spectral-functions}
Let $\phi$ be as stated in Theorem~\ref{thm:weakly-matrix-monotone-functions-lead-to-weakly-submodular-outer-product-matrix-spectral-functions}.
Let $B$ be any symmetric positive semidefinite $n \times n$ matrix, 
and, for $X \subseteq V$, define $B_X$ as the principal submatrix of $B$ 
corresponding to subset $X \subseteq V$. 
Then the 
function $f(X) = \trace{\phi(B_X)}$ 
is monotone non-decreasing $\weakly$-weakly DR submodular.\looseness-1
\end{restatable}
The proof of this is
in Appendix~\ref{sec:proofs-weakly-submodular-matrix-spectral-functions},
page~\pageref{proof:weakly-matrix-monotone-functions-lead-to-weakly-submodular-matrix-spectral-functions}.
This result means $B$ can be any PSD matrix, including those that come from a
kernel
function~\cite{bottou2007large,steinwart2008kernels,scholkopf2002learning,williams2006gaussian}.
Further good news is the following.
\begin{restatable}[Practical functions with $\weakly$-weakly matrix monotone negative gradients]
  {rlem}{someusefulconcavefunctionsareweaklymatrixmonotone} 
\label{lem:some-useful-concave-functions-are-weakly-matrix-monotone}
  The functions
$\mfa(x)$, $\mfb(x)$, and $\mfc(x)$ defined above have $\weakly$-weakly matrix
monotone negative gradients for some $\weakly \in (0,1)$ (an open interval).
Depending on the parameters, $\weakly$ may be close to 1.
\end{restatable}
The proof is in
Appendix~\ref{sec:proofs-weakly-submodular-matrix-spectral-functions},
page~\pageref{proof:some-useful-concave-functions-are-weakly-matrix-monotone}.
In particular, the function $f_\mfnuma(X) = \trace{\mfa(B_X)}$ is a 
weakly DR submodular function that may be seen as form of hybrid
between neural scaling laws and the Vendi score. We
evaluate this function in Section~\ref{app:additional_appraisal_functions}.

\subsection{Fast Matrix Spectral Functions via the Secular Equation}
\label{sec:fast-matrix-spectral-functions}

A variety of algorithms associated with submodular functions need to repeatedly
evaluate queries of the form $f(X \cup \{s\})$ for a fixed subset $X \subseteq
V$ and for $s \in V \setminus X$. A good example is the greedy algorithm for
constrained monotone non-decreasing submodular maximization
(Algorithm~\ref{alg:greedy-max}). If we treat each such evaluation as a separate
query, every matrix spectral evaluation of the form $f(X \cup \{s\}) =
\trace{\phi(B_{X \cup \{s\}})}$ would require $O(m^3)$ cost, and this becomes
prohibitively expensive even if the underlying algorithm is
accelerated~\cite{minoux1978_accelerated_greedy}. On the other hand, there is
redundancy over multiple evaluations $f(X \cup \{s\})$ with $X$ fixed and
$s$ varying, since the only thing that changes is the element $s$. For
many submodular functions, this can be made to run much faster~\cite{pmlr-v89-iyer19b}
owing to this redundancy. 

In the case of matrix spectral functions, the underlying operation is
to quickly compute eigenvalues of rank-1 updates to a positive semidefinite (PSD) matrices.
Let $B_X$ be a PSD matrix of rank $r \in [0,m]$ 
corresponding to set $X \subseteq V$ and
let $u_s$ be a length-$m$ vector associated with element $s \in V \setminus X$. 
We maintain $B_X$ in spectral factored form
$B = Q \Lambda \mtrans{Q}$ where $Q$ is an orthonormal matrix of 
eigenvectors and $\Lambda$ is a diagonal matrix of eigenvalues, $r$ of them
positive and the rest zero. Each query $\trace{\phi(B_{X \cup \{s\}})}$ requires 
computing the eigenvalues $\tilde B = B + u_s \mtrans{u_s} = 
\tilde Q \tilde \Lambda \mtrans{\tilde Q}$ where $\tilde Q$ and
$\tilde \Lambda$ are the eigenvectors and eigenvalues of the updated matrix
$\tilde B$. We have:
\begin{align}
\tilde B 
&= Q \Lambda \mtrans{Q} + u_s \mtrans{u}_s 
= Q (\Lambda + v_s \mtrans{v}_s) \mtrans{Q}
= Q (\tilde U \tilde \Lambda \mtrans{\tilde U}) \mtrans{Q}
= \tilde Q \tilde \Lambda \mtrans{\tilde Q}
\end{align}
where $v_s = Q^T u_s$ and $\tilde U$ and $\tilde \Lambda$ are the eigenvectors
and eigenvalues of the rank-1 update $\Lambda + v_s \mtrans{v}_s$. The form
$\Lambda + v_s \mtrans{v}_s$ is special as it is a rank-1 update to a diagonal
matrix. The eigenvalues of a rank-1 update to a diagonal matrix can be computed
in $O(m^2)$ time using the secular
equation~\cite{bunch1978rank,demmel1997applied}. Specifically, this involves
solving the characteristic polynomial $p(\mu) = \det(\Lambda + v_s \mtrans{v}_s
- \mu I) = \prod_j (\lambda_j - \mu) f(\mu)$ where $\lambda_j$ is the
$j^\text{th}$ eigenvalue in $\Lambda$ and $f(\mu) = 1 + \sum_j
\frac{v_{s,j}^2}{\lambda_j - \mu}$ is known as the secular function. The roots
of the secular function have a very specific form (see
Figure~\ref{fig:secular_equation_plot}) and give us the eigenvalues of the updated
matrix $\tilde B$. Therefore, we can compute $\tilde \Lambda$ in $O(m^2)$ time.
Since $\tilde Q$ is still an orthonormal matrix, we have $\trace{\phi(B_{X \cup \{s\}})} =
\trace{\phi(\tilde Q \tilde \Lambda \mtrans{\tilde Q})} = \trace{\phi(\tilde
\Lambda)}$, and thus we can compute $f(X \cup \{s\})$ in $O(m^2)$ time rather
than $O(m^3)$ time. On the other hand, there are many additional details to this
since $B_X$ might be rank-deficient and the secular equation only applies when
there are no repeat eigenvalues and no zeros in $v_s$. As a result, a process of
reduction involving Householder reflectors must be performed to manipulate the
problem into a form where the secular equation can be solved. We give complete
details of this process in Appendix~\ref{sec:fast-rank-1-updates-and-downdates},
including a background on Householder reflectors and the secular equation, and show
this still results in an $O(m^2)$ algorithm per query $f(X \cup \{s\})$.
Figure~\ref{fig:greedy-maximization-speedup-random-data} in
Section~\ref{sec:greedy-maximization-speedup} 
then shows our approach leads to
an average speedup of $\randomspeedup$ for greedy maximization of the Vendi
score relative to the traditional oracle query approach.

Previous work~\cite{chen2018fast} has shown that a Cholesky-style approach can
speed up evaluations for determinantal point processes (e.g., or DPPs, which is the same as Gaussian entropy). This result relies on the
$\log(ab) = \log(a) + \log(b)$ property of logarithms, and thus does not apply
to anything other than the log-det function. Our approach applies to any matrix spectral submodular or
weakly submodular function (including DPPs, which sets $\xi_2(x) = \log(t+x)$)
and also applies to the Renyi-entropy form of the Vendi score. Since~\cite{chen2018fast}
is specific to log-det, it yields faster results for that particular function,
but as we show in Section~\ref{sec:experiments}, the DPP is not the best
performing matrix spectral function.

\vspace{-.1in}
\section{Experiments}
\label{sec:experiments}
\vspace{-.1in}

\begin{figure*}[tbh]
    \centering
    \begin{subfigure}[t]{.248\textwidth}
        \centering
        \includegraphics[width=\linewidth]{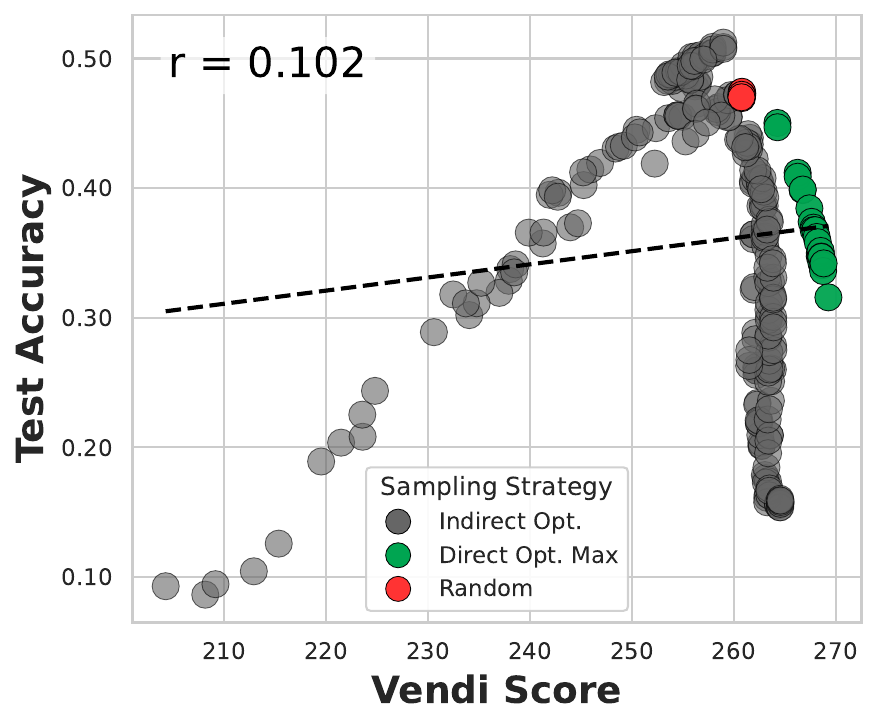}
\caption{\small Vendi Score, Fixed Size}
        \label{fig:fig1a}
    \end{subfigure}\begin{subfigure}[t]{.248\textwidth}
        \centering
        \includegraphics[width=\linewidth]{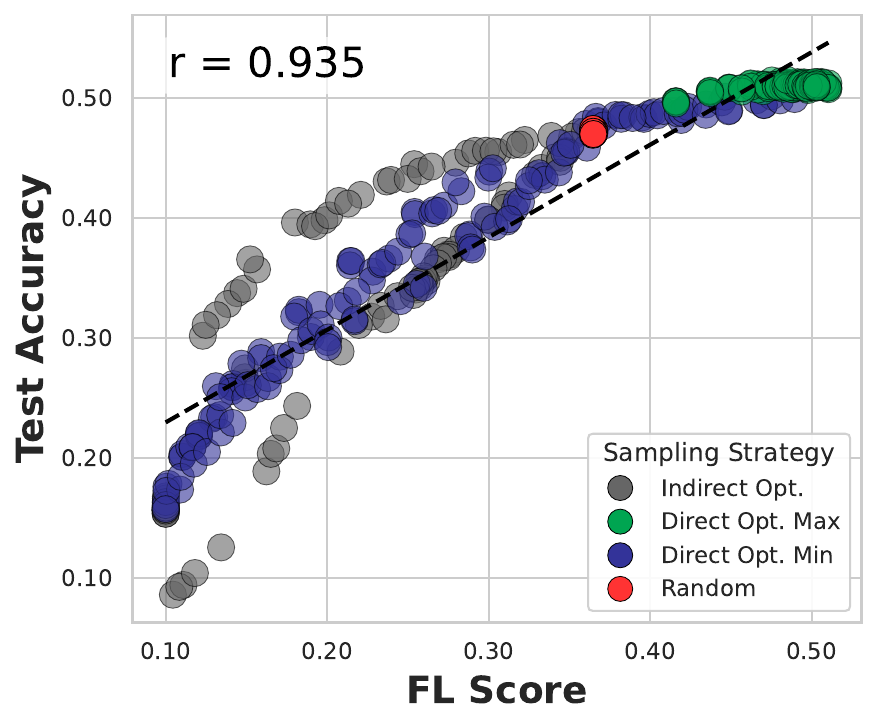}
        \caption{\small FL, Fixed Size}
        \label{fig:fig1b}
    \end{subfigure}\begin{subfigure}[t]{.248\textwidth}
        \centering
        \includegraphics[width=\linewidth]{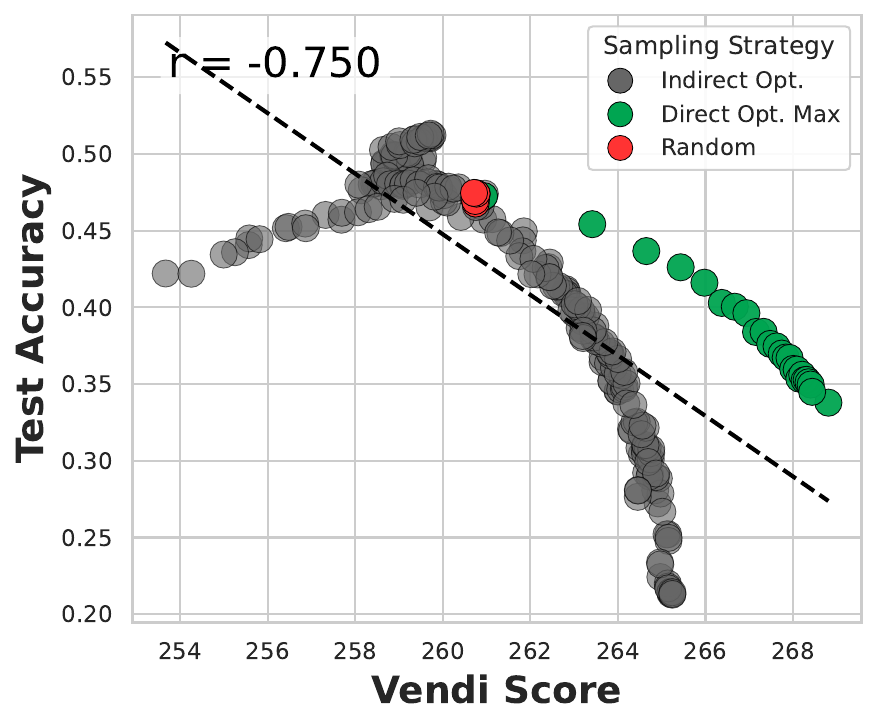}
\caption{Vendi Score, Balanced}
        \label{fig:fig1c}
    \end{subfigure}\begin{subfigure}[t]{.248\textwidth}
        \centering
        \includegraphics[width=\linewidth]{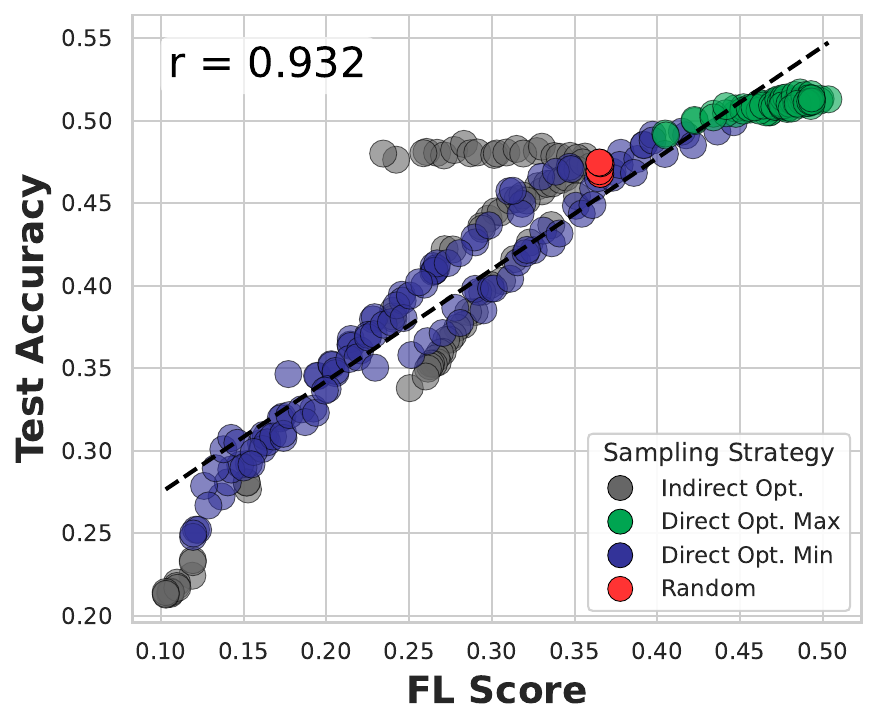}
        \caption{FL, Balanced}
        \label{fig:fig1d}
    \end{subfigure}
    \caption{\small \textbf{Comparing Appraisal Functions} The outcomes of
    several ImageNet-1K runs with identical compute budgets (4M samples seen)
    and training configurations are shown on several sets that are constrained
    to be the same size (10\%) in (a) and (b), and additionally constrained to
    be class-balanced in (c) and (d). Key observations: (1) Vendi correlates
    poorly with test accuracy when evaluated on sets that span a wide range of
    Vendi values. In (a) and (c), note that there is a segment where there is a
    positive correlation with test accuracy and another where there is a
    negative correlation. The latter segment is evaluated on sets that were
    either sampled by maximizing the FL function or directly maximizing the
    Vendi score and thus were previously unexplored
    by~\citep{jung2025prismatic}. (2) random subsets are highly concentrated,
    (3) there is high variance in test performance even within sets that are
    class-balanced and (4) FL score is a  strong predictor of test accuracy.}
    \label{fig:fig1}
\end{figure*}

\begin{figure*}[tbh]
    \centering
    \begin{subfigure}[t]{.248\textwidth}
        \centering
        \includegraphics[width=\linewidth]{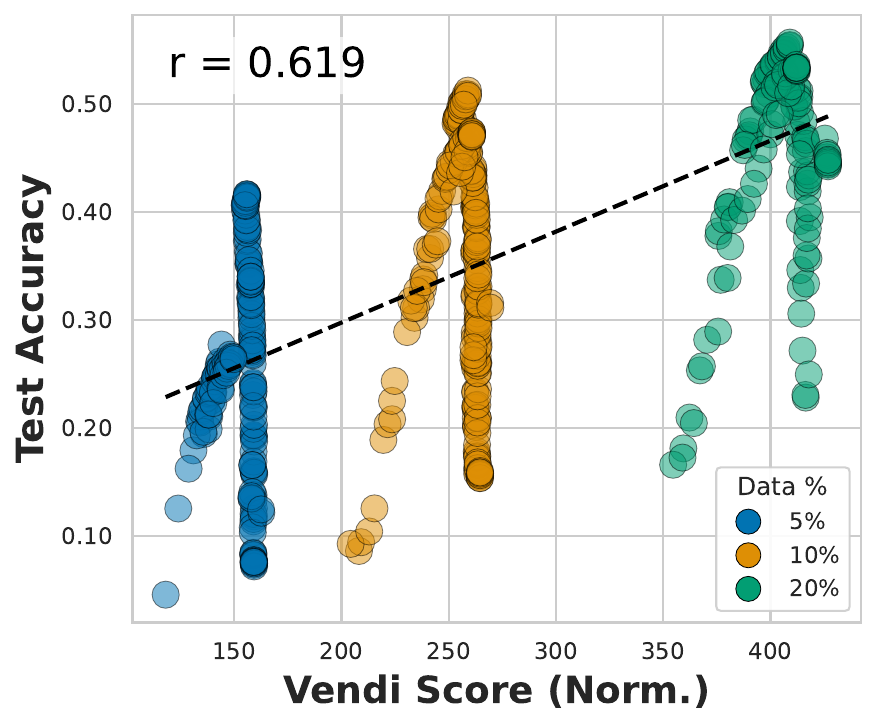}
        \caption{Vendi Score}
        \label{fig:fig2a}
    \end{subfigure}\begin{subfigure}[t]{.248\textwidth}
        \centering
        \includegraphics[width=\linewidth]{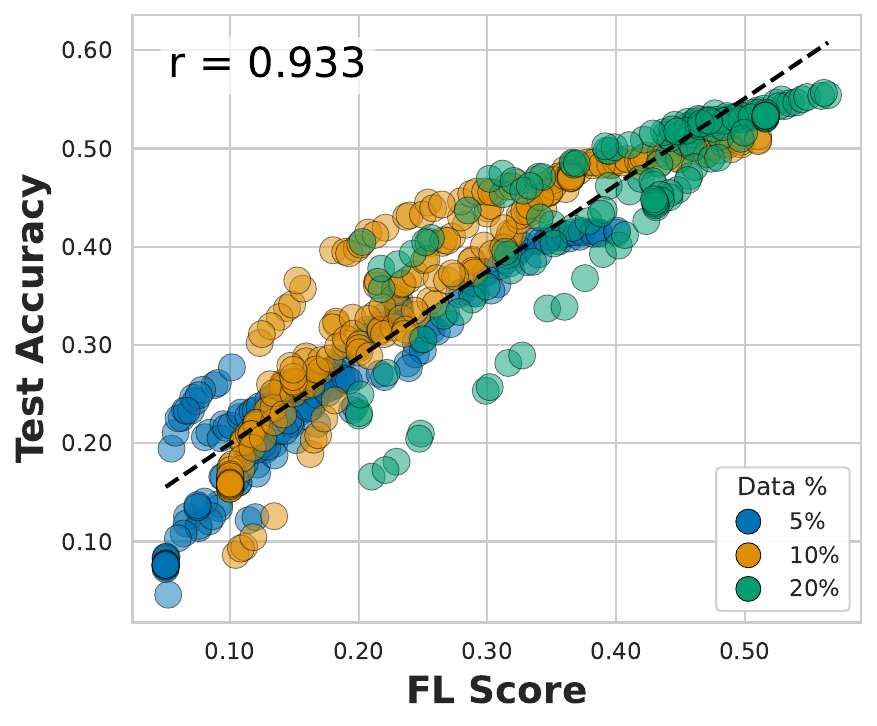}
        \caption{FL}
        \label{fig:fig2b}
    \end{subfigure}\begin{subfigure}[t]{.248\textwidth}
        \centering
        \includegraphics[width=\linewidth]{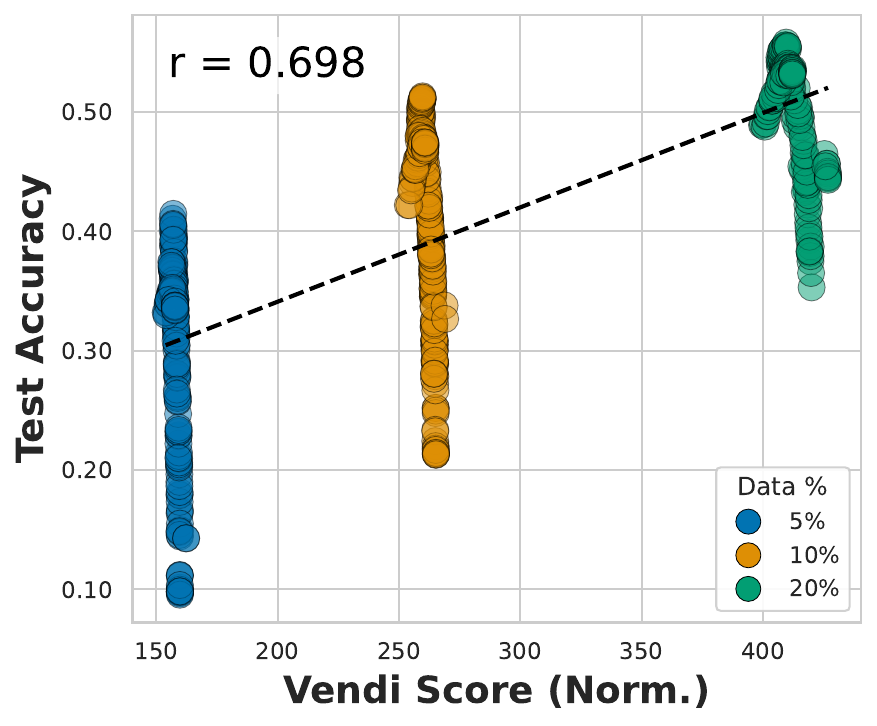}
        \caption{Vendi Score, Balanced}
        \label{fig:fig2c}
    \end{subfigure}\begin{subfigure}[t]{.248\textwidth}
        \centering
        \includegraphics[width=\linewidth]{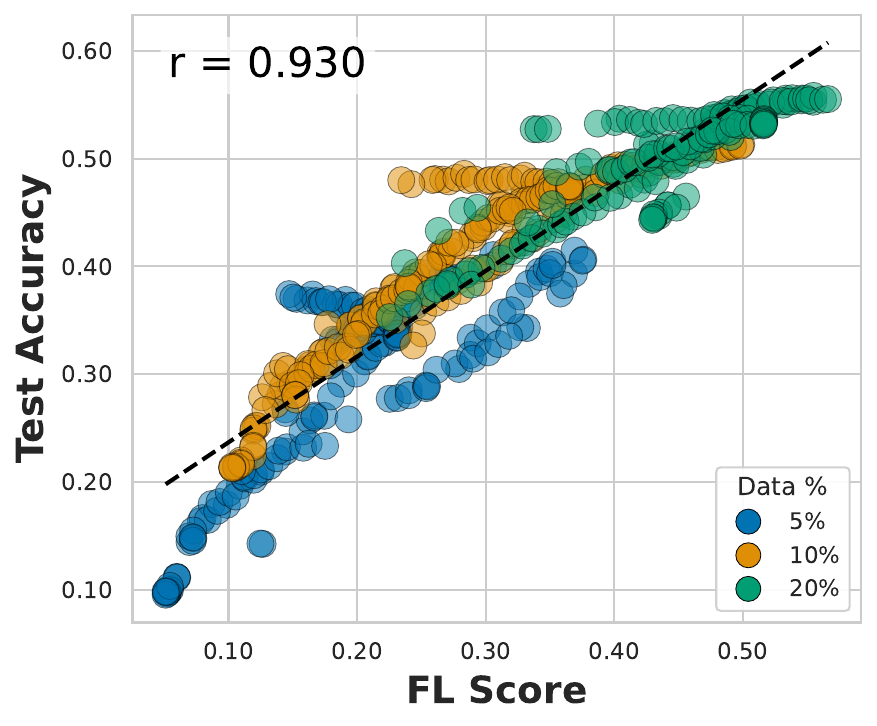}
        \caption{FL, Balanced}
        \label{fig:fig2d}
    \end{subfigure}
    \caption{\small \textbf{Size is a Poor Proxy for Accuracy} The outcomes of
    ImageNet-1K runs with identical compute budgets (4M samples seen) are shown.
    Dataset size is varied within each subplot to 5\%, 10\%, and 20\% of
    $n=1.2$M. We observe that (1) size is a poor proxy for accuracy since there
    are several 5\% subsets that yield higher test performance that 20\% subsets
    and (2) the Vendi score is overly influenced by dataset size. Specifically,
    there are three distinct vertical clusters in panels (a) and (c)
    corresponding to the three subset sizes - the Vendi score stratifies almost
    entirely by how much data was used rather than by how the underlying
    properties correlate to test accuracy.    \looseness-1}
    \label{fig:fig2}
\end{figure*}

We 
compare the utility of various matrix spectral functions for
dataset appraisal with a submodular facility location
(\Cref{eq:facility-location-function}) function. We focus, in this section, on
the Vendi score \& facility location evaluated on ImageNet-1K, but other
functions and datasets (including an Airbnb dataset and 20-newsgroups) are
evaluated in~\Cref{app:additional_datasets}. All test-set evaluations train a
ResNet-18 on a computed subset of ImageNet-1K training data and are tested on
the standard held-out test set (of size 50k). The training hyperparameters over all runs are
fixed except for the learning rate scheduler which is optimized for the total
number of samples seen. Hyperparameter details can be found
in~\Cref{app:in1k_setup}.

To evaluate the functions, we produce a set of subsets $\mathcal S  = \{ S_1,
S_2, \dots \}$ ($S_i \subseteq V$) that span as wide a range of $f(S)$
values, for each $f$, as possible, ranging from very low $f(S)$
to very high, minimizing gaps along the way.
We use three techniques in addition to random selection for
building $\mathcal S$
\textbf{Direct
Optimization (Max).} We run stochastic-greedy~\citep{mirzasoleiman2015lazier}
maximization of $f$, varying the 
stochasticity (and thus approximation quality)
from very low (accurate) to very large (less accurate).
This spans a large sub-range of values of $f(S)$ over the high end of the
total range. \textbf{Direct
Optimization (Heuristic Min).} We use the same stochastic-greedy procedure but at each
step pick the element of \emph{smallest} conditional gain. This is a heuristic
without a mathematical guarantee, but we have found it
to run very fast and it yields very small values of $f(S)$,
spanning a large sub-range of values of $f(S)$ over the low end of the total range.
\textbf{Indirect Optimization.} We use two indirect proxy approaches. First,
following~\citep{jung2025prismatic}, we construct
subsets with low, medium, or high diversity by adding similar examples,
clustering and sampling across clusters, or combining both.
Second,in \emph{cross-function transfer}, a subset optimized directly for $f_1$ is reused
as an indirect sample for other functions $f_2, f_3, \ldots$. Details are in~\Cref{app:in1k_set_sampling_strat}.

To instantiate our ImageNet-1K functions, we use a design matrix $\data$
constructed of per-example normalized last-layer gradients, $\hat x_i = x_i /
\|\hat x_i\|_2$, precisely following~\citet{jung2025prismatic}. We stress that
the facility location objective uses the \emph{same} information available
for all of the matrix spectral functions (including the Vendi Score), ensuring
that no objective receives any inherent advantage. We additionally evaluate
four other matrix spectral functions in~\Cref{app:additional_appraisal_functions}.

Results are summarized in~\Cref{fig:fig1,fig:fig2}. In~\Cref{fig:fig1}, each
point represents the outcome of a training run on a set sampled using one of the
strategies previously discussed. All runs had identical configurations,
including the compute budget. In~\Cref{fig:fig1a,fig:fig1b}, all sets are
constrained to be the same size; in~\Cref{fig:fig1c,fig:fig1d}, they are
additionally constrained to be class-balanced. From these experiments, we make
several empirical observations. \textbf{(1)} The Vendi Score of a training
dataset correlates poorly with test accuracy, as shown
in~\Cref{fig:fig1a,fig:fig1c}. This contradicts the prior results
of~\citep{jung2025prismatic}, but the discrepancy arises because their
experiments do not include samples that span the full range of possible Vendi
Scores. That is, due to our efficient secular-equation based approach 
(\Cref{sec:fast-matrix-spectral-functions})
to matrix spectral functions, 
we are able to examine a much wider range of Vendi score values than what
was possible before; across this broader range, the
correlation with test accuracy is much weaker. \textbf{(2)} The facility location score
correlates well with test accuracy, as shown in~\Cref{fig:fig1b,fig:fig1d}.
Notably, this holds even though the facility location function is computed from
the same embeddings and evaluated on the same training datasets as the Vendi
Score. This finding suggests that facility location is a promising surrogate
objective for data appraisal. \textbf{(3)} There can be high variance even
within subsets of the same size that are perfectly class-balanced as evident
in~\Cref{fig:fig1b,fig:fig1d}. Stratified random sampling is often used as
substitute for diversity sampling, but our results suggests that it is still
suboptimal. \textbf{(4)} Randomly sampled sets are highly concentrated in both
the test accuracies and appraisal scores they attain. We find similar results
for multiple dataset sizes as shown in~\Cref{app:fig1_extension}

In~\Cref{fig:fig2}, all runs again share the same hyperparameter configurations
and compute budget, but each plot now spans training runs across a range of
dataset sizes. Beyond the findings from~\Cref{fig:fig1}, this setting surfaces
two additional observations. \textbf{(5)} Dataset size is a poor proxy for test
accuracy: in every subplot, we observe runs on datasets containing only 5\% of
the full data that outperform runs trained on 20\% subsets, despite identical
compute budgets. \textbf{(6)} The Vendi Score grows with dataset size
(\Cref{fig:fig2a,fig:fig2c}) whereas the facility location score remains
predictive of test accuracy independently of size (\Cref{fig:fig2b,fig:fig2d}).
We show that these findings hold across various compute budgets
in~\Cref{app:fig2_extension}.
In~\Cref{app:additional_in1k}, 
we extend our analysis along three axes: showing
that compute budget alone without taking into account the appropriate quality of the dataset is also a weak predictor of test accuracy (\Cref{app:compute_bad}), evaluating
several alternative matrix spectral methods (\Cref{app:additional_appraisal_functions}), and replicating our findings on
additional datasets (\Cref{app:additional_datasets}).

\subsection{Data Appraisal and Computational Budget}
\label{sec:data-appraisal}

Having established in the previous section,
and further evidenced in~\Cref{app:additional_in1k}
and~\Cref{app:additional_datasets}, that the submodular
facility location function was a better predictor of 
a training subset's ability to achieve a good test-set
accuracy, we in this section consider further
the effect of compute budget on these results. Using
only the facility location function (FL), we consider 
how well it predicts a training set's accuracy based
on both size and quality over a range of compute
budget. These results are shown
in~\Cref{fig:main-teaser} and are further
detailed in~\Cref{app:additional_compute_budgets}.

We use a facility location to produce class-balanced subsets of ImageNet-1K
training data, of size 5\%, 10\%, and 20\% of the full dataset, and this is
performed by maximizing the facility location function subject to a
partition matroid constraint to enforce class balance. This provides our
\textcolor{FigGreen}{green} points in~\Cref{fig:main-teaser}. We also consider
random subsets at those same sizes but also subject to the class balance
constraint, and this is performed by class-specific stratified random sampling,
producing our \textcolor{FigRed}{red} points in the figure. Lastly, we perform
heuristic greedy submodular minimization subject to the same partition matroid
constraint to produce assuredly class balanced subsets that have low facility
location scores, and this is performed by greedily adding the element with the
smallest conditional gain at each step, producing our \textcolor{FigBlue}{blue}
points in the figure. This is a heuristic approach without a mathematical
guarantee, but it runs very fast and produces subsets with low facility location
scores (comparable, we find, to a non-heuristic method using a much more expensive
minimim-norm point algorithm of~\citep{fujishige2005submodular,nagano2011size}).
This provides us with a set of datasets that span a wide range of facility
location scores at different sizes, and thus a wide range of dataset qualities
as valued by the facility location function. 
Visualizations of poritions of both high valued datasets (small
\textcolor{FigGreen}{green \myxshape}) and poor valued datasets (small
\textcolor{FigBlue}{blue \myxshape}) are visulalized
in~\Cref{app:viz_good_bad_sets} to provide further insight into qualitative
differences between the two.

We then train a ResNet-18 on each of these subsets for a range of compute
budgets (from 0.25M upto 32M samples seen, where smaller datasets are repeatedly
trained over multiple epochs to achieve the compute budget, and where the
learning rate schedule is optimized separately for each compute budget), and
evaluate the test accuracy on the standard ImageNet-1K held-out test set (of size 50k). As can
be seen in~\Cref{fig:main-teaser}, there is a strong correlation between the
facility location score and test accuracy across all compute budgets, providing
evidence that this submodular function is a good model of the inherent
information in the subset. Notably, this correlation holds even when controlling
for size, in that there are larger sized subsets (\textcolor{FigBlue}{blue
\mysquareshape} points) with both lower facility location score and lower
test-set accuracy than smaller sized subsets (\textcolor{FigGreen}{green
\myxshape} points). We stress that because all of the created subsets are
perfectly class balanced, the poor \textcolor{FigBlue}{blue} points are {\bf
not} poor due to class imbalance. Also, they are not poor due to a lack of
information in the underlying sample embeddings used to instantiate the FL
function, as the same FL function is used to produce both good and poor subsets
for each size. Rather, these results suggest that it is the value of the
dataset, or perhaps the inherent information in the dataset which is not
determined entirely by size (and as measured by the facility location function)
that is most predictive of test accuracy. This strongly indicates that size
alone is a poor proxy for quality, and should be compared with the neural
scaling laws that, as mentioned in~\Cref{sec:neural-scaling-laws}, value a
dataset based only on size.

We also note that the quality of the dataset is preserved 
when compared across more than two
orders of magnitude of training compute budget (i.e., the highly valued
datasets remain good while the poor valued datasets remain bad, across 
all compute budgets), suggesting that it is not just the amount of compute that
matters. In particular, we note that
there are balanced large low valued datasets 
(i.e., \textcolor{FigBlue}{blue \mysquareshape})
that perform worse in test accuracy than balanced 
small high valued datasets 
(i.e., \textcolor{FigGreen}{green \myxshape}) even at the highest compute budget, 
suggesting that there is more to this story than implied
by~\cite{goyal2024scaling,mohri2026bitter}. Overall, this motivates
more study comparing size, dataset value, compute budget, and resulting
test accuracy on more and different datasets.

An expanded set of results corresponding to~\Cref{fig:main-teaser} are shown
in~\Cref{fig:class-balanced-complete-pyramid}
in~\Cref{app:additional_compute_budgets}. We also provide results in an
unconstrained case, where the facility location function was optimized only for
a particular size, ignoring class balance requirements,
in~\Cref{fig:unconstrained-complete-pyramid}. Unsurprisingly, we find that there
are large poor (\textcolor{FigBlue}{blue \mysquareshape}) subsets, unhindered by
the requirement to be balanced, that are relatively much worse than small good
(\textcolor{FigGreen}{green \myxshape}) subsets that, just because they are
good, will tend to naturally and automatically be more class balanced. 
As yet another reminder, 
please see the visualizations in~\Cref{app:viz_good_bad_sets} which provide further
insight into qualitative differences between high and low valued subsets.

\vspace{-.1in}
\section{Conclusions and Future Work}
\label{sec:conclusions}

We discussed matrix spectral functions, introduced weakly matrix monotone
functions and its relationship to weakly DR submodular functions, and via the
secular equations shown how to make directly optimizing these functions feasible
on large datasets. Our speedups so far apply to the dual eigenspace, where the
fundamental operation boils down to rank-1 updates of a matrix. In future
work, we wish to extend this analysis to arbitrary non-linear
positive definite kernel functions, and derive a new secular-like
expression for that case as well. That would enable
still more expressive matrix spectral functions to become sufficiently
feasible to be evaluated on modern datasets.
Further discussions and conclusions
are given in~\Cref{sec:limitations}.

\section*{References}
\printbibliography[heading=none,nottype=entrytype,title={References}]

@book{demmel1997applied,
  title={Applied numerical linear algebra},
  author={Demmel, James W},
  year={1997},
  publisher={SIAM}
}

@article{golub1973some,
  title={Some modified matrix eigenvalue problems},
  author={Golub, Gene H},
  journal={SIAM review},
  volume={15},
  number={2},
  pages={318--334},
  year={1973},
  publisher={SIAM}
}

@article{bunch1978rank,
  title={Rank-one modification of the symmetric eigenproblem},
  author={Bunch, James R and Nielsen, Christopher P and Sorensen, Danny C},
  journal={Numerische Mathematik},
  volume={31},
  number={1},
  pages={31--48},
  year={1978},
  publisher={Springer}
}

@article{gu1994stable,
  title={A stable and efficient algorithm for the rank-one modification of the symmetric eigenproblem},
  author={Gu, Ming and Eisenstat, Stanley C},
  journal={SIAM journal on Matrix Analysis and Applications},
  volume={15},
  number={4},
  pages={1266--1276},
  year={1994},
  publisher={SIAM}
}

@article{thompson1976behavior,
  title={The behavior of eigenvalues and singular values under perturbations of restricted rank},
  author={Thompson, Robert C},
  journal={Linear Algebra and its Applications},
  volume={13},
  number={1-2},
  pages={69--78},
  year={1976},
  publisher={Elsevier}
}

@Book{wilkinson1965algebraic,
  author={Wilkinson, JH},
  title={{The Algebraic Eigenvalue Problem}},
  publisher={Oxford University Press, Oxford OX2 6DP},
  year={1965},
  OPTkey =       {},
  OPTvolume =    {},
  OPTnumber =    {},
  OPTseries =    {},
  OPTaddress =   {},
  OPTedition =   {},
  OPTmonth =     {},
  OPTnote =      {},
  OPTannote =    {}
}

@inproceedings{dharmasiri2025impact,
  title     = {The Impact of Coreset Selection on Spurious Correlations and Group Robustness},
  author    = {Dharmasiri, Amaya and Yang, William and Kirichenko, Polina and Liu, Lydia T. and Russakovsky, Olga},
  booktitle = {Advances in Neural Information Processing Systems, Datasets and Benchmarks Track},
  year      = {2025}
}

@article{tian2022private,
  title   = {Private Data Valuation and Fair Payment in Data Marketplaces},
  author  = {Tian, Zhihua and Liu, Jian and Li, Jingyu and Cao, Xinle and Jia, Ruoxi and Kong, Jun and Liu, Mengdi and Ren, Kui},
  journal = {arXiv preprint arXiv:2210.08723},
  year    = {2022}
}

@article{hynes2018sterling,
  title   = {A Demonstration of Sterling: A Privacy-Preserving Data Marketplace},
  author  = {Hynes, Nick and Dao, David and Yan, David and Cheng, Raymond and Song, Dawn},
  journal = {Proceedings of the VLDB Endowment},
  volume  = {11},
  number  = {12},
  pages   = {2086--2089},
  year    = {2018}
}

@inproceedings{azcoitia2022try,
  title     = {Try Before You Buy: A Practical Data Purchasing Algorithm for Real-world Data Marketplaces},
  author    = {Azcoitia, Santiago Andr{\'e}s and Laoutaris, Nikolaos},
  booktitle = {Proceedings of the 18th International Conference on Emerging Networking Experiments and Technologies},
  year      = {2022}
}

@article{wong2025privade,
  title   = {{PrivaDE}: Privacy-preserving Data Evaluation for Blockchain-based Data Marketplaces},
  author  = {Wong, Wan Ki and Torkamani, Sahel and Ciampi, Michele and Sarkar, Rik},
  journal = {arXiv preprint arXiv:2510.18109},
  year    = {2025}
}

@techreport{li1993solving,
  title={Solving secular equations stably and efficiently},
  author={Li, Ren-Cang},
  institution =  {Computer Science Dept., University of Tennessee},
  note={Technical Report CS-94-260 (LAPACK Working Note 89.)},
  address =   {Knoxville, TN},
  month=Nov,
  year={1993}
}

@article{melman1997numerical,
  title={A numerical comparison of methods for solving secular equations},
  author={Melman, A},
  journal={Journal of computational and applied mathematics},
  volume={86},
  number={1},
  pages={237--249},
  year={1997},
  publisher={Elsevier}
}

@article{NICHITA2014574,
title = {Rapid and robust resolution of Underwood equations using convex transformations},
journal = {Computers \& Chemical Engineering},
volume = {71},
pages = {574-590},
year = {2014},
issn = {0098-1354},
doi = {https://doi.org/10.1016/j.compchemeng.2014.10.006},
url = {https://www.sciencedirect.com/science/article/pii/S0098135414003019},
author = {Dan Vladimir Nichita and Claude F. Leibovici}
}

@article{brechenmacher2014lagrange,
  title={Lagrange and the secular equation},
  author={Brechenmacher, Fr{\'e}d{\'e}ric},
  journal={Lettera Matematica},
  volume={2},
  number={1},
  pages={79--91},
  year={2014},
  publisher={Springer}
}

@article{ding2007eigenvalues,
  title={Eigenvalues of rank-one updated matrices with some applications},
  author={Ding, Jiu and Zhou, Aihui},
  journal={Applied Mathematics Letters},
  volume={20},
  number={12},
  pages={1223--1226},
  year={2007},
  publisher={Elsevier}
}

@article{paul2021datadiet,
  title={Deep learning on a data diet: Finding important examples early in training},
  author={Paul, Mansheej and Ganguli, Surya and Dziugaite, Gintare Karolina},
  journal={Advances in neural information processing systems},
  volume={34},
  pages={20596--20607},
  year={2021}
}

@inproceedings{toneva2019forgetting,
  title     = {An Empirical Study of Example Forgetting during Deep Neural Network Learning},
  author    = {Toneva, Mariya and Sordoni, Alessandro and Tachet des Combes, Remi and Trischler, Adam and Bengio, Yoshua and Gordon, Geoffrey J.},
  booktitle = {International Conference on Learning Representations},
  year      = {2019}
}

@inproceedings{swayamdipta2020dataset,
  title     = {Dataset Cartography: Mapping and Diagnosing Datasets with Training Dynamics},
  author    = {Swayamdipta, Swabha and Schwartz, Roy and Lourie, Nicholas and Wang, Yizhong and Hajishirzi, Hannaneh and Smith, Noah A. and Choi, Yejin},
  booktitle = {Proceedings of the 2020 Conference on Empirical Methods in Natural Language Processing},
  year      = {2020}
}

@inproceedings{sener2018active,
  title     = {Active Learning for Convolutional Neural Networks: A Core-Set Approach},
  author    = {Sener, Ozan and Savarese, Silvio},
  booktitle = {International Conference on Learning Representations},
  year      = {2018}
}

@article{bachem2017practical,
  title   = {Practical Coreset Constructions for Machine Learning},
  author  = {Bachem, Olivier and Lucic, Mario and Krause, Andreas},
  journal = {arXiv preprint arXiv:1703.06476},
  year    = {2017}
}

@inproceedings{mirzasoleiman2020craig,
  title     = {Coresets for Data-efficient Training of Machine Learning Models},
  author    = {Mirzasoleiman, Baharan and Bilmes, Jeff and Leskovec, Jure},
  booktitle = {Proceedings of the 37th International Conference on Machine Learning},
  year      = {2020}
}

@article{agarwal2005geometric,
  title={Geometric approximation via coresets},
  author={Agarwal, Pankaj K and Har-Peled, Sariel and Varadarajan, Kasturi R and others},
  journal={Combinatorial and computational geometry},
  volume={52},
  number={1},
  pages={1--30},
  year={2005}
}

@inproceedings{mirzasoleiman2015lazier,
  title={Lazier than lazy greedy},
  author={Mirzasoleiman, Baharan and Badanidiyuru, Ashwinkumar and Karbasi, Amin and Vondr{\'a}k, Jan and Krause, Andreas},
  booktitle={Proceedings of the AAAI Conference on Artificial Intelligence},
  volume={29},
  number={1},
  year={2015}
}

@inproceedings{killamsetty2021gradmatch,
  title     = {{GRAD-MATCH}: Gradient Matching Based Data Subset Selection for Efficient Deep Model Training},
  author    = {Killamsetty, Krishnateja and Sivasubramanian, Durga and Ramakrishnan, Ganesh and De, Abir and Iyer, Rishabh},
  booktitle = {Proceedings of the 38th International Conference on Machine Learning},
  year      = {2021}
}

@inproceedings{killamsetty2021glister,
  title     = {{GLISTER}: Generalization Based Data Subset Selection for Efficient and Robust Learning},
  author    = {Killamsetty, Krishnateja and Sivasubramanian, Durga and Ramakrishnan, Ganesh and Iyer, Rishabh},
  booktitle = {Proceedings of the AAAI Conference on Artificial Intelligence},
  year      = {2021}
}

@inproceedings{ghorbani2019datashapley,
  title     = {Data Shapley: Equitable Valuation of Data for Machine Learning},
  author    = {Ghorbani, Amirata and Zou, James},
  booktitle = {Proceedings of the 36th International Conference on Machine Learning},
  year      = {2019}
}

@inproceedings{kwon2022betashapley,
  title     = {Beta Shapley: A Unified and Noise-reduced Data Valuation Framework for Machine Learning},
  author    = {Kwon, Yongchan and Zou, James},
  booktitle = {Proceedings of the 25th International Conference on Artificial Intelligence and Statistics},
  year      = {2022}
}

@inproceedings{yoon2020dvrl,
  title     = {Data Valuation using Reinforcement Learning},
  author    = {Yoon, Jinsung and Arik, Sercan O. and Pfister, Tomas},
  booktitle = {Proceedings of the 37th International Conference on Machine Learning},
  year      = {2020}
}

@article{friedman2022vendi,
  title={The vendi score: A diversity evaluation metric for machine learning},
  author={Friedman, Dan and Dieng, Adji Bousso},
  journal={arXiv preprint arXiv:2210.02410},
  year={2022}
}

@article{friedmanvenditmlr,
  title={The Vendi Score: A Diversity Evaluation Metric for Machine Learning},
  author={Friedman, Dan and Dieng, Adji Bousso},
  journal={Transactions on Machine Learning Research},
  year={2023}
}

@inproceedings{pasarkar2024cousins,
  title={Cousins Of The Vendi Score: A Family Of Similarity-Based Diversity Metrics For Science And Machine Learning},
  author={Pasarkar, Amey P and Dieng, Adji Bousso},
  booktitle={International Conference on Artificial Intelligence and Statistics},
  pages={3808--3816},
  year={2024},
  organization={PMLR}
}

@article{jung2025prismatic,
  title={Prismatic synthesis: Gradient-based data diversification boosts generalization in llm reasoning},
  author={Jung, Jaehun and Han, Seungju and Lu, Ximing and Hallinan, Skyler and Acuna, David and Prabhumoye, Shrimai and Patwary, Mostafa and Shoeybi, Mohammad and Catanzaro, Bryan and Choi, Yejin},
  journal={arXiv preprint arXiv:2505.20161},
  year={2025}
}

@article{pasarkar2026vendi,
  title={Vendi Novelty Scores for Out-of-Distribution Detection},
  author={Pasarkar, Amey P and Dieng, Adji Bousso},
  journal={arXiv preprint arXiv:2602.10062},
  year={2026}
}

@article{nguyen2025vendi,
  title={Vendi information gain: An alternative to mutual information for science and machine learning},
  author={Nguyen, Quan and Dieng, Adji Bousso},
  journal={arXiv preprint arXiv:2505.09007},
  year={2025}
}

@article{sirigiri2026diversity,
  title={Diversity You Can Actually Measure: A Fast, Model-Free Diversity Metric for Robotics Datasets},
  author={Sirigiri, Sreevardhan and de Lara, Nathan Samuel and Agia, Christopher and Shkurti, Florian and Ramos, Fabio},
  journal={arXiv preprint arXiv:2603.11634},
  year={2026}
}

@inproceedings{jalali2026conditional,
  title={Conditional Vendi Score: Prompt-Aware Diversity Evaluation for Generative {AI} Models and {LLM}s},
  author={Mohammad Jalali and Azim Ospanov and Amin Gohari and Farzan Farnia},
  booktitle={The 29th International Conference on Artificial Intelligence and Statistics},
  year={2026},
  url={https://openreview.net/forum?id=iDrZToIsyd}
}

@article{iyer2021generalized,
  title={Generalized submodular information measures: Theoretical properties, examples, optimization algorithms, and applications},
  author={Iyer, Rishabh and Khargonkar, Ninad and Bilmes, Jeff and Asnani, Himanshu},
  journal={IEEE Transactions on Information Theory},
  volume={68},
  number={2},
  pages={752--781},
  year={2021},
  publisher={IEEE}
}

@misc{HestnessEtAl2017,
  author = {Hestness, Joel and Narang, Sharan and Ardalani, Newsha and Diamos, Greg and Jun, Heewoo and Kianinejad, Hassan and Patwary, Md. Mostofa Ali and Yang, Yang and Zhou, Yanqi},
  title  = {Deep Learning Scaling Is Predictable, Empirically},
  year   = {2017},
  howpublished = {arXiv preprint arXiv:1712.00409},
  url    = {https://arxiv.org/abs/1712.00409}
}

@misc{KaplanEtAl2020,
  author = {Kaplan, Jared and McCandlish, Sam and Henighan, Tom and Brown, Tom B. and Chess, Benjamin and Child, Rewon and Gray, Scott and Radford, Alec and Wu, Jeffrey and Amodei, Dario},
  title  = {Scaling Laws for Neural Language Models},
  year   = {2020},
  howpublished = {arXiv preprint arXiv:2001.08361},
  url    = {https://arxiv.org/abs/2001.08361}
}

@article{hoffmann2022training,
  title        = {Training Compute-Optimal Large Language Models},
  author       = {Hoffmann, Jordan and Borgeaud, Sebastian and Mensch, Arthur and Buchatskaya, Elena and Cai, Trevor and Rutherford, Eliza and de Las Casas, Diego and Hendricks, Lisa Anne and Welbl, Johannes and Hennigan, Tom and Noland, Eric and Milligan, Katie and van den Driessche, George and Damoc, Bogdan and Guy, Aurelia and Osindero, Simon and Simonyan, Karen and Elsen, Erich and Rae, Jack W. and Vinyals, Oriol and Sifre, Laurent},
  journal      = {arXiv preprint arXiv:2203.15556},
  year         = {2022},
  eprint       = {2203.15556},
  url          = {https://arxiv.org/abs/2203.15556}
}

@misc{pearce2024reconcilingkaplanchinchillascaling,
      title={Reconciling Kaplan and Chinchilla Scaling Laws}, 
      author={Tim Pearce and Jinyeop Song},
      year={2024},
      eprint={2406.12907},
      archivePrefix={arXiv},
      primaryClass={cs.LG},
      url={https://arxiv.org/abs/2406.12907}, 
      note={discrepancy of Hoffman vs Kaplan due to to Kaplan counting non-embedding parameters only (rather than all parameters) and performing fits at small scale.}
}

@misc{porian2025resolvingdiscrepanciescomputeoptimalscaling,
      title={Resolving Discrepancies in Compute-Optimal Scaling of Language Models}, 
      author={Tomer Porian and Mitchell Wortsman and Jenia Jitsev and Ludwig Schmidt and Yair Carmon},
      year={2025},
      eprint={2406.19146},
      archivePrefix={arXiv},
      primaryClass={cs.LG},
      url={https://arxiv.org/abs/2406.19146}, 
      note={discrepancy of Hoffman vs Kaplan due to: need to correct for three factors (last-layer FLOPs, warmup schedule, and scale-dependent hyperparameter tuning), if done then Kaplan and Hoffmann predictions are much closer}
}

@misc{dey2023cerebrasgptopencomputeoptimallanguage,
      title={{Cerebras-GPT: Open Compute-Optimal Language Models Trained on the Cerebras Wafer-Scale Cluster}}, 
      author={Nolan Dey and Gurpreet Gosal and Zhiming and Chen and Hemant Khachane and William Marshall and Ribhu Pathria and Marvin Tom and Joel Hestness},
      year={2023},
      eprint={2304.03208},
      archivePrefix={arXiv},
      primaryClass={cs.LG},
      url={https://arxiv.org/abs/2304.03208}, 
}

@InProceedings{villalobos2024run,
      title={{Will we run out of data? Limits of LLM scaling based on human-generated data}}, 
      author={Pablo Villalobos and Anson Ho and Jaime Sevilla and Tamay Besiroglu and Lennart Heim and Marius Hobbhahn},
      year={2024},
      eprint={2211.04325},
      archivePrefix={arXiv},
      booktitle={Forty-first International Conference on Machine Learning},
      primaryClass={cs.LG}
}

@misc{digitalstrategy_ec_eu_ai_act_gpaicompute_thresholds,
  author       = {European Commission / Digital Strategy Office},
  title        = {{General Purpose AI Models and Systemic Risk — FAQ (AI Act)}},
  howpublished = {Online FAQ / policy website},
  year         = {2025},
  note         = {Compute thresholds in the EU AI Act, citing scaling stance},
  url          = {https://digital-strategy.ec.europa.eu/en/faqs/general-purpose-ai-models-ai-act-questions-answers}
}

@misc{BusinessInsider_Goldman_Data_2025,
  author       = {Huileng Tan},
  title        = {The {AI} boom has already run out of human-made training data, {G}oldman’s data chief says},
  year         = {2025},
  month        = {10},
  day          = {02},
  url          = {https://www.businessinsider.com/ai-boom-has-run-out-of-human-made-training-data-goldman-2025-10},
  organization = {Business Insider}
}

@misc{usco_part3_ai_training_2025,
  author       = {U.S. Copyright Office},
  title        = {{Copyright and Artificial Intelligence, Part 3: Generative AI Training (Pre-Publication Version)}},
  year         = {2025},
  month        = {May},
  day          = {9},
  howpublished = {U.S. Copyright Office Report},
  url          = {https://www.copyright.gov/ai/Copyright-and-Artificial-Intelligence-Part-3-Generative-AI-Training-Report-Pre-Publication-Version.pdf},
  note         = {Pre-publication version addressing training AI models on copyrighted works}
}

@misc{kang2025autoscalescaleawaredatamixing,
      Title={AutoScale: Scale-Aware Data Mixing for Pre-Training LLMs}, 
      author={Feiyang Kang and Yifan Sun and Bingbing Wen and Si Chen and Dawn Song and Rafid Mahmood and Ruoxi Jia},
      year={2025},
      eprint={2407.20177},
      archivePrefix={arXiv},
      primaryClass={cs.LG},
      url={https://arxiv.org/abs/2407.20177}, 
}

@Article{friedland2011submodular,
  title		= {Submodular spectral functions of principal submatrices of
		  a hermitian matrix, extensions and applications},
  author	= {Friedland, S and Gaubert, S},
  journal	= {Linear Algebra and its Applications},
  year		= {2011},
  publisher	= {Elsevier}
}

@article{loewner1934monotone,
  title={{\"U}ber monotone matrixfunktionen},
  author={L{\"o}wner, Karl},
  journal={Mathematische Zeitschrift},
  volume={38},
  number={1},
  pages={177--216},
  year={1934},
  publisher={Springer}
}

@article{bendat1955monotone,
  title={Monotone and convex operator functions},
  author={Bendat, Julius and Sherman, Seymour},
  journal={Transactions of the American Mathematical Society},
  volume={79},
  number={1},
  pages={58--71},
  year={1955},
  publisher={JSTOR},
  note={Good description of Loewner's result and gives an easy test, from Loewner, to find a counterexample for function being matrix monotone}
}

@book{higham2008functions,
  title={Functions of matrices: theory and computation},
  author={Higham, Nicholas J},
  year={2008},
  publisher={SIAM}
}

@book{lancaster1985theory,
  title={The theory of matrices: with applications},
  author={Lancaster, Peter and Tismenetsky, Miron},
  year={1985},
  publisher={Elsevier}
}

@book{bhatia1997matrix,
  title={Matrix analysis},
  author={Bhatia, Rajendra},
  year={1997},
  publisher={Springer Science \& Business Media}
}

@incollection{bhatia2009positive,
  title={Positive definite matrices},
  author={Bhatia, Rajendra},
  booktitle={Positive Definite Matrices},
  year={2009},
  publisher={Princeton university press}
}

@article{bhatia2009loewner,
  title={Loewner matrices and operator convexity},
  author={Bhatia, Rajendra and Sano, Takashi},
  journal={Mathematische Annalen},
  volume={344},
  number={3},
  pages={703--716},
  year={2009},
  publisher={Springer},
  note={Another very good description of Loewner's result and gives an easy test, from Loewner, to find a counterexample for function being matrix monotone}
}

@book{simon2019loewner,
  title={Loewner's theorem on monotone matrix functions},
  author={Simon, Barry},
  volume={10},
  year={2019},
  publisher={Springer},
  note={yet one more other very good description of Loewner's result and gives an easy test, from Loewner, to find a counterexample for function being matrix monotone}
}

@article{bennett2002quantum,
  title={Quantum information theory},
  author={Bennett, Charles H. and Shor, Peter W.},
  journal={IEEE transactions on information theory},
  volume={44},
  number={6},
  pages={2724--2742},
  year={2002},
  publisher={IEEE}
}

@book{petz2007quantum,
  title={Quantum information theory and quantum statistics},
  author={Petz, D{\'e}nes},
  year={2007},
  publisher={Springer Science \& Business Media}
}

@article{lieb1973proof,
  title={Proof of the strong subadditivity of quantum-mechanical entropy},
  author={Lieb, Elliott H and Ruskai, Mary Beth},
  journal={Les rencontres physiciens-math{\'e}maticiens de Strasbourg-RCP25},
  volume={19},
  pages={36--55},
  year={1973}
}

@article{audenaert2010strongly,
  title={Strongly subadditive functions},
  author={Audenaert, Koenraad and Hiai, Fumio and Petz, D{\'e}nes},
  journal={Acta Mathematica Hungarica},
  volume={128},
  number={4},
  pages={386--394},
  year={2010},
  publisher={Akad{\'e}miai Kiad{\'o}, co-published with Springer Science+ Business Media BV~…}
}

@article{lewin2014family,
  title={A family of monotone quantum relative entropies},
  author={Lewin, Mathieu and Sabin, Julien},
  journal={Letters in Mathematical Physics},
  volume={104},
  number={6},
  pages={691--705},
  year={2014},
  publisher={Springer}
}

@book{nielsen2010quantum,
  title={Quantum computation and Quantum information},
  author={Nielsen, Michael A and Chuang, Isaac L},
  year={2010},
  publisher={Cambridge university press}
}

@article{kulesza2012determinantal,
  title={Determinantal point processes for machine learning},
  author={Kulesza, Alex and Taskar, Ben},
  journal={Foundations and Trends{\textregistered} in Machine Learning},
  volume={5},
  number={2-3},
  pages={123--286},
  year={2012},
  publisher={Emerald Publishing Limited}
}

@book{fujishige2005submodular,
  title={Submodular functions and optimization},
  author={Fujishige, Satoru},
  volume={58},
  year={2005},
  publisher={Elsevier}
}

@inproceedings{nagano2011size,
  title={Size-constrained submodular minimization through minimum norm base},
  author={Nagano, Kiyohito and Kawahara, Yoshinobu and Aihara, Kazuyuki},
  booktitle={Proceedings of the 28th International Conference on Machine Learning (ICML-11)},
  pages={977--984},
  year={2011}
}

@article{bilmes2022submodularity,
  title={Submodularity in machine learning and artificial intelligence},
  author={Bilmes, Jeff},
  journal={arXiv preprint arXiv:2202.00132},
  year={2022}
}

@article{bilmes2017deep,
  title={Deep submodular functions},
  author={Bilmes, Jeffrey and Bai, Wenruo},
  journal={arXiv preprint arXiv:1701.08939},
  year={2017}
}

@book{topkis1998supermodularity,
  title={Supermodularity and complementarity},
  author={Topkis, Donald M},
  year={1998},
  publisher={Princeton university press}
}

@article{nemhauser1978analysis,
  title={An analysis of approximations for maximizing submodular set functions—I},
  author={Nemhauser, George L and Wolsey, Laurence A and Fisher, Marshall L},
  journal={Mathematical programming},
  volume={14},
  number={1},
  pages={265--294},
  year={1978},
  publisher={Springer}
}

@incollection{minoux1978_accelerated_greedy,
   author = {Minoux, Michel},
   affiliation = {Ecole Nationale Supérieure de Techniques Avancées Paris France Paris France},
   title = {Accelerated greedy algorithms for maximizing submodular set functions},
   booktitle = {Optimization Techniques},
   series = {Lecture Notes in Control and Information Sciences},
   editor = {Stoer, J.},
   publisher = {Springer Berlin / Heidelberg},
   isbn = {},
   pages = {234-243},
   volume = {7},
   url = {http://dx.doi.org/10.1007/BFb0006528},
   note = {10.1007/BFb0006528},
   year = {1978}
}

@article{tohidi2020submodularity,
  title={Submodularity in action: From machine learning to signal processing applications},
  author={Tohidi, Ehsan and Amiri, Rouhollah and Coutino, Mario and Gesbert, David and Leus, Geert and Karbasi, Amin},
  journal={IEEE Signal Processing Magazine},
  volume={37},
  number={5},
  pages={120--133},
  year={2020},
  publisher={IEEE}
}

@inproceedings{kumari2024bumblebee,
  title={Bumblebee: Dynamic kv-cache streaming submodular summarization for infinite-context transformers},
  author={Kumari, Lilly and Wang, Shengjie and Zhou, Tianyi and Sarda, Nikhil and Rowe, Anthony and Bilmes, Jeff},
  booktitle={First Conference on Language Modeling},
  year={2024}
}

@inproceedings{kumari2024end,
  title={An end-to-end submodular framework for data-efficient in-context learning},
  author={Kumari, Lilly and Wang, Shengjie and Das, Arnav and Zhou, Tianyi and Bilmes, Jeff},
  booktitle={Findings of the Association for Computational Linguistics: NAACL 2024},
  pages={3293--3308},
  year={2024}
}

@book{bottou2007large,
  title={Large-scale kernel machines},
  author={Bottou, L{\'e}on},
  year={2007},
  publisher={MIT press}
}

@incollection{steinwart2008kernels,
  title={Kernels and reproducing kernel hilbert spaces},
  author={Steinwart, Ingo and Christmann, Andreas},
  booktitle={Support Vector Machines},
  pages={110--163},
  year={2008},
  publisher={Springer}
}

@book{williams2006gaussian,
  title={Gaussian processes for machine learning},
  author={Williams, Christopher KI and Rasmussen, Carl Edward},
  volume={2},
  number={3},
  year={2006},
  publisher={MIT press Cambridge, MA}
}

@book{scholkopf2002learning,
  title={Learning with kernels: support vector machines, regularization, optimization, and beyond},
  author={Sch{\"o}lkopf, Bernhard and Smola, Alexander J},
  year={2002},
  publisher={MIT press}
}

@article{gatmiry2018non,
  title={Non-submodular function maximization subject to a matroid constraint, with applications},
  author={Gatmiry, Khashayar and Gomez-Rodriguez, Manuel},
  journal={arXiv preprint arXiv:1811.07863},
  year={2018}
}

@inproceedings{bogunovic2018robust,
  title={Robust maximization of non-submodular objectives},
  author={Bogunovic, Ilija and Zhao, Junyao and Cevher, Volkan},
  booktitle={International Conference on Artificial Intelligence and Statistics},
  pages={890--899},
  year={2018},
  organization={PMLR}
}

@InProceedings{pmlr-v84-el-halabi18a,
  title = 	 {Combinatorial Penalties: Which structures are preserved by convex relaxations?},
  author = 	 {El Halabi, Marwa and Bach, Francis and Cevher, Volkan},
  booktitle = 	 {Proceedings of the Twenty-First International Conference on Artificial Intelligence and Statistics},
  pages = 	 {1551--1560},
  year = 	 {2018},
  editor = 	 {Storkey, Amos and Perez-Cruz, Fernando},
  volume = 	 {84},
  series = 	 {Proceedings of Machine Learning Research},
  month = 	 {09--11 Apr},
  publisher =    {PMLR},
  url = 	 {https://proceedings.mlr.press/v84/el-halabi18a.html}
}

@inproceedings{lehmann2001combinatorial,
  title={Combinatorial auctions with decreasing marginal utilities},
  author={Lehmann, Benny and Lehmann, Daniel and Nisan, Noam},
  booktitle={Proceedings of the 3rd ACM conference on Electronic Commerce},
  pages={18--28},
  year={2001},
  note={earlier version of lehmann2006combinatorial}
}

@article{chen2018fast,
  title={Fast greedy map inference for determinantal point process to improve recommendation diversity},
  author={Chen, Laming and Zhang, Guoxin and Zhou, Eric},
  journal={Advances in neural information processing systems},
  volume={31},
  year={2018}
}

@article{das2018approximate,
  title={Approximate submodularity and its applications: Subset selection, sparse approximation and dictionary selection},
  author={Das, Abhimanyu and Kempe, David},
  journal={Journal of Machine Learning Research},
  volume={19},
  number={3},
  pages={1--34},
  year={2018}
}

@inproceedings{das2011submodular,
  title={Submodular meets spectral: greedy algorithms for subset selection, sparse approximation and dictionary selection},
  author={Das, Abhimanyu and Kempe, David},
  booktitle={Proceedings of the 28th International Conference on International Conference on Machine Learning},
  pages={1057--1064},
  year={2011}
}

@inproceedings{el2020optimal,
  title={Optimal approximation for unconstrained non-submodular minimization},
  author={El Halabi, Marwa and Jegelka, Stefanie},
  booktitle={International Conference on Machine Learning},
  pages={3961--3972},
  year={2020},
  organization={PMLR}
}

@inproceedings{bian2017guarantees,
  title={Guarantees for greedy maximization of non-submodular functions with applications},
  author={Bian, Andrew An and Buhmann, Joachim M and Krause, Andreas and Tschiatschek, Sebastian},
  booktitle={International conference on machine learning},
  pages={498--507},
  year={2017},
  organization={PMLR}
}

@article{hassani2017gradient,
  title={Gradient methods for submodular maximization},
  author={Hassani, Hamed and Soltanolkotabi, Mahdi and Karbasi, Amin},
  journal={Advances in Neural Information Processing Systems},
  volume={30},
  year={2017}
}

@inproceedings{el2018combinatorial,
  title={Combinatorial Penalties: Which structures are preserved by convex relaxations?},
  author={El Halabi, Marwa and Bach, Francis and Cevher, Volkan},
  booktitle={International Conference on Artificial Intelligence and Statistics},
  pages={1551--1560},
  year={2018},
  organization={PMLR}
}

@InProceedings{pmlr-v89-iyer19b,
  title = 	 {A Memoization Framework for Scaling Submodular Optimization to Large Scale Problems},
  author =       {Iyer, Rishabh and Bilmes, Jeffrey},
  booktitle = 	 {Proceedings of the Twenty-Second International Conference on Artificial Intelligence and Statistics},
  pages = 	 {2340--2349},
  year = 	 {2019},
  editor = 	 {Chaudhuri, Kamalika and Sugiyama, Masashi},
  volume = 	 {89},
  series = 	 {Proceedings of Machine Learning Research},
  month = 	 {16--18 Apr},
  publisher =    {PMLR},
  url = 	 {https://proceedings.mlr.press/v89/iyer19b.html}
}

@misc{leclerc2022ffcv,
    author = {Guillaume Leclerc and Andrew Ilyas and Logan Engstrom and Sung Min Park and Hadi Salman and Aleksander Madry},
    title = {ffcv},
    year = {2022},
    howpublished = {\url{https://github.com/libffcv/ffcv/}},
    note = {commit xxxxxxx}
}

@inproceedings{goyal2024scaling,
  title={Scaling Laws for Data Filtering--Data Curation cannot be Compute Agnostic},
  author={Goyal, Sachin and Maini, Pratyush and Lipton, Zachary C and Raghunathan, Aditi and Kolter, J Zico},
  booktitle={Proceedings of the IEEE/CVF Conference on Computer Vision and Pattern Recognition},
  pages={22702--22711},
  year={2024}
}

@article{mohri2026bitter,
  title={A Bitter Lesson for Data Filtering},
  author={Mohri, Christopher and Duchi, John and Hashimoto, Tatsunori},
  journal={arXiv preprint arXiv:2605.19407},
  year={2026}
}

@article{peirce1906prolegomena,
  title={Prolegomena to an apology for pragmaticism},
  author={Peirce, Charles Santiago Sanders},
  journal={The Monist},
  pages={492--546},
  year={1906},
  publisher={JSTOR}
}

@Misc{bilmes2026-submarine,
  author = 	 {Jeff Bilmes},
  title = 	 {Submarine: {SUBM}odularity for {AR}tificial {IN}telligenc{E} and machine learning},
  howpublished = {Online Software System},
  year = 	 {2026},
  note = {\url{https://submarine.page}},
}

@InProceedings{lin2009-submod-active-seq,
  author = 	 {Hui Lin and Jeff A. Bilmes},
  title = 	 {How to Select a Good Training-data Subset for Transcription: Submodular Active Selection for Sequences},
  booktitle =    {Proc.\ Annual Conference of the International Speech Communication Association (INTERSPEECH)},
  year = 	 {2009},
  address = 	 {Brighton, UK},
  month = {September},
}

@InProceedings{lin2009-submodsum,
  author = 	 {Hui Lin and Jeff Bilmes and Shasha Xie},
  title = 	 {Graph-based Submodular Selection for Extractive Summarization},
  booktitle =    {Proc.\ IEEE Automatic Speech Recognition and Understanding (ASRU)},
  year = 	 {2009},
  month = 	 {December},
  address = 	 {Merano, Italy},
}

@InProceedings{lin2011-class-submod-sum,
  author = 	 {Hui Lin and Jeff Bilmes},
  title = 	 {A Class of Submodular Functions for Document Summarization},
  booktitle = 	 {The 49th Annual Meeting of the Association for Computational Linguistics: Human Language Technologies (ACL/HLT-2011)},
  year = 	 {2011},
  address = 	 {Portland, OR},
  month = 	 {June},
  note = {(long paper)},
}

@inproceedings{kisel2025flawsofimagenet,
  author = {Kisel, Nikita and Volkov, Illia and Hanzelková, Kateřina and Janouskova, Klara and Matas, Jiri},
  title = {Flaws of ImageNet, Computer Vision's Favorite Dataset},
  booktitle = {ICLR Blogposts 2025},
  year = {2025},
  date = {April 28, 2025},
  note = {https://iclr-blogposts.github.io/2025/blog/imagenet-flaws/},
  url  = {https://iclr-blogposts.github.io/2025/blog/imagenet-flaws/}
}

@inproceedings{yun2021re,
  title={Re-labeling imagenet: from single to multi-labels, from global to localized labels},
  author={Yun, Sangdoo and Oh, Seong Joon and Heo, Byeongho and Han, Dongyoon and Choe, Junsuk and Chun, Sanghyuk},
  booktitle={Proceedings of the IEEE/CVF conference on computer vision and pattern recognition},
  pages={2340--2350},
  year={2021}
}

@article{kaiser2017one,
  title={One model to learn them all},
  author={Kaiser, Lukasz and Gomez, Aidan N and Shazeer, Noam and Vaswani, Ashish and Parmar, Niki and Jones, Llion and Uszkoreit, Jakob},
  journal={arXiv preprint arXiv:1706.05137},
  year={2017}
}

@article{kaiser2017depthwise,
  title={Depthwise separable convolutions for neural machine translation},
  author={Kaiser, Lukasz and Gomez, Aidan N and Chollet, Francois},
  journal={arXiv preprint arXiv:1706.03059},
  year={2017}
}

\appendix

\section{Paper Table of Contents}
\label{sec:paper-toc}

\begingroup
\setlength{\parskip}{0pt}
\setlength{\parindent}{0pt}
\setstretch{1.3} \makeatletter
\@starttoc{toc}
\makeatother
\endgroup

\section{Background on Submodularity}
\label{sec:background-submodularity}

In this section, we provide a brief introduction to submodularity and its
properties and its applications to machine learning relevant to this paper. A much more comprehensive
introduction to submodularity for machine learning and artificial intelligence can be found
in~\cite{bilmes2022submodularity,tohidi2020submodularity}, and a
comprehensive discussion of submodular theory can be found
in~\citet{fujishige2005submodular}.

We start with a finite set $V = \{1, 2, \dots, n\}$ and a set function $f: 2^V
\to \R$ that maps subsets of $V$ to real numbers. We say that $f$ is submodular
if for all subsets $X \subseteq Y \subseteq V$ and for all elements $s \in V
\setminus Y$, we have that $f(X \cup \{s\}) - f(X) \geq f(Y \cup \{s\}) - f(Y)$.
This means that the marginal gain of adding an element $s$ to a smaller set $X$
is at least as large as the marginal gain of adding the same element $s$ to a
larger set $Y$. Intuitively, this captures the idea of diminishing returns: as
we add more elements to a set in going from $X$ to $Y$, the additional benefit
of adding another element decreases. There are many phenomena in nature that
have this property. For example: the value of a set of items to a consumer may
exhibit diminishing returns as the consumer acquires more items; the value of a
set of features to a machine learning model may exhibit diminishing returns as
the model is trained on more features; the value of a set of training data
points to a model may exhibit diminishing returns as the model is trained on
more data points; the value of a set of tokens in a large language model's
context may exhibit diminishing returns as the model is given more tokens in its
context~\cite{kumari2024end,kumari2024bumblebee}; the value of a set of friends
to an individual may exhibit diminishing returns as the individual acquires more
friends; the value of a set of sensors to a monitoring system may exhibit
diminishing returns as the system is equipped with more sensors; the value of a
set of locations of facilities may exhibit diminishing returns as the facilities
are located in more locations; and so on and so forth. Indeed, the list of examples where
submodularity naturally arises is vast and practically endless.

Submodular functions have many nice properties and are widely used in various
fields such as combinatorial optimization, machine
learning~\citep{fujishige2005submodular,bilmes2022submodularity,tohidi2020submodularity},
and economics~\citep{topkis1998supermodularity}. For example, they can be
efficiently maximized using greedy algorithms~\citep{nemhauser1978analysis} with
provable approximation guarantees, and they can also be minimized using
polynomial-time algorithms~\citep{fujishige2005submodular} despite the fact that
any submodular function has $O(2^n)$ parameters and thus is intractable to
represent explicitly. Submodularity is a powerful concept that allows us to
model and solve a wide range of problems involving set functions, and is widely
applicable to problems in machine
learning~\cite{bilmes2022submodularity,tohidi2020submodularity}, including for
problems of 
data summarization~\cite{lin2009-submod-active-seq,lin2009-submodsum,lin2011-class-submod-sum}, 
coreset selection~\cite{agarwal2005geometric,mirzasoleiman2020craig}, 
and data filtering 
(three topics that we consider to be synonymous~\cite{bilmes2022submodularity})
as well as data appraisal. For example, it has been noted that submodular functions are good
models of information. In fact, the Shannon entropy function as well as the Von
Neumann entropy function (as we demonstrate
in~\Cref{sec:von-neumann-entropy}) are examples of submodular
functions. This means that if we want to select a subset of data points that has
high information content, we might perform the optimization $\max_{X \subseteq V
: |X| \leq k} f(X)$ where $f$ is a submodular function that models the
information content of a set of data points. If we already have a set of points
$X$ and are considering the value of a new set $Y$, the conditional valuation
$f(Y|X) = f(X \cup Y) - f(X)$ can do this as it provides a conditional data
valuation of $Y$ given that we already have $X$. In fact, the function $f: 2^{V
\setminus X} \to \R$ defined by $f(Y) = f(Y|X)$ is also still submodular 
(this is further discussed in~\Cref{sec:limitations}).

The greedy algorithm for cardinality constrained submodular maximization (which
approximates $\max_{X \subseteq V : |X| \leq k} f(X)$), in particular, is shown
in Algorithm~\ref{alg:greedy-max}.
\begin{algorithm}
\caption{The greedy algorithm for submodular maximization}
\label{alg:greedy-max}
\begin{algorithmic}[1]
\Require A submodular function $f: 2^V \to \R$, a cardinality constraint $k$
\Ensure A subset $X \subseteq V$ of size at most $k$ that approximately maximizes $f(X)$
\State Initialize $X \gets \emptyset$
\For{$i = 1$ to $k$}
    \State $s^* \gets \arg\max_{s \in V \setminus X} f(X \cup \{s\}) - f(X)$ \Comment{Find the element with the largest marginal gain} \label{line:greedy-marginal-gain}  
    \State $X \gets X \cup \{s^*\}$ \Comment{Add the element to the set}
\EndFor
\State \Return $X$
\end{algorithmic}
\end{algorithm}
This algorithm iteratively builds a set $X$ by adding the element that provides
the largest marginal gain at each step, until it reaches the cardinality
constraint $k$. For a monotone non-decreasing submodular function, this greedy
algorithm guarantees~\cite{nemhauser1978analysis} a solution $\tilde X \subseteq V$
such that $f(\tilde X) \geq (1 - 1/e) \cdot \text{OPT}$ where
$\text{OPT} = \max_{X' \subseteq V : |X'| \leq k} f(X')$ is the optimal value of
the submodular function with the given cardinality constraint. 

As can be seen, in line~\ref{line:greedy-marginal-gain} of the algorithm, it
repeatedly needs to evaluate $f(X \cup \{s\})$ for fixed $X$ and different $s
\in V \setminus X$. For some submodular functions, this is quite easy. For
example, with $f(A) = \sqrt{|A|}$, we can compute $f(X \cup \{s\})$ in constant
time by just computing $\sqrt{|X| + 1}$. For other submodular functions, such as
the \emph{facility location function} defined as:
\begin{align}
\label{eq:facility-location-function}
f(A) = \sum_{j \in V} \max_{i \in A} s_{ij},
\end{align}
where $s_{ij} \geq 0$ is a similarity score between elements $i$ and $j$, we can
compute $f(X \cup \{s\})$ in $O(n)$ time by just computing the maximum
similarity of each element to the new element $s$ and comparing it to the
existing maximum similarity to the set $X$, so this can also be made to run
reasonably quickly. Note that the facility location function is submodular, and
is normalized ($f(\emptyset) = 0$) and monotone non-decreasing ($f(A) \leq f(B)$
for $A \subseteq B$) and thus the greedy algorithm above has the $1-1/e$
approximation guarantee. The facility location often works well, but its
downside is that it requires an $n \times n$ similarity matrix, which can be
prohibitively expensive to compute and store for large datasets. In our
ImageNet-1K experiments, we have $n=1.28$ million, so the similarity matrix, if
it was dense, would have had ${1.28}^2$ trillion entries. Therefore, we utilized
a sparse similarity matrix by keeping only the top approximately 2000
similarities for each element, which is still large but is manageable. One of
the motivations of our matrix spectral functions is to find a submodular
function that performs as well as the facility location function without the
$O(n^2)$ memory cost. Note that a matrix spectral function as describes requires
only $\data \in \R^{n \times m}$ and typically $m \ll n$ which is much more
manageable.

However, for submodular functions such as matrix spectral functions, evaluating
$f(X \cup \{s\})$ itself can be much more expensive, since it requires computing
the eigenvalues of a matrix that depends on the set $X \cup \{s\}$. That is, for
matrix spectral functions, $f(X \cup \{s\})$ requires a computation of $O(km^2 +
m^3) = O(nm^2 + m^3)$ where $k=|X|+1 \in O(n)$ to construct the $m \times m$
matrix $B_{X \cup \{s\}}$ and then compute its eigenvalues. When one has access
only to oracle calls, this cost is repeated for each $s \in V \setminus X$ in
greedy even though all but one element of the set $A \cup \{ s\}$ is the same
over different $s$.

In Section~\ref{sec:fast-matrix-spectral-functions}
(and with full details in Appendix~\ref{sec:fast-rank-1-updates-and-downdates}),
we show how to avoid this waste and that
this evaluation can be reduced to $O(m^2)$ by using the secular equation to compute 
the eigenvalues of $B_{X \cup \{s\}}$ from the pre-factored eigenvalues and eigenvectors of $B_X$ and the vector corresponding 
to the new element $s$. Thus, in the case of matrix spectral functions,
we have a speedup of $O(m)$ for each evaluation of $f(X \cup \{s\})$ and thus a speedup of $O(m)$ for 
the entire greedy algorithm, since we need to evaluate $f(X \cup \{s\})$ for $O(n)$ different elements $s$ at 
each of the $k$ iterations. When we use a non-linear kernel, the
cost of constructing $B_{X \cup \{s\}}$ is $O(mk^2 )$, and the cost of computing its eigenvalues is $O(k^3) = O(n^3)$, 
which can be very expensive when $k$ is large, so we rely on the use of the linear kernel
to make this efficient.

Interestingly, the set function $f$ does not need to be perfectly submodular in
order for the greedy algorithm above to offer an approximation guarantee. There
have been a number of forms of "approximate" or "weak" submodularity that have
been proposed in the literature, and for which the greedy algorithm still has an
approximation guarantee. One of the more celebrated approaches is called
the \emph{submodularity ratio} of a set function $f: 2^V \to \R$, which is defined as follows:
\begin{restatable}[Submodularity Ratio~\citep{das2011submodular}]{rdefn}{submodularity-ratio}
\label{defn:submodularity-ratio}
The submodularity ratio of a set function $f: 2^V \to \R$ is defined as
the largest scalar $\gamma \in [0,1]$ such that for all
disjoint subsets $S, T \subseteq V$, we have that:
\begin{align}
\frac{ \sum_{ i \in T} f( i | S ) }{ f(T | S) } \geq \gamma
\end{align}
where $f(i | S) = f(S \cup \{i\}) - f(S)$ is the marginal gain of 
adding element $i$ to set $S$, and $f(T | S) = f(S \cup T) - f(S)$ is the marginal 
gain of adding the set $T$ to set $S$.
\end{restatable}
This means that the total marginal gain contributions of adding the elements of
$T$ one at a time is at least $\gamma$ times the marginal gain of adding the
entire set $T$ at once. When $\gamma = 1$, this is equivalent to the standard
definition of submodularity, but when $\gamma < 1$ this allows for a relaxation
away from submodularity to a certain degree. Indeed, as has been shown
in~\citep{das2011submodular,das2018approximate} a given a set function with a
submodular ratio of $\gamma$, the greedy algorithm for submodular maximization
offers an approximation guarantee of $f(X_\text{greedy}) \geq (1 - e^{-\gamma})
\cdot \text{OPT}$ where $X_\text{greedy}$ is the set returned by the greedy
algorithm and $\text{OPT}$ is the optimal value of the maximization problem.

Another notion of approximate submodularity
is called $\weakly$-weakly DR submodularity~\citep{el2020optimal} which is defined
as
\weaklyDRsubmodularity*
This means that the marginal gain of adding an element $s$ to a smaller set $X$
is at least $\weakly$ times the marginal gain of adding the same element $s$ to
a larger set $Y \supset X$. This notion was first
defined in~\cite{lehmann2001combinatorial} and its name
is adopted via a similar notion used
in continuous forms of submodularity~\cite{hassani2017gradient},
and related ideas are discussed in~\cite{pmlr-v84-el-halabi18a,gatmiry2018non,bian2017guarantees}.
When $\weakly = 1$, this reduces to the standard
definition of submodularity, but when $\weakly < 1$, this allows for some
violation of the diminishing returns property, but still provides some structure
that can be exploited by algorithms.

If a function is $\weakly$-weakly DR submodular, then the function
has a submodularity ratio of $\gamma \geq \weakly$ and thus the greedy algorithm for submodular maximization
offers an approximation guarantee of $f(X_\text{greedy}) \geq (1 - e^{-\weakly}) \cdot \text{OPT}$ where $X_\text{greedy}$ 
is the set returned by the greedy algorithm and $\text{OPT}$ is the optimal value of the maximization problem.
This is formalized in the following.
\begin{restatable}[$\weakly$-weakly DR submodularity implies $\gamma$-submodularity~\citep{el2018combinatorial,bogunovic2018robust}]{rthm}{weaklyDRimpliesSubmodularity}
\label{thm:weakly-DR-implies-submodularity}
If a set function $f: 2^V \to \R$ is $\weakly$-weakly DR submodular for some
$\weakly \in [0,1]$, then $f$ has a submodularity ratio of $\gamma \geq \weakly$.
\end{restatable}
\begin{proof}
\label{proof:weakly-DR-implies-submodularity}
Assume that $f$ is $\weakly$-weakly DR submodular for some $\weakly \in [0,1]$.
Let $S, T \subseteq V$ be any disjoint subsets.
We want to show that $\sum_{i \in T}
f(i | S) \geq \weakly f(T | S)$. We can write $T = \{t_1, t_2, \dots, t_k\}$ under any
order, and define $T_j = \{t_1, t_2, \dots, t_j\}$ for $j = 0, 1, \dots, k$ where $T_0 =
\emptyset$. Then we have:
\begin{align}
\sum_{i \in T} f(i | S) &= \sum_{j=1}^k f(t_j | S) 
\geq \sum_{j=1}^k \weakly f(t_j | S \cup T_{j-1}) 
= \weakly \sum_{j=1}^k f(t_j | S \cup T_{j-1}) =  \weakly f(T | S)
\end{align}
where the first inequality follows from the definition of
$\weakly$-weakly DR submodularity,
and the final equality follows via telescoping summation and
the chain rule of submodularity.
Thus, $\sum_{i \in T} f(i | S) \geq \weakly f(T | S)$ for any
disjoint $S,T$, which means that the
submodularity ratio $\gamma$ of $f$ is at least $\weakly$.
\end{proof}

This justifies our use of the $\weakly$-weakly DR submodularity property 
to motivate our definition of $\weakly$ matrix monotone functions for $-\phi'$ 
as done in Section~\ref{sec:weakly-matrix-monotone}.

Submodular functions (and their approximate brethren) have other attractions,
including that the cardinality constraint is not the only constraint for which
there are approximation algorithms. There are also many approximation algorithms
for other constraints, even under the greedy algorithm, such as matroid
constraints. A matroid is a general combinatorial structure, at a very high
level, consists of a set of sets $\mathcal C \subseteq 2^V$ that satisfy certain
properties. There are many types of matroids, and matroids are quite flexible,
but one particularly useful matroid that is useful when one wishes to obtain
balanced subsets is a partition matroid constraint. A partition matroid defines
$\mathcal C$ based on an underlying partition of the ground set $V$ int a set of
$L$ disjoint subsets $V = \{ V_1, V_2, \dots, V_L\}$ and a set of $L$
non-negative integer-valued limits $k_1, k_2, \dots, k_L$ such that $\mathcal C = \{ A
\subseteq V : |A \cap V_i| \leq k_i \text{ for all } i = 1, 2, \dots, L\}$. This
means that the feasible sets of the partition matroid are those subsets of
$V$ that contain at most $k_i$ elements from each block $V_i$ of the partition. The 
simple greedy algorithm for submodular maximization under a partition matroid constraint has
an approximation guarantee of $f(X_\text{greedy}) \geq 1/2 \cdot
\text{OPT}$. We use matroid-constrained optimization to obtain
our balanced sets discussed in Section~\ref{sec:experiments}. 

We note that such a matroid constraint can be even used heuristically to produced balanced
low-scoring subsets by using a heuristic submodular minimization procedure,
which is the same as the above greedy algorithm but where we take the smallest
rather than the largest element at each step. While this heuristic greedy
submodular minimization procedure has no mathematical guarantees, we find when
we perform this procedure, using our functions, and compare it to a much slower
minimum-norm-point algorithm of~\cite{fujishige2005submodular,nagano2011size},
the heuristic greedy procedure runs much faster, but still produces subsets
whose value is practically the same as the more expensive algorithm with some
mathematical guarantees. Hence, all of our ``poor'' subsets
(\textcolor{FigBlue}{blue points})
in~\Cref{fig:main-teaser,fig:class-balanced-complete-pyramid,fig:unconstrained-complete-pyramid},
and other such plots that require low-valued subsets use this procedure.

All submodular optimization procedures in this paper were performed
using~\cite{bilmes2026-submarine}.

\section{Background on Data Appraisal and Coreset Selection}
\label{sec:background-data-appraisal}

What makes a good and what makes a poor machine learning training
dataset? Classically, a good dataset will be informative,
relevant, non-redundant (i.e., diverse), clean (free of mistakes),
complete (high coverage), noise free (mostly), and
high-resolution. Overall, a good dataset should efficiently and
properly represent a problem that is needing to be solved.  A poor
dataset would do the opposite.
We visualize good and poor data subsets for different
classes of ImageNet-1K in~\Cref{app:viz_good_bad_sets}.

Data appraisal is the process of evaluating the value, quality, and
relevance of data, often to determine its utility for machine
learning, sometimes its monetary worth, and sometimes to determine
whether it should be retained for long-term storage.  Data appraisal
is also strongly related to coresets, a process where we wish to find
a small highly valuable data subset.  These are important problems, as
good data is one of the most important ingredients for achieving
high-quality artificial intelligence.  Data appraisal has become
important not just for AI, but also for healthcare and many other
non-machine-learning fields as well.  As a result, over the years,
besides the costly brute force approach of training on a dataset and
then testing it on a held-out dataset, there have been many approaches
to produce a computationally fast surrogate for the problem of data
appraisal.

There are at least four distinct axes regarding assessing, appraising, scoring,
or choosing a subset (i.e., a coreset, summary, or purchasable unit) of data.
The present paper focuses exclusively on a sub-range (marked by red
stars\textcolor{red}{$^\star$}) of the four axes mentioned below.

First, the underlying score or objective
may be {\bf additive} or 
{\bf non-additive}\textcolor{red}{$^\star$}. 
An additive measure is
one where each sample has its own score that does not change depending
on what other samples are being simultaneously considered. Examples
include~\citep{paul2021datadiet,toneva2019forgetting,dharmasiri2025impact}. 
A non-additive measure scores a subset jointly, or makes a sample's
marginal worth depend on what has already been selected. Examples
include geometric coreset coverage, submodular or gradient-matching
objectives 
and bilevel subset-selection
methods~
\citep{sener2018active,mirzasoleiman2020craig,killamsetty2021gradmatch,
  killamsetty2021glister,ghorbani2019datashapley}. 
Data Shapley~\citep{ghorbani2019datashapley} is actually
additive: it starts from a potentially
non-additive coalition measure, but each sample's value
becomes its average marginal contribution across all coalitions.
Also, we consider
stratified or quota-based approaches as non-additive, because one can
define a non-additive set score whose marginal gain becomes zero once
a quota within a stratum has been hit. The stratified and
group-balanced policies in~\citet{dharmasiri2025impact} are examples.
Second, the score may be fixed, learned in a two-stage pipeline\textcolor{red}{$^\star$}, or
truly bilevel. Fixed objectives include random sampling, sensitivity
sampling, and classical coreset constructions~\citep{bachem2017practical,sener2018active}. Two-stage
methods first train or use a model to produce scores and then freeze
those scores before selecting data, as in~\citep{paul2021datadiet,toneva2019forgetting,swayamdipta2020dataset}.
True bilevel methods are the most computationally expensive and choose
the subset while explicitly accounting for the model that would be
trained on that subset.
Examples include~\citep{killamsetty2021glister,yoon2020dvrl}.
A third axis is the data-access model. That is, the data may be
locally available and openly inspectable\textcolor{red}{$^\star$}, as in standard coreset
selection and data valuation, or it may be privately held by different
parties (e.g., buyer and seller), requiring privacy-preserving data
appraisal, secure multiparty computation, 
zero-knowledge protocols, and so on.
Examples include~\citep{ghorbani2019datashapley,kwon2022betashapley,tian2022private,
  hynes2018sterling,azcoitia2022try,wong2025privade}.
Fourth, the selection process may be deterministic\textcolor{red}{$^\star$} or randomized\textcolor{red}{$^\star$}. Top-$k$ score
ranking, greedy selection, and many bilevel or
gradient-matching procedures are often deterministic once the model
state and tie-breaking strategies has been fixed. 
Sensitivity sampling, random
baselines, stratified random selection, and
probability-proportional-to-score sampling are
randomized~\citep{bachem2017practical,sener2018active,dharmasiri2025impact}.
This axis is orthogonal to additivity as a randomized method can still be
additive if it samples each point in proportion to an individual
score, while a deterministic quota, 
or determinantal 
point process, is typically non-additive.

\section{Proofs that Neural Scaling Laws are Submodular}
\label{sec:proofs-neural-scaling-laws}

Here we show that certain neural scaling laws, when viewed as set functions, are
submodular. In the below, we use the standard
result~\cite{fujishige2005submodular,bilmes2017deep,bilmes2022submodularity}
that, given that function $\phi: \R_+ \to \R$ is concave,
any set function of the form $f(A) = \phi(|A|)$,
a concave function of the cardinality of the set $A$, is
submodular. Also, if we take $\phi(x) = 1 - \psi(x)$, then if $\psi(x)$ is
convex, then $\phi(x)$ is concave, and thus $f(A) = \phi(|A|)$ is submodular if
$\psi$ is convex.

We again emphasize that, normally, scaling laws measure loss, or cost. When we
say that scaling laws ``value'' data with a submodular function, we implicitly
convert a scaling law ``loss/cost'' expression to a scaling law ``value'' expression
by taking a constant minus loss. Were we not to take this constant minus
loss, the same proofs below would hold but would show that the original loss-based
scaling laws are supermodular rather than submodular functions. A function $g: 2^V \to \R$
is said to be supermodular whenever $-g$ is submodular.

\chinchillaissubmodular*This is originally stated in Section~\ref{sec:neural-scaling-laws}, page~\pageref{lem:chinchilla-submodular}.\vspace{-1.0em}
\begin{proof}[Proof of Lemma~\ref{lem:chinchilla-submodular}]
\label{proof:chinchilla-submodular}
$\chinacr(\data) = c' - b |\data|^{-\beta}$ takes the form of a constant plus a function
of the form $\phi(x) = - bx^{-\beta}$. Taking the
second derivative with respect to $x$
yields $\phi''(x) = - b \beta (\beta + 1) x^{-\beta - 2}$, which is negative for all $x > 0$ 
whenever $\beta > 0$ and $b \geq 0$. 
Also, $\lim_{x \to 0^+} \phi''(x) = -\infty$.
Hence, $\phi(x)$, for $x \geq 0$, is a concave function,
$\chinacr(\data)$ applies that to
the cardinality of $\data$, and thus $\chinacr(\data)$  is submodular.
\end{proof}

\clusterscalingissubmodular*Originally stated in Section~\ref{sec:neural-scaling-laws}, page~\pageref{cor:cluster-scaling-submodular}.\vspace{-1.0em}
\begin{proof}[Proof of Corollary~\ref{cor:cluster-scaling-submodular}]
\label{proof:cluster-scaling-submodular}
We have that the function is $\chinacr(\data) = c + \sum_{k=1}^K g_i(\data \cap \data_i)$ where
$g_i(\setfmt{A}) = -(c_i + |\setfmt{A}|)^{-\beta_i} = \phi(|\setfmt{A}|)$. This is a set function 
where $\phi(x) = - (c_i + x)^{-\beta_i}$ is a concave function of $x$ for $x \geq 0$ whenever $\beta_i > 0$ and $c_i \geq 0$. 
Hence, $\phi(|\setfmt{A}|)$
is a concave function composed with the cardinality of $\setfmt{A}$, and $\chinacr(\data)$ is a
sum of a constant and non-negative weighted sum of submodular functions, which is submodular.
\end{proof}

We next show that epoch-based scaling laws are submodular. The proof of
this next theorem is significantly more involved than the proofs of the above
first two theorems, but the basic approach is the same, to identify
a monotone non-decreasing concave function of the cardinality of a set to which
this scaling law corresponds to.
\epochscalingissubmodular*Originally stated in Section~\ref{sec:neural-scaling-laws}, page~\pageref{thm:epoch-scaling-submodular}.\vspace{-1.0em}
\begin{proof}[Proof of Theorem~\ref{thm:epoch-scaling-submodular}]
\label{proof:epoch-scaling-submodular}

We first restate this epoch-based scaling law~\cite{goyal2024scaling} and set up some notation for the proof.
We have that
\begin{align}
\loss_\computeamount (\datasize) 
= c + b \datasize_1^{-\beta_1}\prod_{j=2}^k \left(\frac{\datasize_j}{\datasize_{j-1}}\right)^{-\beta_j}
\end{align}
where $\datasize_j = j \datasize $ is number of samples seen at the end of
$j^\text{th}$ epoch, $b > 0$, $c > 0$, $\datasize \geq 0$ is the base amount of
training data (i.e., $\datasize = |\data|$), $\computeamount$ is the total compute budget,
and where $\beta_{j} = \beta
\left(\frac{1}{2}\right)^{(j-1)/\tau}$ for $j \geq 1$ where $\beta > 0$ is an
initial scaling exponent and $\tau > 0$ is a half-life parameter. 
Thus,
\begin{align}
-\beta = -\beta_1 < -\beta_2 < -\beta_3 < \dots < -\beta_j < \dots < 0
\end{align}
is a strictly increasing sequence of negative numbers that converges to 0 as $j \to \infty$. 
We point out that $d_j$ is not the number of \emph{unique} samples seen at the end of $j^\text{th}$
epoch, but rather the total number of samples that have been trained
on which includes the case that each sample has been trained on, at this
point, $j$ times. Using a ``types'' vs ``tokens'' analogy~\citep{peirce1906prolegomena}, $d_j$ is the number of 
sample ``tokens'' seen at the end of the $j^\text{th}$ epoch, while $\datasize$ is the number of sample ``types'' 
in the dataset. The point of this scaling law, in fact, is to represent
the case that a sample becomes less valuable the more times it is trained on, which 
is why the scaling exponent $\beta_j$ decays as $j$ increases (i.e., the
loss reduction rate decreases when using samples that have been trained on more times).
Also implicit in the above is that $\computeamount = d_k = k \datasize$ is the total compute amount, 
which is the number of samples seen (i.e., trained on) at the end of the final ($k^\text{th}$) epoch.

Since $c$ is only a constant shift, and does not affect submodularity, we
define and work with
$\hat \loss_\computeamount (\datasize) = \loss_\computeamount (\datasize) - c$.

We thus have
\begin{align}
\hat \loss_\computeamount (\datasize) 
&= b \datasize_1^{-\beta_1}\prod_{j=2}^k \left(\frac{\datasize_j}{\datasize_{j-1}}\right)^{-\beta_j} \\
&= b \datasize^{-\beta} \prod_{j=2}^k \left(\frac{j \datasize}{ (j-1) \datasize}\right)^{-\beta_j} \\
&= b \datasize^{-\beta} \prod_{j=2}^k \left(\frac{j }{j - 1}\right)^{-\beta_j} \label{eq:epoch-scaling-loss-ratio-of-j}
\end{align}
Now if $k$ was constant, we would have that 
$\hat \loss_\computeamount (\datasize)  \propto b d^{-\beta}$ which,
since $\beta > 0$
and 
$\hat \loss_\computeamount'' (\datasize) \propto b \beta (\beta + 1) \datasize^{-\beta - 2} > 0$,
can be seen as a convex function of $d$ and
so $f(\data) =  - \hat \loss_\computeamount ( | \data |)$ would immediately
be submodular. However, there is a nuance here which is that
$k$ is not constant, rather it is a function of $d$ since
$k = \computeamount / \datasize$. Moreover,
to show that $\hat \loss_\computeamount (\datasize)$
is convex in $\datasize$ we need to relax this
function to work for any real value $\datasize \geq 0$ and
allow for $\computeamount / \datasize$ being a non-integer. We thus
define $\bar k = \lfloor \computeamount / \datasize \rfloor$ 
to correspond to the number of complete epochs, and also where
the final epoch may be incomplete and consists
of $\computeamount - \bar k \datasize$ samples.

When considering Equation~\eqref{eq:epoch-scaling-loss-ratio-of-j},
we note that scaling exponent $\beta_j$ applies to the ratio of
data seen at the end of epoch $j$ to data seen at the
end of epoch $j-1$, which is just $d_j/d_{j-1} = j/(j-1)$. Thus, the final partial epoch $\bar k + 1$ has ratio
$\computeamount / (\bar k \datasize)$, and thus we can write the
epoch scaling law as follows:
\begin{align}
\hat \loss_\computeamount (\datasize) 
&= b \datasize^{-\beta} \prod_{j=2}^{\bar k} \left(\frac{j }{j - 1}\right)^{-\beta_j}
\cdot \left(\frac{\computeamount }{\bar k \datasize}\right)^{-\beta_{\bar k + 1}} \\
&= 
b \datasize^{-\beta + \beta_{\bar k + 1}} \prod_{j=2}^{\bar k} \left(\frac{j }{j - 1}\right)^{-\beta_j}
\cdot \left(\frac{\computeamount }{\bar k }\right)^{-\beta_{\bar k + 1}}
\end{align}
Note that if $\bar k = \computeamount / \datasize$, the final factor 
$\left(\frac{\computeamount }{\bar k \datasize}\right)^{-\beta_{\bar k + 1}}$
vanishes.

\begin{figure}[tbh]
\centerline{\includegraphics[width=0.8\textwidth]{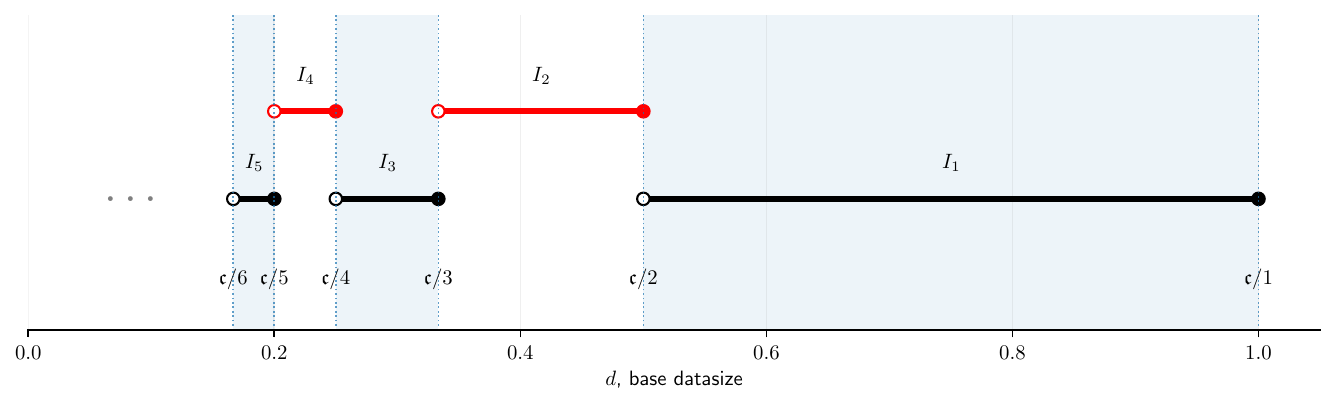}}
\caption{A visualization of the half-open intervals $I_j = \left(
\frac{\computeamount}{j+1}, \frac{\computeamount}{j} \right]$ for $j = 1, 2,
\dots$ normalized by setting $\computeamount = 1$.}
\label{fig:half-open-intervals}
\end{figure}

In order to handle the above, we consider the following series of half-open intervals into
which $\datasize$ may fall and which covers the range $0 < \datasize \leq \computeamount$:
\begin{align}
\dots,
\left( \frac{\computeamount}{j+2}, \frac{\computeamount}{j+1} \right],
\left( \frac{\computeamount}{j+1}, \frac{\computeamount}{j} \right],
\left( \frac{\computeamount}{j}, \frac{\computeamount}{j-1} \right], 
\dots,
\left( \frac{\computeamount}{3}, \frac{\computeamount}{2} \right],
\left( \frac{\computeamount}{2}, \computeamount \right]
\end{align}
We can identify $I_j = \left( \frac{\computeamount}{j+1},
\frac{\computeamount}{j} \right]$, and note that as $j$ increases, $\text{len}(I_j)
= \frac{\computeamount}{j} - \frac{\computeamount}{j+1} =
\frac{\computeamount}{j(j+1)}$ decreases, and $\lim_{j \to \infty}
\text{len}(I_j) = 0$. These intervals are
shown in~\Cref{fig:half-open-intervals}.
Also, for any $\datasize \in I_j$, we have $\bar k = j$ and thus
in any one such interval, we can view $\bar k$ as a constant (i.e., in each interval, there are a fixed constant number of epochs).
If $\datasize \in I_j$ then training proceeds for
$j = \bar k$ complete epochs and then a possible final partial epoch
if $\datasize < \computeamount/{\bar k}$. If $\datasize = \computeamount$ then training
consists of one and only one complete epoch, $\bar k = 1$
and $\hat \loss_\computeamount (\computeamount) = 
b \computeamount^{-\beta}$. Moreover, if $\datasize > \computeamount$ then 
training consists only of a partial epoch, since the compute budget
limits us from using all of the training data. Since the loss
is only a function of datasize, we thus
have that $\hat \loss_\computeamount (\datasize) = \hat \loss_\computeamount (\min( \datasize, \computeamount))$.
Lastly, as $\datasize \to 0^+$, we have $\hat \loss_\computeamount (\datasize) \to +\infty$ since 
the first factor $\datasize^{-\beta}$ dominates and goes to $+\infty$.
Putting the pieces together, we get the following expression
for $\hat \loss_\computeamount : \R_+ \to \R \cup \{ \infty \}$:
\begin{align}
\hat \loss_\computeamount (\datasize) &=
\begin{cases}
+\infty & \text{if } \datasize = 0 \\
b \datasize^{-\beta + \beta_{\bar k + 1}} \prod_{j=2}^{\bar k} \left(\frac{j }{j - 1}\right)^{-\beta_j}
\cdot \left(\frac{\computeamount }{\bar k }\right)^{-\beta_{\bar k + 1}}
& \text{if } 0 < \datasize < \computeamount, \bar k = \lfloor \computeamount / d \rfloor \\
b \computeamount^{-\beta} & \text{if } \datasize \geq \computeamount
\end{cases} \\
&=
\begin{cases}
+\infty & \text{if } \datasize = 0 \\
b \datasize^{-\beta + \beta_{\bar k + 1}} h(\bar k)
& \text{if } 0 < \datasize < \computeamount, \bar k = \lfloor \computeamount / d \rfloor \\
b \computeamount^{-\beta} & \text{if } \datasize \geq \computeamount
\end{cases}
\end{align}
where $h(\bar k)$ is a constant for $\datasize \in I_k$.

We immediately have that $\hat \loss_\computeamount (\datasize)$  is
continuous within each interval $I_j$ for $j \geq 1$. We next show three properties for
$\hat \loss_\computeamount (\datasize)$ as a function of $\datasize \geq 0$:
(1) continuity at the interval boundaries, (2) convexity for $d \geq 0$, and (3) 
monotone non-increasingness for $d \geq 0$.

For {\bf continuity}, we need to consider the boundary of every interval $I_j$ for $j \geq 1$.
First, considering $d \in I_1$,
we have that $\bar k = 1$,
$\hat \loss_\computeamount (\datasize) = b d^{-\beta + \beta_2} \computeamount^{-\beta_2}$,
and $\lim_{\datasize \uparrow \computeamount} \hat \loss_\computeamount (\datasize)
= \hat \loss_\computeamount (\computeamount) = b \computeamount^{-\beta}$. Next,
consider any interval $I_j$ for $j \geq 2$ and the point
$\datasize_{(j)} = \computeamount / j \in I_j$ at the boundary of the right side
of $I_j$ and the left side of $I_{j-1}$. On the right side of $I_j$, we have
\begin{align}
\hat \loss_\computeamount (\datasize_{(j)})
&= b \datasize_{(j)}^{-\beta + \beta_{j + 1}} \prod_{i=2}^{j} \left(\frac{i }{i - 1}\right)^{-\beta_i} 
\cdot \left(\frac{\computeamount }{j }\right)^{-\beta_{j + 1}} \\
&= b \left( \frac{ \computeamount }{j} \right)^{-\beta}
   \prod_{i=2}^{j} \left(\frac{i }{i - 1}\right)^{-\beta_i}
\end{align}
On the left side of $I_{j-1}$ we have
\begin{align}
\lim_{\datasize \downarrow \datasize_{(j)}} \hat \loss_\computeamount (\datasize)
&= \lim_{\datasize \downarrow \datasize_{(j)}} b \datasize^{-\beta + \beta_{j}} \prod_{i=2}^{j-1} \left(\frac{i }{i - 1}\right)^{-\beta_i}
\cdot \left(\frac{\computeamount }{j - 1 }\right)^{-\beta_{j}} \\
&= b {\left( \frac{\computeamount}{j} \right)}^{-\beta}
\prod_{i=2}^{j-1} \left(\frac{i }{i - 1}\right)^{-\beta_i}
\cdot 
{\left( \frac{\computeamount}{j} \right)}^{\beta_j}
\left(\frac{\computeamount }{j - 1 }\right)^{-\beta_{j}}  \\
&= b \left( \frac{ \computeamount }{j} \right)^{-\beta}
   \prod_{i=2}^{j} \left(\frac{i }{i - 1}\right)^{-\beta_i}
\end{align}
Thus, we have continuity at this boundary.
This establishes continuity for $d \geq 0$.

Next, we consider {\bf convexity}. Within every interval $d \in I_j$ we have
$\hat \loss_\computeamount (\datasize) = b \datasize^{-\beta + \beta_{\bar k +
1}} h(\bar k)$ where $h(\bar k) > 0$ is a positive constant for this range of
$d$. Since $\beta > \beta_{\bar k + 1}$, $-\beta + \beta_{\bar k + 1} < 0$. Set
$\zeta_{\bar k} = -\beta + \beta_{\bar k + 1} < 0$, and thus we have $\hat
\loss_\computeamount '' (\datasize) = b \zeta_{\bar k} (\zeta_{\bar k}-1) \datasize^{ \zeta_{\bar k}
- 2} h(\bar k)$. Since $b > 0$, $\zeta_{\bar k} (\zeta_{\bar k}-1) > 0$, $\datasize^{ \zeta_{\bar k}
- 2} > 0$ and $h(\bar k) > 0$ we have that $\hat \loss_\computeamount ''
(\datasize) > 0$ establishing convexity for $d \in I_j$, that is
within each interval. We must still identify
what happens to the slopes at the interval boundaries, to establish
convexity for $\datasize \geq 0$.
We again consider the
point $\datasize_{(j)} = \computeamount / j$ at the boundary of the right side of $I_j$ and
the left side of $I_{j-1}$.
For $\datasize \in I_j$, 
$\hat \loss_\computeamount ' (\datasize) = b \zeta_{\bar k} \datasize^{\zeta_{\bar k} - 1} h(\bar k)
= \frac{\zeta_{\bar k}} { \datasize} \hat \loss_\computeamount (\datasize)$
and therefore,
$\hat \loss_\computeamount ' (\datasize_{(j)}) = 
\frac{\zeta_{\bar k}} { \datasize_{(j)}} \hat \loss_\computeamount (\datasize_{(j)})$.
For $\datasize \in I_{j-1}$, 
\begin{align}
\lim_{\datasize \downarrow \datasize_{(j-1)}} \hat \loss_\computeamount ' (\datasize) &= 
\lim_{\datasize \downarrow \datasize_{(j-1)}} \frac{\zeta_{\bar k - 1}} { \datasize} \hat \loss_\computeamount (\datasize)  \\
&= \frac{\zeta_{\bar k - 1}} { \datasize_{(j)}} \hat \loss_\computeamount (\datasize_{(j)}).
\end{align}
Now $\zeta_{\bar k} = -\beta + \beta_{\bar k + 1} 
< -\beta + \beta_{\bar k} = \zeta_{\bar k - 1}$ since $\beta_j$ is strictly 
decreasing in $j$. This means that the derivative of 
$\hat \loss_\computeamount (\datasize)$ makes a
positive increment at each interval boundary. This establishes
that the slope of $\hat \loss_\computeamount (\datasize)$ is
never decreasing for $d \geq 0$, and thus $\hat \loss_\computeamount (\datasize)$ is convex for $d \geq 0$.

Lastly, we consider {\bf monotone non-increasingness},
and note that $d > 0$,
$\hat \loss_\computeamount (\datasize) > 0$ 
and that $\zeta_j = - \beta + \beta_{j + 1} < 0$ for all $j$,
we have that
$\hat \loss_\computeamount'(\datasize) = \frac{\zeta_{\bar k}} { \datasize} \hat \loss_\computeamount (\datasize) < 0$ for all $d > 0$.
This means that $\hat \loss_\computeamount (\datasize)$ is monotone non-increasing for $\datasize \geq 0$.

Putting the pieces together, we have that $\hat \loss_\computeamount
(\datasize)$ is a convex, monotone non-increasing function of $\datasize$ for
$\datasize \geq 0$. Thus, $\chinacr(\data) = c' - 
\loss_\computeamount (|\data|)$, for any constant $c'$, 
is a monotone non-decreasing submodular function
of $\data$.

\end{proof}

The function $\loss_\computeamount(\datasize)$ can be seen as a piecewise
function where each piece is a powerlaw-like function of $\datasize$ with a
different exponent, and where the average exponent $-\beta$ is increasing as $\datasize$
decreases. The reason is that for small $\datasize$,
we perform more epochs (as $\bar k = \lfloor \computeamount / \datasize \rfloor$ gets larger)
and successive epochs model the loss decreasing more slowly with a larger $-\beta$. With
larger $\datasize$, however, we perform fewer epochs and the 
loss decreases faster with a smaller (more negative) $-\beta$.
With large enough $\datasize$, the function eventually bottoms out as $\datasize \geq
\computeamount$. We also note that while $\lim_{\datasize \to 0^+}
\loss_\computeamount(\datasize) = +\infty$, any actual meaningful datasize will
not be close to 0, so we can consider some $\datasize_\text{min} > 0$ as the minimum
meaningful datasize value. This means $\loss_\computeamount(\datasize)$ is
bounded and forming the concave function $\phi(\datasize) = c' -
\loss_\computeamount (\datasize)$, and the corresponding submodular function
$\chinacr(\data) = c' - \loss_\computeamount (|\data|)$ is meaningful and finite
valued and, with appropriate choice of $c'$ and assuming a non-empty
base subset, we can also have
$\chinacr(\emptyset) = 0$. Such a function could be optimized using the greedy
algorithm~\cite{nemhauser1978analysis}. On the other hand, for a scaling
law, it may be sufficient to approximate the piecewise function 
$\loss_\computeamount(\datasize)$ with a single bottom-saturating
convex function of the form $a \exp(- \beta \datasize) + c$ which can
be found to be a reasonable fit. This
helps to motivate our second concave function
$\mfb(x) = \mfnb$, a form of saturating exponential function,
as mentioned at the end of~\Cref{sec:neural-scaling-laws}.

\section{The General Vendi Score based on the Renyi Entropy}
\label{sec:general-vendi-score}

The general Vendi score has recently been proposed as a strategy to measure
the diversity (and hence value) of a given dataset. Most generally,
the Vendi score~\cite{friedmanvenditmlr,pasarkar2024cousins,pasarkar2026vendi,jung2025prismatic,nguyen2025vendi,sirigiri2026diversity,jalali2026conditional}
can be defined via a Renyi entropy formulation as follows:
\begin{align}
\text{Vendi}(X) = \exp\left( \frac{1}{1-\alpha} \log \sum_{i=1}^m \lambda_i(B_X)^\alpha \right)
\end{align}
where $\lambda_i(B_X)$ is the $i^\text{th}$ eigenvalue of $B_X$ and $\alpha > 0$
is a parameter. The case of $\alpha = 1$ corresponds to the Shannon entropy
case, and thus the Vendi score can be expressed as $f(X) = \exp\left(
\trace{-B_X \log(B_X)} \right)$, which is the exponentiated Shannon spectral
entropy of the eigenvalues of $B_X$. The case of $\alpha = 2$ corresponds to the
collision entropy case, and thus the Vendi score can be expressed as $f(X) =
\exp\left( -\log \sum_{i=1}^m \lambda_i(B_X)^2 \right) = 1/\sum_{i=1}^m
\lambda_i(B_X)^2$, which is the inverse of the sum of squares of the eigenvalues
of $B_X$. The case of $\alpha = 0$ corresponds to the log-support case, and thus
the Vendi score can be expressed as $f(X) = \exp\left( \log \sum_{i=1}^m
1_{\lambda_i(B_X) > 0} \right) = \sum_{i=1}^m 1_{\lambda_i(B_X) > 0}$, which
is the number of non-zero eigenvalues of $B_X$.

We can define a set function for any $\alpha \geq 0$ but it is not precisely of
the form $f(X) = \trace{\phi(B_X)}$, i.e., a sum of a function on the individual
eigenvalues for $\alpha \neq 1$. However, we can still define a set function as
above. The present paper established the submodularity of Vendi score with
$\alpha=1$, it remains to be shown if it is submodular, or weakly DR submodular,
for values other than $\alpha=1$. We do point out, however, that the fast secular
equation solution for being able to run the greedy algorithm still holds for any
$\alpha \geq 0$. The reason is that we only need to be able to get the updated
eigenvalues of $B_{X \cup \{s\}}$ from the pre-factored eigenvalues and
eigenvectors of $B_X$ and the vector corresponding to the new element $s$, which is
precisely what the secular derivation does. Hence, this means
that the generalize Vendi score can take advantage of the
fast methods we develop in Section~\ref{sec:fast-matrix-spectral-functions} for any $\alpha \geq 0$,

\subsection{The Vendi Score, Density Matrices, and Monotone Non-Decreasing Properties}
\label{sec:vendi-score-on-density-matrices}

As mentioned elsewhere, the standard Shannon-entropy based log Vendi score is
defined as $f(X) = \trace{-B_X \log(B_X)} = -\sum_{i=1}^m \lambda_i(B_X)
\log(\lambda_i(B_X))$ where $\lambda_i(B_X)$ is the $i^\text{th}$ eigenvalue of
$B_X$. This is an entropic quantity and uses the function (with matrix monotone
negative derivatives) $\phi(x) = - x \log(x)$, which, while concave, is
guaranteed neither to be either monotone non-decreasing nor non-negative (see
the discussion of this function at the beginning
of~\Cref{sec:von-neumann-entropy}. In the case of Shannon entropy, it applies to
probability distributions and so $\phi(x)$ never reaches its negative region. In
the case of the Vendi score, however, $B_X$ could easily have eigenvalues that
are greater than 1, and thus $\phi(x)$ could easily be negative for some of the
eigenvalues.

In practce, the matrix $B$ (for which $B_X$ is a principal submatrix of) is
normalized in a way so that this does not happen. Firstly, if it is the case
that the $n \times m$ design matrix $\data$ is row-normalized (meaning every
row $x_i$ of $\data$ has $\| x_i \| = 1$) then we know that $\trace{ \data
\mtrans{\data}} = \sum_{i=1}^n \| x_i \|^2 = n$. Then, if we set $B = \frac{1}{n} \data
\mtrans{\data}$ (i.e., or setting $\data \gets \frac{1}{\sqrt{n}}\data$),
we get that $\trace{B} = 1$ and thus the eigenvalues of $B$
sum to 1, and since they are all non-negative (due to positive semidefiniteness),
the $B$ matrix acts like a proper density matrix.
This means that the eigenvalues of $B_X$ also sum to at most 1, and
thus $\phi(x) = - x \log(x)$ is non-negative for all eigenvalues of $B_X$.

If, instead, the kernel matrix $K$ is used from a PSD kernel function, then it is
common for all of the diagonal entries to be equal to one. If, once again,
this is normalized by $n$, so that $B = K/n$, again we have non-negative
eigenvalues that sum to 1.

In either of these normalized cases, when viewed as a set function, the log Vendi
score, as a set function, therefore is non-negative since none of the eigenvalues
are ever in the negative region of $\phi(x) = - x \log(x)$ (i.e., where $x \in [0,1]$).

However, since $\phi'(x) = -\log(x) - 1$ may be negative for some $x \in (0,1)$,
in this case the log Vendi score is not guaranteed to be monotone non-decreasing
since for $x > 1/e$ we have $\phi'(x) < 0$. However, if we normalize
the eigenvalues of $B$ by $ e \times \max_i \lambda_i(B)$ (i.e., the resulting
$B$ has as its maximum eigenvalue $1/e$) then the Vendi score will
be a monotone non-decreasing submodular function.

\section{Proofs of Weakly DR Submodular Matrix Spectral Functions}
\label{sec:proofs-weakly-submodular-matrix-spectral-functions}

In this section, we offer proofs of the
results mentioned in Section~\ref{sec:weakly-matrix-monotone}.

First, we note that a result by 
Loewner~\cite{loewner1934monotone,bendat1955monotone,bhatia2009positive}
states that any matrix monotone
function $\psi : \R_+ \to \R$ can be represented as an integral of the form
\begin{align}
\label{eq:integral-representation-matrix-monotone-function}
\psi(t) 
= \alpha + \beta t + \int_0^\infty \frac{t}{\lambda + t} d\mu(\lambda)
= \alpha + \beta t + \int_0^\infty \frac{\lambda t}{\lambda + t} d\nu(\lambda)
\end{align}
where $\alpha$ is a real number, $\beta \geq 0$ is a non-negative real number,
and $\mu$ is a positive measure on the interval $[0, \infty)$ (and $\nu$ is
a minor change in measure).

In the below, we rely on the following theorem
regarding the theory of monotone matrix functions.
\begin{restatable}[Loewner's theorem on matrix monotone functions~\citep{loewner1934monotone,bhatia2009loewner},~\citep{simon2019loewner}-Theorem 5.1]
{rthm}{loewnersmatrixtheorem}
\label{thm:loewnersmatrixtheorem}
Let $\psi: \R_+ \to \R$ be a real-valued function. Then 
$\psi$ is matrix monotone if and only if for all 
$x_1 < x_2 < \dots < x_m$  we have
that the Loewner matrix $L_\psi(x_1, x_2, \dots, x_m) \in \R^{m \times m}$ whose
$i,j$ entry is defined as
\begin{align}
(L_\psi(x_1, x_2, \dots, x_m))_{ij} = \begin{cases}
\frac{\psi(x_i) - \psi(x_j)}{x_i - x_j} & \text{if } i \neq j \\
\psi'(x_i) & \text{if } i = j
\end{cases}
\end{align}
is positive semidefinite for all $m \in \mathbb{N}$.
\end{restatable}
This theorem, amongst other things, gives us a useful strategy to find counterexamples
of matrix monotonicity.

\concavebutnotneggradmatrixmonotone*This is originally stated in
Section~\ref{sec:weakly-matrix-monotone},
page~\pageref{lem:concave-but-not-matrix-monotone}.\vspace{-1.0em}
\begin{proof}
\label{proof:concave-but-not-matrix-monotone}

We restate the three functions here.
First, 
$\mfa(x) = \mfna$,
is a power-law style function mimicking the Chinchilla scaling laws 
from Section~\ref{sec:neural-scaling-laws}. The second
is $\mfb(x) = \mfnb$, which 
is a form of saturating exponential function. A third function
we consider is $\mfc(x) = \mfnc$. We note
that, as stated, $\mfa(x) = \mfc(x)$ 
for $x \geq 0$. However, our parameterized
generalization of these functions renders them
no longer identical for all values of the parameters.
That is, we actually prove a stronger result than the one stated in the lemma, namely
parameterized versions of the above three functions. We thus first
restate the generalized version of the three, as well as give
valid parameter ranges:
\begin{align}
\mfa(x) &= \mfunca, &\text{ for } \parama \label{eq:gen_param_mfa} \\
\mfb(x) &= \mfuncb, &\text{ for } \paramb \label{eq:gen_param_mfb} \\
\mfc(x) &= \mfuncc, &\text{ for } \paramc \label{eq:gen_param_mfc}
\end{align}
We immediately see that each of these functions are concave since
they are all such that $\phi''_i(x) \leq 0$ for $x \geq 0$. Also,
these functions\footnote{For $\mfa(x) = \mfunca$, we must fix $\beta=1$ for this to hold.} 
all have $\phi_i(0) = 0$ and $\phi'_i(x) \geq 0$
for $x \geq 0$ and hence, these are all normalized monotone non-decreasing
concave functions. Considering Theorem~\ref{thm:eigenvalue_updates},
this means that any matrix spectral function using $\phi_i$ for $i \in \{1,2,3\}$
will be a normalized monotone-nondecreasing set function. The next question is if
they lead to submodular matrix spectral functions.

It is important to realize also that, considering their parameterized generalizations,
$\mfa(x)$ and $\mfc(x)$ are no longer identical families of concave functions.

\paragraph{Proof that $-\mfa'$ is not matrix monotone:}
We have that $\mfa(x) = \mfunca$, for $\parama$. Also, we have that
$g_\mfnuma(x) = - \mfa'(x) = \neggradmfunca$. Since $\beta > 0$ and $x \geq 0$,
we can change variables with $y = x+\beta$ and have $g_\mfnuma(y) = -\alpha
y^{-\alpha - 1}$. Secondly, since $\alpha > 0$ and multiplying a function by a
positive constant does not change matrix monotonicity, this further reduces to
$g_\mfnuma(y) = - y^{-(\alpha + 1)}$. Lastly, since $\alpha > 0$, we have
$\alpha' = \alpha + 1 > 1$, and this further reduces
down to $g_\mfnuma(y) = - y^{-\alpha'}$ for $\alpha' > 1$. 
This looks similar to $\xi_4$ in 
Theorem~\ref{thm:simple-analytical-matrix-monotone-functions} but with a negative sign
in the exponent, and it is
known that the only satisfying range of the exponent is from $1$ to $2$,
and hence $\alpha' > 1$ is outside of this range, and thus
the function is not matrix monotone. As an additional check,
we apply Theorem~\ref{thm:loewnersmatrixtheorem} with the following
example. Set $\alpha = 1$ and $\beta = 0$ so that
$g_\mfnuma(y) = - y^{-2}$. 
The Loewner matrix and with the three points $1$, $2$, and $3$. We get
\begin{align}
L_{g_\mfnuma}(1,2,3) = \begin{bmatrix}
g'_\mfnuma(1) & \frac{g_\mfnuma(1) - g_\mfnuma(2)}{1 - 2} & \frac{g_\mfnuma(1) - g_\mfnuma(3)}{1 - 3} \\
\frac{g_\mfnuma(2) - g_\mfnuma(1)}{2 - 1} & g'_\mfnuma(2) & \frac{g_\mfnuma(2) - g_\mfnuma(3)}{2 - 3} \\
\frac{g_\mfnuma(3) - g_\mfnuma(1)}{3 - 1} & \frac{g_\mfnuma(3) - g_\mfnuma(2)}{3 - 2} & g'_\mfnuma(3) \\
\end{bmatrix}
\end{align}
We have that $g'_\mfnuma(y) = 2 y^{-3}$, so the above becomes
\begin{align}
L_{g_\mfnuma}(1,2,3) = \begin{bmatrix}
2 & 3/4 & 4/9 \\
3/4 & 1/4 & 5/36 \\
4/9 & 5/36 & 2/27. \\
\end{bmatrix}
\end{align}
Of the three eigenvalues, one has value $\approx -0.0475019$ which
is negative, thus establishing the non matrix monotonicity of $g_\mfnuma$ and 
the lack of guaranteed submodularity of $f_\mfnuma(X) = \trace{\mfa(B_X)}$.

\paragraph{Proof that $-\mfb'$ is not matrix monotone:}
We have that $\mfb(x) = \mfuncb$, for $\paramb$.
Also, we have that $g_\mfnumb(x) = - \mfb'(x) = -\gradmfuncb$. 
In this case, we simply find an $A,B$
where $0 \preceq A \preceq B$ but not $g_\mfnumb(A) \preceq g_\mfnumb(B)$.
Consider the following
two $3 \times 3$ matrices.
\begin{align}
A = \begin{bmatrix}
1 & 0 & 0 \\
0 & 2 & 0 \\
0 & 0 & 3 \\
\end{bmatrix}
\quad \text{and} \quad
B = \begin{bmatrix}
1.1 & 0.1 & 0.1 \\
0.1 & 2.1 & 0.1 \\
0.1 & 0.1 & 3.1 \\
\end{bmatrix}
\end{align}
We immediately see that these matrices are symmetric positive definite,
$A$ we have immediately since it is diagonal and $B$ 
by the Gershgorin circle theorem. This says that every eigenvalue
of $B$ lies within at least one of the Gershgorin discs defined by the rows of $B$.
Specifically, every eigenvalue of $B$ lies within at least one of
$D(b_{ii}, R^B_i)$ where $b_{ii}$ is the diagonal element of $B$ in row $i$ and 
$R^B_i = \sum_{j \neq i} |b_{ij}|$. We have $R^B_1 = 0.2$, $R^B_2 = 0.2$, and $R^B_3 = 0.2$, so 
the Gershgorin discs are $D(1.1,0.2)$, $D(2.1,0.2)$, and $D(3.1,0.2)$. We see 
all disks include only positive values, and hence all the eigenvalues of $B$ are positive.
Next, consider the matrix $B-A$:
\begin{align}
B - A = \begin{bmatrix}
0.1 & 0.1 & 0.1 \\
0.1 & 0.1 & 0.1 \\
0.1 & 0.1 & 0.1 \\
\end{bmatrix}
= 0.1 \begin{bmatrix}
1 & 1 & 1 \\
1 & 1 & 1 \\
1 & 1 & 1 \\
\end{bmatrix}
\end{align}
which is a rank-1 matrix with eigenvalues $0.3$, $0$, and $0$.
Thus $0 \preceq A \preceq B$ or
$B - A$ is positive semidefinite. We now
consider $g_\mfnumb(A) = - \mfb'(A)$ and $g_\mfnumb(B) = - \mfb'(B)$. 
Resorting to numerics, we have
\begin{align}
g_\mfnumb(B) - g_\mfnumb(A) \approx
\begin{bmatrix}
 0.03289065 & 0.02045755 & 0.01379102 \\
 0.02045755 & 0.01158852 & 0.00712449 \\
 0.01379102 & 0.00712449 & 0.00395176
\end{bmatrix}
\end{align}
with eigenvalues 
approximately 
[ 5.04624571e-02, -2.04197907e-03,  1.04590142e-05]
one of which is negative. Hence, 
$g_\mfnumb(B) - g_\mfnumb(A)$ is not
positive semidefinite, and thus 
$f_\mfnumb(X) = \trace{\mfb(B_X)}$ is not 
guaranteed to be submodular.

\paragraph{Proof that $-\mfc'$ is not matrix monotone:}
We have that $\mfc(x) = \mfuncc$, for $\paramc$.
Also, we have that $g_\mfnumc(x) = - \mfc'(x) = -\gradmfuncc$.
In this case, we consider three examples, one
for $0 < \alpha < 1$, one for $\alpha = 1$, and
one for $\alpha > 1$ and in each case show that
a corresponding Loewner matrix has
a negative eigenvalue, thus establishing the non matrix monotonicity of $g_\mfnumc$ and
the lack of guaranteed submodularity of $f_\mfnumc(X) = \trace{\mfc(B_X)}$.
For $0 < \alpha < 1$,  we take $\alpha = 1/2$
and consider a $2 \times 2$ Loewner matrix
\begin{align}
L_{g_\mfnumc}(x_1,x_2) = \begin{bmatrix}
g'_\mfnumc(x_1) & \frac{g_\mfnumc(x_1) - g_\mfnumc(x_2)}{x_1 - x_2} \\
\frac{g_\mfnumc(x_2) - g_\mfnumc(x_1)}{x_2 - x_1} & g'_\mfnumc(x_2) \\
\end{bmatrix}.
\end{align}
With $\alpha = 1/2$ 
we have $g_\mfnumc(x) = -(1+\sqrt{x})^{-3}$
and $g'_\mfnumc(x) = 3(1+\sqrt{x})^{-4}/(2\sqrt{x})$.
We set $x_1 =1$ $x_2 = 9$ and get
\begin{align}
L_{g_\mfnumc}(1,9) 
= \begin{bmatrix}
\frac{3}{32} & \frac{7}{512} \\
\frac{7}{512} & \frac{1}{512} \\
\end{bmatrix}
\end{align}
which has a negative determinant 
of $\approx -3.81 \times 10^{-6}$
and hence is not positive semidefinite.
For $\alpha=1$, $g_\mfnumc(x) = -(1+x)^{-2}$ and $g'_\mfnumc(x) = 2(1+x)^{-3}$.
In this case, we again use a $2 \times 2$ Loewner matrix with $x_1 = 1$ and $x_2 = 2$ 
and after some algebra get a negative determinant.
For $\alpha > 1$, we take $\alpha = 2$ where
$g_\mfnumc(x) = -(1+x^2)^{-3/2}$ and $g'_\mfnumc(x) = 3 x^2 (1+x^2)^{-5/2}$.
Yet again, we can get away with a $2 \times 2$ Loewner matrix and get a negative determinant
again evaluating with $x_1 = 1$ and $x_2 = 2$.
Thus,
$f_\mfnumc(X) = \trace{\mfc(B_X)}$ is not 
guaranteed to be submodular.

\end{proof}

We restate the following definition for convenience:
\weaklyMatrixMonotone*

\weaklyMatrixMonotoneFunctionsLeadToWeaklySubmodularOuterProductMatrixSpectralFunctions*
This is originally stated in
Section~\ref{sec:weakly-matrix-monotone},
page~\pageref{thm:weakly-matrix-monotone-functions-lead-to-weakly-submodular-outer-product-matrix-spectral-functions}.\vspace{-1.0em}
\begin{proof}
\label{proof:weakly-matrix-monotone-functions-lead-to-weakly-submodular-outer-product-matrix-spectral-functions}
We have that 
$B_X = \mtrans{\data[X]}\data[X] = \sum_{i \in X} u_i \mtrans{u}_i$
where $u_i \in \R^m$ is the $i^\text{th}$ row of $\data \in \R^{n \times m}$.
Then for $s \in V \setminus X$,
the matrix $B_{X + s} = B_X + u_s \mtrans{u}_s$ is a rank-1 update of $B_X$. 
Consider the function $\chi(t) 
= \trace{\phi(B_X + t u_s \mtrans{u}_s)}$ for $t \in [0,1]$.
We have $f(X) = \chi(0) = \trace{\phi(B_X)}$,
$f(X + s) = \chi(1) = \trace{\phi(B_{X + s})}$, and 
$\frac{d}{dt} \chi(t) = \chi'(t) = \trace{\phi'(B_X + t u_s \mtrans{u}_s) u_s \mtrans{u}_s}$.
Then, by the fundamental theorem of calculus, we have that
\begin{align}
f(s|X) &= f(X + s) - f(X) = \chi(1) - \chi(0) \\
       &= \int_0^1 \chi'(t) dt = \int_0^1 \trace{\phi'(B_X + t u_s \mtrans{u}_s) u_s \mtrans{u}_s} dt. 
\end{align}
The trace trick says for any three matrices $A$, $B$, and $C$ of appropriate dimensions, we have 
$\trace{ABC} = \trace{CAB}$. Therefore
\begin{align}
\trace{\phi'(B_X + t u_s \mtrans{u}_s) u_s \mtrans{u}_s} 
&= \trace{\mtrans{u}_s \phi'(B_X + t u_s \mtrans{u}_s) u_s} \\
&= \mtrans{u}_s \phi'(B_X + t u_s \mtrans{u}_s) u_s
\end{align}
And therefore,
\begin{align}
f(s|X) = \int_0^1 \mtrans{u}_s \phi'(B_X + t u_s \mtrans{u}_s) u_s dt.
\end{align}
Now, recalling~\Cref{defn:weakly-matrix-monotone},
$g = -\phi'$ being $\weakly$-weakly
matrix monotone is the same
as $\phi'$ being $\weakly$-weakly matrix antitone, 
which means that for any $A \preceq B$ we have $\phi'(A) \succeq \weakly \phi'(B)$.
We have moreover that for any $X \subseteq Y \subseteq V$,
$B_X \preceq B_Y$ since $B_Y = B_X + B_{Y \setminus X}$
and $0 \preceq B_{Y \setminus X}$. Therefore, 
for all $u \in \R^m$ and $t \in [0,1]$, since $\phi'(B_X) \succeq \weakly \phi'(B_Y)$, we have
\begin{align}
\mtrans{u} \phi'(B_X + t u \mtrans{u}) u \geq \weakly \mtrans{u} \phi'(B_Y + t u \mtrans{u}) u
\end{align}
Integrating over $t \in [0,1]$, we get
\begin{align}
f(s|X) &= \int_0^1 \mtrans{u}_s \phi'(B_X + t u_s \mtrans{u}_s) u_s dt \\
      &\geq \weakly \int_0^1 \mtrans{u}_s \phi'(B_Y + t u_s \mtrans{u}_s) u_s dt \\
      &= \weakly f(s|Y)
\end{align}
Thus, $f(s|X) \geq \weakly f(s|Y)$ establishing the $\weakly$-weakly DR submodularity of $f$.

The monotone non-decreasing property of $f$ follows from
the fact that the eigenvalues of $B_Y$ are at least
as large as the eigenvalues of $B_X$ (which
follows from Theorem~\ref{thm:eigenvalue_updates}) and
the fact that $\phi'(x) \geq 0$ for $x \geq 0$.
\end{proof}

\weaklyMatrixMonotoneFunctionsLeadToWeaklySubmodularMatrixSpectralFunctions*
This is originally stated in
Section~\ref{sec:weakly-matrix-monotone},
page~\pageref{cor:weakly-matrix-monotone-functions-lead-to-weakly-submodular-matrix-spectral-functions}.\vspace{-1.0em}
\begin{proof}
\label{proof:weakly-matrix-monotone-functions-lead-to-weakly-submodular-matrix-spectral-functions}
Any PSD matrix $B \in \R^{n \times n}$ with finite $n$
can be represented as $B = D D^T$ where $D$ is some $n \times r$ 
matrix with $r \leq n$. If $B$ is the
gram matrix according to a positive definite 
kernel function~\cite{bottou2007large,steinwart2008kernels,scholkopf2002learning,williams2006gaussian},
it is likely that $r = n$ and that
$r$ grows with $n$ but this does not matter.
For $X \subseteq V$, with $B_X$ is the principal row/column
submatrix of $B$,
then $B_X = D[X] D[X]^T$ where 
$D[X]$ is the $|X| \times r$ 
row sub-matrix of $D$. 
From eigenvalue duality,
$B_X$ has the same eigenvalues as the matrix $D[X]^T D[X]$ 
(in general, $B_X$ has at most $|X|$
strictly positive eigenvalues, but
the number of positive eigenvalues 
grows with $|X|$ up to $n=|V|$).
Thus, we have
$f(X) = \trace{\phi(B_X)} = \trace{\phi(D[X]^T D[X])}$.
Now, $D[X]^T D[X]$ is a sum of rank-1 matrices,
each formed by outer products of vectors of length $r$.
Thus, the result immediately follows from 
Theorem~\ref{thm:weakly-matrix-monotone-functions-lead-to-weakly-submodular-outer-product-matrix-spectral-functions}.
\end{proof}
We note that 
Theorem~\ref{thm:weakly-matrix-monotone-functions-lead-to-weakly-submodular-outer-product-matrix-spectral-functions}
and Corollary~\ref{cor:weakly-matrix-monotone-functions-lead-to-weakly-submodular-matrix-spectral-functions}
is a generalization of~\cite{friedland2011submodular,audenaert2010strongly,lewin2014family}
since it not only establishes submodularity when $\weakly=1$
it also establishes weak submodularity when $\weakly < 1$. The proof technique, however,
is quite different than~\cite{friedland2011submodular,audenaert2010strongly,lewin2014family}
which relies on Loewner's representation.

Now moving to the next sub-theme, 
suppose we have a scalar non-negative function $g:[0,R]\to[0,\infty)$ that is 
non-increasing, meaning that if $x \leq y$, then $g(x) \geq g(y)$. This is
analogous to, but not the same as, matrix antitonicity.
Let $A$ and $B$ be positive semidefinite matrices such that $0\preceq A\preceq B$.
Using spectral decompositions, we get:
\begin{align}
A &= Q_A \Lambda_A Q_A^T,\\
B &= Q_B \Lambda_B Q_B^T,
\end{align}
and we can define
\begin{align}
g(A) &= Q_A g(\Lambda_A) Q_A^T,\\
g(B) &= Q_B g(\Lambda_B) Q_B^T.
\end{align}
Here $g(\Lambda_A)$, like any function on a matrix in this paper, means 
that $g$ is applied entrywise to the diagonal
entries of $\Lambda_A$, and similarly for $g(\Lambda_B)$.

Since $0\preceq A\preceq B$, the eigenvalues of $A$ are no larger than the
corresponding eigenvalues of $B$, when both are ordered increasingly or
decreasingly. Since $g$ is non-increasing, the eigenvalues of $g(A)$ are
therefore no smaller than the corresponding eigenvalues of $g(B)$.
However, this does {\bf not} imply
\begin{align}
g(A)\succeq g(B),
\end{align}
the reason for this is that $A$ and $B$ can have different
eigenvectors (or a different eigenbasis).
Thus, importantly, scalar
non-increasingness of $g$ does not imply matrix antitonicity. 

If, however, $A$ and $B$ are simultaneously diagonalizable (which
means they commute), then they share an eigenbasis. In this case we
can write
\begin{align}
A &= Q\Lambda_A Q^T,\\
B &= Q\Lambda_B Q^T.
\end{align}
Then
\begin{align}
g(A)-g(B) &= Qg(\Lambda_A)Q^T - Qg(\Lambda_B)Q^T \\
         &= Q\bigl[g(\Lambda_A)-g(\Lambda_B)\bigr]Q^T.
\end{align}
Since $A\preceq B$ implies $\Lambda_A\preceq \Lambda_B$ entrywise, and since
$g$ is non-increasing, the diagonal matrix
\begin{align}
g(\Lambda_A)-g(\Lambda_B)
\end{align}
has nonnegative diagonal entries. Hence
\begin{align}
g(A)-g(B)\succeq 0.
\end{align}
So in this special commuting case, $g$ behaves as if it were matrix
antitone.

Now let
\begin{align}
\varrho_B = \lambda_{\max}(B)
\end{align}
be the spectral radius of $B$. Since $0\preceq A\preceq B$, every eigenvalue
of $A$ and $B$ lies in the range $[0,\varrho_B]$. Therefore, because $g$ is non-increasing,
\begin{align}
g(\lambda_i(A))\ge g(\varrho_B)
\end{align}
for every eigenvalue $\lambda_i(A)$ of $A$.

Using the spectral decomposition $A=Q_A\Lambda_A Q_A^T$, we have
\begin{align}
g(A)-g(\varrho_B)I
&= Q_A g(\Lambda_A) Q_A^T - g(\varrho_B)Q_AQ_A^T\\
&= Q_A\bigl[g(\Lambda_A)-g(\varrho_B)I\bigr]Q_A^T.
\end{align}
As we established above, the diagonal matrix $g(\Lambda_A)-g(\varrho_B)I$ has nonnegative diagonal entries,
so it is positive semidefinite. Therefore
\begin{align}
g(A)-g(\varrho_B)I \succeq 0.
\end{align}
Equivalently,
\begin{align}
g(A)\succeq g(\varrho_B)I.
\end{align}

Thus, even though scalar non-increasingness of $g$ does not imply full matrix
antitonicity, it does imply the useful lower (and by the same reasoning, upper) bounds:
\begin{align}
g(0)I \succeq g(B) \text{ and } g(A)\succeq g(\lambda_{\max}(B))I
\end{align}
whenever $0\preceq A\preceq B$. This helps
to motivate the following lemma.

\begin{restatable}[Bound on $\weakly$]{rlem}{boundonweakly}
\label{lem:bound-on-weakly}
Let $\phi$ be a monotone non-decreasing non-negative concave
function that is normalized $\phi(0)=0$ with positive gradient,
i.e., $\phi'(x) \geq 0$ for $x \geq 0$ and $\phi'(0) > 0$. Then 
the set function $f(X) = \trace{\phi(B_X)}$ is $\weakly$-weakly DR submodular 
for $\weakly \geq \phi'(\spectralradius)/\phi'(0)$ where 
$\spectralradius_V = \spectralradius(B_V)$ is the spectral radius (i.e., largest eigenvalue) of $B_V$.
\end{restatable}
\begin{proof}
Since $\phi$ is concave, $\phi''(x) \leq 0$ for $x \geq 0$. Hence
$\phi'(x)$ is a non-increasing function in this domain. Define $g(x) = \phi'(x)$ for convenience.
Then we have for $0 \leq x \leq y$, $g(0) \geq g(x) \geq g(y)$. Note
also that by Theorem~\ref{thm:eigenvalue_updates}, we have
$0 \leq \spectralradius(B_X) \leq \spectralradius(B_Y) \leq \spectralradius_V$
for all $X \subseteq Y \subseteq V$. Therefore, from the discussion above,
we have
\begin{align}
  g(\spectralradius_V) I \preceq g(B_X) \text{ and } g(B_Y) \preceq g(0) I
\end{align}
where $I$ is the $m \times m$ identity matrix.
Multiplying the right hand side by $g(\spectralradius_V)/g(0)$, we get
\begin{align}
  g(\spectralradius_V) I \succeq \frac{g(\spectralradius_V)}{g(0)} g(B_Y).
\end{align}
Combining the above two inequalities, we get
\begin{align}
  g(B_X) \succeq g(\spectralradius_V) I \succeq \frac{g(\spectralradius_V)}{g(0)} g(B_Y).
\end{align}
Thus, $g(B_X) \succeq \weakly g(B_Y)$ for $\weakly \geq
g(\spectralradius_V)/g(0)$, and hence $g$ is $\weakly$-weakly matrix antitone,
and thus $f(X) = \trace{\phi(B_X)}$ is $\weakly$-weakly DR submodular for $\weakly \geq g(\spectralradius_V)/g(0)$. 
\end{proof}

We note that $\spectralradius$ is a finite value since $B_V$ is a finite matrix.
Also, in practice $\spectralradius$ is often not too large. For example, it is typical
when applying the Vendi score to normalize the design matrix so that
the trace of the gram matrix to be 1 (much like a density matrix in quantum mechanics). This
means that, if $\lambda_i$ is the $i^\text{th}$ eigenvalue of 
$B_V$, then $0 \leq \gamma_i \leq 1$ and $\sum_{i=1}^m \lambda_i = 1$. While
this means that $\spectralradius$ could be as large as $1$, for typical datasets,
the eigenvalues would be more balanced (more like a high-entropy distribution) and
hence $\spectralradius$ would be similar to $1/m$ where $m$ is the feature
vector dimension. A reasonable upper estimate of the
spectral radius might be $100/m$ and with $m=1024$ this
gives $\tilde \spectralradius \approx 0.1$ 

\someusefulconcavefunctionsareweaklymatrixmonotone*
This is originally stated in
Section~\ref{sec:weakly-matrix-monotone},
page~\pageref{lem:some-useful-concave-functions-are-weakly-matrix-monotone}.\vspace{-1.0em}
\begin{proof}
\label{proof:some-useful-concave-functions-are-weakly-matrix-monotone}
We consider each of the three functions and their gradients individually.
We first note that each of these functions satisfy the
conditions of Lemma~\ref{lem:bound-on-weakly}. Thus, we can apply that lemma 
to get a lower bound on $\weakly$ for each one. In computing the specific
bounds for the instances
below, we use default options of $\alpha=\beta=1$, but note that this is just
for an example; these values are not optimized to provide the best
approximation bound.

First, we have
\begin{align}
\mfa(x) &= \mfunca, \text{ for } \parama
\quad \text{ and } \quad \mfa'(x) = \gradmfunca.
\end{align}

Using Lemma~\ref{lem:bound-on-weakly}, we have
$\weakly \geq \mfa'(\spectralradius)/\mfa'(0)$.
Here, $\mfa'(0) = \alpha \beta^{- \alpha - 1}$
and $\mfa'(\spectralradius) = \gradmfuncaarg{\spectralradius}$.
With $\alpha = \beta = 1$ and $\tilde \spectralradius \approx 0.1$ as discussed above, we have
$\weakly \geq \mfa'(0.1)/\mfa'(0) \approx 0.826$ and hence 
$1 - e^{-\weakly} \approx 0.5623$.

Next, we have
\begin{align}
\mfb(x) &= \mfuncb, \text{ for } \paramb
\quad \text{ and } \quad \mfb'(x) = \gradmfuncb.
\end{align}
Using Lemma~\ref{lem:bound-on-weakly}, we have
$\weakly \geq \mfb'(\spectralradius)/\mfb'(0) = e^{-\spectralradius}$.
This gives a multiplicative approximation bound for the greedy algorithm
of $1-e^{-\weakly} \geq 1-e^{-e^{-\spectralradius}}$. 
With $\tilde \spectralradius \approx 0.1$ as discussed above,
and $\weakly \geq e^{-\spectralradius} \approx 0.905$ and 
$1-e^{-e^{-\spectralradius}} \approx 0.595$.
This is not much worse than the $1-e^{-1} \approx 0.632$ for fully submodular functions.
In general, however, we have $\weakly \geq e^{- \spectralradius}$.

Third, we have
\begin{align}
\mfc(x) &= \mfuncc. \text{ for } \paramc
\quad \text{ and } \quad \mfc'(x) = \gradmfuncc.
\end{align}
Once again, using Lemma~\ref{lem:bound-on-weakly}, we have
$\weakly \geq \mfc'(\spectralradius)/\mfc'(0)$.
Now $\mfc'(0) = 1$ and $\mfc'(\spectralradius) = \gradmfunccarg{\spectralradius}$.
With $\alpha = 1$ and $\tilde \spectralradius \approx 0.1$ as discussed above, we have
$\weakly \geq \mfc'(0.1) \approx 0.826$ and hence 
$1 - e^{-\weakly} \approx 0.5623$.

\end{proof}

\section{Background on Householder Reflectors and Rank-1 Updates}
\label{sec:extr-background-review}

\subsection{Preliminaries and Notation}
\label{sec:extr-prel}

Let $Q$ be an $m \times m$ matrix of orthonormal column vectors thus spanning an
$m$-dimensional space. Then $I_m = Q^T Q = QQ^T$. In this section, we also offer
subsidiary explanations to clarify certain points but to not interrupt the flow of the main text,
they may be skipped by readers who are already familiar with this material.

\begin{explanation}{}
Why is $I_m = Q^T Q = QQ^T$?
We immediately have that $I_m = Q^T Q$ since $Q$ has orthonormal columns.
Also $I_m = Q Q^T$ which follows
since if $A$ and $B$ are any $m \times m$ matrices
with $AB=I$ then $A = IA = (AB)A = A(BA) = AI$
so $BA=I$ as well. Hence, an orthonormal matrix has
both orthonormal columns and rows.
\end{explanation}

When $Q$ be an $m \times r$ matrix of orthonormal column vectors with
$r < m$ we still have $Q^T Q = I_r$ but it is no longer the case that
$Q Q^T = I_m$. However, $Q Q^T$ can be seen as a projection matrix
(sometimes called a projector) since given any vector $x \in \R^m$, the vector
$y= Q Q^Tx$ is the projection of $x$ into the space spanned by the
column vectors in $Q$. Also, if $x \in \text{span}(Q)$, then
$x = Q Q^T x$ and so $Q Q^T$ does act like an identity matrix for
those $x$ already in the column span of $Q$. A projector can even
be formed from a vector outer product. I.e., if $\bar u$ is a unit
vector, then $\bar u {\bar u}^T$ is a projector matrix onto
the direction of $\bar u$ since $\bar u {\bar u}^T x = \alpha \bar u$
where $\alpha$ is the length of $x$'s projection onto $\bar u$.

Of note, if $q$ is an eigenvector of matrix $B$ with eigenvalue
$\lambda$, then $-q$ is also eigenvector of matrix $B$ with eigenvalue
$\lambda$ which follows since
\begin{align}
  B(-q) = - (Bq) = - (\lambda q) = \lambda (-q)
\end{align}
Hence, when computing eigenvalues it is often arbitrary if we return
either $q$ or $-q$. When $Q$ is a matrix with eigenvectors as columns,
we have $Q^T Q = I$ regardless of if a column of $Q$ is multiplied
by $-1$.

\subsection{Elementary reflector matrices}
\label{sec:elem-refl-matr}

An elementary reflector matrix is also known as a Householder
transformation, which is a square matrix that describes a reflection
about a hyperplane.

Given non-zero vector $u \in \R^m$, elementary
reflector matrix $H$ is defined as
\begin{align}
H = I - 2 \frac{u u^T}{u^Tu}
\end{align}
where $I$ is the $m \times m$ identity matrix,
and $\frac{uu^T}{u^Tu}$ is a projection matrix onto
the vector $u$. Note that we can define
a normalized version 
$\bar u = u/\|u\|_2$ so that
\begin{align}
H = I - 2 \bar u {\bar u}^T
\end{align}
which justifies why we see 
$\frac{uu^T}{u^Tu}$ as a projection matrix onto $u$,
namely if $x \in \R^m$ then 
$\bar u {\bar u}^T x = \alpha \bar u$ where
$\alpha={\bar u}^T x$ is the component of $x$ in the
direction of unit vector $\bar u$. 

\begin{figure}
\centering
\includegraphics[width=0.8\textwidth]{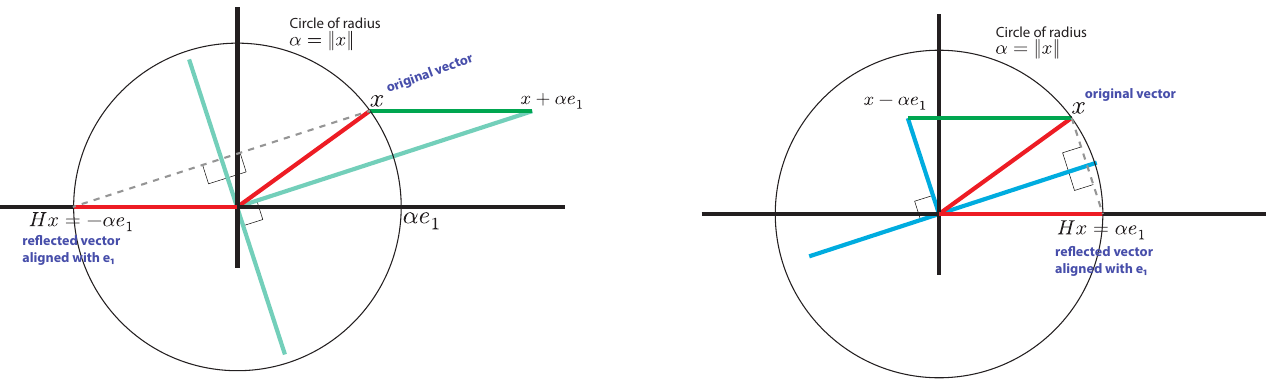}
\caption{2D description of Householder reflector where
an original vector $x$ is reflected to 
a new vector $Hx$ that is aligned with the coordinate axis $e_1$.
There are two choices: (1) for 
$u = x + \| x \|_2 e_1$ (on the left)
which reflects $x$ to $-\alpha e_1$;
and (2) $u = x - \| x \|_2 e_1$ (on the right)
which reflects $x$ to $\alpha e_1$.
Here, $u$ is an unnormalized version of $\bar u$. In general,
for $x \in \R^m$ we can reflect it to any $\pm \alpha e_j$ for $j \in [m]$ 
by using $u = x \pm \| x \|_2 e_j$.
}
\label{fig:householderreflector}
\end{figure}

The name ``reflector'' comes from the fact that $Hx$ is a reflection
of $x$ across the hyperplane
that is perpendicular to $u$ and that goes through 0
(see Figure~\ref{fig:householderreflector}). I.e.,
we can think of $Hx = x - 2 \bar u {\bar u}^T x$
which is $x$ with twice $x$'s projection onto $u$ removed,
and hence is a reflection.

Some properties of $H$ are that:
\begin{itemize}

\item Symmetric: $H = H^T$ which is immediate. In
fact it is Hermitian in that $H = H^*$.

\item Orthonormality: $H$ is an orthonormal matrix
which means its columns/rows form an
orthonormal basis. To see this, we compute:
\begin{align}
H^T H &= 
\left(I - 2 \frac{u u^T}{u^Tu}\right)^T
(I - 2 \frac{u u^T}{u^Tu})
= I - 2 \frac{u u^T}{u^Tu} 
- 2 \frac{u u^T}{u^Tu} 
+ 4 \frac{u u^T}{u^Tu} \frac{u u^T}{u^Tu}  \\
&= I - 2 \frac{u u^T}{u^Tu} 
- 2 \frac{u u^T}{u^Tu} 
+ 4 \frac{u (u^T u) u^T}{(u^Tu)^2}  = I
\end{align}
and also $HH^T = I$ so that $H^{-1} = H^T$ (i.e., $H$
is a unitary matrix)
and $H^2 = I$.

Also note that if $D = dI$ (a diagonal matrix with
a constant along the diagonal) then $H^T D H = dI$.

However, if $D$ is any arbitrary diagonal matrix, we have
\begin{align}
H^T D H &= 
\left(I - 2 \frac{u u^T}{u^Tu}  \right )^T D
 (I - 2 \frac{u u^T}{u^Tu}) \\
&= 
\left(D - 2 \frac{u u^T}{u^Tu} D \right )^T
 (I - 2 \frac{u u^T}{u^Tu})
= D  - 2 D \frac{u u^T}{u^Tu} 
- 2 \frac{u u^T}{u^Tu} D
+ 4 \frac{u u^T}{u^Tu} D \frac{u u^T}{u^Tu}  \\
&= I - 2 D\frac{u u^T}{u^Tu} 
- 2 \frac{u u^T}{u^Tu} D
+ 4 \frac{u (u^T D u) u^T}{(u^Tu)^2}  \\
&= I - 2 \frac{u_d u^T}{u^Tu} 
- 2 \frac{u u_d^T}{u^Tu}
+ 4 c \frac{u u^T}{(u^Tu)^2}
\end{align}
where $u_d = Du$ and $c = u^T D u$. Thus
$H^T D H \neq D$ unless $D=dI$.

\item Involutory: $H$ is its own inverse since $H^2 = I$.

\item Eigenvalues: any Householder matrix $H$ has eigenvalues of 1 and -1.

To see this, let $H = I - 2 \bar u {\bar u}^T$
and let $\bar x$ be a vector orthogonal to $\bar u$
so that $\langle \bar x, \bar u \rangle = 0$.
Then
\begin{align}
H \bar x = 
(I - 2 \bar u {\bar u}^T) \bar x
= \bar x - 2 \bar u \langle u , \bar x \rangle
= \bar x
\end{align}
and so $\bar x$ is an Eigenvector with eigenvalue 1.
Also, the multiplicity of eigenvalue 1 is $m-1$
the reason being that there are $m-1$ independent
orthogonal choices for $x$ in an $m$-dimensional space.
For the final eigenvalue, the last direction is $\bar u$,
and consider
\begin{align}
H \bar u = 
(I - 2 \bar u {\bar u}^T) \bar u
= \bar u - 2 \bar u = - \bar u
\end{align}
Hence, $\bar u$ is also an Eigenvector with eigenvalue $-1$.

\item For the above reason, $\det(H)=-1$. Hence, 
as also implied above, $\text{rank}(H)=m$ and $H$ is invertible.

\item Norm preservation. $H$ preserves norms and angles of
vectors when it is used as a premultiplier. I.e.,
\begin{align}
\| H x \|^2 &= 
\| (I - 2 \bar u {\bar u}^T) x \|^2 = 
\| x - 2 \bar u {\bar u}^T x \|^2 \\
&= (x - 2 \bar u {\bar u}^T x)^T(x - 2 \bar u {\bar u}^T x) \\
&= 
x^T x - 2 x^T \bar u {\bar u}^T x  
- 2 (\bar u {\bar u}^T x)^T x
+ 4 (\bar u {\bar u}^T x)^T \bar u {\bar u}^T x \\
&=
x^T x - 2 x^T \bar u {\bar u}^T x  
       - 2 x^T \bar u {\bar u}^T  x
+ 4 x^T \bar u {\bar u}^T \bar u {\bar u}^T x \\
&= x^T x - 4 x^T \bar u {\bar u}^T x  + 4 x^T \bar u {\bar u}^T x = x^T x = \| x \|^2
\end{align}
and hence $\| H x \| = \| x \|$.

Also, for $x,y$ vectors,
\begin{align}
\langle Hx , Hy \rangle
= (Hx)^THy = x^T H^T H y = x^T y
\end{align}
so inner-products and thus angles are preserved.

\item $H$ preserves orthogonality when used as a post-multipler. That
is, if $Q$ is an $m \times r$ matrix of orthonormal column vectors, then
$QH$ is still orthonormal. We have this since
\begin{align}
\langle QH, QH \rangle
= H^T Q^T Q H = H^T H = I
\end{align}
However, if $Q$ is just orthogonal with $Q^T Q = D$ (a diagonal matrix), then 
\begin{align}
\langle QH, QH \rangle
= H^T Q^T Q H = H^T D H
\end{align}
which, as mentioned above, is not $D$ in general unless $D=dI$.

\item Symmetry of reflections. If $y = Hx$ then
$Hy = H^2x = Ix = x$ so that $x = Hy$.
So if $H$ maps $x$ to $y$ (a reflection)
then we can reflect it back by another
application of $H$, consistent with our
intuition of what a reflection should do.

\end{itemize}

One way to construct a reflector matrix is to find one that reflects
any vector $x$ to a vector that is aligned with a coordinate axis such
as the first basis vector $e_1$ (this is used in QR factorization).  That is,
to construct an $x$-dependent Householder matrix $H$ such that $Hx = \pm \alpha e_1$
where $\alpha = \| x \|_2$, we can consider the following. We want
$Hx = (I-2 uu^T)x = x - 2u(u^Tx) = \pm \alpha e_1$ or that
$u = \frac{1}{2u^Tx}(x - \pm \alpha e_1)$.  Hence, we want a unit
vector in the direction $x - \pm \alpha e_1$ for one of the two
signs. We also note that this value of $\alpha$ is the only choice of
constant since if $Hx = \pm c e_1$ for some $c \in \R$ then thanks to
$H$'s norm preservation property we have
$\| x \| = \| H x \| = \| c e_1 \| = |c|$ or that
$c = \pm \| x \|_2 = \pm \alpha$. This is further
described in Explanation~\ref{explanation:householder-reflector}.

\begin{explanation}[label=explanation:householder-reflector]{}
Lets consider the positive sign case first,
i.e., $u = x + \| x \|_2 e_1$ with
$\bar u = u / \| u \|_2$ (see Fig.~\ref{fig:householderreflector}-left),
and fully derive why this works.
First, set $\alpha= \| x\|_2$ and
note that
\begin{align}
\| u \|_2^2
= ( x + \alpha e_1)^T ( x + \alpha e_1)
= x^T x + 2 \alpha x_1 + \| x\|^2
= 2( \alpha^2 + \alpha x_1).
\end{align}
Then, with $H = I - 2 \bar u \bar u^T$, we have
\begin{align}
Hx 
&=  \left[ I - \frac{2}{ \|  u \|_2^2} \left(x + \| x \|_2 e_1\right) \left(x + \| x \|_2 e_1\right)^T \right] x \\
&=  \left[ I - \frac{2}{ \|  u \|_2^2} \left(x +  \alpha  e_1\right) \left(x +  \alpha  e_1\right)^T \right] x \\
&= 
\left[ I - \frac{2}{ \|  u \|_2^2} \left( x x^T + \alpha x e_1^T + \alpha e_1 x^T + \alpha^2 e_1 e_1^T \right)\right] x \\
&= x - \frac{2}{ \|  u \|_2^2} 
\left( x \alpha^2 + \alpha x x_1 + \alpha e_1 \alpha^2 + \alpha^2 e_1 x_1 \right) \\
&= 
x - \frac{1}{ \|  u \|_2^2} 
\left( 2x (\alpha^2 + \alpha x_1) + 2\alpha e_1 (\alpha^2 + \alpha x_1) \right) \\
&= 
x - x - \alpha e_1 = -\alpha e_1
\end{align}
as desired.

For the negative sign case,
i.e., $ u = x - \| x \|_2 e_1$
with $\bar u = u / \|  u \|_2$ (see Fig.~\ref{fig:householderreflector}-right).
In this case, we note that
\begin{align}
\|  u \|_2^2
= ( x - \alpha e_1)^T ( x - \alpha e_1)
= x^T x - 2 \alpha x_1 + \| x\|^2
= 2( \alpha^2 - \alpha x_1).
\end{align}
Then we have
\begin{align}
Hx 
&=  \left[ I - \frac{2}{ \|  u \|_2^2} \left(x - \| x \|_2 e_1\right) \left(x - \| x \|_2 e_1\right)^T \right] x \\
&=  \left[ I - \frac{2}{ \|  u \|_2^2} \left(x - \alpha e_1\right) \left(x - \alpha e_1\right)^T \right] x \\
&= 
\left[ I - \frac{2}{ \|  u \|_2^2} \left( x x^T - \alpha x e_1^T - \alpha e_1 x^T + \alpha^2 e_1 e_1^T \right)\right] x \\
&= x - \frac{2}{ \|  u \|_2^2} 
\left( x \alpha^2 - \alpha x x_1 - \alpha e_1 \alpha^2 + \alpha^2 e_1 x_1 \right) \\
&= 
x - \frac{1}{ \|  u \|_2^2} 
\left( 2x (\alpha^2 - \alpha x_1) - 2\alpha e_1 (\alpha^2 - \alpha x_1) \right) \\
&= 
x - x + \alpha e_1 = \alpha e_1
\end{align}
as desired.
\end{explanation}

If we only care about finding $H$ so that $Hx = c e_1$ for 
$c \in \R$ (which necessarily means $c = \pm \| x \| = \pm \alpha$) but
the sign of $c$ does not matter mathematically for a problem, then
in practice it is typical to take
$u = x + \text{sgn}(x_1) \| x \|_2 e_1$ and then
$\bar u = u / \| u \|_2$. The use of $\text{sgn}(x_1)$ helps
numerical stability by avoiding excessive floating point errors associated with
cancellation of the first element of $u$ for certain $x$. For example, otherwise, if, say
$x \approx -e_1$, we could get $u \approx -e_1 + e_1 = 0$ for the wrong sign. Indeed,
with this sign correction, we have
that $\| u \|^2 = 2 \| x \| ( \| x \| + |x_1|)$
which begs the question, why always project onto the first axis? If instead
we project onto axis $j$ where $|x_j|$ is largest, then we would have
$\| u \|^2 = 2 \| x \| ( \| x| + |x_j|)$ which is larger and 
thus more numerically stable when forming $\bar u = u / \| u \|$. 
Hence, in practice, we use $j \in \argmax_i |x_i|$ and 
$u = x + \text{sgn}(x_j) \| x \|_2 e_j$.
Again, 
mathematically any choice of sign will work, as we saw above, to get $Hx = c e_j$ for
$c \in \pm \alpha$ and some $j$.

\subsubsection{Another property: post Householder reflector multiplied matrices}
\label{sec:anoth-prop-refl}

Suppose we are given a positive definite (PD) matrix $B = Q\Lambda Q^T \in \R^{m \times m}$ where
$Q$ is an orthonormal matrix and $\Lambda = \text{diag}(\lambda_i)$ is a diagonal
matrix of non-negative eigenvalues. Suppose also that $r$ (with $1 < r \leq m$) of the eigenvalues are the same
(i.e., there is a multiplicity of $r$ for some eigenvalue $\lambda$).
We permute and partition the columns of $Q$ and permute $\Lambda$
so that $Q=(Q_1,Q_2)$ with $Q_1 \in \R^{m \times r}$ 
and where the first $r$ eigenvalues in $\Lambda$ correspond to 
those duplicate eigenvalues. Thus,
\begin{align}
\Lambda = 
\begin{pmatrix}
\lambda I_r & 0 \\
0 & \Lambda_{r+1:m}
\end{pmatrix}
\end{align}
where $\lambda$ is
the common eigenvalue and
$\Lambda_{r+1:m}$ is the lower 
right $(m-r) \times (m-r)$ block of $\Lambda$ (which is diagonal),
and $I_r$ is the $r \times r$ identity matrix.
We
next replace $Q_1$ with $\bar Q_1 = Q_1 H^T$ with $H$ being any $r \times r$
Householder reflector. Then we have:
\begin{itemize}
\item $\bar Q_1^T \bar Q_1 = H Q_1^T Q_1 H^T = H H^T = I_r$
\item $\bar Q_1^T Q_2 = H Q_1^T Q_2 = 0$.
\end{itemize}
Thus, $\bar Q_1$ still consists of a set of orthonormal column vectors
all of which are still orthogonal to $Q_2$.
We now replace $Q$ with $\bar Q = (\bar Q_1, Q_2) = Q \bar H$
where $\bar H$ is the block matrix defined by
\begin{align}
\bar H = \begin{pmatrix}
H & 0 \\
0 & I_{m-r}
\end{pmatrix}
\end{align}
with $I_{m-r}$ the $(m-r) \times (m-r)$ identity matrix,
and hence $\bar Q^T \bar Q = \bar Q \bar Q^T = I_m$.

Now, considering the matrix $\bar Q^T \Lambda \bar Q$. We get
\begin{align}
\bar Q \Lambda \bar Q^T 
&= Q \bar H \Lambda \bar H^T Q^T \\
&= Q
\begin{pmatrix}
  H & 0 \\
  0 & I 
\end{pmatrix}
\Lambda
\begin{pmatrix}
  H & 0 \\
  0 & I 
\end{pmatrix}
Q^T \\
&=  Q
\begin{pmatrix}
  H & 0 \\
  0 & I 
\end{pmatrix}
\begin{pmatrix}
\lambda I_r & 0 \\
0 & \Lambda_{r+1:m}
\end{pmatrix}
\begin{pmatrix}
  H & 0 \\
  0 & I 
\end{pmatrix}
Q^T \\
&=  Q
\begin{pmatrix}
  \lambda H^2  & 0 \\
  0 & \Lambda_{r+1:m} 
\end{pmatrix}
Q^T
=  Q
\begin{pmatrix}
  \lambda I_r  & 0 \\
  0 & \Lambda_{r+1:m} 
\end{pmatrix}
Q^T  \\
&= Q \Lambda Q^T = B
\end{align}
Thus, whenever there is a multiplicity of eigenvalues of $B$, we
are free to replace the eigenvectors
associated with them with a post Householder reflector multiplied
version of them while maintaining a valid spectral factorization
of $B = Q \Lambda Q^T = \bar Q^T \Lambda \bar Q$. This property
is used in~\Cref{sec:case-i-2} when solving the secular equation
for the case of duplicate eigenvalues.

It should be noted that the above property used only the
fact that $H$ was an orthonormal matrix, and thus the same property holds for any
orthonormal matrix $H$ (not just a Householder reflector) and thus we have a
much larger class of transformations that can be applied to the eigenvectors
associated with duplicate eigenvalues without changing the spectral factorization of $B$.
However, as mentioned above, with a Householder reflector we
also can form it so that $Hx = \pm \| x \| e_1$ for any $x$ which
will be useful below in~\Cref{sec:case-i-2}.

\subsection{Factored (Eigen) Decomposition $B = U A U^T$ and rank-1 updates}
\label{sec:fact-eigen-decomp}

Any symmetric matrix $B$ (of size $m\times m$ and rank $r \leq m$) can be
represented in a factored form 
\begin{align}
B = U A U^T,
\end{align}
where $U$ is an $m\times r$ matrix with orthonormal columns and $A$ is
an $r\times r$ symmetric matrix.  In general, $U$ need not be the
matrix of eigenvector columns (which is discussed below); rather, it can be any
orthonormal basis spanning the column space (range) of $B$, with
$A = U^T B U$ capturing the representation of $B$ in that basis.  This
representation is obtained by choosing $U$ as \textit{any} orthonormal
basis for the range of $B$, and defining $A = U^T B
U$. Because $U^T U = I_r$ and because $UU^T$ is a projection matrix
into the column-subspace spanned by $B$, given any $z \in \text{span}(B)$ (i.e.,
$z$ in the column space of $B$)
we have $z = UU^Tz$ meaning that $UU^T$ acts like an identity for such $z$. In
particular, $UU^TB = B$.  Also, $A$ is symmetric whenever
$B$ is symmetric follows since
$A^T = (U^T B U)^T = U^T B^T U = U^T B U = A$.  Also, this $A$
recovers $B$ since
$B =
UU^TB =
{(UU^TB)}^T =
B UU^T =
U U^T B U U^T =
U A U^T$.

If $B$ is full rank ($r=m$) and $U$ is chosen as the matrix of
eigenvectors of $B$, then $A$ becomes diagonal (the matrix of
eigenvalues), recovering the standard eigen-decomposition
$B = Q \Lambda Q^T$ as a special case where $Q$ is a matrix with
columns as eigenvectors. In fact, any time an $m \times m$ matrix $B$
is in the form $Q \Lambda Q^T$ when $Q$ is a matrix of $m$ orthonormal
vectors and $\Lambda$ is an $m \times m$ diagonal matrix, it means that the values
of $\Lambda$ on the diagonal are eigenvalues for the corresponding
columns of $Q$ eigenvectors.  The reason is, with $q_i$ being the
$i^\text{th}$ column of $Q$, we have
\begin{align}
Q \Lambda Q^T
= \sum_{i=1}^m \lambda_i q_i q_i^T
\label{eq:diagonalization_eigen_decomposition}
\end{align}
and thus 
\begin{align}
Q \Lambda Q^T q_j
= \sum_{i=1}^m \lambda_i q_i q_i^T q_j = \lambda_j q_j
\label{eq:diagonalization_eigen_decomposition_eigenvectors}
\end{align}

If now instead we have $B = Q \Lambda Q^T$ where $Q \in \R^{m \times r}$
with $r < m$ and $\Lambda$ is an $r \times r$ diagonal matrix, but
still the $r$ columns of $Q$ are orthonormal, it means the rank of
$Q \Lambda Q^T$ is now only $r$. Also, the columns of $Q$ are no
longer called eigenvectors, and the diagonal entries $\Lambda$ are no
longer called eigenvalues, rather the columns of $Q$ and the
diagonal entries $\Lambda$ are called the \emph{principal components}
and \emph{principal values} (or eigenspectrum or \emph{nonzero
  spectral components}) of the rank-$r$ symmetric matrix $B$, and
correspond to its nonzero eigenpairs.  Sometimes in this context,
people refer to the decomposition $B = Q \Lambda Q^T$ as a spectral
decomposition, or eigendecomposition, or diagonalization of a low-rank
symmetric PSD matrix $B$, but the term ``eigenvector'' for the columns of
$Q$ only strictly applies when $Q$ is full rank.  The columns of $Q$
in the low rank case do span the invariant subspace of $B$ associated
with its nonzero eigenvalues, so they still carry eigenvalue
structure, just not the full spectrum,
and equations such as
Eqn.~\eqref{eq:diagonalization_eigen_decomposition}
and~\eqref{eq:diagonalization_eigen_decomposition_eigenvectors} with $r<m$
and summing from $1$ to $r$ still hold.
More generally still,
any (complex valued) matrix $B$ having $B\ctrans{B} = \ctrans{B} B$ 
(where $\ctrans{B}$ is the conjugate transpose of $B$) 
is
called a normal matrix (which includes real-valued symmetric
matrices and Hermitian matrices, important
for Von Neumann (i.e., quantum) entropy as 
shown in~\Cref{sec:von-neumann-entropy}). A matrix is diagonalizable
iff it is normal. Going forward, even if $r < m$ we abuse
terminology and 
with $B = Q \Lambda Q^T$ refer to the columns of $Q$ as
eigenvectors with corresponding eigenvalues in $\Lambda$.

Importantly, $B=U A U^T$ and $A \in \R^{r \times r}$ (with $A$ being full
rank $r$) share the same non-zero eigenvalues even when $A$ is not
diagonal.
This follows from the fact that $U A U^T$ and $A$ have the same rank
$r$, and we next show that the spectrum of $U A U^T$ consists of the $r$
eigenvalues of $A$ plus $m-r$ zeros. 
We can see via a spectral decomposition of $A$: we represent
$A = Q \Lambda Q^T$ with $\Lambda \in \R^{r \times r}$ and
$Q \in \R^{r \times r}$ with $Q$ a matrix of orthogonal column
vectors. If $A$ is already diagonal, $Q=I_r$.  We thus have the
representation $B = (UQ) \Lambda {(UQ)}^T = \bar Q \Lambda {\bar Q}^T$
with $\bar Q = UQ$ and then we note that $\bar Q \in \R^{m \times r}$ is a matrix of $r$
orthonormal columns which follows since
${\bar Q}^T \bar Q = {(UQ)}^T UQ = Q^T U^T U Q = Q^T Q = I_r$.  Thus,
we see that $A$'s eigenvalues, namely the diagonal matrix $\Lambda$,
are the same as $B$'s eigenvalues, which follows from
Eqn.\eqref{eq:diagonalization_eigen_decomposition}.  If we wish, when
$r < m$, we can always extend $\Lambda$ to an $m \times m$ matrix by
padding $m-r$ zeros on the lower right and also concatenating zero
column-vectors to the right of $\bar Q$.

\subsubsection{Rank-1 updates}
\label{sec:rank-1-updates}

Suppose now we have a matrix $B$ of rank $r$ with $0 \leq r \leq m$
and we represent $B$ as $B = Q \Lambda Q^T$ where
$Q \in \R^{m \times r}$ and $\Lambda \in \R^{r \times r}$ are the
non-zero spectral components. We next perform a rank-1 update and form
$\tilde B = B + \rho u u^T$ where $u \in \R^m$ is a column vector and
$\rho \in \R$. 

We define 
$u_\parallel = Q Q^T u$ 
and
$u_\perp = (I - QQ^T)u = u - QQ^T u = u - u_\parallel$
and thus we have the decomposition $u =  u_\perp + u_\parallel$.
Intuitively, $u_\perp$ and $u_\parallel$ should be orthogonal vectors
which follows since:
\begin{align}
\langle u_\parallel, u_\perp \rangle 
&=  (QQ^Tu)^T(I - QQ^T)u \\
&=u^TQQ^Tu - u^T Q Q^T Q Q^T u \\
&= u^T QQ^T u - u^T Q Q^T u = 0
\end{align}

In this section, we assume $u_\perp = 0$.
This could happen either because $r=m$ (i.e., $B$ is full rank
and so $Q$'s columns is a set of $m$ orthonormal vectors, and
$QQ^T = Q^TQ = I_m$)
or because $u$ is in the column space of $B$ (in which
case $u = QQ^Tu$). We discuss the case where 
$u_\perp \neq 0$ 
(i.e., $\rank(\tilde B) = \rank(B)+1$, which must mean $r < m$)
in
Section~\ref{sec:case-b-psd}.

When $\rho > 0$, we have $\rank(\tilde B) = \rank(B)$
(since $(I - QQ^T)u = 0$).
When $\rho < 0$ rank might decrease, we discuss this further below
in Section~\ref{sec:downdate-with-rho}.
In either case, we get
\begin{align}
\tilde B&= Q\Lambda Q^T + \rho uu^T = Q\Lambda Q^T + \rho Q Q^T uu^TQ Q^T \\
  &= Q( \Lambda + \rho Q^T uu^TQ ) Q^T = Q( \Lambda + \rho vv^T ) Q^T
\end{align}
where $v = Q^Tu \in \R^r$ is the
representation of vector $u$ in the $r$-dimensional Eigenbasis $Q$.

Now, the spectral components (eigenvalues if $r=m$) of $\tilde B$ are
the spectral components of $\Lambda + \rho vv^T$ which is a rank-1
update to a diagonal $r \times r$ matrix $\Lambda$.  That is, if
$B = Q \Lambda Q^T$ then there is a factorization
$\tilde B = \tilde Q \tilde \Lambda {\tilde Q}^T$ where
$\tilde \Lambda$ is the diagonal matrix of eigenvalues
$\{ \tilde \lambda_i \}_i$ of both $\tilde B$ and
$\Lambda + \rho vv^T$, and $\tilde Q$ are the eigenvectors of
$\tilde B$ but not of $\Lambda + \rho v v^T$. While this follows from
the above, justification in this specific case can be elucidating,
and is further discussed in Explanation~\ref{expln:updated_eigenvalues_eigenvectors}.

\begin{explanation}[label=expln:updated_eigenvalues_eigenvectors]{}

To see this, consider the following:
First, suppose that $(x,\lambda)$ is an eigenpair for $\tilde B$ so $\tilde B x = \lambda x$. Then 
set $y = Q^Tx$. We have
\begin{align}
\lambda x = \tilde B x = Q( \Lambda + \rho vv^T ) Q^T x
 = Q( \Lambda + \rho vv^T ) y.
\end{align}
Premultiplying both sides by $Q^T$ gives
\begin{align}
\lambda Q^T x = \lambda y = ( \Lambda + \rho vv^T ) y
\end{align}
and hence $\lambda$ is an Eigenvalue of $( \Lambda + \rho vv^T )$ 
with Eigenvector $y = Q^Tx$.

Conversely, suppose $(y,\lambda)$ is an eigenpair for
$\Lambda + vv^T$ with $y \in \R^r$,
so
$(\Lambda + \rho vv^T)y = \lambda y$.
We take $x = Qy$ which is in the subspace spanned by $Q$.
For any such $x$, $Q Q^T$ acts like an identity
(i.e., $Q Q^T x = Q Q^T Qy = Qy = x$)
and also $Q^T x = Q^T Q y = y$.
Then 
\begin{align}
  \tilde Bx = Q( \Lambda + \rho vv^T ) Q^T x
      = Q( \Lambda + \rho vv^T ) y
      = \lambda Q y
      = \lambda Q Q^T x = \lambda x
\end{align}
so that $\lambda$ is an Eigenvalue for $\tilde B$ with Eigenvector $x$.

Notice that the Eigenvectors of $\Lambda + \rho vv^T$
are related to the Eigenvectors
of $\tilde B$ through $Q$ (the Eigenvectors of $B$), so that if 
\begin{align}
 \tilde B = Q( \Lambda + \rho vv^T ) Q^T  = Q(U \tilde \Lambda U^T)Q^T
 = \tilde Q \tilde \Lambda {\tilde Q}^T
\end{align}
where $U$ is the orthonormal Eigenvectors of 
$\Lambda + \rho vv^T$ with diagonal matrix $\tilde \Lambda$ we can
get the Eigenvectors of $\tilde B$ via $\tilde Q = QU$. 
Indeed,
we have
\begin{align}
{\tilde Q }^T \tilde Q 
&= {(QU)}^T QU \\
&= U^T Q^T Q U \\
&= U^T U = I,
\end{align}
meaning $\tilde Q$ is an orthonormal matrix of eigenvectors of $\tilde B$.
Note that this dense matrix-matrix multiply $QU$ is an $O(m^3)$ operation
in the worst case but it can be made to run very fast.

\end{explanation}

In much of the below, we will assume that the eigenvalues $\Lambda$ of
$B$ are sorted ascending $\lambda_1 \leq \lambda_2 \leq \dots$ and the
columns of $Q$ are sorted correspondingly. Once an update occurs to
give $\tilde B = \tilde Q \tilde \Lambda { \tilde Q}^T$ things will
need to be resorted.

\section{Fast Rank-1 Updates and Downdates: Positive and Semidefinite}
\label{sec:fast-rank-1-updates-and-downdates}

In this section, we fully derive and then prove that
it is possible to perform gain computations for matrix
spectral functions via fast rank-1 updates via the
secular\footnote{The reader might wonder why this
is called the "secular" equation. The reason is
due to an early use of the term regarding
"belonging to the world"
and planetary motion. The equation
refered to in this paper
is distantly related to those concepts.
There is a good listing for this term in the Oxford English Dictionary.} 
equation. We derive these results from first principles,
and we also give a self-contained proof of the secular equation
and the associated eigenvalue properties of rank-1 updates,
including when reduction and deflation (defined below) is necessary.

We start with an $m \times m$ symmetric real-valued matrix
$B \in \R^{m \times m}$ that is either positive definite (PD) or
positive semi-definite (PSD). It is possible that $B$ could be an all
zero matrix, but this is fine since in this case it is still PSD. 
In general, $B$ comes from the inner product of a row-submatrix 
of some design matrix $\data$. I.e., $B = \data[X]^T \data[X]$ where $\data \in \R^{n \times m}$
and where $\data[X]$ is a $|X| \times m$ row submatrix (corresponding to
rows $X \subseteq V$ of $\data$ with $V=[n]$) of some larger design matrix $\data$.
We presumably have the list of Eigenvalues of $B$ precomputed (which if
$B=0$ is just a list of zeros).

Our goal is to quickly get the updated Eigenvalues to the matrix
$\tilde B = B + \rho uu^T$ where $u \in \R^m$ is non-zero $m \times 1$ real
valued vector, and we need to do this repeatedly for many different
possible $u$ vectors, check on these updated eigenvalues, and based on
this check, decide on a final $u$, commit to that $u$, and then perform a final update
$B \gets B + \rho uu^T$ where $u$ is the final $u$ selected from the set of
possible $u$ values. This is the core matrix part of
the greedy algorithm in Algorithm~\ref{alg:greedy-max} when 
operating on any matrix spectral function.

Critically, these checks and the final update needs to be fast, no
more than $O(m^2)$, meaning that both the check on the Eigenvectors of
$\tilde B = B + \rho uu^T$ and the update needs to be $O(m^2)$.  In other words,
the step of checking on the updated Eigenvalues of $\tilde B$ and then doing
the final update must not be $O(m^3)$ (which would be the approach of
just recomputing the Eigenvalues from scratch from, say, a general Hermitian
matrix eigen solver).

One fact about the eigenvalues of a rank one update is that
the sum of the new minus the sum of the old eigenvalues is
the norm squared of $u$. Indeed, recall for
PSD matrix $B$ that $\tr{B} = \sum_i \lambda_i$
where $\lambda_i \geq 0$ are the eigenvalues of $B$,
while $\det{B} = \prod_i \lambda_i$. Therefore, if
$\tilde \lambda_i$ are the eigenvalues of $\tilde B = B + \rho uu^T$ then
\begin{align}
\sum_i \tilde \lambda_i
- \sum_i \lambda_i = \tr{\tilde B} - \tr{B} = \sum_i \rho u_i^2 = \rho \| u \|^2 > 0
\label{eqn:incremental_eigenvalue_energy}
\end{align}
Thus, when performing a rank-1 update on a PSD matrix (i.e., with non-negative
eigenvalues), and when $\rho > 0$ and $\| u \| \neq 0$, the total eigenvalue ``energy''
can only increase. Furthermore, we see below in Theorem~\ref{thm:eigenvalue_updates}
that each individual eigenvalue can never decrease, meaning
that $\lambda_i \leq \tilde \lambda_i$ when $\rho > 0$. 
In fact, the theorem says for $1 \leq i < m$,
$\lambda_i \leq \tilde \lambda_i  \leq \tilde \lambda_{i+1}$.
When $\rho < 0$
the opposite is true. Also, the above means we have a bound on the
new largest eigenvalue, i.e., specifically we have:
\begin{align}
\max_i \tilde \lambda_i
= \tilde \lambda_m
\leq \lambda_m + \rho \| u\|^2
= \max_i \lambda_i + \rho \| u\|^2.
\end{align}
This bound is useful for computing numerical thresholds in practice.

While we derive the case for arbitrary $\rho$, in practice
we will only need $\rho = \pm 1$. 

Also, when seen as a matrix spectral set function, the linear
function (which is concave) applied to the eigenvalues
gives a modular (i.e., additive) matrix spectral function and
the contribution of $u$ is the same regardless of the context $B$
into which it is added.

We first consider $\rho > 0$. Our approach is to compute the new
eigenvalues of $\tilde B$ by maintaining a factored form of $B$.
There are two broad cases, when $B$ is already PD and so has $m$
positive Eigenvalues and when $B$ is only PSD and so has fewer than
$m$ (and possibly even no) positive Eigenvalues. Also, in the second
case, when $B$ is only PSD, a rank-1 update might preserve $B$'s rank
(rank preserving) or might increase $B$'s rank by one (rank
increasing).  We consider each case separately, starting with the
simpler case of when $B$ is already PD. Then we consider the PSD case
and the two subcases of rank-preserving and rank-increasing rank-1
update (we'll see that these subcases relate back to the PD case).

After this is all done, we consider $\rho < 0$ in
Section~\ref{sec:downdate-with-rho}.

\subsection{Case A: $\rho > 0$, $B$ is already PD.}
\label{sec:case-a-when}

If $B$ is already PD, then it is full rank with $m$ positive
Eigenvalues. We then we maintain a spectrally factored 
$B = Q\Lambda Q^T$ where $Q$ is an $m \times m$ orthonormal matrix (columns
are orthonormal Eigenvectors) and
$\Lambda = \text{diag}(\lambda_1, \lambda_2, \dots, \lambda_m)$ is a diagonal matrix of 
Eigenvalues of $B$ where $\lambda_i > 0, \forall i$. Thus,
$I_m = Q^T Q = QQ^T$.

Next, lets make the update.
We represent $\tilde B = Q\Lambda Q^T + \rho uu^T$ with $\rho > 0$.
We get
\begin{align}
\tilde B&= Q\Lambda Q^T + \rho uu^T = Q\Lambda Q^T + \rho Q Q^T uu^TQ Q^T \\
  &= Q( \Lambda + \rho Q^T uu^TQ ) Q^T = Q( \Lambda + \rho vv^T ) Q^T \\ 
  &= Q(\tilde U \tilde \Lambda {\tilde U}^T) Q^T = \tilde Q \tilde \Lambda {\tilde Q}^T
\end{align}
where $v = Q^Tu$ is the
representation of vector $u$ in the Eigenbasis $Q$ (and where $u = Qv$),
where $\tilde U \tilde \Lambda {\tilde U}^T = \Lambda + \rho vv^T$ is the spectral
factorization of the rank-1 update the diagonal,
and where $\tilde Q = Q \tilde U$.
As we saw
above, the Eigenvalues of $\tilde B$ are $\tilde \Lambda$ which 
are the Eigenvalues of $\Lambda + \rho vv^T$ (a rank-1 update to a diagonal matrix $\Lambda$). The eigenvectors 
(up to a scalar multiple of $-1.0$)
of $\tilde B$ are $\tilde Q$.
We need not worry if any of the eigenvalues are multiplied
by $-1.0$ for reasons mentioned in Section~\Ref{sec:extr-prel}
and since
\begin{align}
\tilde Q \tilde \Lambda {\tilde Q}^T
= \sum_{i=1}^m \tilde \lambda_i \tilde q_i {\tilde q_i}^T
= \sum_{i=1}^m \tilde \lambda_i (-\tilde q_i) {(-\tilde q_i)}^T
\end{align}

We need to develop two processes
corresponding to the greedy
algorithm in Algorithm~\ref{alg:greedy-max}.
First, we need to quickly compute and
{\bf test the Eigenvalues} of the rank-1 update $\tilde B$ but
preserve $B$, and its factored form, for other tests.  Second, once we have tested multiple
$u$ vectors, we want, for a selected $u$, to quickly {\bf commit the
  update} $\tilde B \gets B + \rho u u^T$, destroying the old $B$. We
consider each in turn.

\paragraph{Scaled eigenvector case}
\label{sec:scal-eigenv-case}

Before we begin, consider the case that $u$ is in precisely the
direction of one of the eigenvectors, say the $j^\text{th}$
eigenvector. Then $u = \| u \| q_j$ and $v = Q^T u = e_j \|u \| = e_j \alpha$
where $\alpha = \| u\|$.  The
update becomes $\Lambda + \rho \alpha^2 e_j e_j^T = \tilde \Lambda$ where
$\lambda_i'=\lambda_i$ for $i \neq j$ and
$\lambda_j' = \lambda_j + \rho \alpha^2$. We
thus have $\tilde B = B + \rho u u^T = Q\tilde \Lambda Q^T$ which is still
in eigenvalue/eigenvector form just with an updated eigenvalue.
This case, therefore is particularly
easy, especially since the eigenvectors stay the same, we simply need
to check for the case where $v = e_j \|u \|$ for some $j$.  In the below,
therefore, we assume that $u$ is not parallel with one of the
eigenvectors, meaning $v \neq e_j \|u \|$ for any $j$. We note that
the reduction approach, described below, for the more general case when
one or more zeros are present in $v$ can also handle this case,
albeit more expensively.

\subsubsection{Case A-I: $\rho > 0$, $B$ PD, test Eigenvalues of $B+uu^T$}
\label{sec:case-i:-h}

Here we compute and test a particular $u$ rank-1 update, but we do not
update $B$ or its factored form. Our goal is to, as quickly as possible, find the
Eigenvalues of $\tilde B = B + uu^T$ which is the same as the
eigenvalues of $\Lambda + v v^T$ where $v = Q^T u$.

\begin{figure}
\centering
\includegraphics[width=0.49\textwidth]{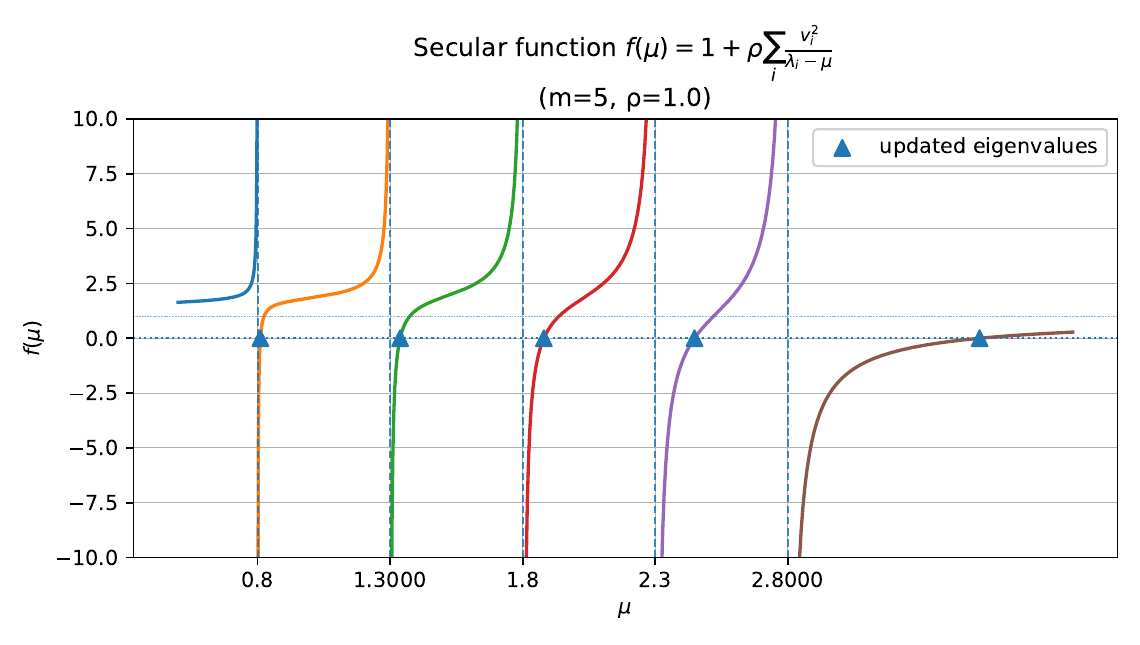}\includegraphics[width=0.49\textwidth]{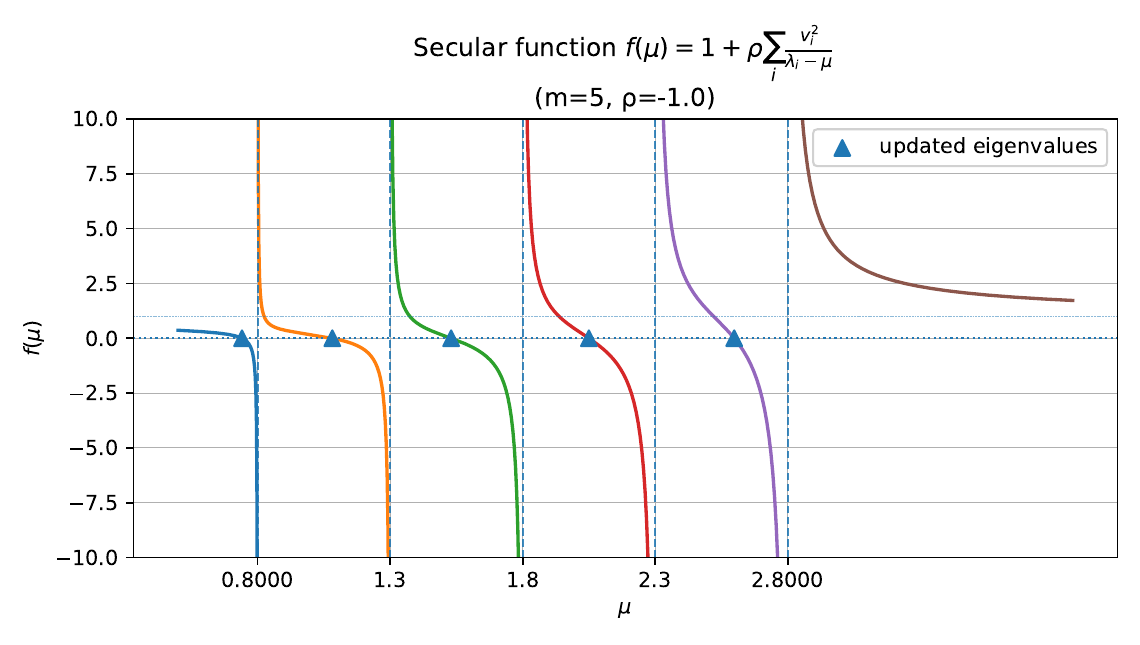}
\caption{Example plot of the secular function $f(\mu)$ with
  $\rho = +1$ (on the left) and $\rho=-1$ (on the right) where the
  vertical dashed line shows the existing $m=5$ eigenvalues and
  the blue triangles show the updated eigenvalues corresponding to the
  roots of $f(\mu)$. The $\rho=+1$ case corresponds to a rank-1 update
  $\Lambda + v v^T$ and the plot shows that the function is monotone
  increasing (between the vertical asymptotes, and between the right
  most (resp.\ left most) vertical asymptote and $+\infty$ (resp.\ $-\infty$)), while
  the $\rho=-1$ case corresponds to a rank-1 downdate
  $\Lambda - v v^T$ and the plot shows the function is monotone
  decreasing within each segment.  }
\label{fig:secular_equation_plot}
\end{figure}

To find the Eigenvalues of $\Lambda + \rho vv^T$, we use the
characteristic equation which says that
\begin{align}
p(\mu) = \det(\Lambda + \rho v v^T - \mu I) = 0
\end{align}
for any new Eigenvalue $\mu$. We get the Eigenvalues by solving for
the roots of $p(\mu)$. We then use the matrix determinant lemma~\cite{ding2007eigenvalues}
which
says that
\begin{align}
\det(A + x y^T) 
= \det\left(A(I + A^{-1}xy^T)\right)
= \det(A) (1 + y^T A^{-1} x).
\end{align}
We set $A = \Lambda - \mu I$ and $x=y=\sqrt{\rho}v$ to give
\begin{align}
p(\mu) 
&= \det( (\Lambda - \mu I) (I + \rho (\Lambda - \mu I)^{-1} v v^T)) \\
&= \det(\Lambda - \mu I) \left(1 + \rho v^T {(\Lambda - \mu I)}^{-1} v \right) \\
&=  \det(\Lambda - \mu I)f(\mu) \\
&= \left[ \prod_j (\lambda_j - \mu)\right] f(\mu)
\end{align}
where the second factor $f(\mu)$ has the form
\begin{align}
f(\mu)
=\left(1 + \rho v^T {(\Lambda - \mu I)}^{-1} v \right)
 = 1 + \rho \sum_{i=1}^m \frac{v_i^2}{\lambda_i - \mu}.
\label{eqn:secular_equation}
\end{align}
This equation, a nonlinear sum of rational components, is known as the {\bf secular
  equation}~\cite{demmel1997applied,melman1997numerical,NICHITA2014574,brechenmacher2014lagrange}.
A plot of this equation is shown in
Figure~\ref{fig:secular_equation_plot} .  We note that $f(\lambda_i)$
is undefined for all $i$ since, for $\rho > 0$,
$\lim_{\mu \uparrow \lambda_i} f(\mu) = +\infty$ while
$\lim_{\mu \downarrow \lambda_i} f(\mu) = -\infty$ (for $\rho < 0$ the
reverse is true).  Hence, each of the original eigenvalues are
vertical asymptotes (sometimes called poles) of $f(\cdot)$.  For the $\rho > 0$ case, the
function is everywhere, between the asymptotes, increasing (the
figure also shows $\rho < 0$ case where the function is everywhere, between the asymptotes,
decreasing).  We can see this by considering the derivative of the
secular equation
\begin{align}
f'(\mu) = \rho \sum_{i=1}^m \frac{v_i^2}{(\lambda_i - \mu)^2}
\end{align}
and see that $f'()$ is either everywhere else (other than the vertical asymptotes)
always positive or always negative depending on the sign of $\rho$.

Right away, we can see that the roots of the characteristic equation
partially depend on the roots of the secular equation.  That is, in the
$\rho > 0$ case, and on the open interval $(\lambda_j, \lambda_{j+1})$
(assuming $\lambda_j < \lambda_{j+1}$), as $\mu \downarrow \lambda_j$
from the right, we have that
$\frac{v_j^2}{\lambda_j - \mu} \to -\infty$ and thus dominates the
equation, while if $\mu \uparrow \lambda_{j+1}$ from the left,
$\frac{v_{j+1}^2}{\lambda_{j+1} - \mu} \to \infty$ and thus dominates,
and so there must be a single (since the gradient is everywhere
positive in the interval) zero crossing in the interval. In the
$\rho > 0$ case there is one more zero to the right of the right-most
vertical asymptote, while if $\rho < 0$ there is one more zero to the left of the
left-most vertical asymptote.

But in order for the roots to entirely (or only partially) depend on the secular equation,
we need to further analyses, hence the secular equation (and how to
solve it) is further discussed below.  What we will do in fact depends
also on the $v$ vector as well as the existing eigenvalues
$\Lambda$.

\paragraph{Case A-I-1: $v_i \neq 0, \forall i$ and $\lambda_i$ all distinct}
\label{sec:case-i-1}

Recall that $v$ is the eigenbasis space representation of $u$, i.e.,
$v = Q^T u$, and we want the eigenvalues of $\tilde B = B + u u^T$ which
are the same as the eigenvalues of $\Lambda + v v^T$.

In this first case, we assume that $v_i \neq 0, \forall i$ and
the existing eigenvalues $\lambda_i$ all distinct. When considering $p(\mu)$ we notice that the
first factor has $\det(\Lambda - \mu I) = 0$ for $\mu = \lambda_i$ for
any $i$ since $\det(\Lambda - \mu I) = \prod_j (\lambda_j - \mu)$,
which makes sense since $\lambda_i$ are Eigenvalues of $\Lambda$. For the
second factor, $\lim_{\mu \to \lambda_i} f(\mu)$ does not
exist for any $i$, 
but when
we multiply them together, we get:
\begin{align}
p(\mu) = \prod_j (\lambda_j - \mu)
+ \rho \sum_{k=1}^m v_k^2 \prod_{j \neq k} (\lambda_j - \mu)
\end{align}
and we see that not only $p(\lambda_i)$ is defined
but
\begin{align}
p(\lambda_i)
= \rho \sum_{k=1}^m v_k^2 \prod_{j \neq k} (\lambda_j - \lambda_i) 
=
\rho v_i^2 \prod_{j \neq i} (\lambda_j - \lambda_i)
\neq 0
\end{align}
since the eigenvalues are all distinct and $v_i \neq 0$. This 
means that none of the original eigenvalues are still
eigenvalues of $\Lambda + \rho v v^T$, in this case,
and we must find $m$ new eigenvalues.

To find the new eigenvalues, we find zeros of $p(\mu)$. Since none of
the original eigenvalues are the same, the new eigenvalues never have
$\prod_j (\lambda_j - \mu) = 0$ and hence the zeros of $p(\mu)$ are
the same as the roots of $f(\mu)=0$, i.e., we need to find all of the
values $\mu$ such that the secular equation is zero.

It is easiest to consider $f(\mu)$ when the original eigenvalues are sorted
ascending, i.e. $\lambda_1 < \lambda_2 < \dots < \lambda_m$ where
the inequality strictness follows since the eigenvalues are distinct,
and the columns of $Q$ are correspondingly sorted.
Away from these vertical asymptotes, the function is everywhere
increasing (when $\rho > 0$, as mentioned above).
We also note that $\lim_{\mu \to -\infty} f(\mu) = 1$
which means that there are no roots to the left of $\lambda_1$,
and also  $\lim_{\mu \to \infty} f(\mu) = 1$
which, along with strict increasing from the right-most vertical asymptote,
means there is one root to the right of $\lambda_m$, and there
are $m$ total roots.

Hence, the roots of $f(\mu)$ (and the new eigenvalues
$\{\tilde \lambda_i \}_i$) have the relationship
\begin{align}
\lambda_1 < \tilde \lambda_1 < 
\lambda_2 < \tilde \lambda_2 < 
\dots < 
\lambda_m < \tilde \lambda_m
\end{align}
This means that while each eigenvalue $\lambda_i$ increases (and can never
decrease for any rank-1 update), it
does not increase by too much as we have
the relationship $\lambda_{i} < \tilde \lambda_i < \lambda_{i+1}$.
The right-most new eigenvalue $\lambda_m$ can increase but not unboundedly
since recall above Eqn~\eqref{eqn:incremental_eigenvalue_energy} that
the total incremental eigenvalue energy is bounded by $\rho \| u \|^2$.
This is further discussed in~\cite{golub1973some,thompson1976behavior}.

This is a special case of the following more general theorem which 
also holds in the PSD case.
The general theorem is proven 
in~\cite{wilkinson1965algebraic,thompson1976behavior,,demmel1997applied}
and also in~\cite{golub1973some,bunch1978rank} (where it is particularly
clear)
although consideration together of the above and what follows further below
(treating the PSD case) also constitutes a complete proof.
\begin{restatable}[Eigenvalue updates for rank-1 updates]{rthm}{eigenvalueupdates}
\label{thm:eigenvalue_updates}
  Let $C = \Lambda + \rho v v^T$ where $\Lambda$ is a diagonal matrix,
  $\| v \| = 1$. Let
  $\lambda_1 \leq \lambda_2 \leq \dots \leq \lambda_m$ be the
  eigenvalues of $\Lambda$ and let
  $\tilde \lambda_1 \leq \tilde \lambda_2 \leq \dots \leq \tilde \lambda_m$ be the
  eigenvalues of $C$. Then we have
  $\tilde \lambda_i = \lambda_i + \rho p_i$ for $i \in [m]$ where
  $\sum_{i=1}^m p_i = 1$ and $0 \leq p_i \leq 1$ (i.e., $p=(p_1, \dots, p_m)$ is a
  distribution), and moreover we
  have that if $\rho > 0$ that
\begin{align}
\lambda_1 \leq \tilde \lambda_1 \leq 
\lambda_2 \leq \tilde \lambda_2 \leq 
\dots \leq 
\lambda_m \leq \tilde \lambda_m \leq \rho \| v \|^2 + \lambda_m
\end{align}
while if $\rho < 0$ that
\begin{align}
\rho \| v^2 \| + \lambda_1
\leq \tilde \lambda_1 \leq  \lambda_1 \leq
\tilde \lambda_2 \leq  \lambda_2 \leq 
\dots \leq 
\tilde \lambda_m \leq  \lambda_m
\end{align}
And if the eigenvalues $\lambda_i$ are all distinct and all of the
elements of $v$ are non-zero, then the eigenvalues of $C$ strictly
separate those of $\Lambda$ (meaning $p_i > 0, \forall i$ and all
the above inequalities are strict).
\end{restatable}

In the above, if $\| v \| \neq 1$ we can replace $v$ with 
$\bar v = v / \| v \|$ and set $\bar \rho \gets \rho \| v \|^2$, 
so that $\bar \rho \bar v \bar v^T = \rho v v^T$.
Thus, $\| v \|^2$ is like the energy of $v$ that
gets introduced with the rank-1 update, and that gets re-distributed
over (or removed from, in the $\rho < 0$ case) the eigenvalues via $p$ with
$\bar \rho p_i$ being the update to the $i^\text{th}$ eigevalue,
where $0 \leq p_i \leq 1$ and $\sum_i p_i = 1$. This
is also consistent
with $\tr{\Lambda + \rho v v^T} - \tr{\Lambda} = 
\sum_i (\tilde \lambda_i - \lambda_i) = \sum_i p_i \bar \rho = \rho \| v \|^2$.

The key takeaway is that finding the updated eigenvalues means finding
the (when the eigenvalues are distinct and $v_i \neq 0$) $m$ roots of $f$. 
We show below that this can be done in $O(n^2)$
time but first consider cases where some of the $v_i$ might be zero and/or some 
of the $\lambda_i$ might not be distinct (and later the PSD case) which is
more involved.

\paragraph{Case A-I-2: $\rho > 0$, some $v_i = 0$ and/or not all
  $\lambda_i$ distinct: deflation}
\label{sec:case-i-2}

In the case where some of the elements of $v$ might be zero, or if some of
the existing Eigenvalues $\Lambda$ have a multiplicity of more than one,
we need to do things slightly differently. It turns out, however,
that these two cases are treated quite similarly. 

We first address the case where some of the elements might be zero.
Recall $v = Q^Tu$ so if $v_i = 0$ this means that
$\langle q_i, u \rangle = 0$ and since $Q$ is full rank, this must
mean either $u =0$ (which we do not consider)
or $u$ has no component in the direction of $q_i$. This therefore
is a generalization of the case we considered in
Section~\ref{sec:scal-eigenv-case},
in that Section $|\{ i : v_i = 0 \}| = m-1$ while
in this Section $|\{ i : v_i = 0 \}| < m-1$.

For any $i$ with $v_i = 0$, then the eigenvalue at $i$ is preserved.
We can
see this immediately. 
Let $(\lambda_i, e_i)$ be an eigenpair of $\Lambda$. Then
\begin{align}
(\Lambda + \rho v v^T) e_i
= \Lambda e_i  + \rho v v^T e_i
= \lambda_i e_i + \rho v v_i
= \lambda_i e_i
\end{align}
and hence $(\lambda_i, e_i)$ is also an eigenpair of $(\Lambda + \rho v v^T)$.

Hence, we can construct a sub-matrix of $\Lambda$
corresponding only the non-zero $v_i$s and solve only the
sub-problem for the new eigenvalues, and copy over the old eigenvalues
for all cases where any $v_i = 0$.

Interestingly, the case of repeated eigenvalues is quite similar.
Let $\Lambda$ be the diagonal matrix of eigenvalues and assume not 
all are distinct. 
Suppose that there is a batch of eigenvalues that are the same with
multiplicity $r_\ell$, and lets say that $E_\ell \subset [m]$ is a
size $|E_\ell|=r_\ell$ subset of integers for which
$\lambda_i = \lambda_j = \lambda_{(\ell)}$ for all $i,j \in E_\ell$
for each $\ell$.  In the below, we do this for one $E = E_\ell$ with
$r=r_\ell$ and avoid the $\ell$ subscript, but note that the process
is done for each such batch of identical eigenvalues.
Also, let the complement of $E$, i.e., $E_c = [m] \setminus E$.

Recall $B = Q \Lambda Q^T$
and we wish the eigenvalues
of $\tilde B = B + u u^T$.
We permute the indices.
and also permute
the corresponding columns of $Q$,
so that the
first $r$ columns (i.e., eigenvectors of) $Q$ have eigenvalue $\lambda$
so that $Q=(Q_E,Q_{E_c})$ where $Q_E$ is the
$r$ eigenvectors corresponding to the $r$ repeated eigenvalues.

Within each of the two blocks the order doesn't matter, as long as the
ordering is consistent amongst the columns of $Q$ and the eigenvalues
--- this partial within $E$ order irrelevance will be
addressed again below.

We considered $u$ in the eigenbasis $Q$ as the  $v$ vector
\begin{align}
v = Q^T u = 
\begin{pmatrix}
Q_E^T u \\
Q_{E_c}^T u
\end{pmatrix}
= 
\begin{pmatrix}
v_E \\
v_{E_c}
\end{pmatrix}
\end{align}
where $v_E \in \R^{r}$
and $v_{E_c} \in \R^{m-r}$.
We next create an $r \times r$ reflector matrix $H$ 
(see Section~\ref{sec:elem-refl-matr})
so that
$H v_E = \alpha e_j$ for $|\alpha| = \| v_E \|$ 
and with $e_j = (0,\dots,1,0,\dots,0) \in \R^{r}$ (with the 1 in the $j^\text{th}$ position).
The sign of $\alpha$ does not
matter for computing the eigenvalues
(see for example the use of $\alpha_\ell$
in Equation~\eqref{eq:final_deflation_update} below).
However, for the purposes
of producing a Householder reflector,
we select the numerically preferred method mentioned
in Section~\ref{sec:anoth-prop-refl}. That is,
we set $H = I - 2 \bar w {\bar w}^T$ where
$w = v_E + \text{sign}(v_{E}(j)) \| v_E \| e_j$ 
with $j \in \argmax_{i \in [r]} |v_E(i)|$
and $\bar w = w / \| w \|$
(where we take $\text{sign}(0) = 1$
and 
where $v_E(j)$ is the $j^\text{th}$ element of $v_E$).
With this choice of $w$, and for some vector $x$,
we have that $H v_E = -\text{sign}(v_E(j)) \| v_E \| e_j$
as per Explanation~\ref{explanation:householder-reflector}
in Section~\ref{sec:anoth-prop-refl}. On
the other hand,
for computing the eigenvectors in
the reduced rank-1 problem 
$\Lambda + \bar v {\bar v}^T$ (see below), 
we need to make sure that any such
sign change is reversed. In other
words, if we have computed $\bar v$
with sign flips for the Householder
blocks below, we need to use
the negated sign flip when computing
the reduced rank-1 eigenvector problem eigenvectors, so that the 
final eigenvectors of $\tilde B$ are correct.
This subtle point is often not mentioned in the 
literature but is important to keep in mind in practice.

We then replace $Q$ with $\bar Q = (\bar Q_E, Q_{E_c}) = Q \bar H$
where $\bar Q_E = Q_E H = Q_E H^T$ which, as mentioned
in Section~\ref{sec:anoth-prop-refl},
preserves $B = Q \Lambda Q^T = \bar Q \Lambda \bar Q^T$
and $\bar Q^T \bar Q = \bar Q \bar Q^T = I_m$. 
In particular,
we have $\bar Q_E^T u = H Q_E^T u = H v_E = \alpha e_1$ and so
\begin{align}
\bar v= \bar Q^T u
=
\begin{pmatrix}
\bar Q_E^T u \\
Q_{E_c}^T u
\end{pmatrix}
= 
\begin{pmatrix}
\alpha \\
0 \\
\vdots \\
0 \\
v_{r+1} \\
\vdots \\
v_m
\end{pmatrix}
\; \begin{array}{l}
\left.\vphantom{\begin{matrix}
\alpha \\ 0 \\ \vdots \\ 0
\end{matrix}}\right\} \; r \\ \left.\vphantom{\begin{matrix}
v_{r+1} \\ \vdots \\ v_m
\end{matrix}}\right\} \; m - r
\end{array}
\end{align}
Thus, what we have done is found an
easy to obtain alternate decomposition of $\tilde B$ where
\begin{align}
\tilde B &= Q \Lambda Q^T + u u^T
= Q (\Lambda + v v^T) Q^T \\
&= \bar Q \Lambda \bar Q^T + \bar Q \bar Q^T u {(\bar Q \bar Q^T u )}^T
= \bar Q (\Lambda + \bar v \bar v^T) \bar Q^T \\
&= Q \bar H (\Lambda + \bar v \bar v^T) \bar H^T Q^T \label{eq:final_deflation_update}
\end{align}
where all of the repeated eigenvalues of $\Lambda$ but
one have a 0 in the corresponding $\bar v$ position and hence
those eigenvalues stay the same. The one remaining repeated 
eigenvalue has a non-zero of value 
$\alpha = \pm \|v_E\|$.

The order of elements in
$E$ does not matter, and this means that when we form $\bar Q_E = Q_E
H$, the column that will correspond to the non-zero entry in $\bar
v$ is also irrelevant as long as it corresponds to one of the repeated
eigenvalues. This means that any one of the repeated eigenvalues can
be chosen to be updated, and the remainder of the repeated eigenvalues
will be retained.  We can do this for each bunch
$E$ for which an eigenvalue has a multiplicity greater than one. 
Importantly, we still note that there is a change to the
eigenvector basis according to the Householder reflector
$H$ and its block extension $\bar H$. If there are multiple
groups of repeated eigenvalues, then 
$\bar H$ will be a combination of a block diagonal matrix
(where each block is a corresponding Householder reflector) as well as an identity matrix.
That is, when there are $L$ groups of repeated eigenvalues, $\bar H$
will have the form
\begin{align}
\bar H =
\begin{pmatrix}
H_1 & 0 & \dots & 0 & 0 \\
0 & H_2 & \dots & 0 & 0 \\
\vdots & \vdots & \ddots & \vdots & \vdots \\
0 & 0 & \dots & H_L & 0 \\
0 & 0 & \dots & 0 & I_{ \bar r}
\end{pmatrix}
\end{align}
where each $H_\ell$ is a Householder reflector for the $\ell^\text{th}$ group of repeated eigenvalues
and $I_{\bar r}$ is the identity matrix of appropriate 
dimension $\bar r = m - \sum_{\ell =1}^L |E_\ell|$ for the non-repeated eigenvalues
and where $\bar r = \bar r_\text{z} + \bar r_\text{nz}$ where 
$\bar r_\text{z}$ is the number of zero $v_i$ and 
$\bar r_\text{nz}$ is the number of non-zero $v_i$ for the non-repeated eigenvalues.
We note that if there is any group $\ell$ for which all corresponding
$v_i$ are zero, then we can skip the Householder reflector for that group 
and just use the identity matrix for that block since all of those eigenvalues will be preserved.

Once we have $\Lambda + \bar v {\bar v}^T$ as
in the form of Equation~\eqref{eq:final_deflation_update},
we rearrange/permute the positions so that the $\bar r_{\text{z}}$ zeros in the original
$v$, as well as the $\sum_{\ell=1}^L |E_\ell| - L$ new zeros corresponding to 
the repeated eigenvalue, are grouped together and placed at one end of the matrix.
The remaining positions correspond to the $\bar m = L + \bar r_{\text{nz}}$ non-zeros
corresponding to the updated eigenvalues, 
After this, we thus have a new sub-problem
with an $\bar m \times \bar m$ rank $\bar m$ matrix 
for which
there are no zeros in the rank-1 update vector and no repeat
eigenvalues, and we then use the method in Section~\ref{sec:case-i-1}.

We can also derive the zero $v_i$ and the repeat eigenvalue case 
very simply by looking
directly at the secular equation (Eqn.~\eqref{eqn:secular_equation})
repeated here for convenience: Recall
\begin{align}
f(\mu) =  1 + \rho \sum_{i=1}^m \frac{v_i^2}{\lambda_i - \mu}.
\end{align}
We derive the more general case for $L \geq 0$ blocks of repeated eigenvalues
$\{ E_\ell \}_{\ell \in [L]}$ each with
repeated eigenvalues with common value $\lambda_\ell$. 
That is, 
$E_\ell = \{ i : \lambda_i = \lambda_\ell 
\text{ and } v_i^2 \neq 0 \}$ with $|E_\ell| > 1$ for each $\ell$.
Also, let $E^c = [m] \setminus ( \bigcup_{\ell=1}^L E_\ell )$.
We then have
\begin{align}
\label{eqn:alt_deflation_derivation}
f(\mu) &=  1 + \rho 
  \sum_{\ell = 1}^L \sum_{i \in E_\ell} \frac{v_i^2}{\lambda_i - \mu}.
+  \rho \sum_{i \in E^c} \frac{v_i^2}{\lambda_i - \mu} \\
&=  1 + \rho 
  \sum_{\ell = 1}^L \sum_{i \in E_\ell} \frac{v_i^2}{\lambda_\ell - \mu}.
+  \rho \sum_{i \in E^c} \frac{v_i^2}{\lambda_i - \mu}  \\
&=  1 + \rho 
  \sum_{\ell =1}^L \frac{1}{\lambda_\ell - \mu} \sum_{i \in E_\ell} v_i^2
+  \rho \sum_{i \in E^c} \frac{v_i^2}{\lambda_i - \mu}  \\
&=  1 + \rho 
  \sum_{\ell=1}^L \frac{\alpha^2_\ell}{\lambda_\ell - \mu} 
+  \rho \sum_{i \in E^c} \frac{v_i^2}{\lambda_i - \mu}  \\
&=  1 + \rho 
  \sum_{\ell=1}^L \frac{\alpha_\ell^2}{\lambda_\ell - \mu} 
+  \rho \sum_{i \in E^c : v_i  \neq 0} \frac{v_i^2}{\lambda_i - \mu} 
\label{eq:reduced_secular}
\end{align}
where, like above, $\alpha_\ell = \| v_{E_\ell} \| = \sqrt{\sum_{i \in E_\ell} v_i^2}$.
When we consider $p(\mu) = \det(\Lambda - \mu I)f(\mu)$ we notice
that all of the terms $i$ above that have been excluded due
to repeat eigenvalues or due to zero valued of $v_i$ give 
$p(\lambda_i) = 0$ due to the first factor, meaning
that those eigenvalues are retained, and we have
a reduced secular equation Eqn.~\eqref{eq:reduced_secular} 
involving only $k < m$ terms to solve
to get the new eigenvalues where 
$k=|\{ \ell \in [L]: \alpha_\ell \neq 0\}| +|\{ i \in E^c : v_i \neq 0 \}|$.

In the repeat eigenvalue case, after the rank-1 update, the
multiplicity of each repeat eigenvalue decreases by one (unless it
happens that $u$ is in precisely the direction of one of the
eigenvectors). This is also related to the PSD case where when we have
some zero-valued eigenvalues which might be repeated, and a rank-1 update
can reduce the number of zero eigenvalues by one (i.e., all but one of the
zero eigenvalues stays the same). However, there is more work
to deal with the PSD case which we turn to 
in Section~\ref{sec:case-b-when}.

\subsubsection{Case A-II: $B$ PD, eigenvectors, commit $\tilde B = B+uu^T$}
\label{sec:case-ii:-b}

At some point, we will need to commit the update and compute the new
matrix $\tilde B = B+uu^T$. However, since we are maintaining a
factored form of $B = Q \Lambda Q^T$, we would prefer to directly
produce a factored form of
$\tilde B = \tilde Q \tilde \Lambda {\tilde Q}^T$ since we do 
not wish first to form a monolithic $\tilde B$ and 
then pay the $O(n^3)$ cost to factor it from
scratch. That is, we want to go directly from $Q \Lambda Q^T$ to
$\tilde Q \tilde \Lambda {\tilde Q}^T$ with a process having $O(n^2)$ cost. We already
have discussed how to get the eigenvalues $\tilde \Lambda$. In this
section we talk about how to get the updated eigenvectors $\tilde Q$.

In fact, we've already encountered the solution
in Explanation~\ref{expln:updated_eigenvalues_eigenvectors} to get the updated
eigenvectors $\tilde Q$ of $\tilde B$ but that relied on having the
eigenvectors $U$ of $\Lambda + \rho v v^T$. Getting this $U$ is
formalized in the following theorem.
\begin{restatable}[Lemma 5.2 in~\cite{demmel1997applied}]{rthm}{demmel52}
\label{thm:demmel52}
If $\mu$ is an eigenvalue of $\Lambda + \rho v v^T$ then ${(\Lambda - \mu I)}^{-1}v$ is
its eigenvector. Since $\Lambda - \mu I$ is diagonal, the cost to compute
this eigenvector is $O(m)$, and the cost is $O(m^2)$ to do this for all eigenvalues.
\end{restatable}
\begin{proof}
We have the following:
\begin{align}
(\Lambda + \rho v v^T)[{(\Lambda - \mu I)}^{-1}v] 
   &= (\Lambda - \mu I + \mu I + \rho v v^T){(\Lambda - \mu I)}^{-1} v \\
   &= v + \mu{(\Lambda - \mu I)}^{-1} v + v[ \rho v^T {(\Lambda - \mu I)}^{-1} v ] \\
   &= v + \mu{(\Lambda - \mu I)}^{-1} v - v \\
   & \qquad \text{since $\rho v^T{(\Lambda - \mu I)}^{-1} v + 1 = f(\mu) = 0$} \\
   &= \mu [{(\Lambda - \mu I)}^{-1} v ] \text{ as desired. }
\end{align}
\end{proof}
With $\tilde Q$ the updated eigenvectors of $\tilde B = B + \rho u u^T$, 
and $U$ the updated eigenvectors of $\Lambda + \rho v v^T$, 
we have $\tilde Q = Q U$.
The $k^\text{th}$ column of $U$
has form $u^{(k)} = {(\Lambda - \tilde \lambda_k I)}^{-1}v$
which elementwise has form:
\begin{align}
u^{(k)}_j = \frac{ v_j } { \lambda_j - \tilde \lambda_k }
\end{align}
Unfortunately, according to~\cite{demmel1997applied}, this formula is not numerically
well behaved owing to problems that occur when there
are two new eigenvalues $\tilde \lambda_k$ and $\tilde \lambda_{k+1}$ that
are close to each other, leading to numerically non-orthogonal computed eigenvectors
$u^{(k)}$. Specifically, the issue is that the eigenvectors
should be orthonormal but this is numerically challenging
since ${(\Lambda - \tilde \lambda_k I)}^{-1}v$
and ${(\Lambda - \tilde \lambda_{k+1} I)}^{-1}v$ are
similar formulas whose difference lies only
in what might be a small difference between $\tilde \lambda_k$ and
$\tilde \lambda_{k+1}$.

Fortunately, there is a way to get numerically well-behaved updated
eigenvectors via the following theorem from Loewner.
\begin{restatable}[Loewner, from~\cite{demmel1997applied} page 224]{rthm}{lowner}
\label{thm:lowner}
Let $\Lambda = \text{diag}(\lambda_1, \lambda_2, \dots, \lambda_m)$ be a
diagonal matrix with $\lambda_m < \lambda_{m-1} < \dots < \lambda_1$. Let
$\tilde \lambda_m < \tilde \lambda_{m-1} < \dots < \tilde \lambda_1$ be given
real numbers that satisfy the following interlacing property:
\begin{align}
\lambda_m < \tilde \lambda_m < \dots < 
\lambda_{i+1} < \tilde \lambda_{i+1} < 
\lambda_i < \tilde \lambda_i < 
\dots < 
\lambda_1 < \tilde \lambda_1
\end{align}
Note all inequalities are all strict. Then there is a vector $\tilde v \in \R^m$ such that
the $\tilde \lambda_i$ values are the exact eigenvalues of 
$\hat \Lambda = \Lambda + \tilde v {\tilde v}^T$. The $i^\text{th}$ entry of $\tilde v$,
for $1 \leq i \leq m$, is given by:
\begin{align}
(\tilde v_i)^2 =
\frac{ \prod_{j=1}^m (\tilde \lambda_j - \lambda_i) }
    { \prod_{j=1, j \neq i}^m (\lambda_j - \lambda_i) }
\label{eqn:lowner}
\end{align}
\end{restatable}
Note that the r.h.s.\ of Eqn~\eqref{eqn:lowner} is positive because
the number of negative factors in the numerator is the same as the
number of negative factors in the denominator, and hence we can take
the square root to get $\hat v_i > 0$.

Thus, as outlined by~\cite{gu1994stable,demmel1997applied}, the process to get the updated
eigenvectors is as follows:

\begin{enumerate}
\item Solve the secular equation (Eqn.~\eqref{eqn:secular_equation})
  to find the roots of $f(\mu) =  1 + \rho \sum_{i=1}^m \frac{v_i^2}{\lambda_i - \mu}$
  which are $\tilde \Lambda = \text{diag}(\tilde \lambda_1, \dots, \tilde \lambda_m)$.
\item Use the above Loewner's theorem to compute $\tilde v$ according to
  Eqn~\eqref{eqn:lowner} so that 
  $(\tilde \lambda_1, \dots, \tilde \lambda_m)$ are the exact
  eigenvalues of $\Lambda + \tilde v {\tilde v}^T$
\item Use Theorem~\ref{thm:demmel52} to compute the eigenvectors $U$
  of $U \tilde \Lambda U^T = \Lambda + \tilde v {\tilde v}^T$ (note, no $\rho$ here).
\item If we need to form the factorization $\tilde B = \tilde Q \tilde \Lambda {\tilde Q}^T$,
  we form $\tilde Q = QU$ as mentioned in Section~\ref{sec:rank-1-updates}.
\end{enumerate}
It is shown in~\cite{demmel1997applied} that,
while $\Lambda + \rho v v^T \approx \Lambda + \tilde v {\tilde v}^T$
and even though we use the same
Theorem~\ref{thm:demmel52} on $\Lambda + \tilde v {\tilde v}^T$,
this procedure is numerically
much better than just using Theorem~\ref{thm:demmel52} 
directly on $\Lambda + \rho v v^T$.

Now, to apply the above and get the eigenvectors when all $\lambda_i$
are distinct, we just apply the results of the above directly.  Also,
when some $u_i = 0$ for some $i$ and/or not all $\lambda_i$ are
distinct, we use the deflation procedure of Section~\ref{sec:case-i-2}
to reduce the eigenvalues to a subset of eigenvalues, yielding a
sub-problem where all eigenvalue are distinct. We then apply the
results of the above on the deflated problem to update some of the
eigenvalues/eigenvectors.

\paragraph{Deflation and Eigenvector Reconstruction}
\label{sec:defl-eigenv}

Theorems~\ref{thm:demmel52} and~\ref{thm:lowner} refer to the
situation in the full $m$ dimensions, but suppose deflation 
as described in Section~\ref{sec:case-i-2}
has
happened and reduced the problem to having only $\bar m < m$ distinct
eigenvalues. This means that in the deflated case the resulting
eigenvectors in Eqn.~\eqref{eqn:lowner} are only $\bar m$-dimensional
rather than $m$ dimensional. The question is, how do these $\bar m$ newly
updated eigenvalues and their corresponding $\bar m$ new $\bar m$-dimensional
eigenvectors combine with the existing preserved non-deflated
$m$-dimensional eigenvectors?

The solution stems from considering the original eigenvectors of
$\Lambda$
in the expression
in Equation~\eqref{eq:final_deflation_update}
repeated here
as $\tilde B = Q \bar H (\Lambda + \bar v \bar v^T) \bar H^T Q^T$.
Here, $\Lambda = I_m \Lambda I_m^T$ where $I_m$ is the
$m \times m$ identity matrix and $\Lambda$
is an $m \times m$ diagonal matrix of eigenvalues,
implying that $e_i$ is the $i^\text{th}$
eigenvector which makes sense since $\Lambda$ is diagonal. 
We then reduce $\Lambda$ and $\bar v$ to 
the $\bar m \times \bar m$ diagonal matrix
$\Lambda_{\bar m}$ and length $\bar m$ vector $\bar v_{\bar m}$ by removing the 
rows/columns corresponding to:
(1) the zero $\bar v_i$ values; and (2) 
for each repeated-eigenvalue group,
all but one representative coordinate
after the Householder reflector $\bar H$ step 
used to get $\bar v$ from $v$.
After reduction, suppose
that $R_{\bar m} \tilde \Lambda R_{\bar m}^T = \Lambda_{\bar m} + \bar v_{\bar m} {\bar v_{\bar m}}^T$ is the
spectral factorization computed above for the deflated problem where
$R_{\bar m} \in \R^{\bar m \times \bar m}$ 
and let $r_\ell \in \R^{\bar m}$ be the
$\ell^\text{th}$ column of $R_{\bar m}$, for $\ell \in [\bar m]$.  Suppose also that
$J = (j_1, j_2, \dots, j_{\bar m}) \subseteq [m]$ is a length-$\bar m$ ordered 
set of active 
indices after the Householder $\bar H$
transformation,
where $j_\ell \in [m]$ for $\ell \in [\bar m]$ that are used in the
deflated problem ---
for non-repeated eigenvalues, these
correspond to original indices, while for repeated eigenvalue
groups they are the chosen active coordinates
after the Householder reflector step.
Let $\bar J = [m] \setminus J$ be the indices of the
preserved eigenvalues/eigenvectors. Hence, for any $i \in \bar J$, the
preserved eigenvector is still $e_i \in \R^m$.

Suppose $j_\ell \in J$ for some $\ell$. We define the expanded reconstructed
eigenvector $\tilde q_{j_\ell} \in \R^m$,
with position $\tilde q_{j_\ell}[s]$ for $s \in [m]$,
as
\begin{align}
\tilde q_{j_\ell}[s]
=
\begin{cases}
r_\ell[t] & \text{ if } s = j_t, t \in [\bar m] \\
0         & \text { else }.
\end{cases}
\end{align}
That is, 
starting with a zero vector,
we fill in a subset of elements of $\tilde q_{j_\ell}$ with the corresponding
elements from $r_\ell$.
We can think of this as zero-padded, or ``lifted'' vector 
$\tilde q_{j_\ell}$ from $r_\ell$. 
Then whenever $\| r_\ell  \| = 1$ (which is assured from the
above), we also have $\| \tilde q_{j_\ell} \| = 1$.
Moreover,  $\langle \tilde q_{j_\ell}, e_i \rangle = 0$ for all $i \neq j_\ell$ since
for such $j_\ell$, $\tilde q_{j_\ell}[i]=0$. 

Hence, we can construct the original
eigenvectors by, starting with the identity $I_m$, replacing each column
$j_\ell \in J$ of $I_m$ with $\tilde q_{j_\ell}$ to get the updated $m \times m$ matrix
of eigenvectors $\tilde Q$ giving $\Lambda + \bar v {\bar v}^T = \tilde Q \tilde \Lambda \tilde Q^T$ 
where $\tilde \Lambda$ is the diagonal $m \times m$ matrix of updated 
and original eigenvalues. This
gives $\tilde B = Q \bar H \tilde Q \tilde \Lambda \tilde Q^T \bar H^T Q^T$ as the updated 
factorization of $\tilde B$ and the final updated eigenvalues
are given by $\tilde \Lambda$ and the final
updated eigenvectors being $Q \bar H \tilde Q$. This is further
described in Explanation~\ref{expln:deflation_eigenvector_reconstruction}.

\begin{explanation}[label=expln:deflation_eigenvector_reconstruction]{}

As a simple example,
let $m=7$, $L=2$,
$E_1 = \{1,2,3\}$, $E_2 = \{4,5\}$,
and $E^c = \{6,7\}$ and so $\bar m = 4$, which
means $J = \{ 1,4, 6,7\}$ and $\bar J = \{2,3,5\}$.
Then $\bar H = \text{blkdiag}(H_1, H_2, 1, 1)$
with $H_1 \in \R^{3 \times 3}$ and $H_2 \in \R^{2 \times 2}$
and in the reduced space
$R_{\bar m} \in \R^{4 \times 4}$ with
coordinates $r_{ij}$.
The $7 \times 7$ block diagonal matrix $\bar H$ with
Householder reflectors blocks,
and lifted $7 \times 7$ matrix $\tilde Q$, have the form
{\tiny
\begin{align}
\!\!\!\!
\bar H
= \begin{pmatrix}
H_1 & 0 & 0 & 0 \\
0 & H_2 & 0 & 0 \\
0 & 0 & 1 & 0 \\
0 & 0 & 0 & 1
\end{pmatrix}
=
\begin{pmatrix}
h^{(1)}_{11} & h^{(1)}_{12} & h^{(1)}_{13} & 0 & 0 & 0 & 0\\
h^{(1)}_{21} & h^{(1)}_{22} & h^{(1)}_{23} & 0 & 0 & 0 & 0\\
h^{(1)}_{31} & h^{(1)}_{32} & h^{(1)}_{33} & 0 & 0 & 0 & 0\\
0 & 0 & 0 & h^{(2)}_{11} & h^{(2)}_{12} & 0 & 0\\
0 & 0 & 0 & h^{(2)}_{21} & h^{(2)}_{22} & 0 & 0\\
0 & 0 & 0 & 0 & 0 & 1 & 0\\
0 & 0 & 0 & 0 & 0 & 0 & 1
\end{pmatrix},
\quad\!\!\!\!
\tilde Q =
\begin{pmatrix}
r_{11} & 0 & 0 & r_{12} & 0 & r_{13} & r_{14}\\
0      & 1 & 0 & 0      & 0 & 0      & 0\\
0      & 0 & 1 & 0      & 0 & 0      & 0\\
r_{21} & 0 & 0 & r_{22} & 0 & r_{23} & r_{24}\\
0      & 0 & 0 & 0      & 1 & 0      & 0\\
r_{31} & 0 & 0 & r_{32} & 0 & r_{33} & r_{34}\\
r_{41} & 0 & 0 & r_{42} & 0 & r_{43} & r_{44}
\end{pmatrix}
\notag
\end{align}
\par}
and then $\bar H \tilde Q$ has the form
\begin{align}
\bar H \tilde Q
=
\begin{bmatrix}
h^{(1)}_{11}r_{11} & h^{(1)}_{12} & h^{(1)}_{13} & h^{(1)}_{11}r_{12} & 0 & h^{(1)}_{11}r_{13} & h^{(1)}_{11}r_{14}\\
h^{(1)}_{21}r_{11} & h^{(1)}_{22} & h^{(1)}_{23} & h^{(1)}_{21}r_{12} & 0 & h^{(1)}_{21}r_{13} & h^{(1)}_{21}r_{14}\\
h^{(1)}_{31}r_{11} & h^{(1)}_{32} & h^{(1)}_{33} & h^{(1)}_{31}r_{12} & 0 & h^{(1)}_{31}r_{13} & h^{(1)}_{31}r_{14}\\
h^{(2)}_{11}r_{21} & 0 & 0 & h^{(2)}_{11}r_{22} & h^{(2)}_{12} & h^{(2)}_{11}r_{23} & h^{(2)}_{11}r_{24}\\
h^{(2)}_{21}r_{21} & 0 & 0 & h^{(2)}_{21}r_{22} & h^{(2)}_{22} & h^{(2)}_{21}r_{23} & h^{(2)}_{21}r_{24}\\
r_{31} & 0 & 0 & r_{32} & 0 & r_{33} & r_{34}\\
r_{41} & 0 & 0 & r_{42} & 0 & r_{43} & r_{44}
\end{bmatrix}
\end{align}

\end{explanation}

\subsection{Solving the Secular Equation: *LAED4}
\label{sec:solv-secul-equat}

To find the updated eigenvalues, we need
to find the roots of the secular equation (Eqn.~\eqref{eqn:secular_equation}).

As we can see from the Figure~\ref{fig:secular_equation_plot}, within
each segment between eigenvalues, we need to find a root. In the
$\rho > 0$ case there is one more root to the right of the largest
existing eigenvalue.  In the $\rho < 0$ case there is one more root to
the left of the smallest existing eigenvalue. Each root is itself a
non-linear function that requires a solver, such as Newton's method.
Newton's method iteratively approximates the function by a linear
function whose zero point is easy to find. 
As we can clearly see, however, by looking at the
secular equation plot in Figure~\ref{fig:secular_equation_plot},
the secular function in each segment is not well approximated by the linear functions
that Newton's method uses.

One solution would use a bracketed Newton method, where we run
Newton's method in each open interval $(\lambda_j,\lambda_{j+1})$
starting with that interval as the bracket. If Newton ever takes us
out of that interval (which is quite possible since the secular
equation can get very flat in the interval), we would take the
mid-point of the current bracketed  region. If Newton does not leave the
interval, or even if it does and we use a midpoint based update, each
time we get an update we adjust the bracket's left/right limits (i.e.,
when $\rho > 0$, if the updated value is negative, we update the left
bracket while if the value is positive we update the right
bracket). This way the solution will eventually converge.

A better option that converges faster and that uses a smarter option
as implemented as *LEAD4 as part of
LAPACK~\cite{demmel1997applied,li1993solving,bunch1978rank}.  The
*LAED4 methods includes DLAED4 for double FP64, and SLAED4 for single
FP32 precision. This also includes *LASD4 which takes in and then
computes the updated singular values (square roots of the eigenvalues,
which is available as part of the LAPACK port to Python) while *LAED4
takes in and computes the updated eigenvalues directly. These methods
do not use Newton directly (and thus do not do a linear approximation)
but instead at each iteration approximate the secular function with
another surrogate function that has vertical asymptotes at the same
location of the current segment boundary, and then sets coefficients
to attempt to match the shape of the secular function within that
segment. Critically, once these coefficients are set, it is relatively
easy to find the root of the approximation, and this process iterates
and quickly finds a numerically good representation of the true root
of the secular equation. The details are described
in~\cite{bunch1978rank,li1993solving,demmel1997applied}.
Notably, *LAED4 does this in time only linear in the number
of eigenvalues, and thus calling it $m$ times takes $O(m^2)$ time.

We note that this routine takes in the eigenvalues $(\lambda_1, \lambda_2, \dots, \lambda_m)$ in strictly increasing
order which means the eigenvalues must be distinct but do not need to
be positive. Thus, if there is a zero eigenvalue it can occur only once.
This is used as a subroutine
also in the deflated case (where we might have a $v_i=0$ or repeated eigenvalue).

The critical thing about this is that this procedure is only $O(m^2)$
per rank-1 update which is necessary.  It should also be noted that
DLAED4 supports pre-shifting and pre-scaling of $\lambda$ and $v$
before DLAED4 is called, and then the results can be un-scaled and
un-shifted. This can be useful when solving the $\rho < 0$ case
below.

Once we get the updated Eigenvalues, if we wish to commit the rank-1
update, we can use an LAPACK DLAED3-like routine to get the updated Eigenvectors also in
$O(m^2)$ time, so we then come up with a new representation. In our
implementation, we did not use DLAED3 (since it has, for our purposes,
an unnecessary additional matrix multiplication) but instead implemented the process ourselves 
using Loewner's theorem and the formula for the updated eigenvectors as described in
Section~\ref{sec:case-ii:-b}.

\subsection{Case B: When $B$ is only PSD.}
\label{sec:case-b-when}

The second case occurs when building up $B$ and, during this time, $B$
is only PSD, meaning it is not full rank. $B$ is an $m \times m$
symmetric matrix and so it has rank $0 \leq r \leq m$, and so we
consider here the case when $r < m$. In fact, $B$ may start out as the
zero matrix with $r=0$. The LAPACK DLAED4 code mentioned above will not work
out-of-the-box here since it requires all distinct eigenvalues, but we
will develop a process where we can still use it with only $r < m$
positive distinct eigenvalues. Hence, the problem is, given a PSD $B$
and a rank-1 update $\tilde B = B + uu^T$, how to compute the updated
eigenvalues and eigenvectors efficiently (i.e., in $O(m^2)$ time and
memory).

When $B$ is only PSD, it is still possible to factor $B$ as
$B = Q \Lambda Q^T$ but where $\Lambda$ is a diagonal matrix with $r$
positive diagonal and $m-r$ zeros along the diagonal. Hence, we can
represent this problem as $B = Q \Lambda Q^T$ where now
$Q \in \R^{m \times r}$ matrix of orthonormal columns and
$\Lambda \in \R^{r \times r}$ matrix of positive eigenvalues where
$r < m$.

What is critical to note is that each rank-1 update that happens might
retain $B$'s rank or it might increase it by one. When building up $B$
incrementally, we need to check on if the new $u$ vector
maintains or increases the rank of $B$. At some point when
the rank of $B$ becomes $m$, we then switch to the PD case above in
Section~\ref{sec:case-a-when}, and since that case requires
maintaining $B$ as $B=Q\Lambda Q^T$ ($Q$ are eigenvectors and
$\Lambda$ eigenvalues), we here need to incrementally build $B$ in a way so
that once $r$ reaches $m$, it is ready for the PD case (meaning, we
avoid any one-time $O(m^3)$ eigen analysis computational cost to get the
representation necessary for the PD case).

Our approach is to maintain $B = Q\Lambda Q^T$ where $Q$ is an
$m \times r$ matrix of orthonormal columns and $\Lambda \in \R^{r \times r}$ is a diagonal
spectral matrix. Then when a new $u$ vector comes along, 
we form the decomposition 
$u =  u_\perp + u_\parallel$
where $u_\parallel = Q Q^T u$ 
and $u_\perp = (I - QQ^T)u = u - QQ^T u = u - u_\parallel$.
The vector
$u_\perp$ is a form of residual ``error'' vector.
Recall, $QQ^T u$ is a projection
of $u$ onto the $r$-dimensional basis given by $Q$, and $u_\perp$ is
the orthogonal vector having norm $\alpha$.  
We define $\alpha = \|u_\perp\|$.  If $\alpha > 0$ (or above some
threshold), we are in the rank-increasing case, while if $\alpha = 0$
we are in the rank-preserving case.

Our approach below works when $B$ is only PSD, and even works when $B$
starts out as the zero matrix, as $u_\perp$ will just equal $u$ in
that case.

\subsubsection{Case B-PSD-RankPreserving}
\label{sec:case-b-psd-2}

If $\alpha =0$ (or is near zero),
then the rank of $B$ would stay the same. In this case, $Q$ remains
$m \times r$, $\Lambda$ remains $r \times r$ and we develop a way to
update $Q$ and $\Lambda$ to reflect the update. In fact, this strategy
closely mimics the PD case in Section~\ref{sec:case-a-when} but will
apply it to difference $r \times r$ matrices.  We already have the
representation $B = Q \Lambda Q^T$ where $Q \in \R^{m \times r}$ and
$\Lambda \in \R^{r \times r}$ so we can form $v = Q^T u \in \R^r$ in
$\tilde B = Q(\Lambda + \rho v v^T)Q^T$ noting that $u = QQ^Tu$ and
compute the updated eigenvalues to $\Lambda + \rho v v^T$ and use the
same strategy already developed in Section~\ref{sec:case-a-when} but
for an $r \times r$ sub-system. Of course we may have to deal with
zeros in $v$ or repeated eigenvalues in $\Lambda$ but that is already
handled above. Therefore, this case is done.

\subsubsection{Case B-PSD Rank Increasing}
\label{sec:case-b-psd-1}

If $\alpha > 0$ (i.e., above a small threshold), then the rank of $B$ increases.

\paragraph{Case B-PSD-RankIncreasing-I: Base case with $r=0$.}

When $r=0$ we start out with an empty matrix $B=0$ and it is particularly
easy. We do the rank-1 update $\tilde B = B + u u^T = uu^T$ and
represent this as $Q \Lambda Q^T$ where $Q \in \R^{m \times 1}$
and $\Lambda \in \R^{1 \times 1}$, we set $\bar u = u/\| u \|$
and $\alpha = \|u\|$. Then we have that
\begin{align}
\tilde B = B + u u^T = \alpha^2 \bar u {\bar u}^T
\end{align}
so we just set $Q=\bar u \in \R^{m \times 1}$ and
$\lambda_1 = \alpha^2$ and $\Lambda$ is set to a $1 \times 1$ matrix.
Thus, we immediately get both the single eigenvalue $\lambda_1$ and
the single eigenvector $\bar u$.

\paragraph{Case B-PSD-RankIncreasing-I: $0 < r < m$, all eigenvalues distinct.}
\label{sec:case-b-psd}

As above, $Q \in \R^{m \times r}$ is a matrix of orthogonal column
eigenvectors, $\Lambda$ is a diagonal matrix, the representation
$B = Q \Lambda Q^T$ holds. Due to the rank increment, we wish to
quickly form
$\tilde B = B + uu^T = \tilde Q \tilde \Lambda \tilde Q^T$ with
eigenvalues $\tilde Q \in \R^{m \times (r+1)}$ and eigenvectors
$\tilde \Lambda \in \R^{(r+1)\times (r+1)}$.  The expression
$Q(\Lambda + \rho v v^T)Q^T$ alone is insufficient since $Q$ needs an
additional column and $\Lambda$ needs to become an
$(r+1) \times (r+1)$ matrix. We must expand the basis in $Q$ by one
direction.

As was done in Section~\ref{sec:rank-1-updates}, we have
$u_\perp = u - QQ^T u$ and thus
$u = u_\perp + QQ^Tu = u_\perp + u_\parallel = u_\perp + Qv$ where
$v = Q^Tu \in \R^r$
and $\langle u_\perp, u_\parallel \rangle = 0$.
We also define $\bar u_\perp = u_\perp/\alpha$ 
with $\alpha = \| u_\perp \|$
and
note that $\| \bar u_\perp \|=1$. 
Let $q_i$ be the $i^\text{th}$ column of
matrix $Q$ for $i = 1, 2, \dots, r$.
The vector $\bar u_\perp$ is used for a new
left-most column of $Q$ which we call
$\hat q_0 = \bar u_\perp$. Thus,
we have a new $m \times (r+1)$ matrix $\hat Q$ where
\begin{align}
\hat Q =
\begin{pmatrix}
\vert  & \vert & \vert &     & \vert \\
\bar u_\perp &  q_1 &  q_2 & \hdots &  q_r \\
\vert  & \vert & \vert &     & \vert
\end{pmatrix}
=
\begin{pmatrix}
\vert  & \vert & \vert &     & \vert \\
\hat q_0 & \hat q_1 & \hat q_2 & \hdots & \hat q_r \\
\vert  & \vert & \vert &     & \vert
\end{pmatrix},
\end{align}
where we have equated $\hat q_0 = \bar u_\perp$ and $\hat q_i = q_i$
for $0 \leq i \leq r$.  We immediately have
$\hat Q^T \hat Q = I_{r+1}$ and that
$\hat Q$ will become the new matrix of eigenvalues.

We also define
\begin{align}
\hat \Lambda = 
\begin{pmatrix}
0         &           &             &         &            \\
          & \lambda_1 &             &         &            \\
          &           & \lambda_2   &         &            \\
          &           &             &  \ddots &            \\
          &           &             &         &  \lambda_r \\
\end{pmatrix}
= \text{diag}(\hat \lambda_0, \hat \lambda_1, \dots, \hat \lambda_r)
\end{align}
where $\hat \lambda_0 = 0$ and $\hat \lambda_i = \lambda_i$ for
$1 \leq i \leq r$.  $\hat \Lambda$ is a new $(r+1)\times(r+1)$
diagonal matrix of eigenvalues, where we have retained the
relationship
$\hat \lambda_0 < \hat \lambda_1 \leq \dots \leq \hat \lambda_r$, and with
the first value being 0 to account for $B$ not spanning the direction
of $u_\perp$ (i.e., $u_\perp$ is in the null-space of $B$).
Thus, we have a representation of $B$ in the updated eigenspace
that includes (but has no component in) the new direction $u_\perp$:
\begin{align}
B = Q \Lambda Q^T = \sum_{i=1}^r \lambda_i q_i q_i^T
= \sum_{i=0}^r \hat \lambda_i \hat q_i {\hat q_i}^T = 
\hat Q \hat \Lambda {\hat Q}^T
\end{align}
Consider 
\begin{align}
{\hat Q}^T u &= {\hat Q}^T(u_\parallel + u_\perp) = {\hat Q}^Tu_\parallel + {\hat Q}^T u_\perp\\
&= 
\begin{pmatrix}
0 \\
\vert \\
v \\
\vert
\end{pmatrix}
+
\begin{pmatrix}
\alpha \\
\vert \\
\mathbf{0} \\
\vert 
\end{pmatrix} 
=
\begin{pmatrix}
\alpha \\
\vert \\
v \\
\vert 
\end{pmatrix} 
\triangleq \hat v \in \R^{r+1}
\end{align}
Where $v = Q^T u = Q^T u_\parallel \in \R^r$.
We thus have
$
\hat Q {\hat Q}^T u = \hat Q \hat v = \alpha \bar u_\perp + u_\parallel = u$
and
\begin{align}
\newcommand{\hvert}{\text{---}}
{\hat Q}^T(u_\parallel + u_\perp){(u_\parallel + u_\perp)}^T \hat Q
= \hat v {\hat v}^T
= \begin{pmatrix}
\alpha^2 & \hvert \; \alpha {v} ^T \hvert \\
\vert    &     \\
\alpha v        & v v^T \\
\vert    & \\
\end{pmatrix}
\end{align}

Thus,
\begin{align}
\tilde B 
&= B + \rho u u^T 
= \hat Q \hat \Lambda {\hat Q}^T + \rho \hat Q {\hat Q}^T u u^T \hat Q {\hat Q}^T \\
&= \hat Q(\hat \Lambda + \rho {\hat Q}^T u u^T \hat Q) {\hat Q}^T
= \hat Q(\hat \Lambda + \rho \hat v {\hat v}^T) {\hat Q}^T.
\end{align}
Thus $\tilde B$ has one new non-zero eigenvalue compared to $B$ as well
as the more usual perturbed original eigenvalues. 
This has the same form, namely a rank-1 update
$\rho \hat v {\hat v}^T$ of a diagonal matrix $\hat \Lambda$. We can
therefore use the approach described in Section~\ref{sec:case-a-when}
for the PD case which can be used as an $O(m^2)$ strategy both to get
the updated eigenvalues and eigenvectors of $\tilde B$. This moreover
applies in when elements of $\hat v$ are zero or there are repeated
eigenvalues in $\hat \Lambda$.

\subsection{Downdate with $\rho < 0$}
\label{sec:downdate-with-rho}

When $\rho < 0$ we have a rank-1 ``downdate'' rather than an update,
which we write implicitly as $\tilde B = B - \rho u u^T$. Firstly, we
make an assumption that $u$ and $\rho$ is such that
$\tilde B \succeq 0$ (i.e., $\tilde B$ is PSD) for any $u$ of
interest. This holds if we start with $B = \sum_u \rho u u^T$ and then
subtract off one at a time.  We can still construct
$\tilde B = Q (\Lambda - \rho v v^T) Q^T$ which converts this to a
diagonal minus rank-1 matrix. If any $v_i=0$ or if there are any
repeat eigenvalues, we can use the same approach to deflate the
problem as described in Section~\ref{sec:case-i-2} (The equations
starting at Eqn~\eqref{eqn:alt_deflation_derivation} makes this
particularly clear). Furthermore, in the PSD case if $B$ is of rank
$r$ then consider $Q \in \R^{m \times r}$ and
$\Lambda \in \R^{r \times r}$.  Hence, going forward we assume we have
a (possibly deflated) problem where all $v_i \neq 0$ and all
$r$ eigenvalues are distinct.

To get the downdated eigenvalues, we can not use LAPACK's *LAED4 directly since that
routine requires $\rho > 0$. However, if we consider a new problem
$\Lambda - \rho v v^T = - (-\Lambda + \rho v v^T)$ we can perform an
update to negative eigenvalues $-\Lambda$ to get a new result which we
then negate. In this case, the same approach mentioned in
Sections~\ref{sec:case-a-when} and~\ref{sec:case-b-when} can be used
to form $- U \tilde \Lambda U = -\Lambda + \rho v v^T$ (specifically,
what will be computed is $U$ and $-\tilde \Lambda$ which can be easily
converted to $\tilde \Lambda$). It is important in this case, while LAPACK's *LAED4 accepts
negative eigenvalues, since *LAED4
(Section~\ref{sec:solv-secul-equat}) requires the eigenvalues to be
sorted ascending, so the sort of the diagonal of $-\Lambda$ for the
downdate must be in the reverse of what we used when updating
$\Lambda$.

\paragraph{Possible rank reduction:} After a downdate occurs, it is possible that the rank
of $B$ decreases and this can be checked as it will be the case that
one of the eigenvalues $\tilde \lambda_i$ will hit zero. Note that
even if $\Lambda$ has repeat eigenvalues, only one of those
repetitions will be modified and hence only at most one of them might
hit zero. Hence, once $\tilde \Lambda$ is computed, we check for a new
zero and if so, we change $\tilde \Lambda$ from $r \times r$ to
$(r-1) \times (r-1)$ and similarly reduce the number of
eigenvectors. This latter step is done by first computing $U$, then
computing $\tilde Q = QU$ and removing the column of $\tilde Q$
corresponding to the eigenvalue that became zero. If none of the
eigenvalues become zero, then the rank is preserved.

\subsection{Rank $k > 1$ updates rather than Rank $k=1$ update}
\label{sec:rank-k-more-than-one}

So far, we have considered only rank 1 updates but what if the update
comes in a form of $U \in \R^{n \times k}$ where $k > 1$ is a set of
column vectors. Let $U = (u_1, u_2, \dots, u_k)$ as a series of column
vectors and we wish to perform $\tilde B = B + \rho U U^T$.

Given the above, the easiest way to handle this is do a series of $k$
rank-1 updates, each time updating both the eigenvalues and
eigenvectors before doing the next update. In other words, we would
first form $v_1 = Q^Tu_1$ and then get
$\tilde B = \tilde Q \tilde \Lambda {\tilde Q}^T$.  Lets call this
\begin{align}
\tilde B  \triangleq \tilde B_1 = \tilde Q_1 \tilde \Lambda_1 {\tilde Q_1}^T
\end{align}
Then we would form $v_2 = Q^T_1 u_2$ and form
$\tilde B_2 = \tilde Q_2 \tilde \Lambda_1 {\tilde Q_2}^T$,
and in general form 
$v_{t} = Q^T_{t-1} u_{t}$ and to form 
$\tilde B_t = \tilde Q_t \tilde \Lambda_t {\tilde Q_t}^T$,
and when $t=k$ we are done. This overall takes $O(km^2)$.

Another option might be to form a modified form of the secular
equation for rank $k$ updates, and solve for the roots. Considering
only the PD case, suppose we have a rank $k$ update matrix
$U \in \R^{m \times k}$.  Using $V = Q^T U$ we form
\begin{align}
\tilde B&= Q\Lambda Q^T + \rho UU^T = Q\Lambda Q^T + \rho Q Q^T UU^TQ Q^T \\
  &= Q( \Lambda + \rho Q^T UU^TQ ) Q^T = Q( \Lambda + \rho VV^T ) Q^T.
\end{align}
We then consider the eigenvalues and eigenvectors of
$\Lambda + \rho VV^T$. From an updated characteristic
equation
\begin{align}
\det(\Lambda + \rho VV^T - \mu I) = \det(\Lambda - \mu I)\det(I_k + V^T(\Lambda - \mu I)^{-1} V)
\end{align}
we get an updated secular equation
\begin{align}
f_k(\mu) = \det(I_k + V^T(\Lambda - \mu I)^{-1} V) =0
\end{align}
whose roots we wish to find. Note that $f_1(\mu) = f(\mu)$.  With
this, one could perform a bracketed Newton like method here to find
the roots of $f_k$ since the *LAED4 method, described above, is good
only for rank-1 updates. Alternatively, we could produce a full rank-k
version of *LEAD4.  Since rank-$k$ for $k > 1$ is less frequent (e.g., the
greedy algorithm for submodular maximization updates only one single element at a time),
we elect not to do that here, and instead, if multiple updates
are required, we resort to the simpler
incremental multiple rank-1 update method that uses existing rank-1
approach $k$ times.

\section{Greedy maximization Speedup for Matrix Spectral Functions: Secular Approach vs. Oracle Access}
\label{sec:greedy-maximization-speedup}

\begin{figure}[tbh]
\centering
\includegraphics[width=0.98\textwidth]{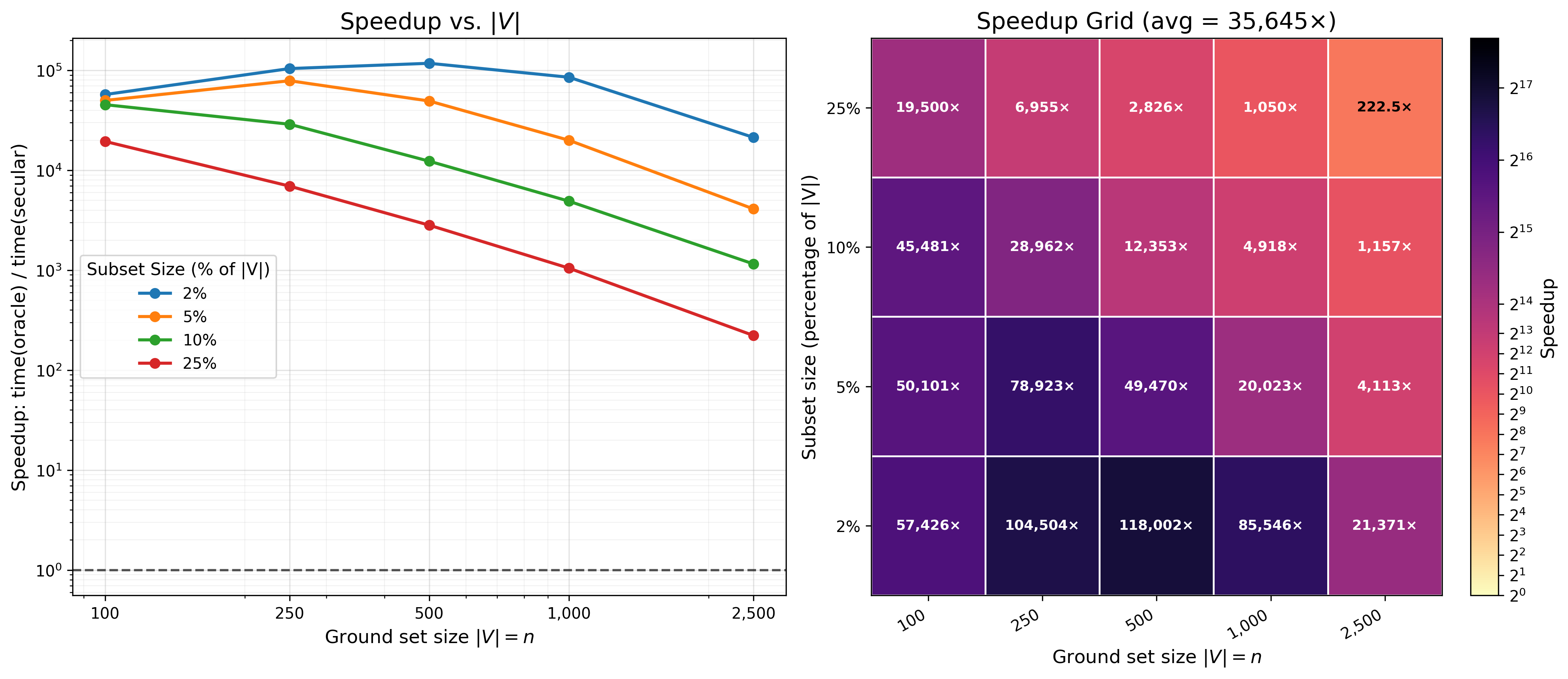}
\caption{Speedup of the secular approach compared to the oracle approach for
greedy maximization of the log Vendi score using an $n \times m$ design matrix $\data$ consisting
of random Gaussian values, where $m$ is fixed at $m=1024$.
Speedup is defined as the time taken
by the oracle approach divided by the time taken for the secular approach. We
run this for various different ground set sizes (i.e., $|V| = n \in \{ 100, 250,
500, 1000, 2\,500 \}$) and for fractions of the ground set selected (i.e.,
$k/|V| \in \{ 0.02, 0.05, 0.1, 0.25 \}$) where $k$ is the cardinality constraint
given in the priority-queue accelerated greedy algorithm (i.e.,
Algorithm~\ref{alg:greedy-max}). The time of each size, fraction pair (for both the secular
and oracle cases) is taken
as the minimum over multiple runs (to avoid an operating system 
interference that could cause the time to be measured higher than it should be).
The resulting speedup is always greater than 1 since the secular approach is always faster.
The same speedup data is shown in two ways, as a standard plot (left) and
as a grid (right).
As can be seen, the secular approach is
significantly faster than the oracle approach (the average speedup
is about $\randomspeedup$). The speedups get slower as $k$ and $n$
get larger due to the priority queue doing a worse job
predicting the maximum. Still, the worst speedup number,
a factor of $222.5$ is quite significant. We were unable to run
larger problem sizes since the slow oracle approach took too long to run (it would
have taken more than a week on a state-of-the-art CPU machine, and was not feasible).
}
\label{fig:greedy-maximization-speedup-random-data}
\end{figure}

In this section, we demonstrate the speedup that can be achieved via the
approach outlined in Section~\ref{sec:fast-matrix-spectral-functions} 
and Section~\ref{sec:fast-rank-1-updates-and-downdates}. We compare
the accelerated greedy maximization algorithm of~\citep{minoux1978_accelerated_greedy}
(which means using a priority queue to keep track of the largest gain thereby
avoiding certain re-evaluations when unnecessary)
on matrix spectral functions with $\phi(x) = - x \log x$ (i.e., the log Vendi score).
We normalize the matrix so this corresponds to a monotone non-decreasing submodular function
(specifically, like a density matrix, the spectrum of $B$ is normalized to sum to 1 
so that the log Vendi score is non-negative and monotone non-decreasing).
We compare the greedy 
algorithm (Algorithm~\ref{alg:greedy-max}) 
when either: (1) each $f(X \cup \{s\})$, $X \subseteq V$, $s \notin X$, must be computed from scratch
for each $s$ even if $X$ is the same, this is known as the
oracle approach, with (2) using the
secular approach so that evaluations of $f(X \cup \{s\})$ for different $s$ but 
the same $X$ can be done with a factor $O(m)$ speedup (in theory) compared to the oracle approach.
We note that the theoretical speedup does not account for any constants associated
with either approaches nor does it account for the behavior of the priority
queue as the problem sizes $n$ grow (this latter phenomenon can
be very problem dependent).

For the oracle approach, we use the Eigen C++ library to compute {\bf just} the
updated eigenvalues for $X \cup \{s\}$ and then apply $\phi$ to those
eigenvalues to get $f(X \cup \{s\})$. The reason we compute just the eigenvalues
is that the eigenvectors are not needed and it would have been slower. We use Eigen's full
dense symmetric positive semidefinite eigendecomposition routine 
(Listing~\ref{lst:oracle-eigenvalues})
to compute the eigenvalues which is known
properly exploit symmetry, uses only the lower triangular part of the matrix
and, according to Eigen's documentation, reduces the symmetric matrix to
tridiagonal form and then uses implicit symmetric QR factorization, and is
faster and more accurate for this class of matrix than other more
general-purpose variants. Eigen's implementation moreover is considered by many
to be a state-of-the-art implementation of this algorithm and is widely used in
the scientific computing community.

For the secular approach, we have implemented our own C++ version of the entire
secular approach as described in detail in
Section~\ref{sec:fast-rank-1-updates-and-downdates}.

\begin{lstlisting}[language=C++,caption={Eigen C++ library oracle eigenvalue code. This bit
  of code shows how we computed eigenvalues for the baseline oracle matrix function call
   access $f(X) = \trace{\phi(\data[X]^T\data[X])}$ where $\data$ is an $n \times m$
     design matrix.
Hence, each oracle call to $f(X)$ needs
      to compute the eigenvalues of an $m \times m$ matrix
         which is a $O(m^3)$ operation.  The Eigen C++ library
        is a state-of-the-art library for linear algebra and is widely used 
        in the scientific computing community, providing a state-of-the-art implementation
       for this problem. The calls are specific to symmetric positive semidefinite matrices.
      This provides evidence that our baseline oracle approach is efficient.
      Despite this, we find that our secular approach for computing rank-1 updates for
     use as gains in the greedy algorithm is significantly faster than this oracle approach.
 },label={lst:oracle-eigenvalues}]
// Bps is the updated matrix for X \cup {s}, i.e., B_{X + s}
Eigen::SelfAdjointEigenSolver<Eigen::MatrixXd> 
                es(Bps,Eigen::ComputeEigenvaluesOnly);
const auto eigenvalues = es.eigenvalues();
...
\end{lstlisting}

We show that the secular approach can significantly improve the speed 
(Figure~\ref{fig:greedy-maximization-speedup-random-data})
of the
accelerated greedy maximization process. We verified, moreover, that the solution
results produced by both the oracle and the secular approach are identical.
This secular code moreover was used to produce all of the
results mentioned in Section~\ref{sec:experiments}.

\section{Quantum (von Neumann) Entropy and SSA from a pure Linear Algebra Perspective and its difference
from Matrix Spectral functions}
\label{sec:von-neumann-entropy}

In this section, we derive (from a linear algebra first principle perspective) the quantum entropy,
and the strongly subadditive (SSA)~\cite{lieb1973proof} property of quantum entropy~\citep{bennett2002quantum,petz2007quantum,nielsen2010quantum}. While SSA is a
general submodularity property (as we show below) it is not the same 
as the matrix spectral (and thus the log Vendi score, or DPP) submodularity property that arises by applying functions whose
negative derivatives are matrix monotone to principal sub-matrices of
an underlying positive semidefinite matrix.

We first restate again this latter principal to make this section
complete. Given a function $\phi: \R_+ \to R$, 
an index set $V = [n]$, 
a subset $S \subseteq V$,
and a symmetric
positive semidefinite matrix $M \in \R^{n \times n}$,
we can
apply this function to any row/column 
submatrix $M_S \in \R^{|S| \times |S|}$ of $M$ via $f(S) = \trace{\phi(M_S)}$ 
where $\phi(M_S)$ is a standard matrix function,
i.e., $\phi(M_S) = Q_S \phi(\Lambda_S) \ctrans{Q}_S$
where $\Lambda_S$ is a diagonal of non-negative eigenvalues
of $M_S$ and $\phi(\Lambda_S)$ applies $\phi(\cdot)$ to
each of the diagonal elements. We have the result
that $f(S)$ is submodular if and only if $-\phi'$ is
matrix monotone over all PSD matrices. 
The function $-\phi'$ being matrix monotone means that for any two
matrices $M_1$ and $M_2$ with $0 \preceq M_1 \preceq M_2$
we have that $-\phi'(M_1) \preceq -\phi'(M_2)$. An
alternate name for this is that $\phi'$ is matrix antitone.
Note that $\phi(M_S)$ operates on only one sub-matrix
of $M$, namely the one created by the indices in $S$.

In the explanation below, we avoid any quantum mechanical or quantum
physics interpretation of constructs, and we also avoid use of the
bra-ket notation (things like $| \psi \rangle$ and $\langle \psi |$ that is often used in quantum mechanics),
keeping things grounded in pure linear algebra
notation. Indeed, the application of the below to quantum physics is
fascinating, and the corresponding quantum mechanical bra-ket notation
is useful, but it is not necessary for the purposes of this section. 
Here our goal is to point out
that the SSA (submodularity) property of quantum entropy is not the same 
as the submodularity property of applying such an $f$ to principal
matrices of a PSD matrix.

The quantum entropy function is defined
as 
\begin{align}
S(\rho) = -\trace{\rho \log \rho}
= - \sum \lambda_i \log \lambda_i
\end{align}
where $\rho$ is a density
matrix (i.e., a positive semidefinite Hermitian matrix with trace 1). 
Specifically, without getting into the details of physics behind quantum mechanics
and quantum information theory, $\rho$ is a quantum state (i.e., a
density matrix)
but for the present purposes we can think of $\rho$ as just a
regular matrix of complex values that is 
positive semi-definite ($\rho \succeq 0$, meaning that $\ctrans{x} \rho x \geq 0$ for all $x$
where $\ctrans{x}$ is the conjugate transpose of $x$), and also
Hermitian ($\rho = \ctrans{\rho}$ meaning $\rho$ is equal to
its conjugate transpose, i.e., $\rho$ is self-adjoint).
The matrix $\rho$
has diagonal entries that are non-negative real values (since if
a complex number is equal to its conjugate, it must be real)
and has unit trace ($\trace{\rho} = 1$). This means that the diagonal entries of $\rho$ must 
be non-negative and sum to 1, 
as is true of any discrete probability distribution.
The off-diagonal entries of $\rho$ may be complex valued (subject to that
the matrix is positive semidefinite and Hermitian).
This also means that the eigenvalues $\{ \lambda_i \}_{i=1}^n$ of $\rho$ are non-negative 
real-valued and also sum to 1, and thus can also be thought of
as a probability distribution.

The quantum entropy function $S(\rho)$ can thus
be expressed as a function on a matrix via
\begin{align}
S(\rho) = -\trace{\rho \log \rho} = \trace{ \phi(\rho)}
= -\sum_i \lambda_i \log \lambda_i
\end{align}
where here $\phi(x) = -x \log x$ and $\phi(\rho) = Q \phi(\Lambda) Q^T$ 
where $\Lambda$ is the diagonal of non-negative eigenvalues of $\rho$ and
$Q$ is a matrix of eigenvectors as column vectors. Recall
that the eigenvalues of a positive semidefinite Hermitian matrix
are always non-negative real valued.
We note
that this $\phi(x)$ is concave, and monotone-nondecreasing 
for $x \in [0, 1/e]$ reaching its peak at $x=1/e$ and is
monotone-nonincreasing for $x > 1/e$ hitting zero
at $x = 1$ after which it becomes negative. Since
the eigenvalues are in the range $[0,1]$,
$\phi(\lambda_i) \geq 0$ and thus $S(\rho) \geq 0$ for 
any density matrix $\rho$.

Now superficially, $S(\rho)$ indeed looks just like any matrix
function here applied to the density matrix $\rho$. 
This seems especially true since with $\phi(x) = -x \log x$ we have that
$-\phi'$ is matrix monotone  and this $\phi$ can be used
to get a submodular function by applying it to principal 
sub-matrices of an underlying PSD matrix, and this is precisely
the Vendi score.
However, the SSA property 
of quantum entropy is quite a different submodularity property
than applying a function $\phi$ to principal sub-matrices of an underlying PSD matrix.
The reason is, as we explain below, the
quantum entropy 
is not the application of $\phi$ to principal sub-matrices of an underlying PSD matrix.

We may think of $\rho \in \mathcal B(\mathcal H)$ as a density matrix on a Hilbert
space $\mathcal H$ 
where, e.g., $\mathcal H = \C^n$, and
where $\rho : \mathcal H \to \mathcal H$ 
is a linear operator, which means essentially that $\rho$ is a matrix on $\mathcal H$.
To be a density matrix, we require that
$\rho \in \mathcal B(\mathcal H)$ is
a bounded linear operator that is positive semidefinite Hermitian (i.e., self-adjoint) 
with trace 1. This means that any $\rho \in \mathcal B(\mathcal H)$ we consider is an $n \times n$ positive
semidefinite Hermitian matrix with trace 1. 

We can think also of $\mathcal H$ as
being a composite Hilbert space $\mathcal H = \mathcal H_A \otimes \mathcal H_B
\otimes \mathcal H_C$ where $\otimes$ is the tensor product and $\mathcal H_A$,
$\mathcal H_B$, and $\mathcal H_C$ are Hilbert spaces of sub-systems $A$, $B$,
and $C$. This means any vector $v \in \mathcal H$ can be written as a linear combination of
{\bf multiple} vectors of the form $v_A \otimes v_B \otimes v_C$ where $v_A \in \mathcal H_A$,
$v_B \in \mathcal H_B$, and $v_C \in \mathcal H_C$.  

Then, the SSA property of quantum entropy is usually expressed in the context of a
such a composite Hilbert space $\mathcal H$ as a tensor product of three
sub-systems $A$, $B$, and $C$ (i.e., $\mathcal H = \mathcal H_A \otimes \mathcal H_B \otimes \mathcal H_C$
as mentioned above).  
We define reduced density matrices $\rho_{AB} = \ptrace{C}{\rho_{ABC}}$,
$\rho_{BC} = \ptrace{A}{\rho_{ABC}}$, and $\rho_B = \ptrace{AC}{\rho_{ABC}}$ where
$\ptrace{X}{\cdot}$ is the partial trace (explained below) over subsystem $X$. The SSA property of quantum
entropy, then, is that for any such tripartite density matrix $\rho_{ABC}$ we have
\begin{align}
  S(\rho_{AB}) + S(\rho_{BC}) \geq S(\rho_{ABC}) + S(\rho_B)
\end{align}
which is a submodularity property of the quantum entropy function $S(\cdot)$. 

Again, this is not the same submodularity property as applying a function $\phi$ to
principal sub-matrix of an underlying PSD matrix $M$, as is the case for the Vendi score.
In order to see this, we first generalize and then specialize.
Let us now consider a finite ground set $V=[n]$ with $|V|=n$. We have a collection of
$n$ Hilbert spaces $\mathcal H_{(i)}$ for $i \in V$ 
where the dimension 
\footnote{We could say $d_{(i)} \geq 1$ here but w.l.o.g.\ we can just say $\geq 2$ 
since a 1-dimensional Hilbert space would multiply by a 
scalar complex number, as $\mathcal H \otimes \C = \mathcal H$, and thus not change anything.}
of $\mathcal H_{(i)}$ is $d_{(i)} \geq 2$.
We define a 
tensor product Hilbert space 
$\mathcal H^V = \bigotimes_{i=1}^n \mathcal H_{(i)}$ 
where the total dimension of $\mathcal H^V$
is $d_V = \prod_{i=1}^n d_i$.

To define the SSA property, it is necessary to fully understand the partial
trace property. One way to understand partial trace is to consider
the fact that
any $M \in \mathcal B(\mathcal H^V)$ can be written
as a finite linear combination of tensor products. i.e.:
\begin{align}
M = \sum_{i_1, i_2, \dots, i_n} w_{i_1, i_2, \dots, i_n} M_{(1), i_1} \otimes M_{(2), i_2} \otimes \dots \otimes M_{(n), i_n} 
\label{eqn:general_M_as_sum_of_tensor_products}
\end{align}
where $w_{i_1, i_2, \dots, i_n}$ are complex numbers 
and where,
for each $i$ and $j$,
$M_{(i), j} \in \mathcal B(\mathcal H_{(i)})$ is a matrix on $\mathcal H_{(i)}$.
Also, note that the notation $\otimes$ corresponds
to the Kronecker product of matrices, a generalization of
the outer product, and thus if $M_{(B)} \in \C^{2 \times 2}$
and $M_{(D)} \in \C^{3 \times 3}$ 
then $M_{(B)} \otimes M_{(D)} \in \C^{6 \times 6}$.

Then the partial trace out of, say, subsystem $(k)$ can be written as:
\begin{align}
\ptrace{(k)}{M} = 
\sum_{i_1, i_2, \dots, i_n} w_{i_1, i_2, \dots, i_n} 
M_{(1), i_1} \otimes M_{(2), i_2} \otimes \dots \otimes \trace{M_{(k), i_k}} \otimes \dots \otimes M_{(n), i_n}
\end{align}  
where $\trace{M_{(k), i_k}} \in \C$ is the standard matrix trace of $M_{(k), i_k}$,
where this local factor might have a complex trace 
(although, again, we always have $\trace{M} = 1$ for density matrices).
For example, with $M \in \C^{d_V \times d_V}$, 
then $\ptrace{(k)}{M} \in \C^{(d_V/d_k) \times (d_V/d_k)}$ 
as we are tracing out the subsystem $(k)$ which has dimension $d_k$. 
This follows since given $M_{(i)} \in \mathcal B(\mathcal H_{(i)})$ for $i \in V$, 
\begin{align}
\ptrace{(k)}{M_{(1)} \otimes M_{(2)} \otimes \dots \otimes M_{(n)}} = 
\trace{M_{(k)}}  M_{(1)} \otimes M_{(2)} \otimes \dots 
M_{(k-1)} \otimes M_{(k+1)} \otimes \dots \otimes M_{(n)},
\end{align}
meaning that the $k^\text{th}$ factor is removed and replaced with its trace. If we have
three subsystems, A,B, and C, and a given $\rho_{ABC} \in 
\mathcal B(\mathcal H_A \otimes \mathcal H_B \otimes \mathcal H_C)$, 
where we take $C$ as the inner and $A$ as the outer subsystem (and index),
we can also write the partial trace via
\begin{align}
\rho_{AB,(i,j),(i',j')} =
\ptrace{C}{\rho_{ABC}}_{(i,j),(i',j')} 
= \sum_{k=1}^{d_C} \rho_{ABC, (i,j,k),(i',j',k)}
\label{eqn:partial_trace_as_sum_over_inner_product}.
\end{align}

As mentioned above, eigenvalues of a positive semidefinite Hermitian
matrix are non-negative real valued, and when $M$ is positive
semidefinite Hermitian, then any partial trace of $M$ is also positive
semidefinite Hermitian. Hence, the eigenvalues of any partial trace of
such a matrix are also non-negative and real valued.

We note that the number of
free parameters in
$M_{(1)} \otimes M_{(2)} \otimes \dots \otimes M_{(n)}$
is $O(\sum_{i=1}^n d_{(i)}^2)$
but the number of free parameters
in $M$ is $O( d_V^2 ) = O( \prod_{i=1}^n d_{(i)}^2 )$ which is much larger.
The number of terms in the sum in~\eqref{eqn:general_M_as_sum_of_tensor_products} 
may be $O( \prod_{i=1}^n d_{(i)}^2 )$ in order
to achieve this.

\begin{figure}[tbh]
\centerline{\includegraphics[page=1,width=1.00\textwidth]{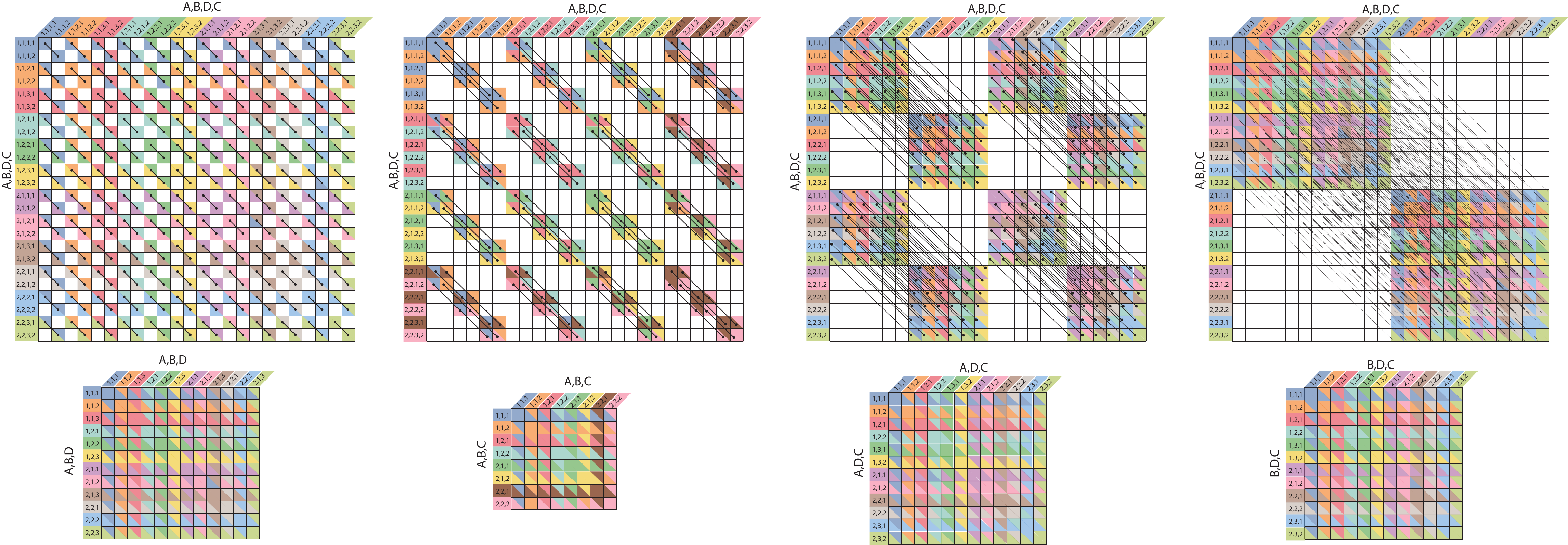}}
\caption{Partial trace diagramatically explained: Example $\mathcal H = \mathcal
H_A \otimes \mathcal H_B \otimes \mathcal H_D \otimes \mathcal H_C$ with $\dim
\mathcal H_A = 2, \dim \mathcal H_B = 2, \dim \mathcal H_D = 3, \dim \mathcal
H_C = 2$, so the matrix is $(2 \times 2 \times 3 \times 2) \times (2 \times 2
\times 3 \times 2) = 24 \times 24$. We use indices $i$ for $A$, $j$ for $B$, $k$
for $D$, and $\ell$ for $C$, so each row (column) is a four-tuple $(i,j,k,\ell)$
with $i,j,\ell \in \{1,2\}$ and $k \in \{1,2,3\}$. The top row shows four views
of $\rho_{ABDC} \in \mathcal B(\mathcal H)$ where the rows and columns are
marked with colors, and the four indices correspond to the subsystems $A$, $B$, $D$, and $C$
where $C$ is the inner-most index and $A$ is the outer-most index. We deliberately
placed $D$ in the middle to assist in the explanation of 
the partial trace and how the SSA property is a general submodularity property
in Equation~\eqref{eqn:SSA_as_submodularity}.
The downward
diagonal line segments show the summation operations of the partial trace, and the small
dots on each line segment indicate the number of elements that are summed together
to get the resulting single entry in the matrix indicated immediately below.
Each cell
in $\rho_{ABDC}$ in each case has two colors corresponding to its row and column, and the
colors are set in each instance of $\rho_{ABCD}$ so that the line segments sum only
cell entries of the same color, and hence the cell colors are preserved in the partial
trace matrices shown in the bottom row.
The bottom left shows $\rho_{ABD} = \ptrace{C}{\rho_{ABDC}}$, the bottom second from
left shows $\rho_{ABC} = \ptrace{D}{\rho_{ABDC}}$, the bottom second from right
shows $\rho_{ADC} = \ptrace{B}{\rho_{ABDC}}$, and the bottom right shows
$\rho_{BCD} = \ptrace{A}{\rho_{ABDC}}$. In each case, each row and column are
marked with a color, and cells with the same row/column color are "traced out"
(i.e., summed over) to get the reduced density matrix. A black line shows the
elements in the original $\rho \in \mathcal B (\mathcal H)$ that are summed over
to get the element in the reduced density matrix where dots along the black line
mark the cells that are summed over and that also show the same row/column color
as the reduced density matrix.
\label{fig:quantum_four_hilbert_product_space_matrix}
}
\end{figure}

To make this concrete, we demonstrate this using a product of four Hilbert spaces
$\mathcal H = \mathcal H_A \otimes \mathcal H_B \otimes \mathcal H_D \otimes
\mathcal H_C$ where the dimensions of each of the four Hilbert spaces,
respectively is $d_A=2$, $d_B=2$, $d_D=3$, and $d_C=2$ respectively so
that any $\rho \in \mathcal B(\mathcal H)$ is a $24 \times 24$ density matrix.
Note that we deliberately put $D$ in the middle here to make a particular
point a bit further below. 
We can then
define the partial trace over any one of the four sub-systems, e.g.,
$\ptrace{D}{\cdot}$, which is a linear operator that takes a density matrix
$\rho_{ABDC}$ on $\mathcal H$ and produces a density matrix $\rho_{ABC} =
\ptrace{D}{\rho_{ABDC}}$ on $\mathcal H_A \otimes \mathcal H_B \otimes \mathcal
H_C$. This means that the partial trace is
essentially "tracing out" the subsystem $D$ and leaving only the subsystems $A$,
$B$, and $C$. An example of a four Hilbert space density
matrix $\rho \in \mathcal B(\mathcal H)$,
as well as partial traces 
$\rho_{ABD} = \ptrace{C}{\rho}$, 
$\rho_{ABC} = \ptrace{D}{\rho}$, 
$\rho_{ADC} = \ptrace{B}{\rho}$, 
and $\rho_{BCD} = \ptrace{A}{\rho}$,
is shown in Figure~\ref{fig:quantum_four_hilbert_product_space_matrix}.

To see how SSA is a submodularity property,
consider
a positive semi-definite Hermitian 
density matrix $\rho \in \mathcal B( \mathcal H^V)$ 
(i.e., a positive semidefinite Hermitian matrix with trace 1). 
Again, in quantum information theory, $\rho$ is a quantum state, but 
here we consider it just as a complex valued matrix
$\rho \in \C^{d_V \times d_V}$.
We also define the reduced density matrices $\rho_S = \ptrace{V\setminus S}{\rho}$ for 
any set $S \subseteq V$ 
where $\ptrace{V\setminus S}{\cdot}$ is the partial trace over the subsystems in $V\setminus S$
(meaning we "trace out" the subsystems in $V\setminus S$ leaving only $S$). 
We consider a real valued function $\phi: \R_+ \to \R$
where again $\phi(x) = - x \log x$. As mentioned
above, we know this function is concave on $(0, \infty)$, and also that $-\phi'$ is matrix monotone,
but this is not what is critical here. 

Now we can define a set function $f: 2^V \to \R$ defined as follows, for any $S \subseteq V$
\begin{align}
f(S) = \trace{ \phi( \ptrace{V \setminus S}{\rho} ) } = \trace{ \phi( \rho_S ) } = S(\rho_S)
\label{eqn:quantum_entropy_as_set_function}
\end{align}
That is $f: 2^V \to \R$ is a set function that first takes the partial trace of
$\rho$ over the subsystems in $V \setminus S$ to get a reduced density matrix
$\rho_S = \ptrace{V \setminus S}{\rho}$, then applies the matrix function $\phi(\cdot)$
to this reduced density matrix, and then takes the normal trace of the resulting matrix
to get a real value.

The SSA property of quantum entropy is then the same as
the submodularity property of $f$, namely that is for any
$S, T \subseteq V$, we have that
\begin{align}
f(S) + f(T) \geq f(S \cup T) + f(S \cap T)
\end{align}
The usual three-part statement of SSA is recovered by
first tracing out the complement of $S \cup T$
and then defining 
three disjoint groups
$A = S \setminus T$, $B = S \cap T$, and $C = T \setminus S$,
and then writing the SSA inequality as
\begin{align}
\label{eqn:SSA_as_submodularity}  
  f(A \cup B) + f(B \cup C) \geq f(A \cup B \cup C) + f(B)
\end{align}
which is the same as
\begin{align}
  S(\rho_{AB}) + S(\rho_{BC}) \geq S(\rho_{ABC}) + S(\rho_B)
\end{align}
This is shown in Figure~\ref{fig:quantum_four_hilbert_product_space_matrix}
where the SSA inequality is shown where
$S \cup T = A \cup B \cup C$ and $V \setminus (S \cup T) = D$
where $\dim \mathcal H_A = 2$, 
$\dim \mathcal H_B = 2$, 
$\dim \mathcal H_D = 3$, 
and $\dim \mathcal H_C = 2$. Hence, 
any $\rho_{ABC} \in \mathcal B(\mathcal H_A \otimes \mathcal H_B \otimes \mathcal H_C)$ is a $8 \times 8$ density matrix
as shown in the figure (bottom row, second from left).

Considering Eqn~\eqref{eqn:quantum_entropy_as_set_function}, 
$S = \emptyset$, then
$\rho_\emptyset = \ptrace{V}{\rho} = \trace{\rho} = 1$ and thus $f(\emptyset) = \trace{\phi(1)} = 0$.
Thus, this $f(\cdot)$ is normalized in the sense that $f(\emptyset) = 0$. 
On the other hand, quantum entropy is not necessarily monotone-nondecreasing. For
example, consider the vector $x = \ctrans{(1, 0, 0, 1)}$ and the matrix formed
as $\rho = \frac{1}{2} x \ctrans{x}$ which is a valid density matrix, and a
ground set $V=\{1,2\}$ (some may recognize this as a Bell state).
This is a positive semi-definite $4 \times 4$ density
matrix, but since it is formed from a single rank-1 update, it has one non-zero
eigenvalue of value 1. Hence $f(V) = S(\rho) = \trace{\phi(\rho)} =
\trace{\phi(1)} = 0$. However, if we take the partial trace of $\rho$ over either
of the two subsystems, using the order $00,01,10,11$, we get $\rho_{\{1\}} =
\ptrace{\{2\}}{\rho} = \frac{1}{2} \begin{pmatrix}1 & 0 \\ 0 & 1 \end{pmatrix} =
\rho_{\{2\}}$ and thus $f(\{1\}) = f(\{2\}) = \trace{\phi(\rho_{\{1\}})} =
\trace{\phi(\rho_{\{2\}})} = \log(2)$. Thus,
while submodularity holds for this particular pair of sets ($f(A) + f(B) = 2\log(2) > f(A \cup B) + f(A \cap B) = f(V)
+ f(\emptyset) = 0$), monotonicity does not since
$f(\{1\}) = f(\{2\}) = \log(2) > f(V) = 0$.

Now, while it is the case that $-\phi'$ is matrix monotone on $(0,\infty)$, we can show that
$-\phi'$ being matrix monotone is not a sufficient condition for 
$f(S) = \trace{ \phi( \ptrace{V \setminus S}{\rho} ) }$ to be submodular.
As a counterexample, consider another function $\phi(x) = \sqrt{x}$; and
it is 
mentioned in~\Cref{sec:matrix_spectral_functions}
that this $-\phi'$ is also matrix monotone. However, the
function $f(S) = \trace{ \phi( \ptrace{V \setminus S}{\rho} ) }$ is
not submodular, and here is a simple demonstrating example where $|V|=3$,
$V=\{1,2,3\}$, $\rho$ is
a $8 \times 8$ density matrix, and $d_{(i)} = 2$ for $i=1,2,3$ (i.e., each subsystem is a qubit).
We define the diagonal matrix
\begin{align}
\rho
= \begin{pmatrix}
1/4 & 0 & 0 & 0 & 0 & 0 & 0 & 0 \\
0 & 1/4 & 0 & 0 & 0 & 0 & 0 & 0 \\
0 & 0 & 0 & 0 & 0 & 0 & 0 & 0 \\
0 & 0 & 0 & 0 & 0 & 0 & 0 & 0 \\
0 & 0 & 0 & 0 & 1/4 & 0 & 0 & 0 \\
0 & 0 & 0 & 0 & 0 & 1/4 & 0 & 0 \\
0 & 0 & 0 & 0 & 0 & 0 & 0 & 0 \\
0 & 0 & 0 & 0 & 0 & 0 & 0 & 0 \\
\end{pmatrix}
\end{align}
with the basis ordered as $000, 001, 010, 011, 100, 101, 110, 111$ where the
first bit is subsystem 1, the second bit is subsystem 2, and the third bit is
subsystem 3. 

As above, we define $f(S) = \trace{ \phi( \ptrace{V \setminus S}{\rho}) } = \trace{
\sqrt{ \rho_S } }$ where $\ptrace{V \setminus S}{\rho}$ is the partial trace of
$\rho$ over the subsystems in $V \setminus S$
($V \setminus S$ is "traced out" leaving only $S$).
We can compute $f(S)$ for all $S
\subseteq V$ and check the submodularity inequality.
Specifically, we check if $f(S) + f(T) \geq f(S \cup T) + f(S \cap T)$. If this inequality does
not hold for some $S,T$, then $f$ is not submodular despite that $-\phi'$
is matrix monotone. We set $S=\{1,2\}$, $T=\{2,3\}$,
$S \cup T = V = \{ 1,2,3 \}$,
and $S \cap T = \{ 2 \}$.
We have that $f(S) = \trace{ \sqrt{ \ptrace{\{3\}}{\rho} } } = \trace{ \sqrt{ \rho_{\{1,2\}} } }
= \sqrt{1/2} + \sqrt{1/2} = \sqrt{2}$.
$f(T) = \trace{ \sqrt{ \ptrace{\{1\}}{\rho} } } = \trace{ \sqrt{ \rho_{\{2,3\}} } }
= \sqrt{1/2} + \sqrt{1/2} = \sqrt{2}$.
$f(S \cup T) = f(V) = \trace{ \sqrt{ \rho }} = 4\sqrt{1/4} = 2$.
And $f(S \cap T) = f(\{2\}) = \trace{ \sqrt{ \ptrace{\{1,3\}}{\rho} } } = \trace{ \sqrt{ \rho_{\{2\}} } }
= \sqrt{1} + \sqrt{0} = 1$.
Thus, we have $f(S) + f(T) = \sqrt{2} + \sqrt{2} = 2\sqrt{2} \approx 2.828
< f(S \cup T) + f(S \cap T) = 2 + 1 = 3$, thus violating submodularity.
Therefore, matrix monotonicity of $-\phi'$ is not sufficient for submodularity of $f(S) = \trace{ \phi( \ptrace{V \setminus S}{\rho} ) }$.

On the other hand, we see rather trivially that a matrix spectral
function defined with $\phi(x) = \sqrt{x}$, 
$M = \rho$ and $M_S$ now being the principal sub-matrix of $M$
according to $S$ (recall $\rho_S = \ptrace{V \setminus S}{\rho}$ is not 
a sub-matrix of $\rho$ but rather partial trace).
Hence, we can consider a now size-8 ground set $V = [8]$
over this matrix, 
and define $f(S) = \trace{ \sqrt{M_S} }$.
This is trivially submodular, not only since $-\phi'$ is matrix monotone,
but in this particular case, since $M$ is just a diagonal
matrix, $f(S) =  \sum_{i \in S} \sqrt{d(i)} $ where $d(\cdot)$ is
the modular vector of diagonal elements of $M$. Hence, this is a simple
modular (and thus submodular) function.
Matrix spectral functions are monotone-nondecreasing 
as long as their corresponding $\phi$ function
is normalized ($\phi(0) = 0$) and monotone nondecreasing for the range of eigenvalues.

\section{Additional Experimental Details and Results on ImageNet}
\label{app:additional_in1k}

We now expand on the three set sampling strategies and then describe the training setup. The next two subsections extend the main-paper figures to all combinations of subset size ($5\%$, $10\%$, $20\%$) and compute budget ($1\text{M}$, $2\text{M}$, $4\text{M}$ samples seen). We then report an additional analysis on compute as a proxy for test accuracy. Finally, we evaluate four additional appraisal functions (DPP, $\xi_3$, $\mfa$, $\mfc$) and the unnormalized-gradient variant of Vendi, in addition to the Facility Location and Vendi (normalized) comparison from the main paper. For the definitions of \mfa,\mfc, and $\xi_3$, refer to Section~\ref{sec:weakly-matrix-monotone} and for more discussion on weakly matrix monotone functions.

\subsection{ImageNet-1K Experimental Setup}
\label{app:in1k_setup}
All training runs are conducted on a single machine with four NVIDIA A100 GPUs, 40 vCPUs, and 512\, GB of RAM. We use the FFCV (Fast Forward Computer Vision) library~\cite{leclerc2022ffcv} for data loading and training. Apart from the number of training steps, all hyperparameters match those of the reference ResNet-18 16-epoch configuration provided in the official FFCV ImageNet repository.\footnote{\url{https://github.com/libffcv/ffcv-imagenet/blob/main/rn18_configs/rn18_16_epochs.yaml}}

\subsection{Set Sampling Strategies}
\label{app:in1k_set_sampling_strat}

Here we expand on the three sampling techniques summarized in the main paper. Throughout, we hold the subset size $k$ fixed and use random selection alongside the techniques below to construct $\mathcal{S}$.

\paragraph{Direct Optimization (Max).} We run stochastic-greedy~\citep{mirzasoleiman2015lazier} maximization of $f$, varying the stochasticity (and thus approximation quality) from very low (accurate) to very large (less accurate). This spans a large sub-range of values of $f(S)$ over the high end of the total range.

\paragraph{Direct Optimization (Heuristic Min).} We use the same stochastic-greedy procedure but at each step pick the element of \emph{smallest} conditional gain. This is a heuristic without a mathematical guarantee, but it runs very fast and yields very small values of $f(S)$, spanning a large sub-range of values of $f(S)$ over the low end of the total range. To further fill in the middle of the range, we prefix-seed the procedure: instead of starting from $S = \emptyset$, we initialize $S$ with a subset of size $k_0 < k$ and run Heuristic Min for the remaining $k - k_0$ steps, with $k_0 = 0$ recovering the unseeded case. The seed is either a uniformly random size-$k_0$ subset or a stochastic-greedy maximizer of $f$ at size $k_0$. Varying $k_0$ and the seed type gives subsets that interpolate between the random-sample band and the Min anchors. For computational reasons, a large fraction of $\mathcal{S}$ is generated by Direct Max and Heuristic Min on Facility Location, supplemented by smaller sweeps of Direct Max on each remaining function so that every $f$ in consideration contributes its own anchors. Each candidate subset is scored under every function evaluated on its dataset, so subsets generated by sampling on $f_1$ also serve as valid data points for the correlation plot of any other $f_2$.

\paragraph{Indirect Optimization.} We use two indirect proxy approaches. The first follows Prismatic Synthesis~\citep{jung2025prismatic}, which constructs subsets at three diversity levels. Low-diversity subsets are grown by repeatedly adding examples most similar to those already chosen. High-diversity subsets are obtained by clustering the ground set and sampling uniformly across clusters (using ground-truth labels in the case of ImageNet). Medium-diversity subsets combine both. The second is cross-function transfer: a subset optimized directly for $f_1$ is reused as an indirect sample for other functions $f_2, f_3, \ldots$, so we score it under every function we consider.

\subsection{Additional Dataset Size Results}
\label{app:fig1_extension}
\begin{figure*}[h!]
    \centering
    \begin{subfigure}[t]{.248\textwidth}
        \centering
        \includegraphics[width=\linewidth]
        {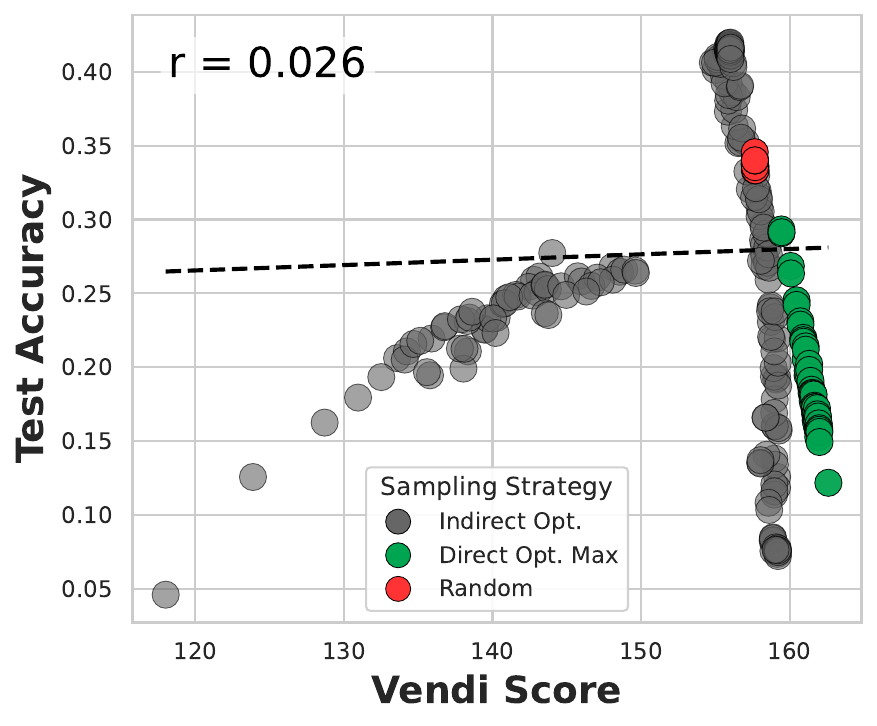}
        \caption{\small Vendi Score, Fixed Size}
        \label{fig:fig1a_supp5}
    \end{subfigure}\begin{subfigure}[t]{.248\textwidth}
        \centering
        \includegraphics[width=\linewidth]{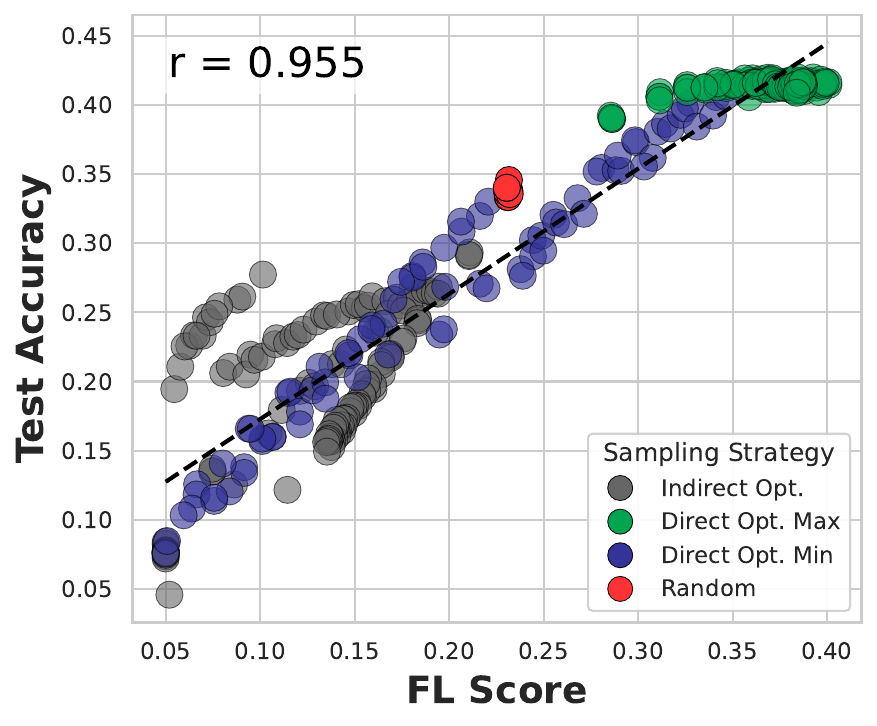}
        \caption{\small FL, Fixed Size}
        \label{fig:fig1b_supp5}
    \end{subfigure}\begin{subfigure}[t]{.248\textwidth}
        \centering
        \includegraphics[width=\linewidth]
        {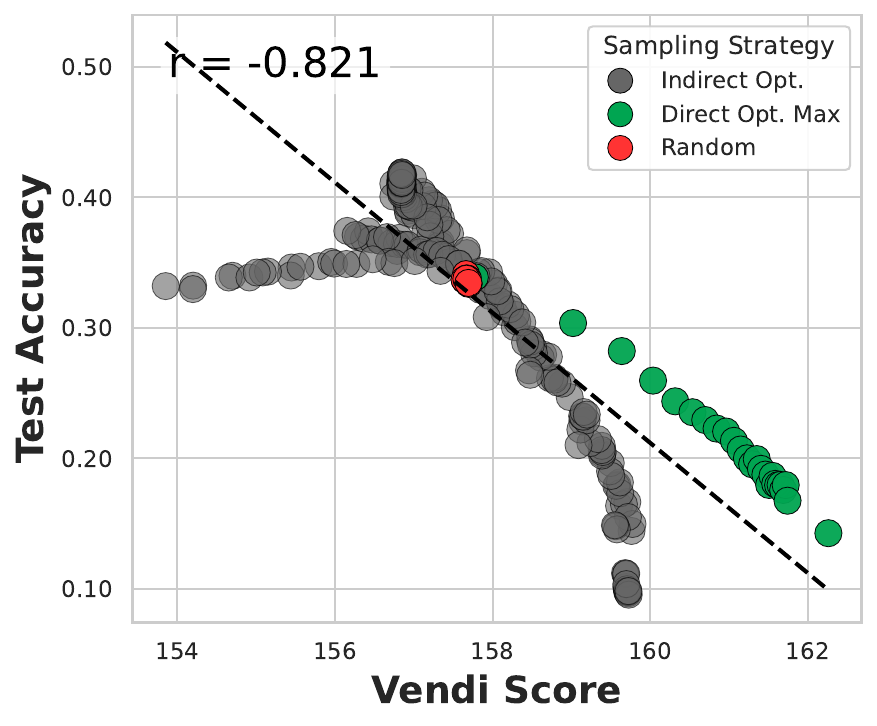}
        \caption{Vendi Score, Balanced}
        \label{fig:fig1c_supp5}
    \end{subfigure}\begin{subfigure}[t]{.248\textwidth}
        \centering
        \includegraphics[width=\linewidth]{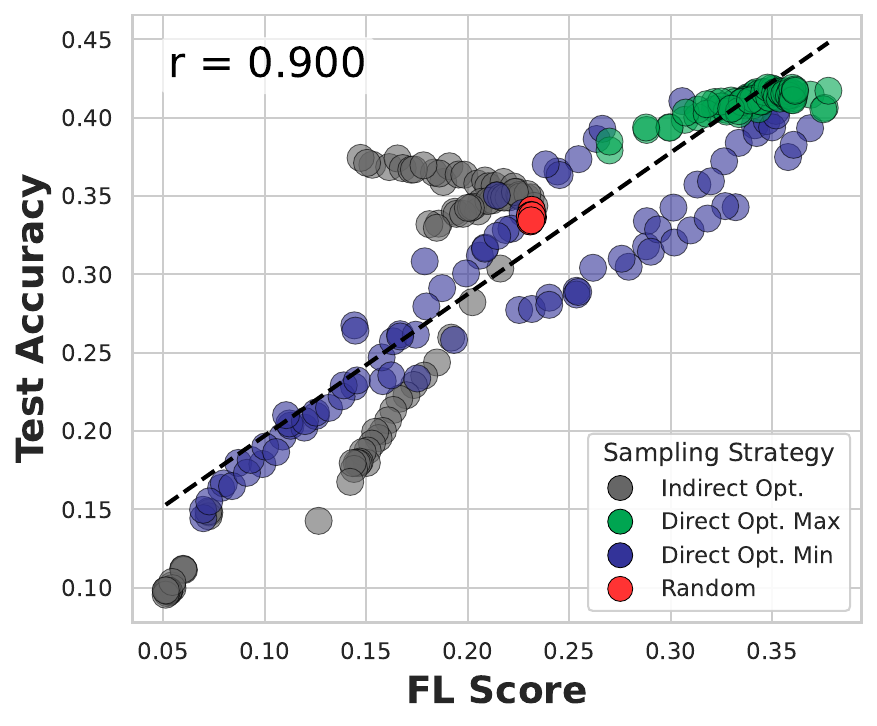}
        \caption{FL, Balanced}
        \label{fig:fig1d_supp5}
    \end{subfigure}
    \caption{\small \textbf{5\% Dataset Size}. Each run represents the outcome of an ImageNet-1K training run with the same configuration on different 5\% subsets. The datasets in (c) and (d) are constrained to be perfectly class-balanced. We observe that (1) sampling sets with high Vendi scores via direct optimization (shown in green) reveals that Vendi is poorly correlated with test acc. (2) random sets (shown in red) are highly concentrated and (3) the FL value is much more predictive of test accuracy.}
    \label{fig:fig1_supp5}
\end{figure*}

\begin{figure*}[h!]
    \centering
    \begin{subfigure}[t]{.248\textwidth}
        \centering
        \includegraphics[width=\linewidth]
        {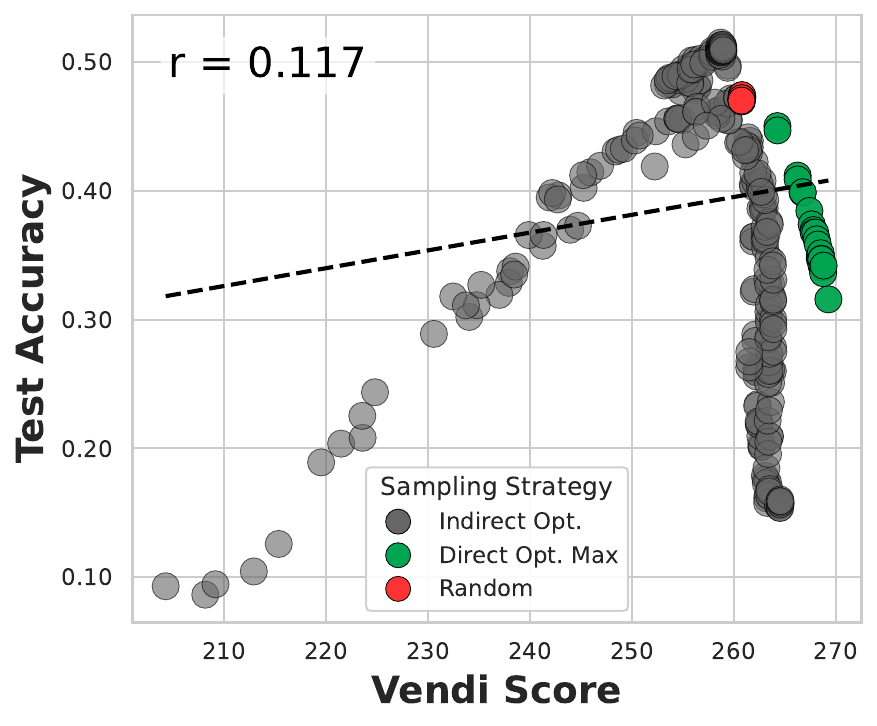}
        \caption{\small Vendi Score, Fixed Size}
        \label{fig:fig1a_supp10}
    \end{subfigure}\begin{subfigure}[t]{.248\textwidth}
        \centering
        \includegraphics[width=\linewidth]{figs/single_panels_all/single_panels_10pct/fl_cardinality_10pct_40ep.pdf}
        \caption{\small FL, Fixed Size}
        \label{fig:fig1b_supp10}
    \end{subfigure}\begin{subfigure}[t]{.248\textwidth}
        \centering
        \includegraphics[width=\linewidth]
        {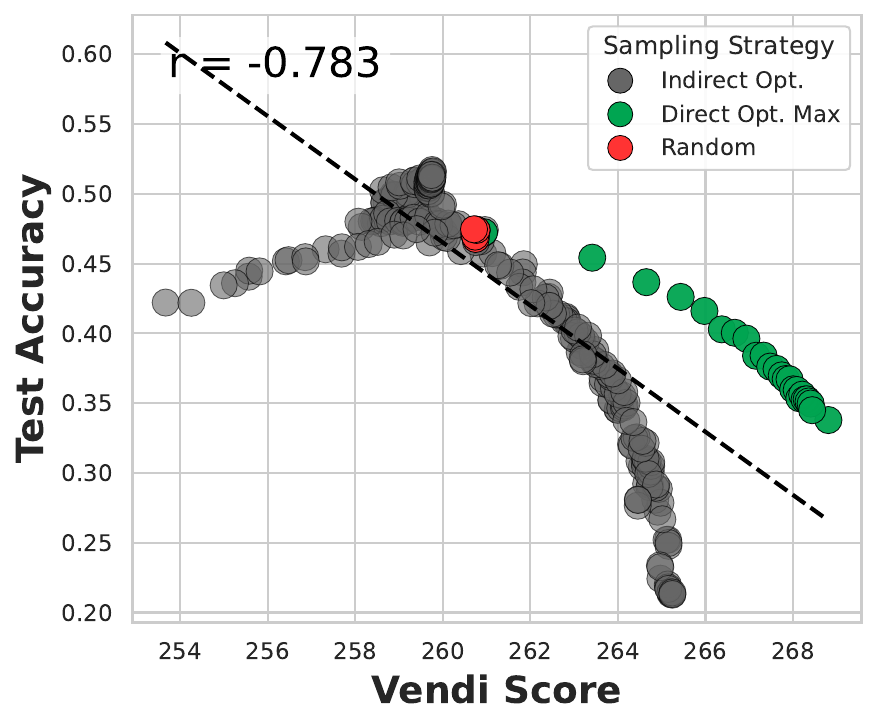}
        \caption{Vendi Score, Balanced}
        \label{fig:fig1c_supp10}
    \end{subfigure}\begin{subfigure}[t]{.248\textwidth}
        \centering
        \includegraphics[width=\linewidth]{figs/single_panels_all/single_panels_10pct/fl_balance_10pct_40ep.pdf}
        \caption{FL, Balanced}
        \label{fig:fig1d_supp10}
    \end{subfigure}
    \caption{\small \textbf{10\% Dataset Size}. Each run represents the outcome of an ImageNet-1K training run with the same configuration on different 10\% subsets. The datasets in (c) and (d) are constrained to be perfectly class-balanced. We observe that (1) sampling sets with high Vendi scores via direct optimization (shown in green) reveals that Vendi is poorly correlated with test acc. (2) random sets (shown in red) are highly concentrated and (3) the FL value is much more predictive of test accuracy.}
    \label{fig:fig1_supp10}
\end{figure*}

\begin{figure*}[h!]
    \centering
    \begin{subfigure}[t]{.248\textwidth}
        \centering
        \includegraphics[width=\linewidth]
        {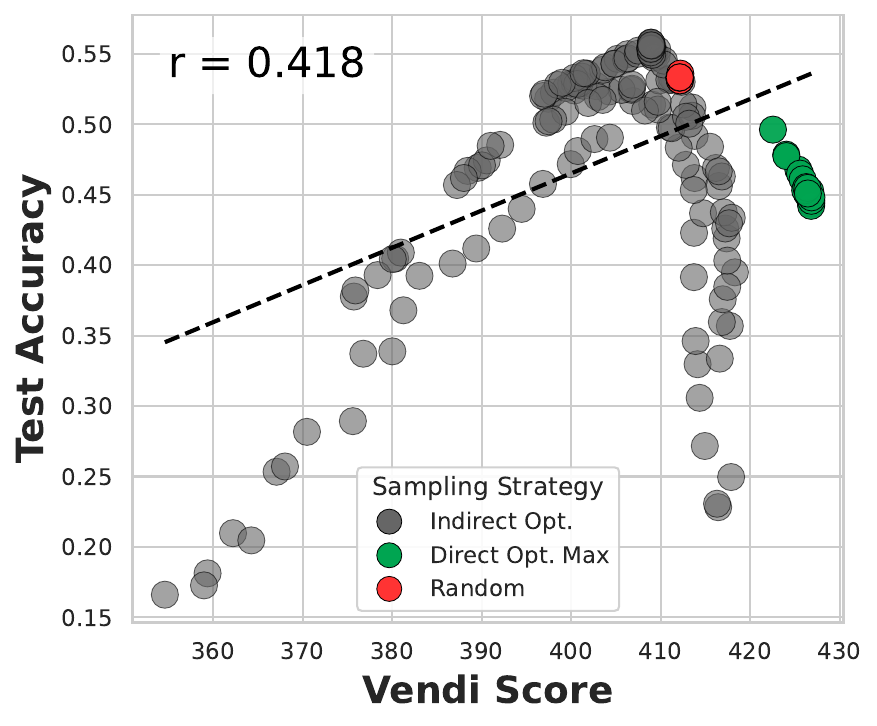}
        \caption{\small Vendi Score, Fixed Size}
        \label{fig:fig1a_supp20}
    \end{subfigure}\begin{subfigure}[t]{.248\textwidth}
        \centering
        \includegraphics[width=\linewidth]{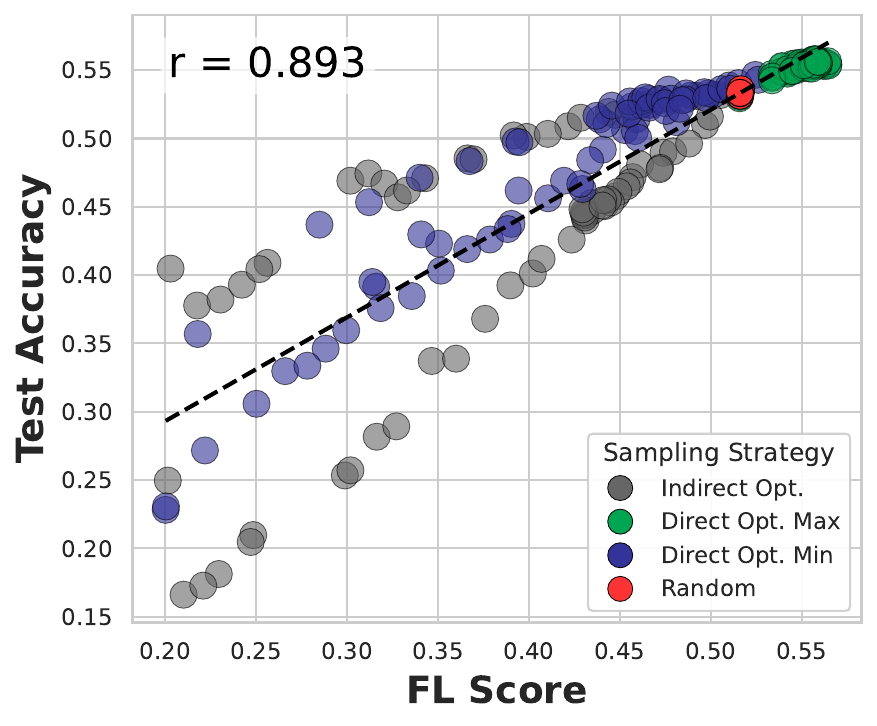}
        \caption{\small FL, Fixed Size}
        \label{fig:fig1b_supp20}
    \end{subfigure}\begin{subfigure}[t]{.248\textwidth}
        \centering
        \includegraphics[width=\linewidth]
        {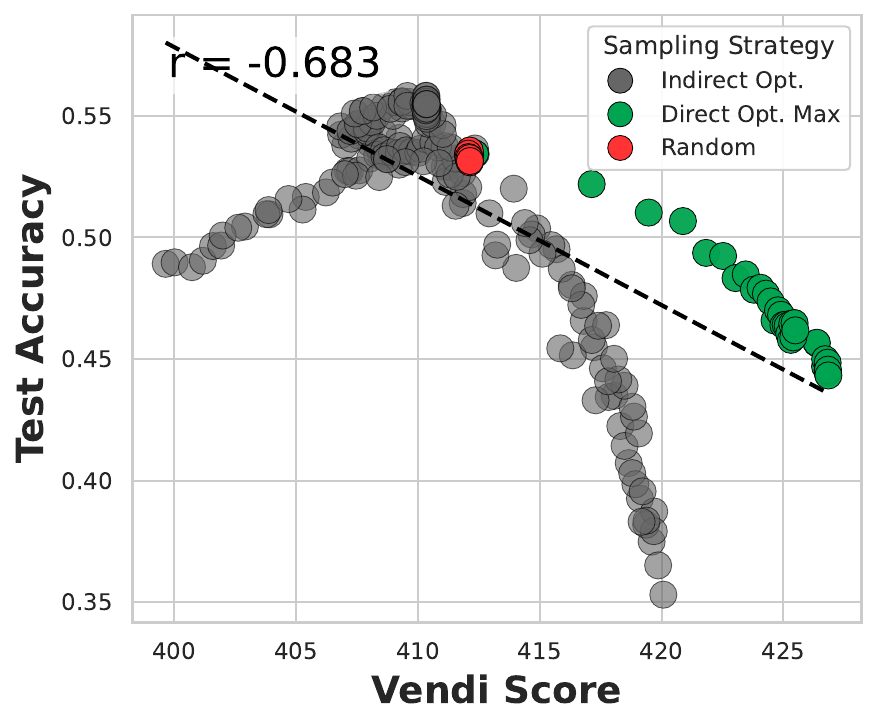}
        \caption{Vendi Score, Balanced}
        \label{fig:fig1c_supp20}
    \end{subfigure}\begin{subfigure}[t]{.248\textwidth}
        \centering
        \includegraphics[width=\linewidth]{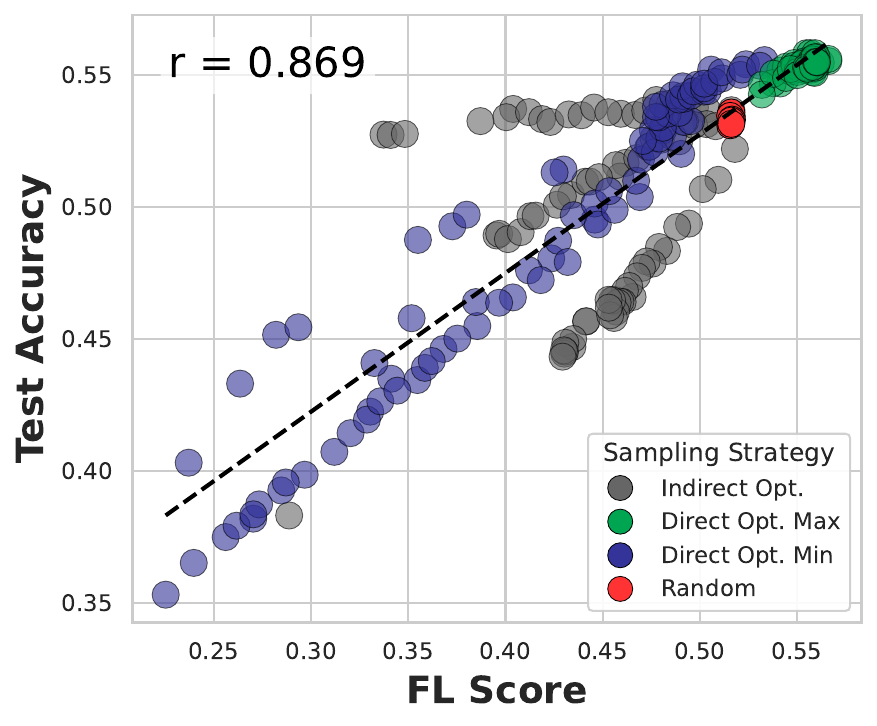}
        \caption{FL, Balanced}
        \label{fig:fig1d_supp20}
    \end{subfigure}
    \caption{\small \textbf{20\% Dataset Size}. Each run represents the outcome of an ImageNet-1K training run with the same configuration on different 20\% subsets. The datasets in (c) and (d) are constrained to be perfectly class-balanced. We observe that (1) sampling sets with high Vendi scores via direct optimization (shown in green) reveals that Vendi is poorly correlated with test acc. (2) random sets (shown in red) are highly concentrated and (3) the FL value is much more predictive of test accuracy.}
    \label{fig:fig1_supp20}
\end{figure*}

In this section, we extend the results of~\Cref{fig:fig1} to include other dataset sizes as well. As shown in~\Cref{fig:fig1_supp5,fig:fig1_supp10,fig:fig1_supp20}, the same key observations hold true.

\subsection{Additional Compute Constrained Results}
\label{app:fig2_extension}
\begin{figure*}[tbh!]
    \centering
    \begin{subfigure}[t]{.248\textwidth}
        \centering
        \includegraphics[width=\linewidth]{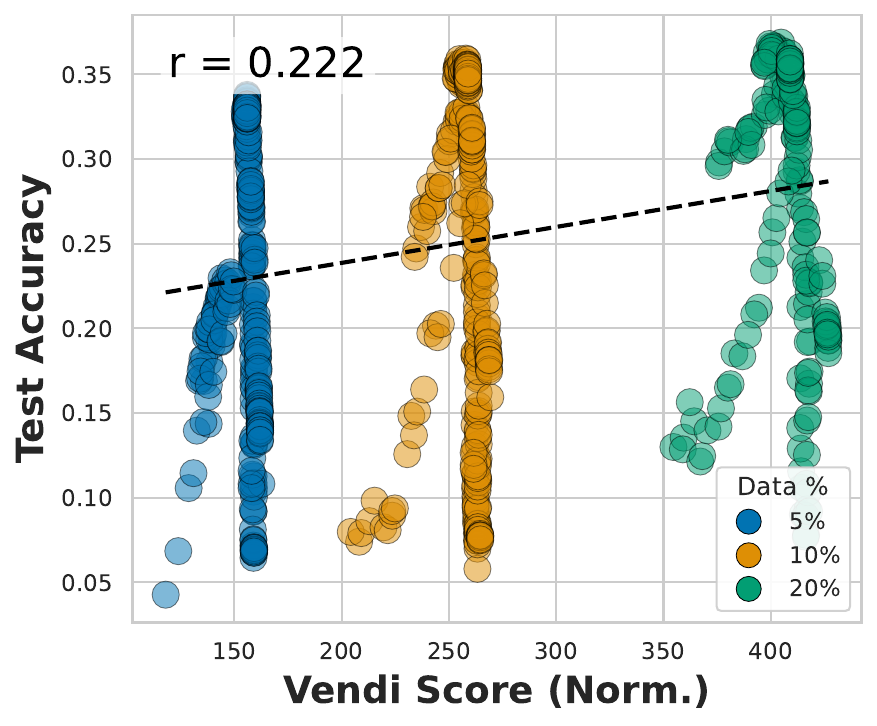}
        \caption{Vendi Score}
        \label{fig:fig2a_1M}
    \end{subfigure}\begin{subfigure}[t]{.248\textwidth}
        \centering
        \includegraphics[width=\linewidth]{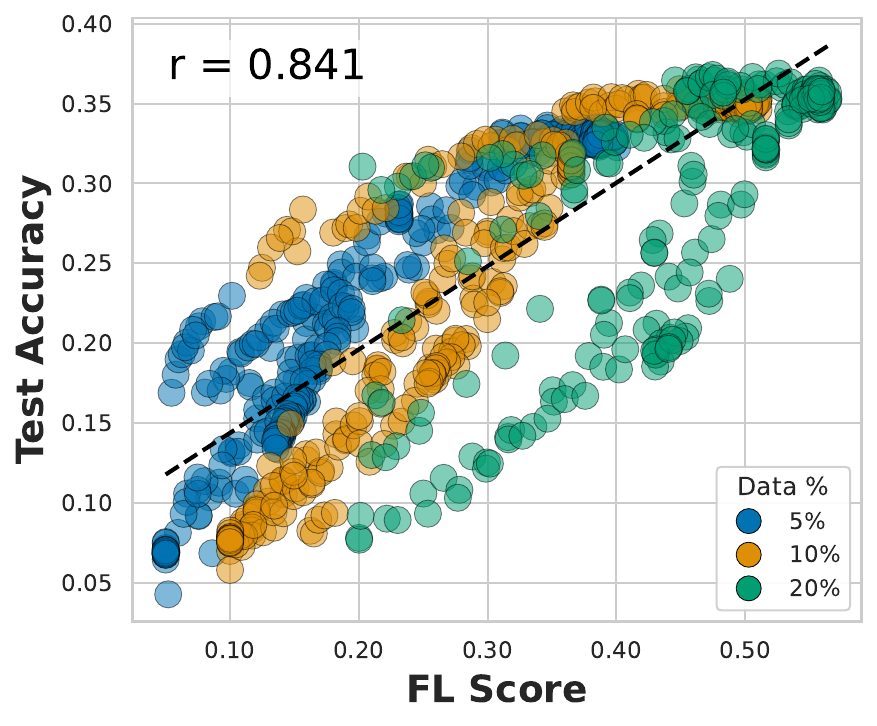}
        \caption{FL}
        \label{fig:fig2b_1M}
    \end{subfigure}\begin{subfigure}[t]{.248\textwidth}
        \centering
        \includegraphics[width=\linewidth]{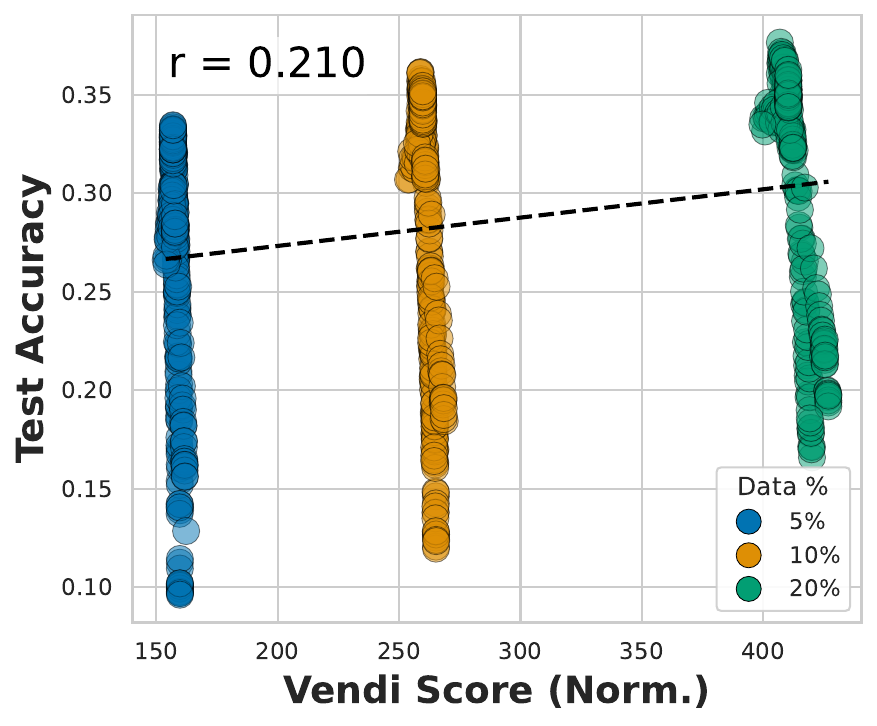}
        \caption{Vendi Score, Balanced}
        \label{fig:fig2c_1M}
    \end{subfigure}\begin{subfigure}[t]{.248\textwidth}
        \centering
        \includegraphics[width=\linewidth]{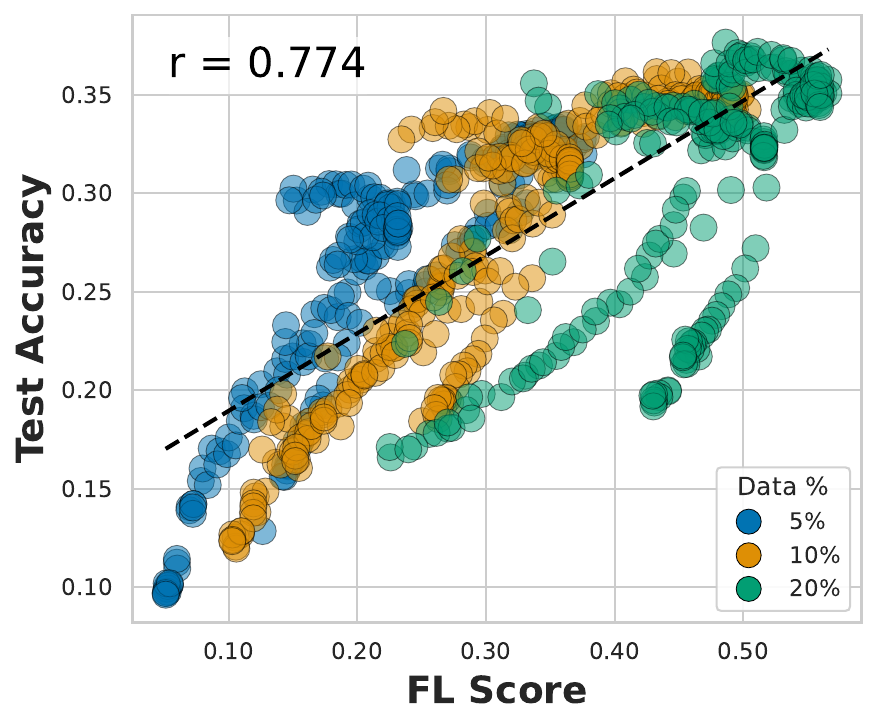}
        \caption{FL, Balanced}
        \label{fig:fig2d_1M}
    \end{subfigure}
    \caption{\small \textbf{Compute Budget Fixed to 1M Samples Seen}. Each run represents the outcome of an ImageNet-1K training run with the compute budget which is fixed to 1M samples seen irrespective of the dataset size. The datasets in (c) and (d) are constrained to be perfectly class-balanced. Critically, there are many instances of runs trained on 5\% subsets that outperform ones that were trained on 20\% size subsets. Moreover, Vendi score is highly biased by size while FL remains correlated with test accuracy. }
    \label{fig:fig2_1M}
\end{figure*}

\begin{figure*}[tbh!]
    \centering
    \begin{subfigure}[t]{.248\textwidth}
        \centering
        \includegraphics[width=\linewidth]{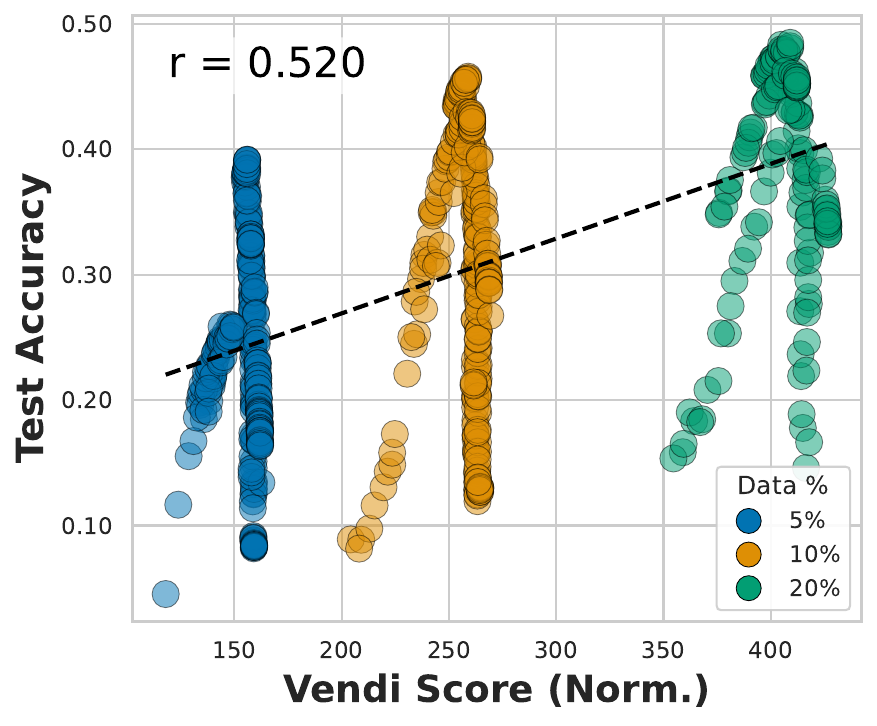}
        \caption{Vendi Score}
        \label{fig:fig2a_2M}
    \end{subfigure}\begin{subfigure}[t]{.248\textwidth}
        \centering
        \includegraphics[width=\linewidth]{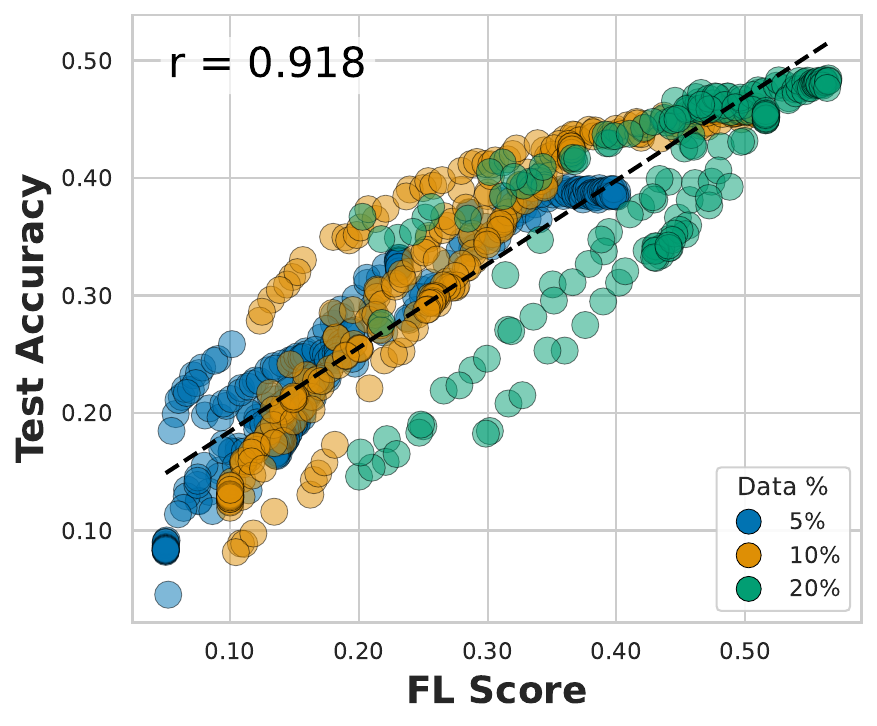}
        \caption{FL}
        \label{fig:fig2b_2M}
    \end{subfigure}\begin{subfigure}[t]{.248\textwidth}
        \centering
        \includegraphics[width=\linewidth]{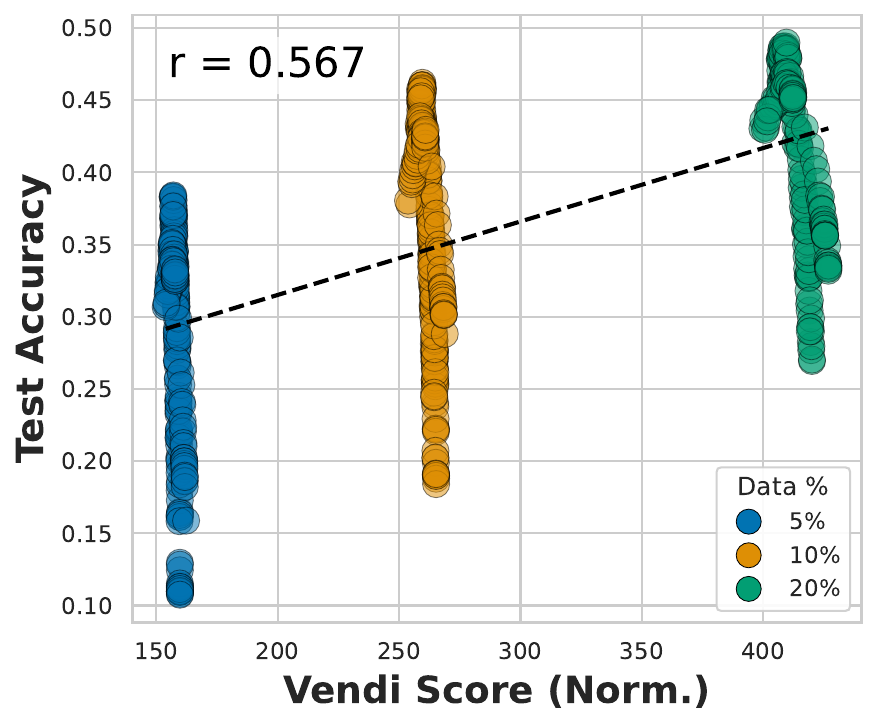}
        \caption{Vendi Score, Balanced}
        \label{fig:fig2c_2M}
    \end{subfigure}\begin{subfigure}[t]{.248\textwidth}
        \centering
        \includegraphics[width=\linewidth]{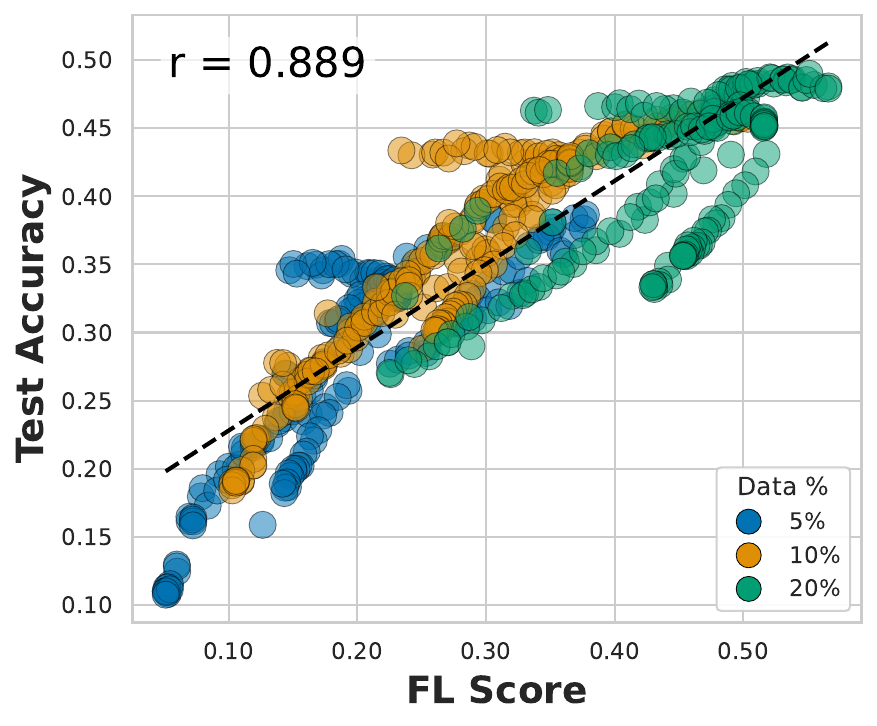}
        \caption{FL, Balanced}
        \label{fig:fig2d_2M}
    \end{subfigure}
    \caption{\small \textbf{Compute Budget Fixed to 2M Samples Seen}. Each run represents the outcome of an ImageNet-1K training run with the compute budget which is fixed to 2M samples seen irrespective of the dataset size. The datasets in (c) and (d) are constrained to be perfectly class-balanced. Critically, there are many instances of runs trained on 5\% subsets that outperform ones that were trained on 20\% size subsets. Moreover, Vendi score is highly biased by size while FL remains correlated with test accuracy.}
    \label{fig:fig2_2M}
\end{figure*}

\begin{figure*}[tbh]
    \centering
    \begin{subfigure}[t]{.248\textwidth}
        \centering
        \includegraphics[width=\linewidth]{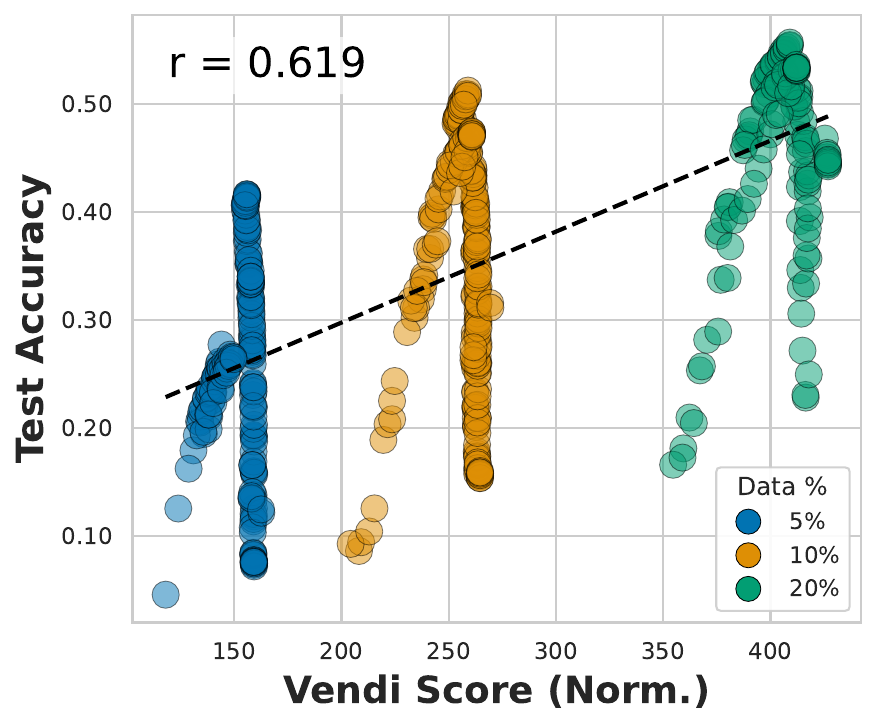}
        \caption{Vendi Score}
        \label{fig:fig2a_4M}
    \end{subfigure}\begin{subfigure}[t]{.248\textwidth}
        \centering
        \includegraphics[width=\linewidth]{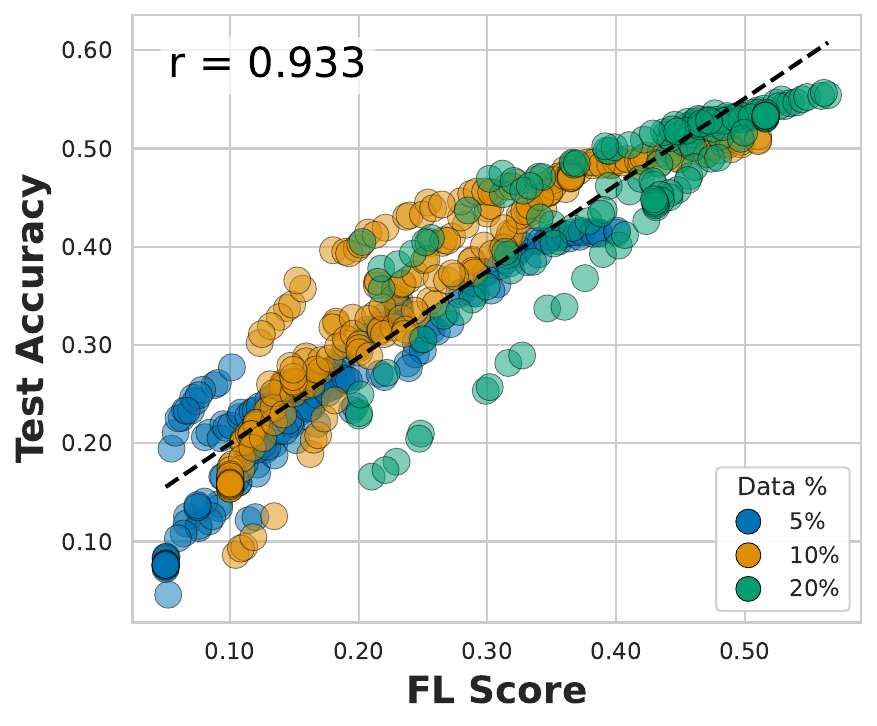}
        \caption{FL}
        \label{fig:fig2b_4M}
    \end{subfigure}\begin{subfigure}[t]{.248\textwidth}
        \centering
        \includegraphics[width=\linewidth]{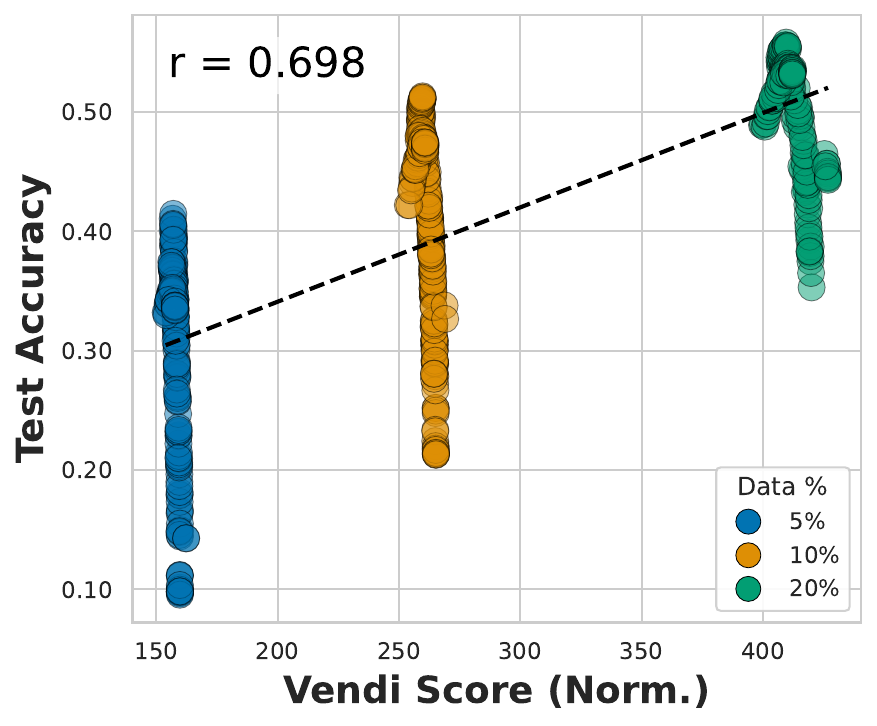}
        \caption{Vendi Score, Balanced}
        \label{fig:fig2c_4M}
    \end{subfigure}\begin{subfigure}[t]{.248\textwidth}
        \centering
        \includegraphics[width=\linewidth]{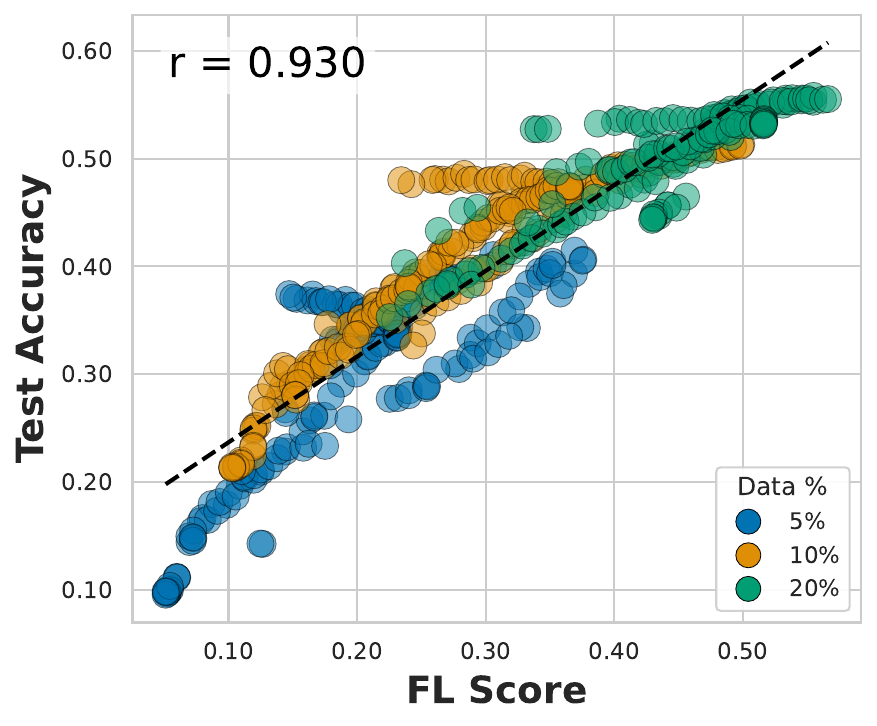}
        \caption{FL, Balanced}
        \label{fig:fig2d_4M}
    \end{subfigure}
    \caption{\small \textbf{Compute Budget Fixed to 4M Samples Seen}. Each run represents the outcome of an ImageNet-1K training run with the compute budget which is fixed to 4M samples seen irrespective of the dataset size. The datasets in (c) and (d) are constrained to be perfectly class-balanced. Critically, there are many instances of runs trained on 5\% subsets that outperform ones that were trained on 20\% size subsets. Moreover, Vendi score is highly biased by size while FL remains correlated with test accuracy.}
    \label{fig:fig2_4M}
\end{figure*}

In~\Cref{fig:fig2_1M,fig:fig2_2M,fig:fig2_4M}, we extend the results presented in~\Cref{fig:fig2} to other compute budgets. We find that the key conclusions continue to hold true in these additional settings as well.

\subsection{Compute is a Poor Proxy for Test Accuracy}
\label{app:compute_bad}
\begin{figure*}[tbh!]
    \centering
    \begin{subfigure}[t]{.248\textwidth}
        \centering
        \includegraphics[width=\linewidth]{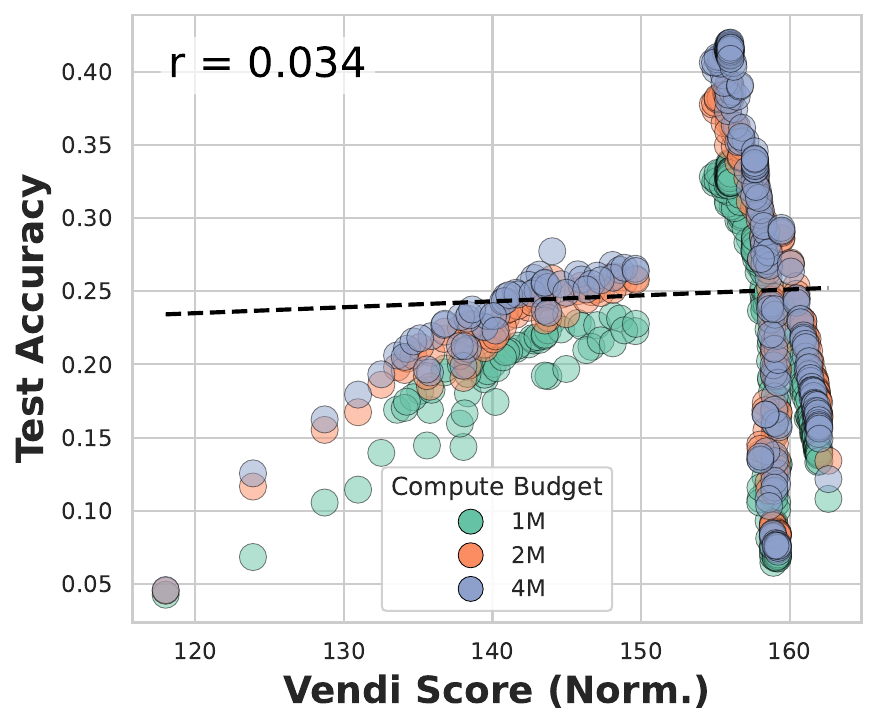}
        \caption{Vendi Score}
        \label{fig:fig3a_5pct}
    \end{subfigure}\begin{subfigure}[t]{.248\textwidth}
        \centering
        \includegraphics[width=\linewidth]{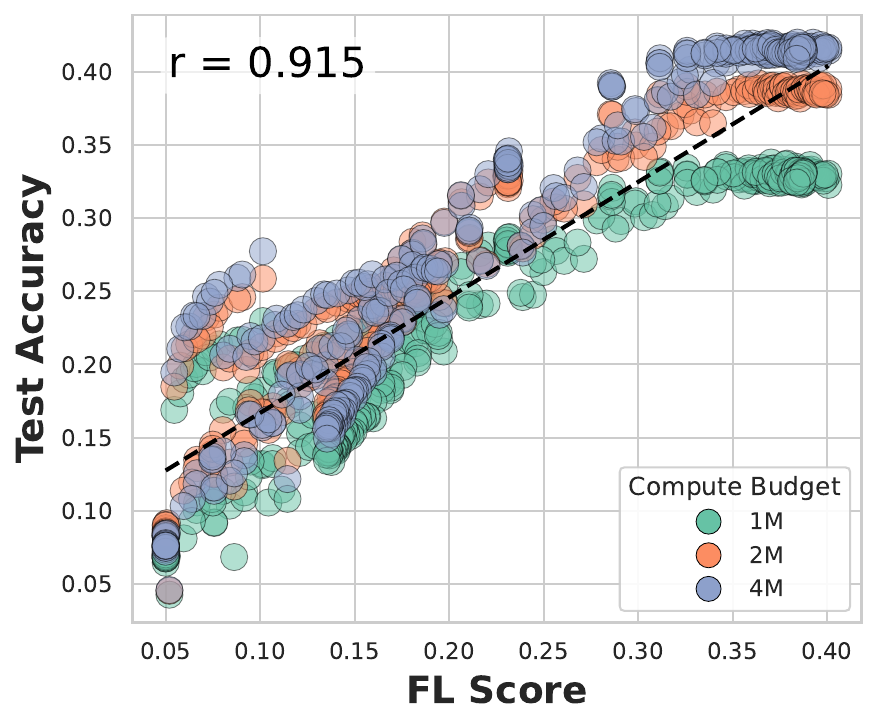}
        \caption{FL}
        \label{fig:fig3b_5pct}
    \end{subfigure}\begin{subfigure}[t]{.248\textwidth}
        \centering
        \includegraphics[width=\linewidth]{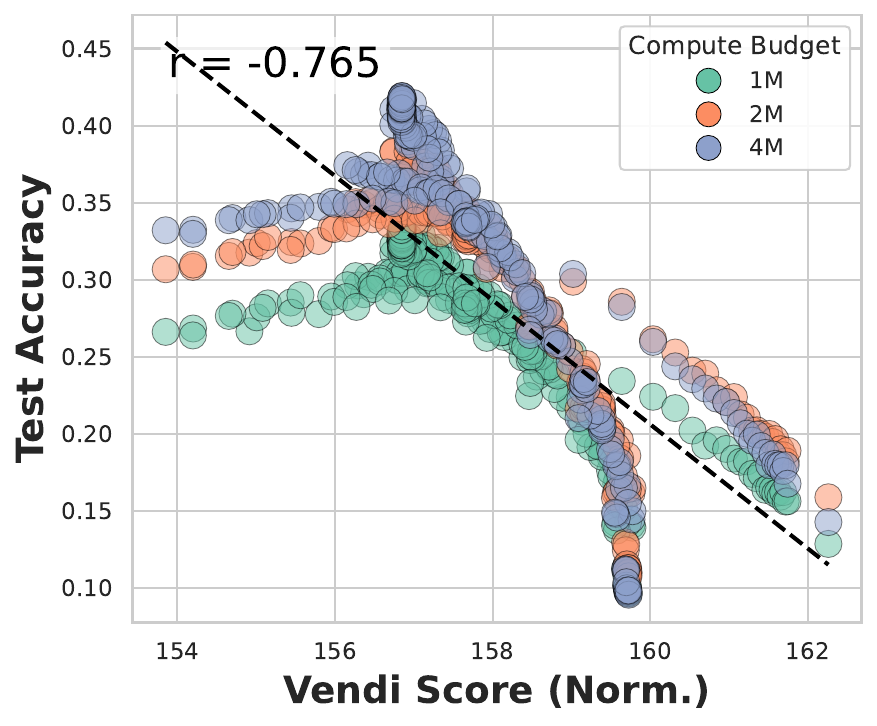}
        \caption{Vendi Score, Balanced}
        \label{fig:fig3c_5pct}
    \end{subfigure}\begin{subfigure}[t]{.248\textwidth}
        \centering
        \includegraphics[width=\linewidth]{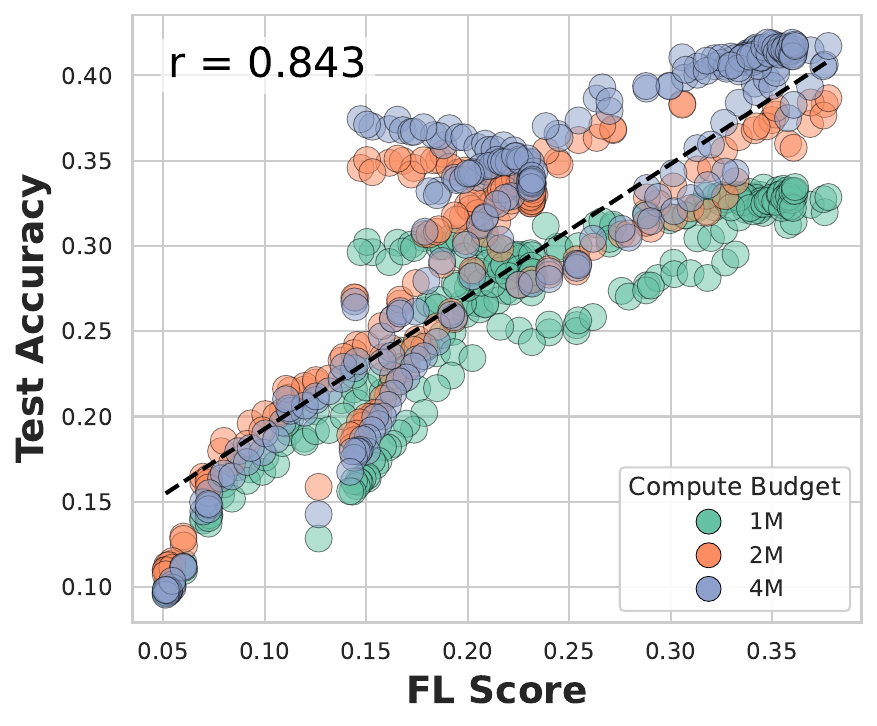}
        \caption{FL, Balanced}
        \label{fig:fig3d_5pct}
    \end{subfigure}
    \caption{\small \textbf{5\% Dataset Size} Each run represents the outcome of an ImageNet-1K training runs spanning 3 different compute budgets on 5\% subsets. The subsets in (c) and (d) are all perfectly class-balanced. We observe that there are many instances where a run with a low compute budget outperforms ones with a high compute budgets, suggesting that compute alone is a poor proxy for test accuracy.}
    \label{fig:fig3_5pct}
\end{figure*}

\begin{figure*}[tbh!]
    \centering
    \begin{subfigure}[t]{.248\textwidth}
        \centering
        \includegraphics[width=\linewidth]{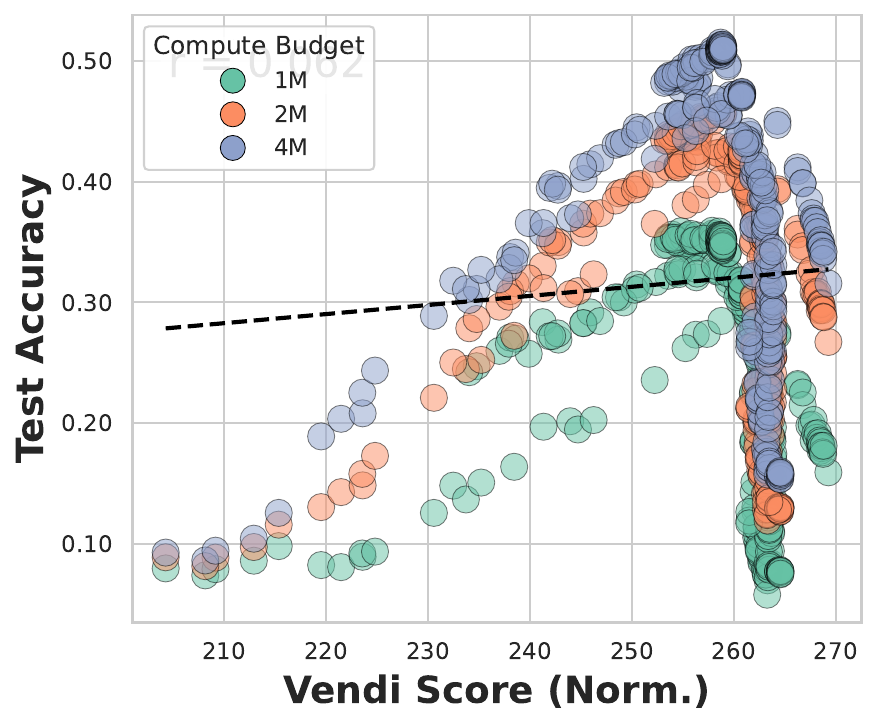}
        \caption{Vendi Score}
        \label{fig:fig3a_10pct}
    \end{subfigure}\begin{subfigure}[t]{.248\textwidth}
        \centering
        \includegraphics[width=\linewidth]{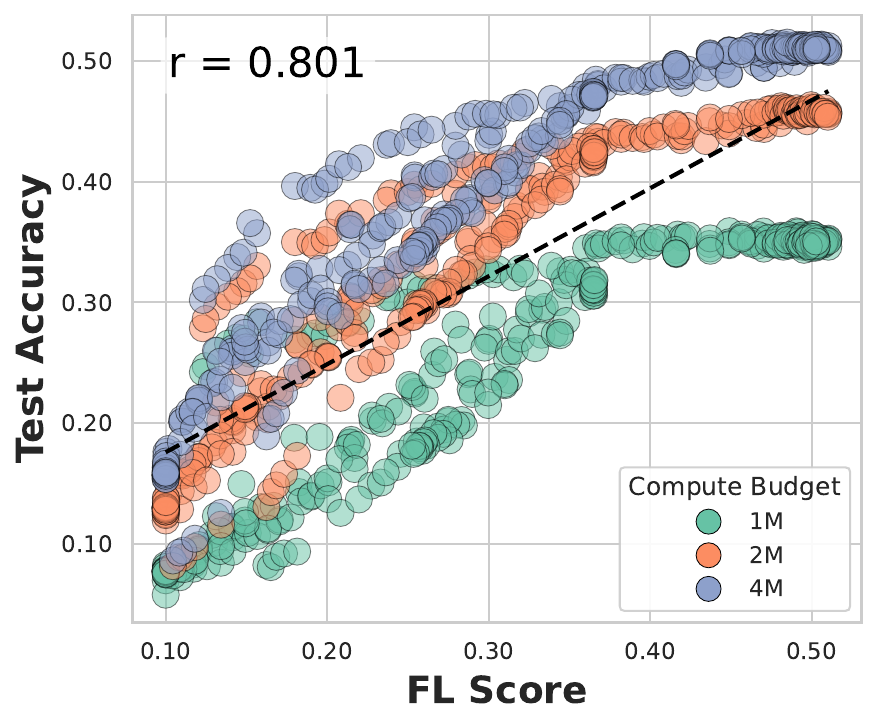}
        \caption{FL}
        \label{fig:fig3b_10pct}
    \end{subfigure}\begin{subfigure}[t]{.248\textwidth}
        \centering
        \includegraphics[width=\linewidth]{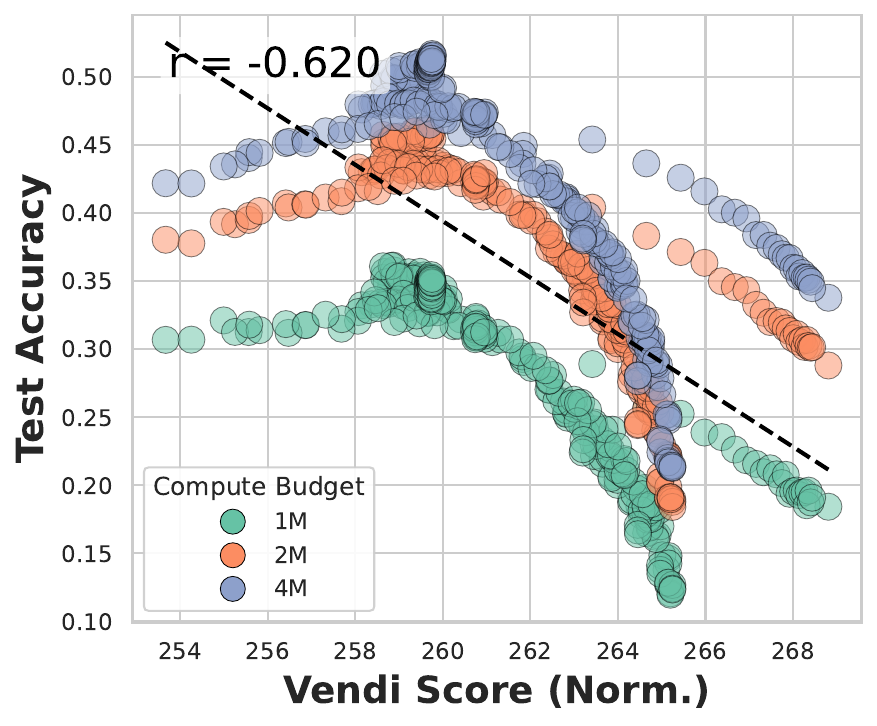}
        \caption{Vendi Score, Balanced}
        \label{fig:fig3c_10pct}
    \end{subfigure}\begin{subfigure}[t]{.248\textwidth}
        \centering
        \includegraphics[width=\linewidth]{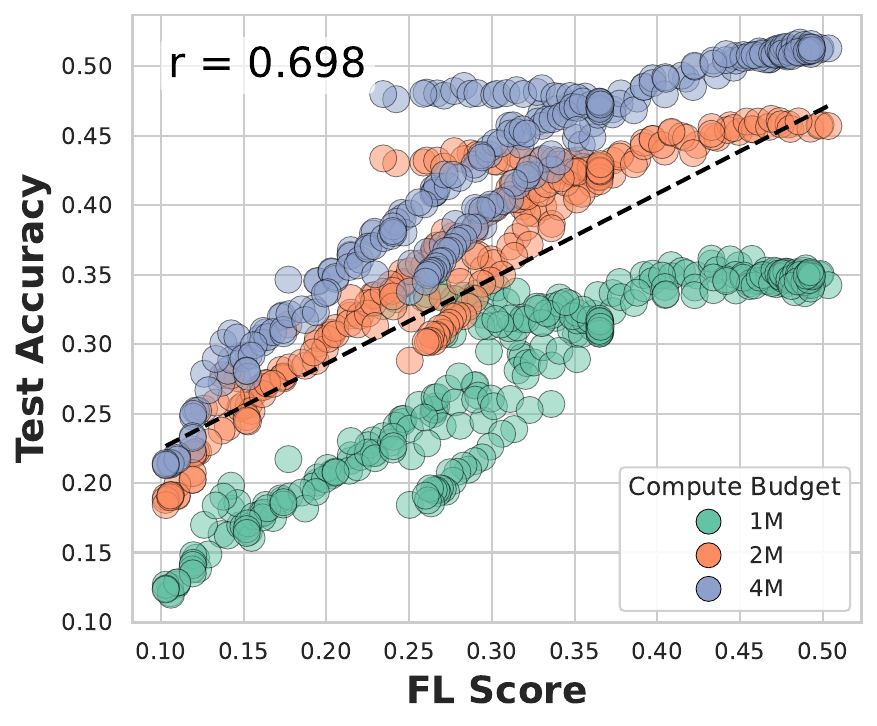}
        \caption{FL, Balanced}
        \label{fig:fig3d_10pct}
    \end{subfigure}
    \caption{\small \textbf{10\% Dataset Size}  Each run represents the outcome of an ImageNet-1K training runs spanning 3 different compute budgets on 10\% subsets. The subsets in (c) and (d) are all perfectly class-balanced.  We observe that there are many instances where a run with a low compute budget outperforms ones with a high compute budgets, suggesting that compute alone is a poor proxy for test accuracy.}
    \label{fig:fig3_10pct}
\end{figure*}

\begin{figure*}[tbh]
    \centering
    \begin{subfigure}[t]{.248\textwidth}
        \centering
        \includegraphics[width=\linewidth]{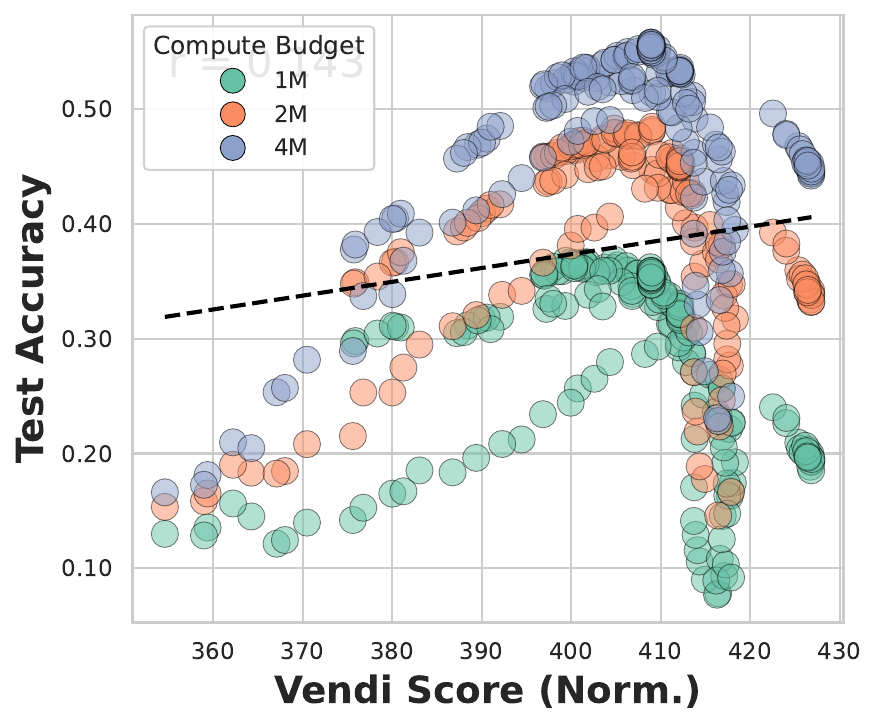}
        \caption{Vendi Score}
        \label{fig:fig3a_20pct}
    \end{subfigure}\begin{subfigure}[t]{.248\textwidth}
        \centering
        \includegraphics[width=\linewidth]{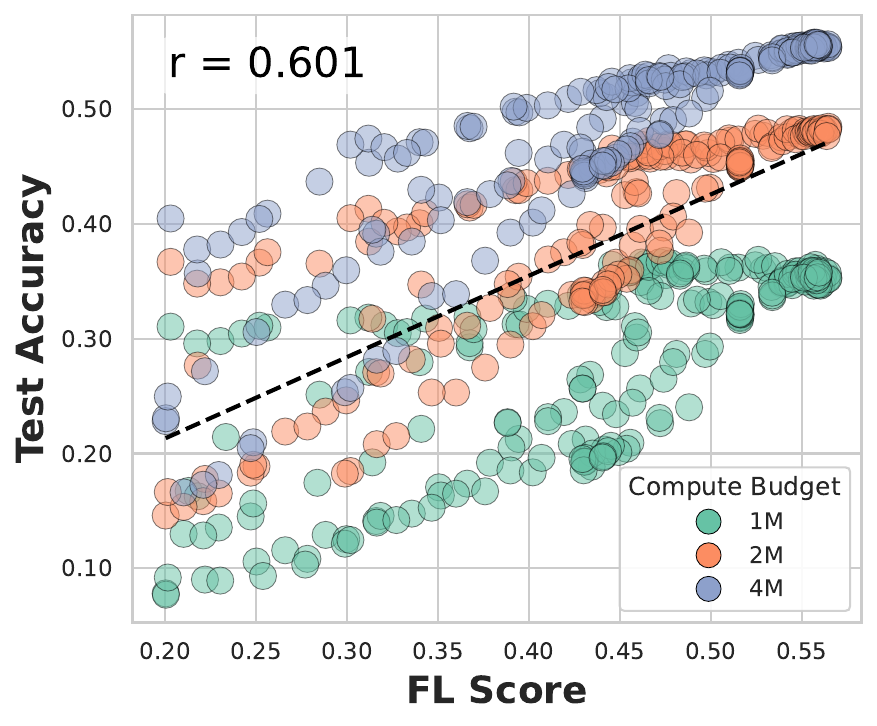}
        \caption{FL}
        \label{fig:fig3b_20pct}
    \end{subfigure}\begin{subfigure}[t]{.248\textwidth}
        \centering
        \includegraphics[width=\linewidth]{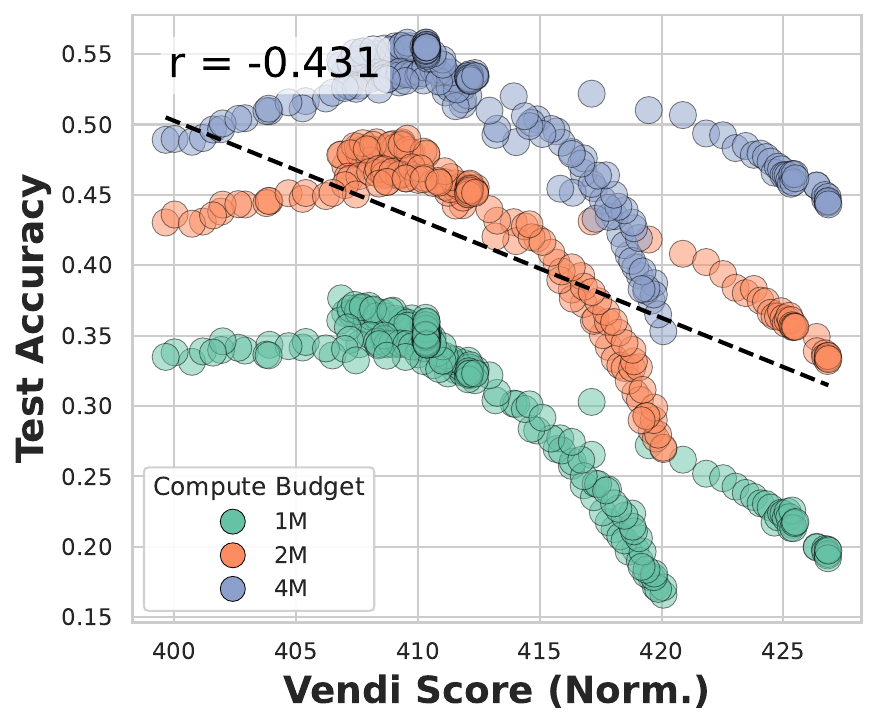}
        \caption{Vendi Score, Balanced}
        \label{fig:fig3c_20pct}
    \end{subfigure}\begin{subfigure}[t]{.248\textwidth}
        \centering
        \includegraphics[width=\linewidth]{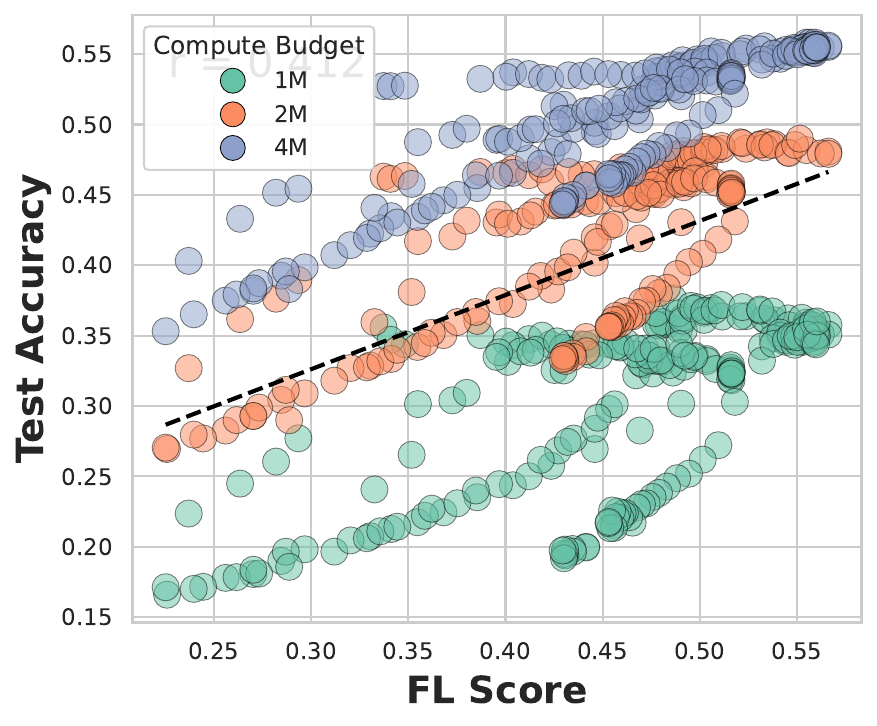}
        \caption{FL, Balanced}
        \label{fig:fig3d_20pct}
    \end{subfigure}
    \caption{\small \textbf{20\% Dataset Size}  Each run represents the outcome of an ImageNet-1K training runs spanning 3 different compute budgets on 20\% subsets. The subsets in (c) and (d) are all perfectly class-balanced.  We observe that there are many instances where a run with a low compute budget outperforms ones with a high compute budgets, suggesting that compute alone is a poor proxy for test accuracy.}
    \label{fig:fig3_20pct}
\end{figure*}

In~\Cref{fig:fig3_5pct,fig:fig3_10pct,fig:fig3_20pct}, all runs share identical hyperparameter configurations except for the compute budget, which is varied across 1M, 2M, and 4M samples seen. Within each plot, all subsets have the same cardinality; in plots (c) and (d), they are additionally constrained to be class-balanced. As before, FL exhibits a significantly higher correlation with test accuracy than the Vendi score. We further observe that compute is a poor proxy for test accuracy: across all subplots and dataset sizes, there are several instances where training on a poor subset with a 4M-sample compute budget yields worse performance than training on a good subset with a 1M-sample budget.

\subsection{Experiments with Additional Appraisal Functions}
\label{app:additional_appraisal_functions}
In the main paper, we focused on two functions, Vendi (with normalized gradients as the design matrix) and Facility Location. We now study several additional functions and, for completeness, restate the ones from the main paper. Given a design matrix \data, we form the linear-kernel Gram matrix $B= \data\data^\top$, and for any subset $X \subseteq V$ with $V = [n]$ we write $B_X = \data[X] \data[X]^\top$, where $\data[X]$ is the row-submatrix of \data,  indexed by $X$. Each matrix spectral function takes the form $f(X) = \mathrm{tr}[\phi(B_X)]$ for some scalar $\phi : \mathbb{R}_+ \to \mathbb{R}$. 

We consider the following six functions, in addition to the Facility Location function, where the notations are taken from Section~\ref{sec:weakly-matrix-monotone}. 

\begin{itemize}
    \item \textbf{Facility Location} (not a matrix spectral function):
    $f_{\mathrm{FL}}(X) = \sum_{j \in V} \max_{i \in X} s_{ij}$, where $s_{ij} \ge 0$ is a similarity.
    \item \textbf{Vendi Score} $\xi_1(x) = -(t+x)\log(t+x)$, giving $f(X) = \mathrm{tr}[-B_X \log B_X]$.
    \item \textbf{DPP} $\xi_2(x) = \log(t + x)$ for $t \geq 0$, giving $f(X) = \log\det(tI + B_X)$.
    \item $\xi_3(x) = x^\eta$ for $0 < \eta \le 1$.
    \item $\mfa(x) = 1 - (x + \beta)^{-\alpha}$ for $\alpha, \beta > 0$.
    \item $\mfc(x) = x / (1 + x^\alpha)^{1/\alpha}$ for $\alpha > 0$.
\end{itemize}

For the Facility Location function, we use the RBF Kernel as a similarity metric. Since FL doesn't require a symmetric similarity matrix, to make it scalable at the ImageNet scale, we sparsify the rows of the matrix by retaining only the top-k values. This significantly reduces the memory load. 

\textbf{Choice of design matrix on ImageNet.} The functions above are defined
for any design matrix. In the main paper, the design matrix consists of
per-example normalized gradients $\hat x_i = x_i / \|x_i\|_2$, following
\citet{jung2025prismatic}. Normalization discards the gradient magnitude
$\|x_i\|_2$, which is itself informative: examples with large gradients are
harder for the reference model to fit and arguably contribute more during
training. Retaining the magnitude lets matrix spectral functions weight examples
by importance rather than by direction alone, analogous to the practice in DPPs
of scaling the kernel diagonal by per-example quality scores. Normalization can
also amplify noise on examples whose gradients are small to begin with, since
dividing a near-zero vector by its norm magnifies whatever directional component
happens to dominate. Finally, submodularity of the matrix spectral functions
does not depend on the row norms of $\data$, so the approximation guarantees of
greedy maximization are preserved. We therefore additionally evaluate Vendi on
unnormalized gradients. In subsequent sections, Vendi (Unnorm.) refers to this
variant of the design matrix, and we will refer to Vendi (Norm.) where we use
normalized gradients. 

\begin{figure*}[tbh]
    \centering
\begin{subfigure}[t]{.32\textwidth}
        \centering
        \includegraphics[width=\linewidth]{figs/single_panels_all/single_panels_10pct/vendi_norm_cardinality_10pct_40ep.pdf}
        \caption{Vendi (Normalized)}
        \label{fig:vendi_norm_cardinality}
    \end{subfigure}\hfill
    \begin{subfigure}[t]{.32\textwidth}
        \centering
        \includegraphics[width=\linewidth]{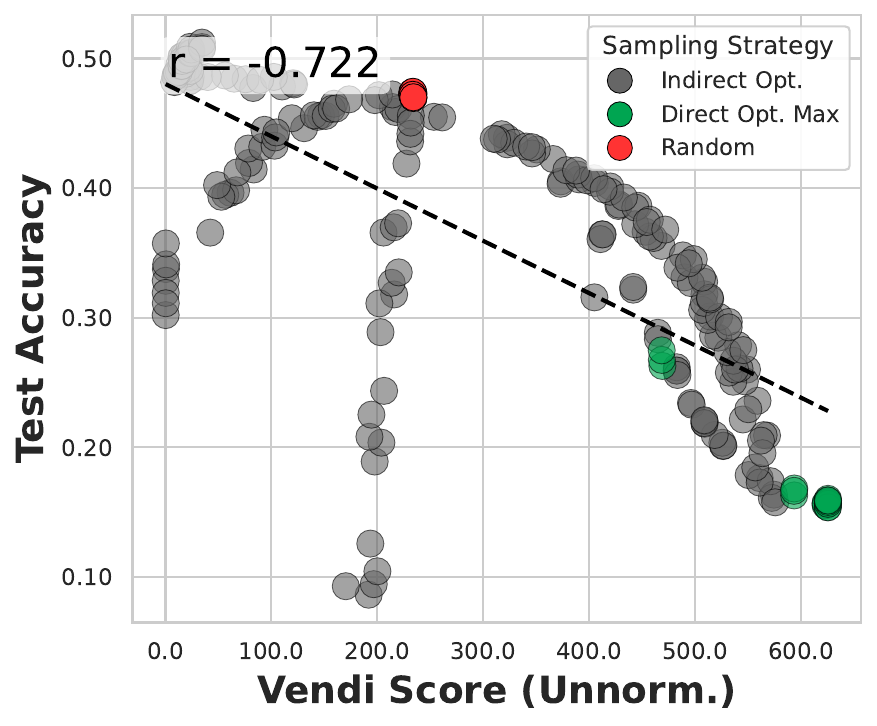}
        \caption{Vendi (Unnormalized)}
        \label{fig:vendi_unnorm_cardinality}
    \end{subfigure}\hfill
    \begin{subfigure}[t]{.32\textwidth}
        \centering
        \includegraphics[width=\linewidth]{figs/single_panels_all/single_panels_10pct/fl_cardinality_10pct_40ep.pdf}
        \caption{FL}
        \label{fig:fl_cardinality}
    \end{subfigure}
\begin{subfigure}[t]{.32\textwidth}
        \centering
        \includegraphics[width=\linewidth]{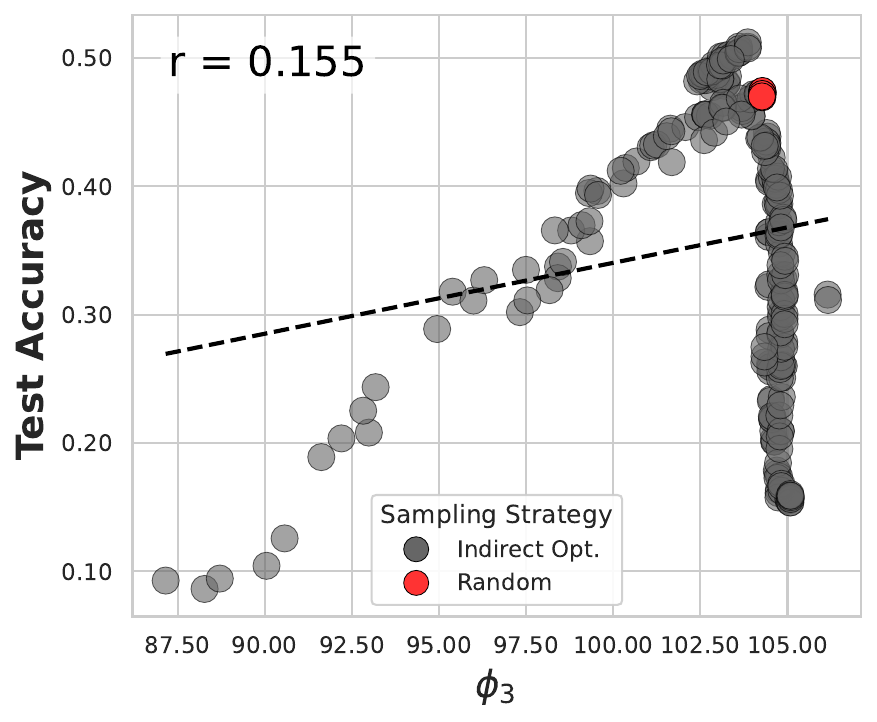}
        \caption{\mfc}
        \label{fig:bmin_cardinality}
    \end{subfigure}\hfill
    \begin{subfigure}[t]{.32\textwidth}
        \centering
        \includegraphics[width=\linewidth]{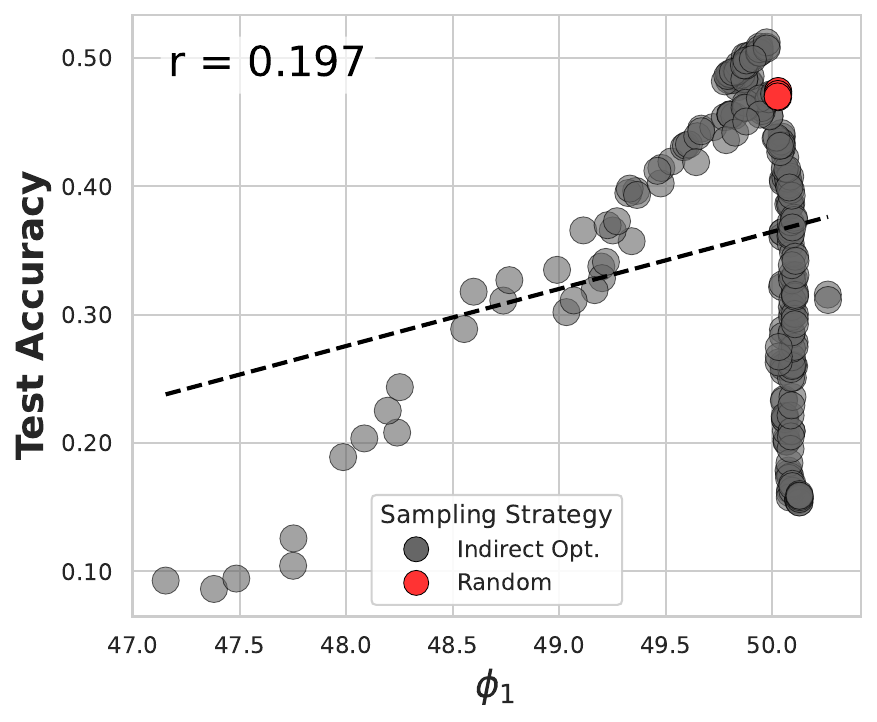}
        \caption{\mfa}
        \label{fig:plaw_cardinality}
    \end{subfigure}\hfill
    \begin{subfigure}[t]{.32\textwidth}
        \centering
        \includegraphics[width=\linewidth]{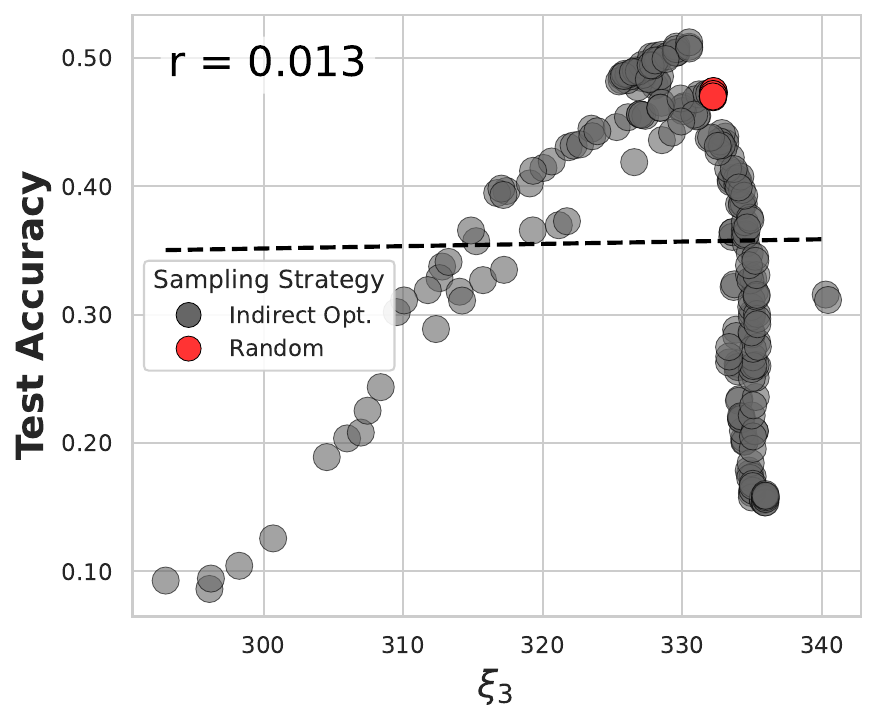}
        \caption{$\xi_3$}
        \label{fig:pow_cardinality}
    \end{subfigure}
    \begin{subfigure}[t]{.32\textwidth}
        \centering
        \includegraphics[width=\linewidth]{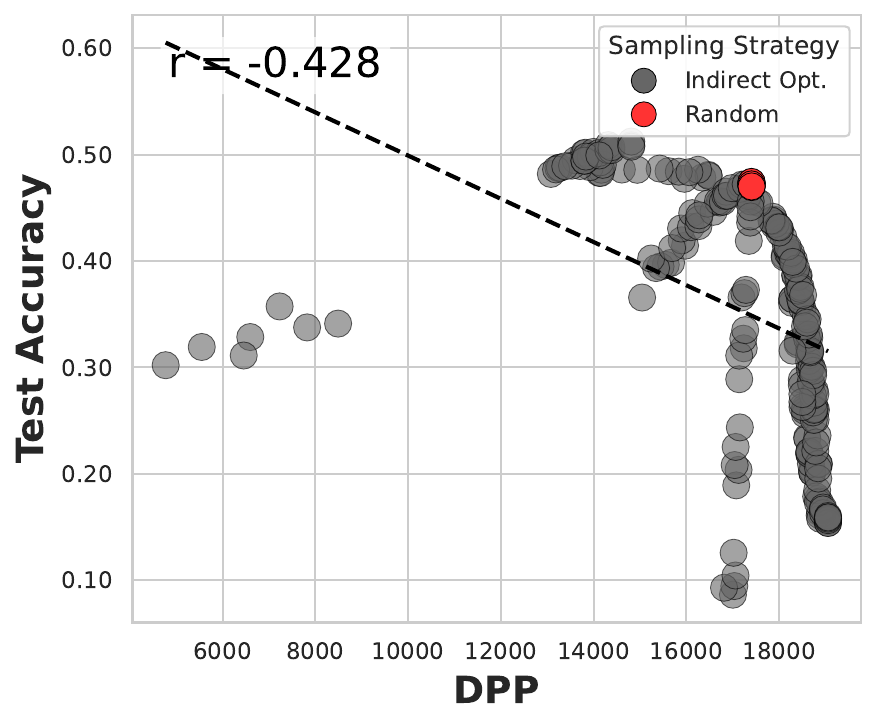}
        \caption{DPP}
        \label{fig:dpp_cardinality}
    \end{subfigure}
    \vspace{-.1in}

\caption{\small \textbf{Comparison of All Appraisal Functions on ImageNet, 10\% subsets, 4M samples seen.} Each run represents the outcome of an ImageNet-1K training run. Facility Location is the only function strongly correlated with accuracy ($r = 0.917$). Vendi (Normalized) is essentially uncorrelated, and Vendi (Unnormalized) is negatively correlated: subsets with higher Vendi-Unnorm scores yield lower test accuracy. The weakly matrix monotone functions $\phi_3$, $\phi_1$, and $\xi_3$ all sit near zero. Across the matrix spectral panels, we see an inverted-U shape: accuracy rises with score up to a peak and then drops sharply. For the Vendi score in particular, inverted-U also offers a likely explanation for the positive correlations reported in Prismatic Synthesis~\citep{jung2025prismatic}: the indirect samples from that work populate only the ascending branch of the curve, and a positive trend on that side does not extrapolate past the peak.  The latter segment is evaluated on sets that were either sampled by maximizing the FL function or directly maximizing the Vendi score and previously unexplored by \citep{jung2025prismatic}. Reaching the highest value of Vendi score was not previously feasible at this scale and is enabled here by the secular-equation of Section~\ref{sec:solv-secul-equat}. Moreover, \textbf{random subsets are tightly concentrated} in both test accuracy and function value. For the definitions of $\phi_1$, $\phi_3$, and $\xi_3$, see Section~\ref{sec:weakly-matrix-monotone}.}
\label{fig:cardinality_all_funcs}
\end{figure*}

\begin{figure*}[tbh]
    \centering
\begin{subfigure}[t]{.32\textwidth}
        \centering
        \includegraphics[width=\linewidth]{figs/single_panels_all/single_panels_10pct/vendi_norm_balance_10pct_4Msteps.pdf}
        \caption{Vendi (Normalized)}
        \label{fig:vendi_norm_balance}
    \end{subfigure}\hfill
    \begin{subfigure}[t]{.32\textwidth}
        \centering
        \includegraphics[width=\linewidth]{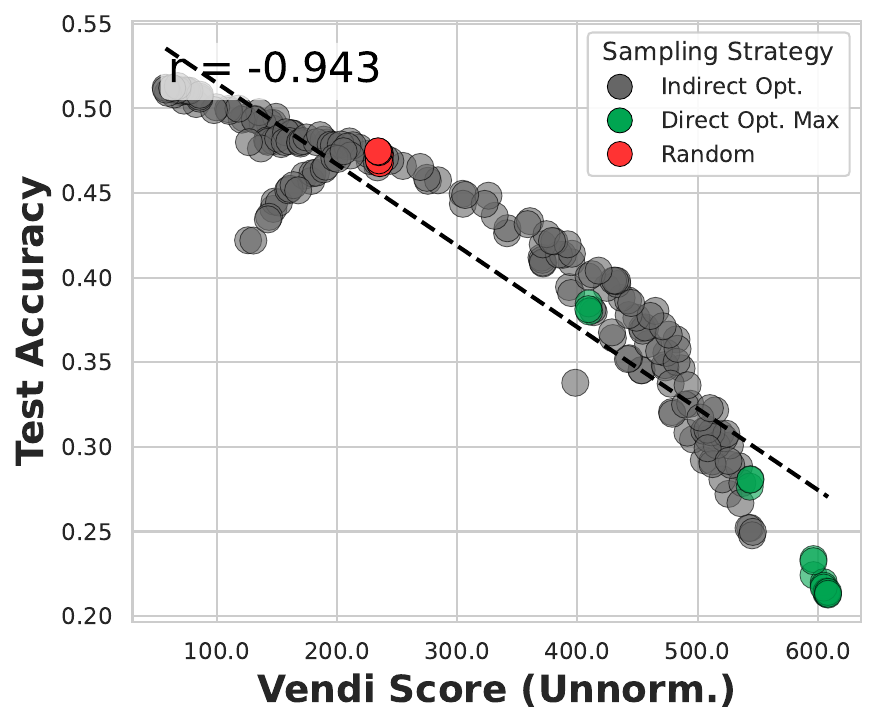}
        \caption{Vendi (Unnormalized)}
        \label{fig:vendi_unnorm_balance}
    \end{subfigure}\hfill
    \begin{subfigure}[t]{.32\textwidth}
        \centering
        \includegraphics[width=\linewidth]{figs/single_panels_all/single_panels_10pct/fl_balance_10pct_40ep.pdf}
        \caption{FL}
        \label{fig:fl_balance}
    \end{subfigure}

    \vspace{0.5em}

\begin{subfigure}[t]{.32\textwidth}
        \centering
        \includegraphics[width=\linewidth]{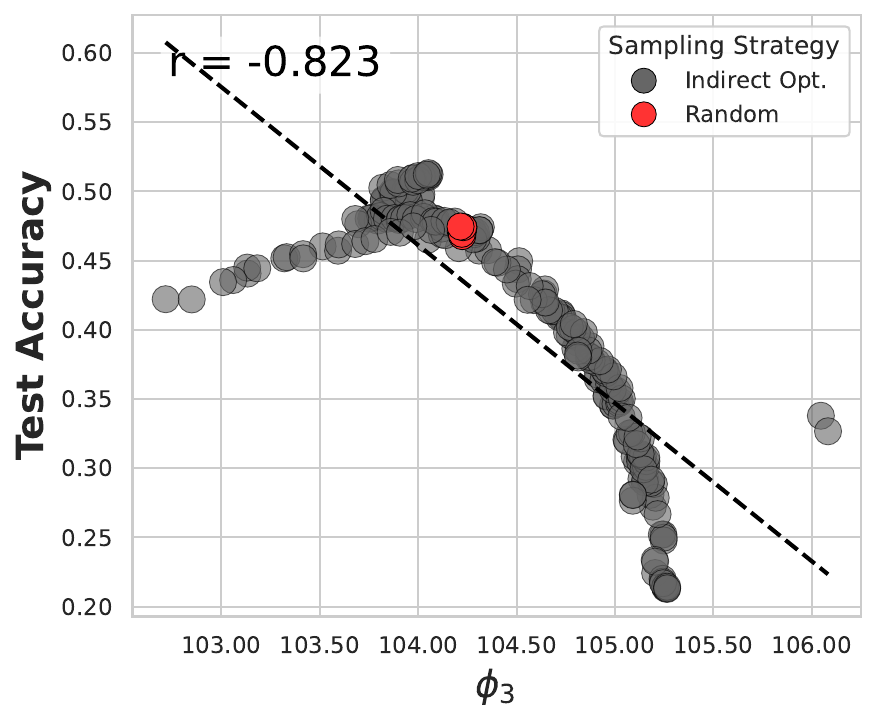}
        \caption{\mfc}
        \label{fig:bmin_balance}
    \end{subfigure}\hfill
    \begin{subfigure}[t]{.32\textwidth}
        \centering
        \includegraphics[width=\linewidth]{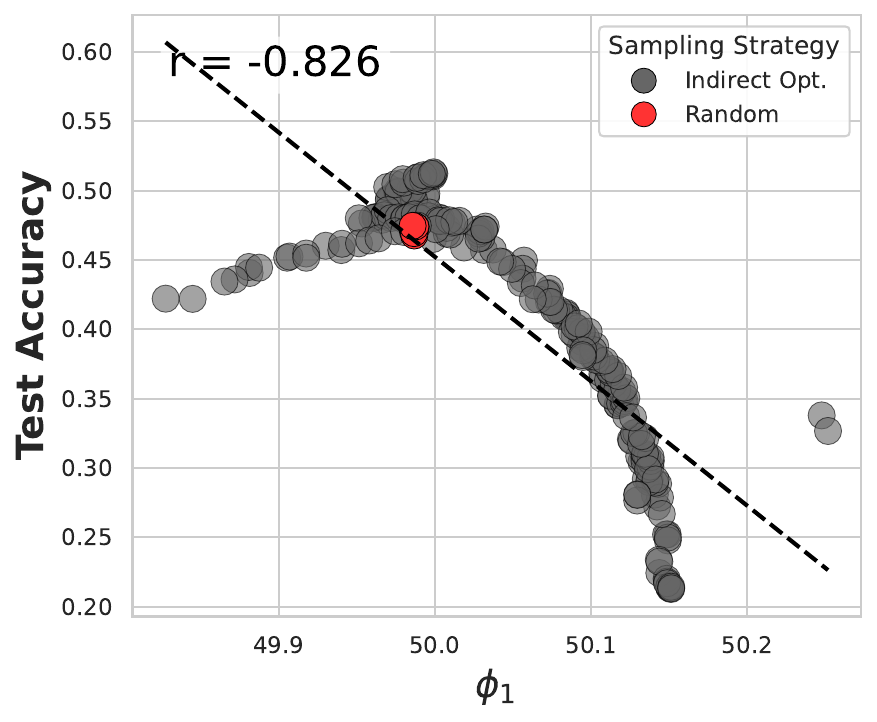}
        \caption{\mfa}
        \label{fig:plaw_balance}
    \end{subfigure}\hfill
    \begin{subfigure}[t]{.32\textwidth}
        \centering
        \includegraphics[width=\linewidth]{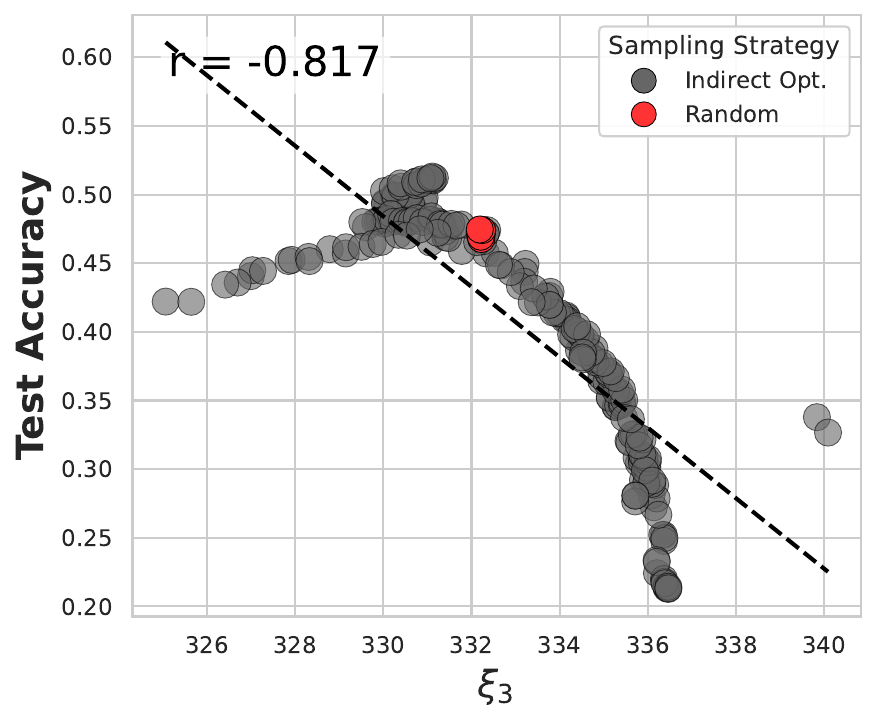}
        \caption{$\xi_3$}
        \label{fig:pow_balance}
    \end{subfigure}
        \begin{subfigure}[t]{.32\textwidth}
        \centering
        \includegraphics[width=\linewidth]{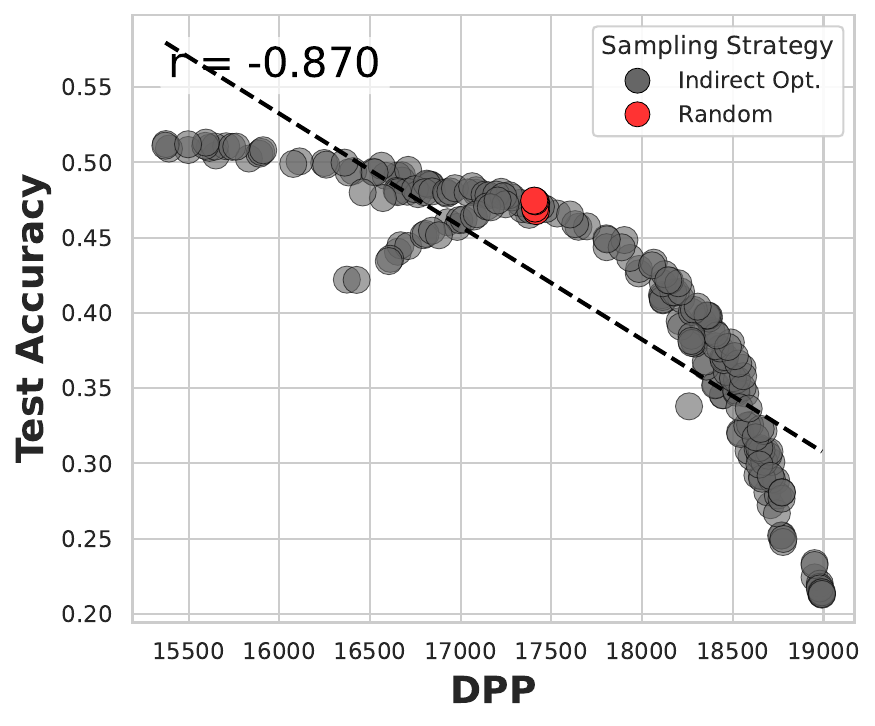}
        \caption{DPP}
        \label{fig:dpp_balance}
    \end{subfigure}
    \vspace{-.1in}
\caption{\small \textbf{Comparison of All Appraisal Functions on ImageNet, 10\%
class-balanced subsets, 4M samples seen.} Each run represents the outcome of an
ImageNet-1K training run. Same setup as Figure~\ref{fig:cardinality_all_funcs},
but with all subsets constrained to be perfectly class-balanced. As in
Figure~\ref{fig:cardinality_all_funcs}, each panel plots function score (x-axis)
against ResNet-18 test accuracy (y-axis), with points colored by sampling
strategy and Pearson $r$ shown. Facility Location remains strongly positively
correlated with accuracy ($r = 0.930$), consistent with the unconstrained
setting. Fixing class balance effectively removes the ascending branch seen
previously, leaving a near-monotone regime where higher spectral scores predict
lower accuracy, while Facility Location remains aligned with downstream
performance. Therefore, highlighting the fact that \textbf{class balance alone
is not enough} for achieving good performance. Moreover, \textbf{random
class-balanced subsets are tightly concentrated} in both test accuracy and
function value. For definitions of $\phi_1$, $\phi_3$, and $\xi_3$, see
Section~\ref{sec:weakly-matrix-monotone}.}
    \label{fig:balance_all_funcs}
\end{figure*}

Figures~\ref{fig:cardinality_all_funcs} and~\ref{fig:balance_all_funcs} show two consistent patterns. First, Facility Location is the only function whose score predicts downstream test accuracy across both the cardinality-constrained and class-balanced settings ($r = 0.917$ and $r = 0.930$). Every matrix spectral function we tested fails on at least one of the two settings: in the cardinality-constrained case, the matrix spectral functions are weakly correlated or uncorrelated with accuracy, while in the class-balanced case, they flip sign and become strongly negatively correlated. Second, the matrix spectral panels in Figure~\ref{fig:cardinality_all_funcs} show a clear inverted-U: accuracy rises with the function value up to a peak and then collapses, with the highest-scoring subsets (those produced by Direct Max) landing on the descending side. Maximizing Vendi or any of the weakly matrix monotone functions, therefore, actively selects subsets that perform worse than other subsets under the same function. The effect is strongest for Vendi (Unnormalized), where the Direct Max subsets sit close to the bottom of the test-accuracy axis. The inverted-U also offers a likely explanation for the positive correlations reported in Prismatic Synthesis~\citep{jung2025prismatic}: the indirect samples from that work populate only the ascending branch of the curve, and a positive trend on that side does not extrapolate past the peak. The latter segment is evaluated on sets that were either sampled by maximizing the FL function or directly maximizing the Vendi score and previously unexplored by \citep{jung2025prismatic}. Reaching the highest value of Vendi score was not previously feasible at this scale and is enabled here by the secular-equation of Section~\ref{sec:solv-secul-equat}. We do not run Direct Max on $\phi_3$, $\phi_1$, or $\xi_3$ because hyperparameter tuning at ImageNet scale is prohibitively expensive. The indirect-sample inverted-U is already clear in these panels, and the same pattern is expected to hold.

\clearpage

\subsection{Qualitative Analysis of High and Low-Valued Subsets}
\label{app:viz_good_bad_sets}

\begin{figure}[h]
    \centering
    \begin{subfigure}[t]{0.48\textwidth}
        \centering
        \includegraphics[width=\linewidth]{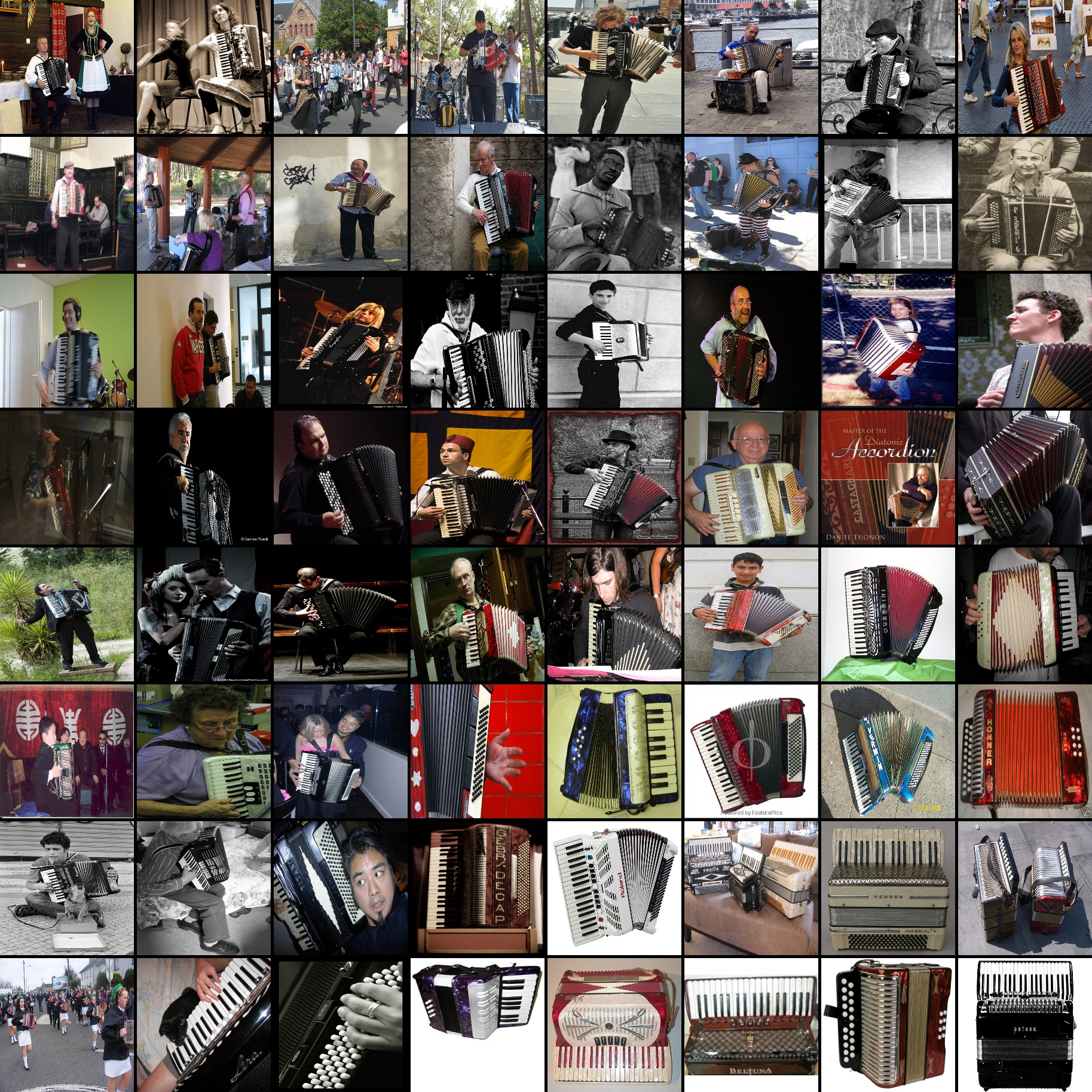}
        \caption{High-value (high facility location score) subset (\textcolor{FigGreen}{green \myxshape} 
        in~\Cref{fig:main-teaser}
          just for accordion)}
    \end{subfigure}\hspace{0.02\textwidth}
    \begin{subfigure}[t]{0.48\textwidth}
        \centering
        \includegraphics[width=\linewidth]{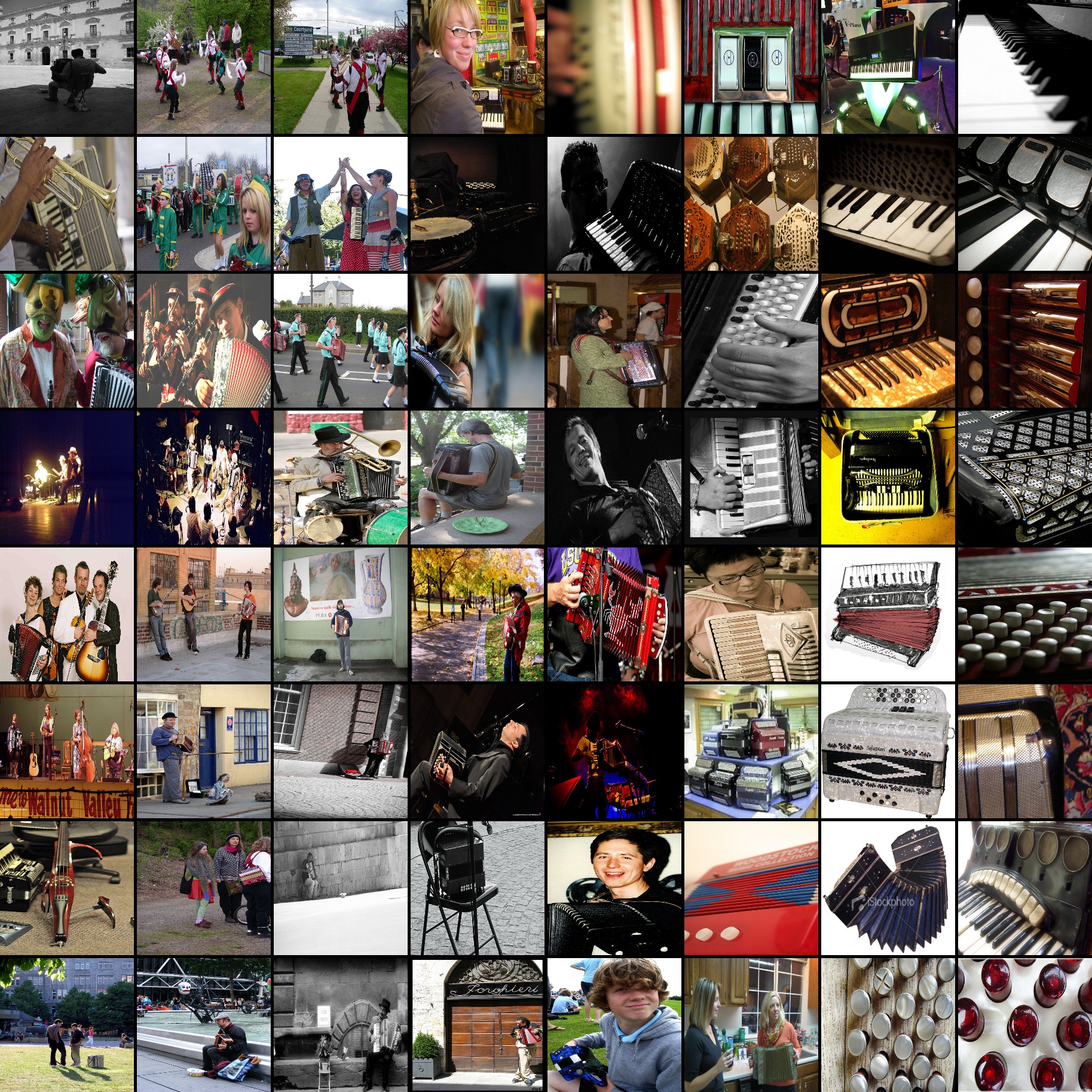}
        \caption{Low-value (low facility location score) subset (\textcolor{FigBlue}{blue \myxshape} 
        in~\Cref{fig:main-teaser}
          just for accordion)}
    \end{subfigure}
    \caption{\small \textbf{Accordion.} Images in the high-value (high facility
    location score) subset (above left $8 \times 8$ grid, and \textcolor{FigGreen}{green \myxshape}
    in~\Cref{fig:main-teaser}) are canonical examples in which an accordion
    is clearly the, or at least a, central subject. In contrast, the low-value (low facility
    location score) subset (above right $8 \times 8$ grid, and \textcolor{FigBlue}{blue \myxshape} in~\Cref{fig:main-teaser}) contains many much more atypical images with confounding objects. For example the
    image in row 6 column 1, contains several other instruments and the
    accordion is hardly visible. In row 5, column 1, we see an entire
    band which includes guitar and people, and a better single label for that
    image might be ``band'' or ``musical group'' (which are not ImageNet-1K classes).
    Other images, such as the ones shown in row 8
    column 7, row 8 column 8, and row 1 column 5, appear to not contain anything resembling the
    totality of the accordion ``concept'' at all and may simply be instances of label noise,
    or might be extreme closeups of the buttons in a button accordion, which are not visually similar 
    to the more canonical or typical examples of what defines an accordion visually.
    }
    \label{fig:viz_accordian}
\end{figure}

\begin{figure}[h]
    \centering
    \begin{subfigure}[t]{0.48\textwidth}
        \centering
        \includegraphics[width=\linewidth]{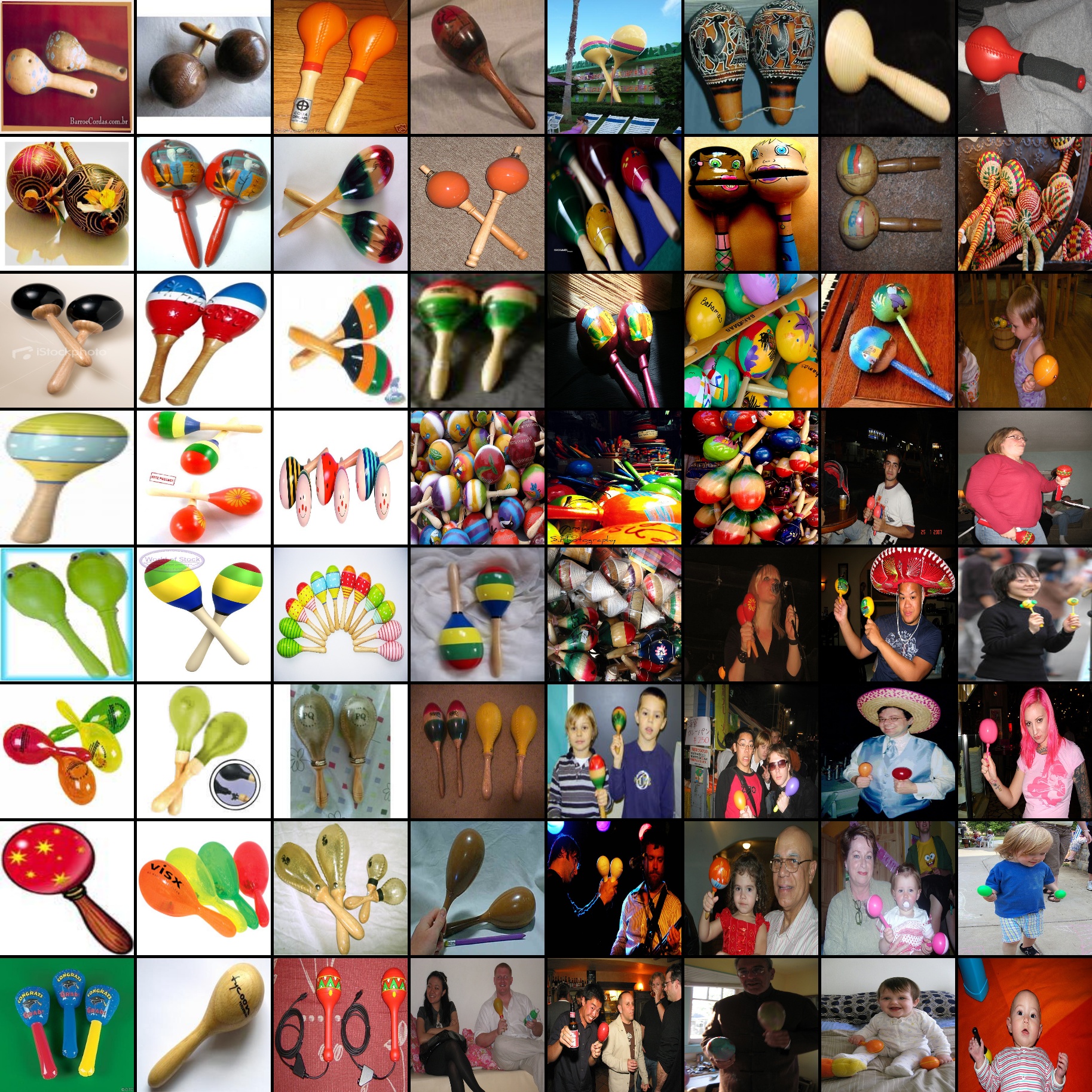}
        \caption{High-value (high facility location score) subset (\textcolor{FigGreen}{green \myxshape}
        in~\Cref{fig:main-teaser}
          just for maracas)}
    \end{subfigure}\hspace{0.02\textwidth}
    \begin{subfigure}[t]{0.48\textwidth}
        \centering
        \includegraphics[width=\linewidth]{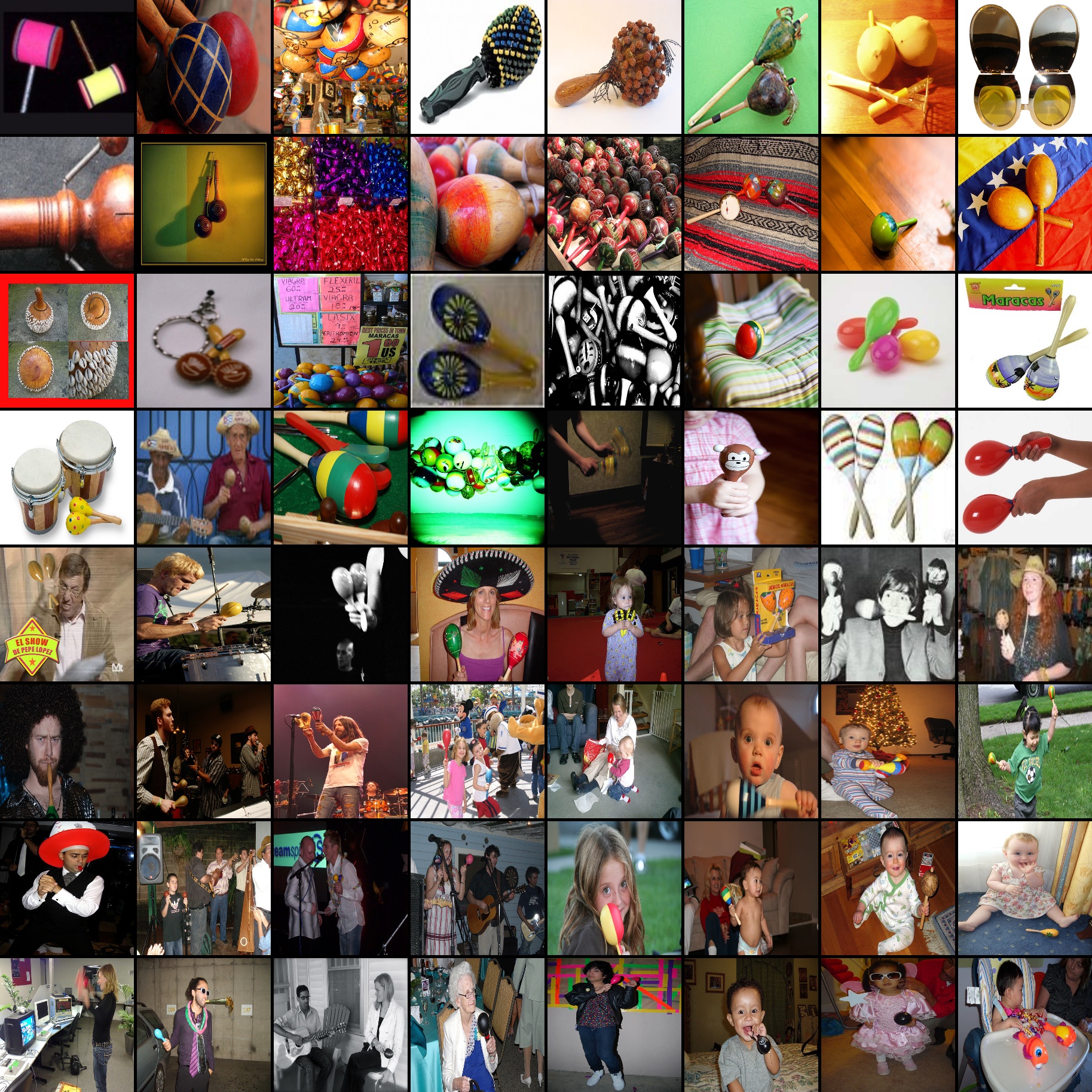}
        \caption{Low-value (low facility location score) subset (\textcolor{FigBlue}{blue \myxshape} 
        in~\Cref{fig:main-teaser} just for maracas)}
    \end{subfigure}
    \caption{\small \textbf{Maracas.} Images in the high-value (high facility
    location score) subset (above left $8 \times 8$ grid, and
    \textcolor{FigGreen}{green \myxshape} in~\Cref{fig:main-teaser}) are
    canonical examples in which the maracas are clearly the central subject. In
    contrast, the low-value (low facility location score) subset (above right $8
    \times 8$ grid, and \textcolor{FigBlue}{blue \myxshape}
    in~\Cref{fig:main-teaser})
    contains images focused on the players rather than the instrument itself
    (see bottom four rows, right grid). There are also clear instances of label
    noise, such as row~1, column~1, which depicts a pair of mallets with no
    maracas present. In other images, such as row~4, column~1, confounding
    objects from other ImageNet-1K classes are prominently visible, such as
    bongos.}
    \label{fig:viz_maracas}
\end{figure}

\begin{figure}[th]
    \centering
    \begin{subfigure}[t]{0.48\textwidth}
        \centering
        \includegraphics[width=\linewidth]{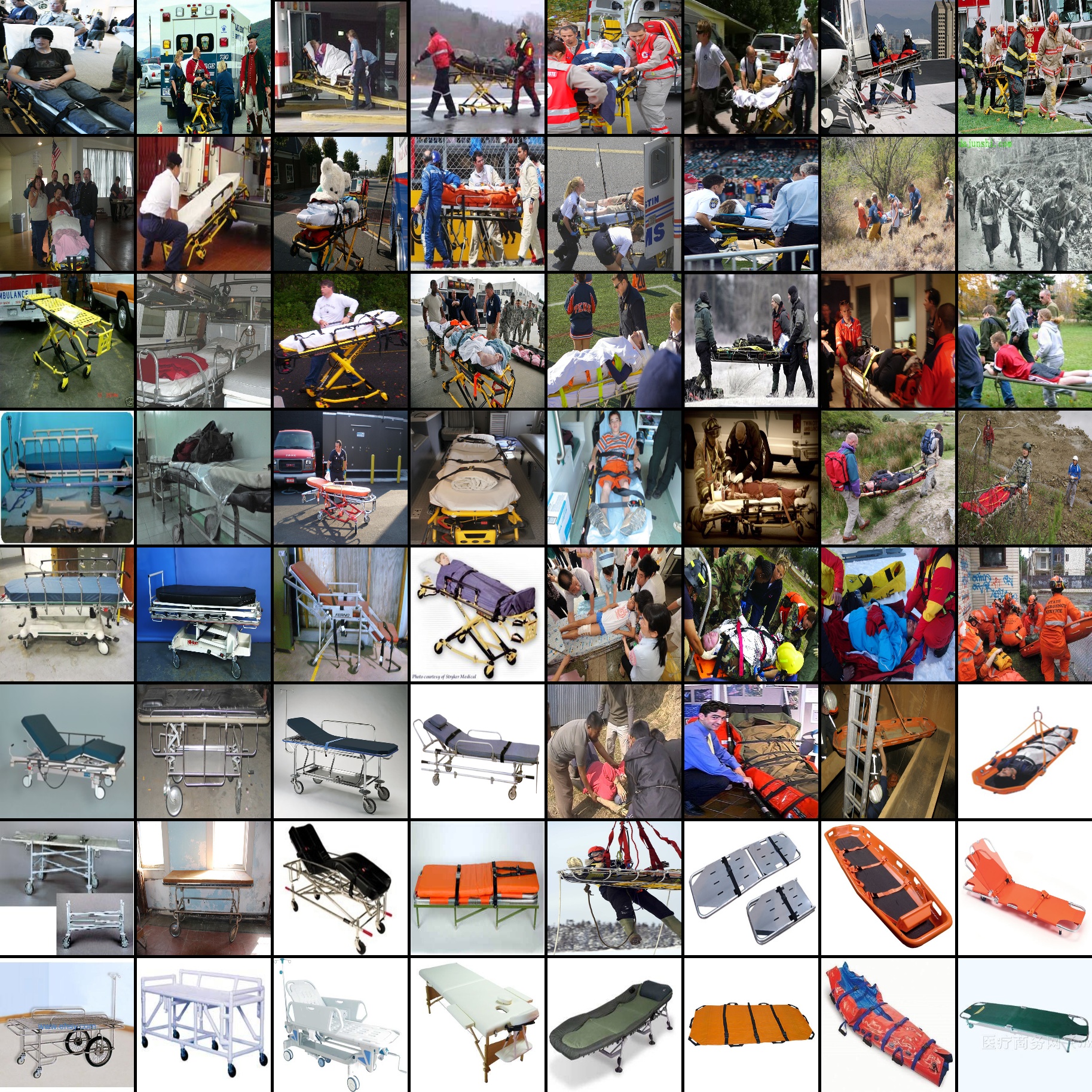}
        \caption{High-value (high facility location score) subset (\textcolor{FigGreen}{green \myxshape} 
        in~\Cref{fig:main-teaser} just for stretcher)}
    \end{subfigure}\hspace{0.02\textwidth}
    \begin{subfigure}[t]{0.48\textwidth}
        \centering
        \includegraphics[width=\linewidth]{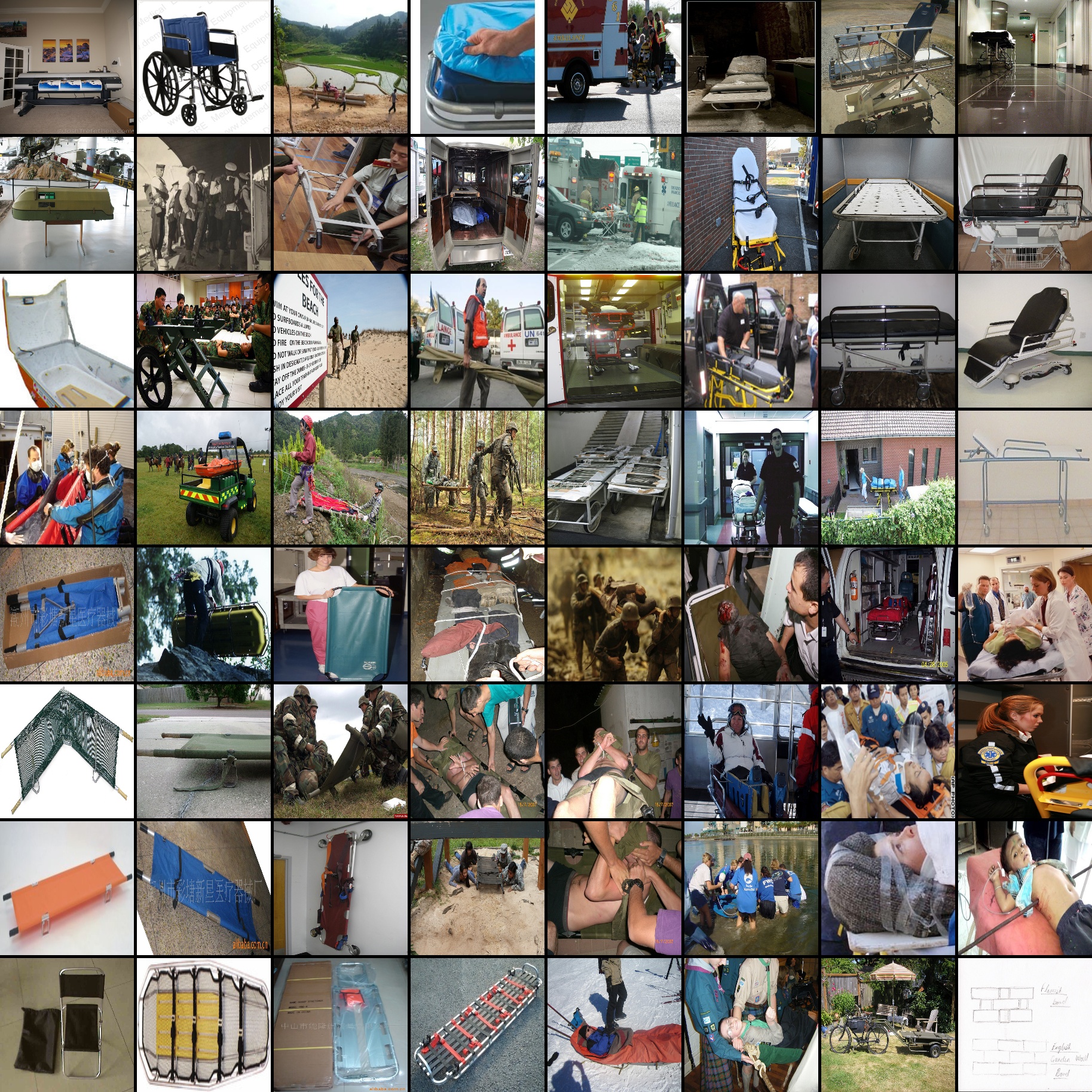}
        \caption{Low-value (low facility location score) subset (\textcolor{FigBlue}{blue \myxshape} 
        in~\Cref{fig:main-teaser} just for stretcher)}
    \end{subfigure}
    \caption{\small \textbf{Stretcher.} Images in the high-value (high facility
    location score) subset (above left $8 \times 8$ grid, and \textcolor{FigGreen}{green \myxshape}
    in~\Cref{fig:main-teaser}) are canonical examples in which a stretcher
    is clearly the central subject. In contrast, the low-value (low facility
    location score) subset (above right $8 \times 8$ grid, and \textcolor{FigBlue}{blue \myxshape}
    in~\Cref{fig:main-teaser}) contains label noise (e.g., the image in row~1
    column~2 which contains a wheelchair mislabeled as a stretcher or row~8
    column~8 which contains a drawing of something that does not resemble a
    stretcher and instead seems to be a hand-drawing of a brick wall). 
    Other images, such as row~7 column~7, are focused on patients
    and the stretcher is hardly visible.}
    \label{fig:viz_stretcher}
\end{figure}

As demonstrated in~\Cref{fig:class-balanced-all-merged-teaser} and elsewhere,
there is a significant difference in test accuracy between subsets with low and
high Facility Location values even when subsets are perfectly class-balanced.
In~\Cref{fig:viz_accordian,fig:viz_maracas,fig:viz_stretcher}, we visualize all
samples from three classes drawn from a class-balanced 5\% subset, contrasting
the high-value high facility location score (left) and low-value low facility
location score (right) selections in each case. We note that some of the 1000
categories in ImageNet-1K are much more homogeneous and canonical than others,
so the visual differences between high and low valued subsets in those
categories might be less pronounced. The three categories
we selected below, however, are likely \textbf{not} examples where there are the
most pronounced differences between high and low valued; these three were selected
by randomly selecting 30 out of the 1000 categories, and then picking three
with clearer visual differences. 

In all of the examples, high-valued subsets tend to consist of clean, canonical
images in which the target object is prominently featured and centered, whereas
low-valued subsets consist of less typical examples, which includes cases such
as potential label noise, images containing multiple objects, and images where
the target object is small, peripheral, and not centrally focused. The existence
of mislabeled images, and images that should have multiple labels, in
ImageNet-1K is a phenomenon that has been well
documented~\cite{yun2021re,kisel2025flawsofimagenet}; we posit that applying
heuristic greedy minimization using a facility location objective, or some other
appropriate submodular function, might be an effective general strategy to
identify such samples. Regarding the facility location function, recall that it
has the form given in~\Cref{eq:facility-location-function}, and when maximized
tends to select sets that are not only diverse but also that are representative
of other samples. When performing submodular min on this function, it appears
that the selected samples are not necessarily less diverse but seem likely to be
less representative of other samples. Without examining all of the 1M images in
the dataset, it is hard to visualize representativeness. Nevertheless,
considering that the low-valued subsets appear to contain more atypical examples
along with the fact that the low-valued subsets perform poorly when used as a
training set for a classifier (i.e.,~\Cref{fig:main-teaser}), provide evidence
that, in addition to diversity (which is the same as lack of redundancy), what
also is required to make a good training dataset is representativeness.
In general, a good training dataset needs to be a compact, non-redundant, representation
of the information that should be learned. That submodularity
is a good model for these concepts is further
discussed in~\cite{bilmes2022submodularity}.

\section{Additional Datasets: Categorical Coverage}
\label{app:additional_datasets}

Our ImageNet results show that class balance alone is not sufficient for
downstream performance, since even balanced subsets vary widely in test
accuracy. Producing those balanced subsets, however, required an explicit
partition matroid constraint to enforce per-class quotas. A good diversity
function, submodular or otherwise, should not need such scaffolding. If the
function genuinely captures coverage, then unconstrained greedy maximization
should already spread mass across the underlying categories. We probe this
directly on two datasets where the downstream metric is the categorical coverage
of the selected summary.

\textbf{AirBnB Duplicate Image Dataset.} The task is to pick a fixed-size
summary that represents as many of the 356 distinct rooms in the dataset as
possible. Listings are embedded with a \texttt{ViT-L/14 DINOv3} backbone, giving
feature vectors in $\mathbb{R}^m$ with $m = 1260$. We sweep two summary sizes $k
\in \{356, 600\}$. The dataset is publicly available on
Kaggle\footnote{\url{https://www.kaggle.com/datasets/barelydedicated/airbnb-duplicate-image-detection}}. 

\textbf{20 Newsgroups.} The task is to pick a fixed-size summary that covers as
many of the 20 newsgroups as possible. Each post is embedded with the Mixedbread
AI large model (\texttt{mxbai-embed-large-v1}), giving vectors of dimension $m =
1024$. We sweep two summary sizes $k \in \{20, 40\}$. The dataset is publicly
available over Sklearn.

On both datasets, we evaluate all six matrix spectral functions, tuning their
hyperparameters by maximizing the coverage achieved by the greedy summary at
each target budget, where the lowest budget corresponds to the distinct
categories in each. $\mfa$ and $\mfc$ are only weakly submodular, but greedy
maximization still admits an approximation guarantee in this case. Please refer
to Section~\ref{sec:weakly-matrix-monotone} for more details.

\subsection{Results: Categorical Coverage v/s Function Value}
\label{app:coverage_v_func_value}

\begin{figure*}[tbh]
    \centering
\begin{subfigure}[t]{.32\textwidth}
        \centering
        \includegraphics[width=\linewidth]{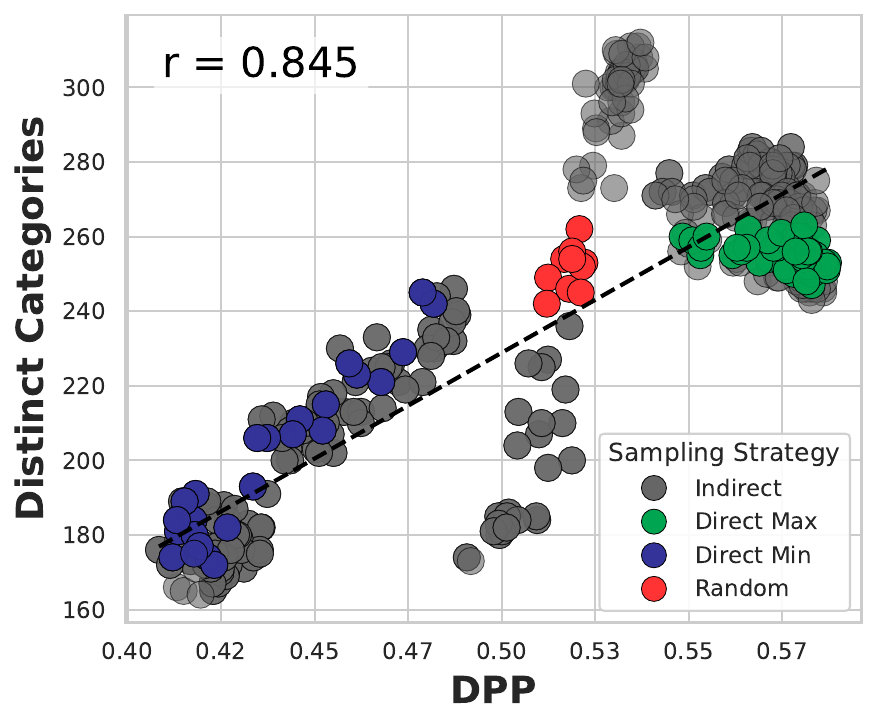}
        \caption{DPP}
        \label{fig:ldsop_budget356}
    \end{subfigure}\hfill
    \begin{subfigure}[t]{.32\textwidth}
        \centering
        \includegraphics[width=\linewidth]{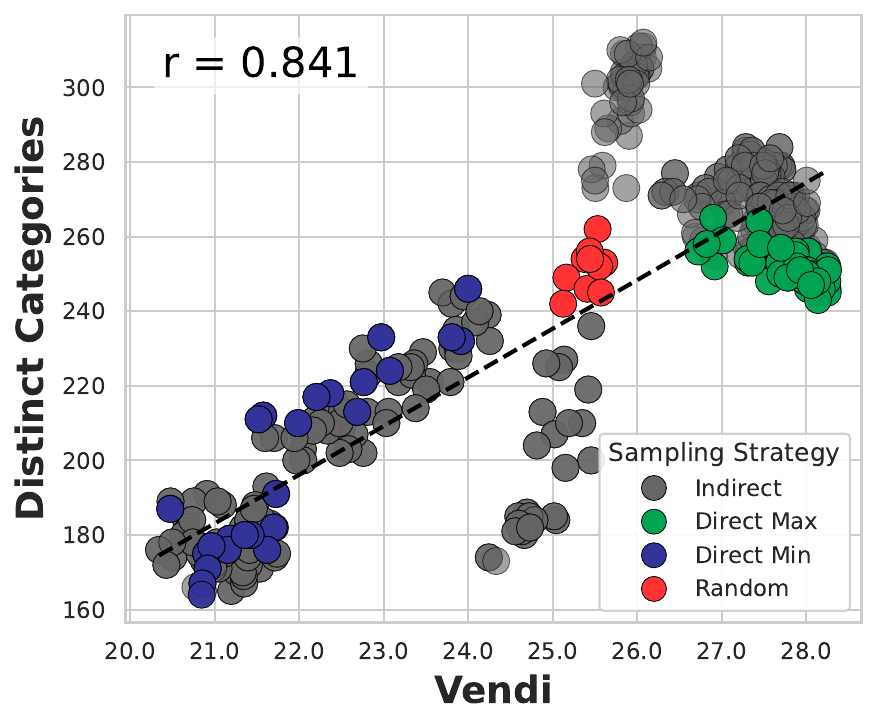}
        \caption{Vendi Score }
        \label{fig:nxlogx_budget356}
    \end{subfigure}\hfill
    \begin{subfigure}[t]{.32\textwidth}
        \centering
        \includegraphics[width=\linewidth]{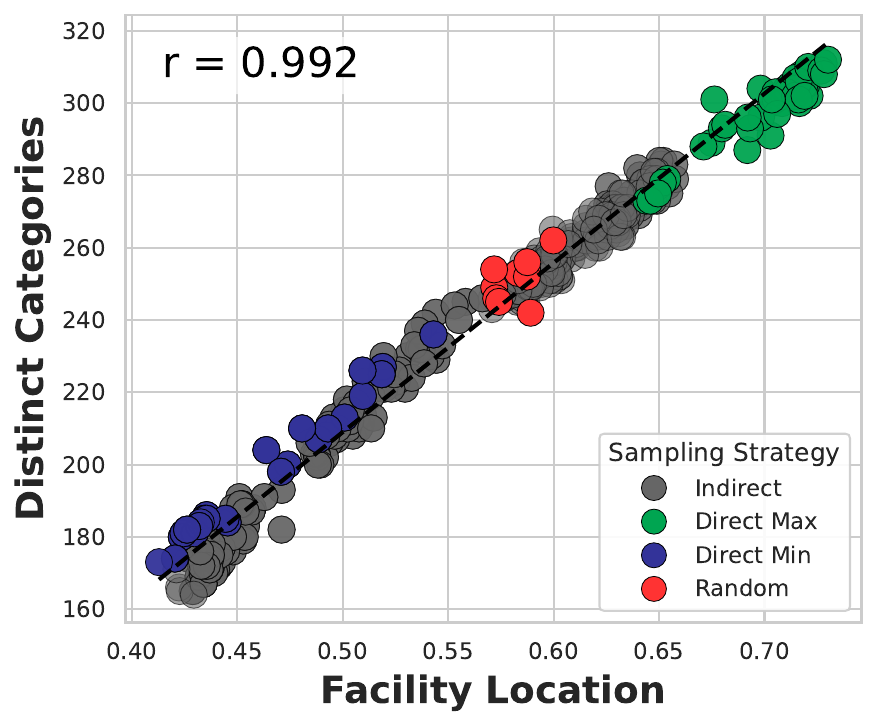}
        \caption{FL}
        \label{fig:fl_budget356}
    \end{subfigure}
\begin{subfigure}[t]{.32\textwidth}
        \centering
        \includegraphics[width=\linewidth]{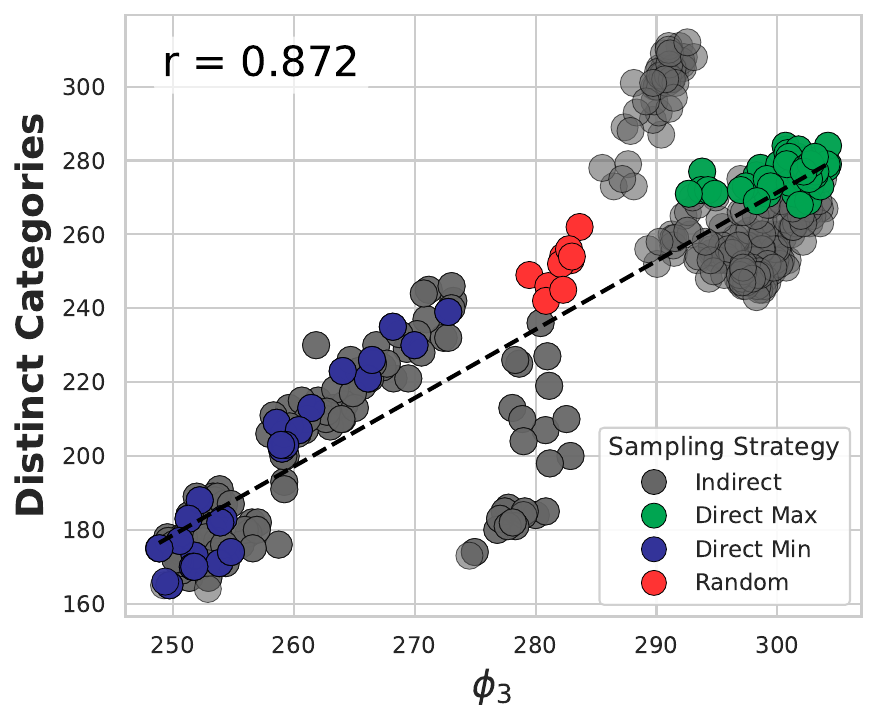}
        \caption{\mfc}
        \label{fig:bmin_budget356}
    \end{subfigure}\hfill
    \begin{subfigure}[t]{.32\textwidth}
        \centering
        \includegraphics[width=\linewidth]{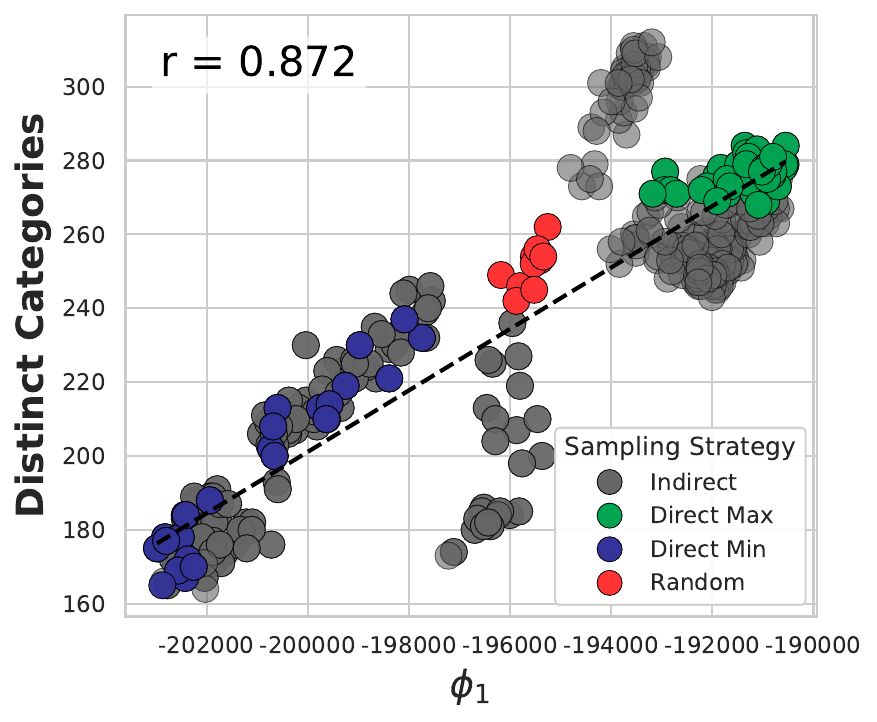}
        \caption{\mfa}
        \label{fig:plaw_budget356}
    \end{subfigure}\hfill
    \begin{subfigure}[t]{.32\textwidth}
        \centering
        \includegraphics[width=\linewidth]{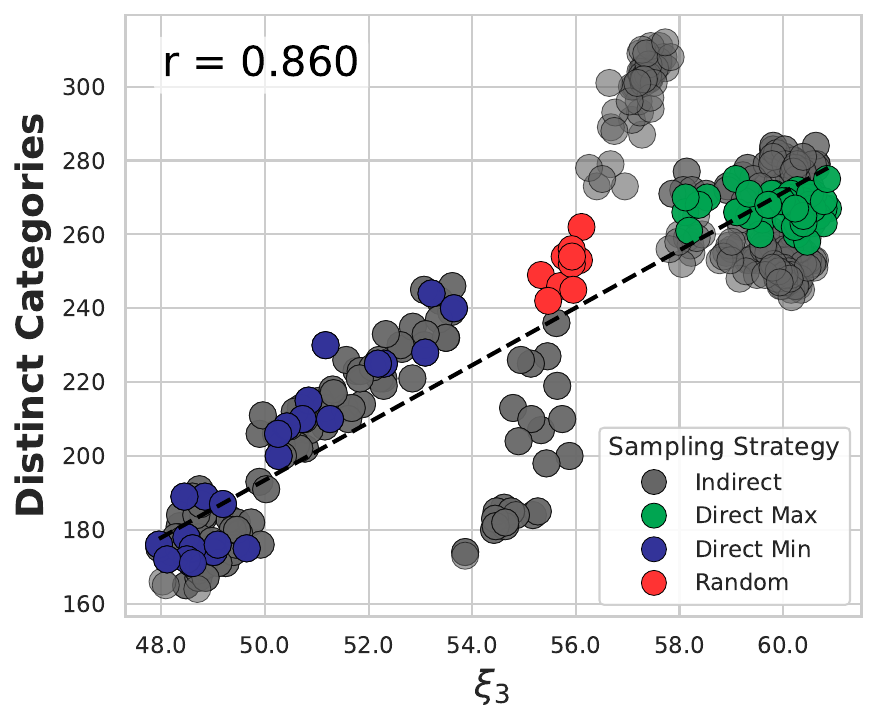}
        \caption{$\xi_3$}
        \label{fig:pow_budget356}
    \end{subfigure}
\caption{\small \textbf{Comparison of All Appraisal Functions, Airbnb Dataset, Budget = 356.} Each panel plots a function's score (x-axis) against the number of distinct rooms covered (y-axis); points are colored by sampling strategy, and the Pearson correlation $r$ is shown in the top-left. Facility Location is the strongest correlate to coverage ($r = 0.992$); $\phi_3$, $\phi_1$, and $\xi_3$ sit at $r \in [0.86, 0.87]$, slightly above DPP and Vendi at $r \approx 0.84$. Vendi's Direct Max subsets (green) cover fewer rooms than Indirect samples (gray) at slightly lower Vendi scores. For the definitions of $\phi_1$, $\phi_3$, and $\xi_3$, see Section~\ref{sec:weakly-matrix-monotone}.}
    \label{fig:airbnb_budget356}
\end{figure*}

\begin{figure*}[tbh]
    \centering
\begin{subfigure}[t]{.32\textwidth}
        \centering
        \includegraphics[width=\linewidth]{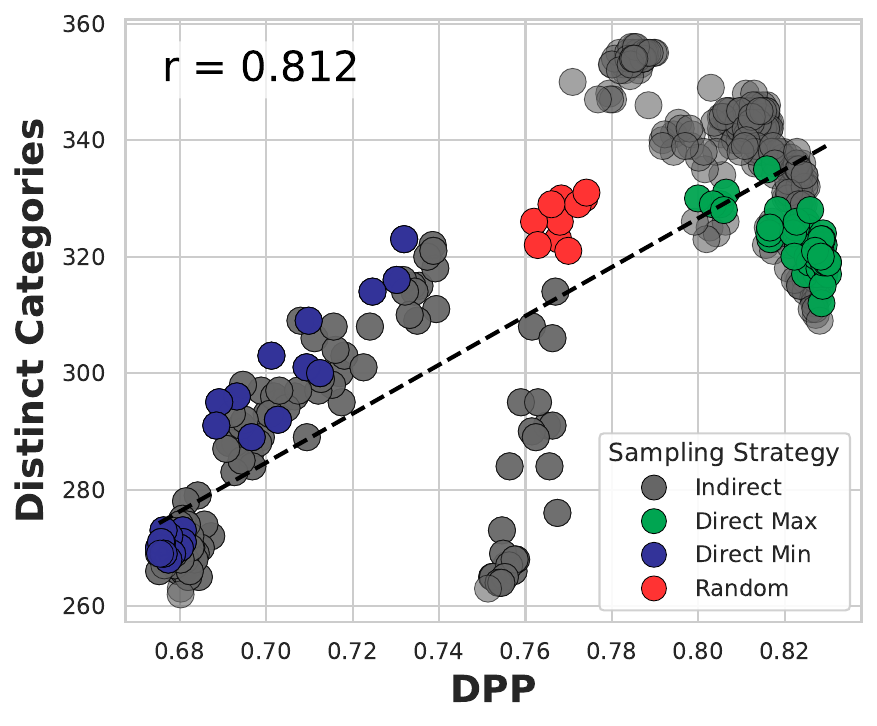}
        \caption{DPP}
        \label{fig:ldsop_budget600}
    \end{subfigure}\hfill
    \begin{subfigure}[t]{.32\textwidth}
        \centering
        \includegraphics[width=\linewidth]{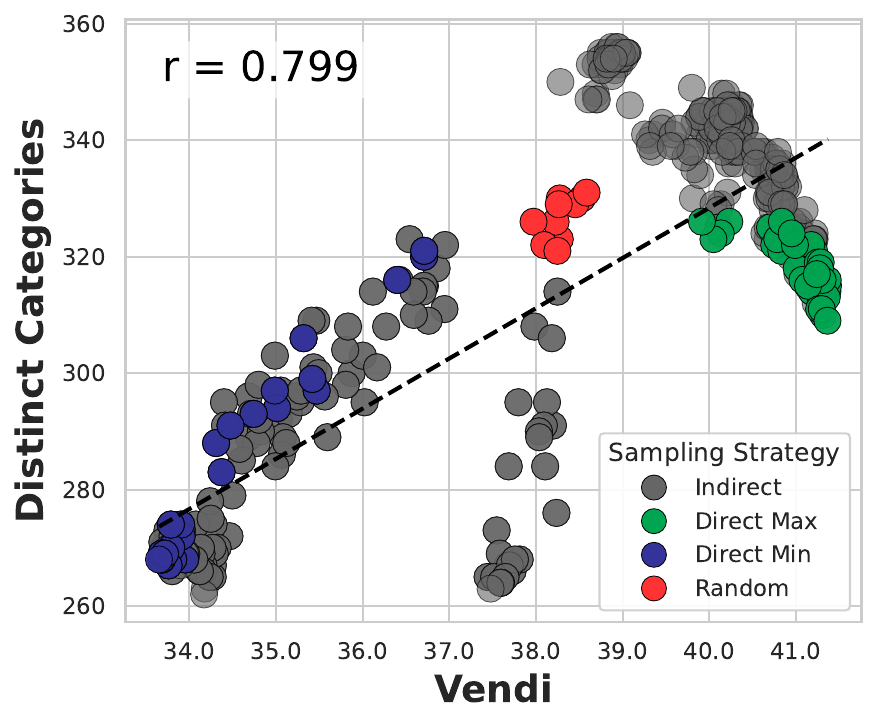}
        \caption{Vendi Score }
        \label{fig:nxlogx_budget600}
    \end{subfigure}\hfill
    \begin{subfigure}[t]{.32\textwidth}
        \centering
        \includegraphics[width=\linewidth]{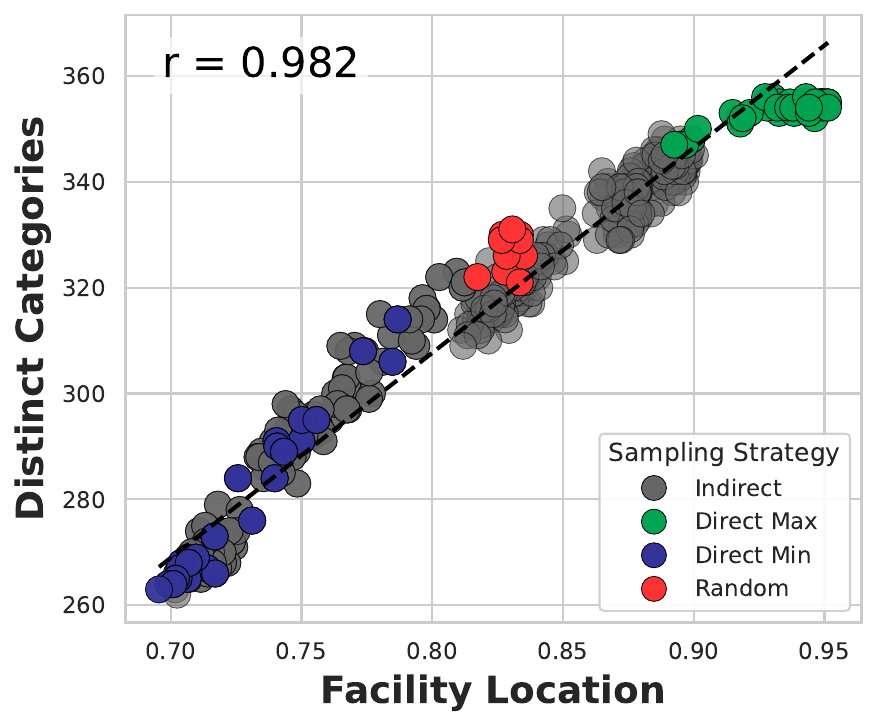}
        \caption{FL}
        \label{fig:fl_budget600}
    \end{subfigure}
\begin{subfigure}[t]{.32\textwidth}
        \centering
        \includegraphics[width=\linewidth]{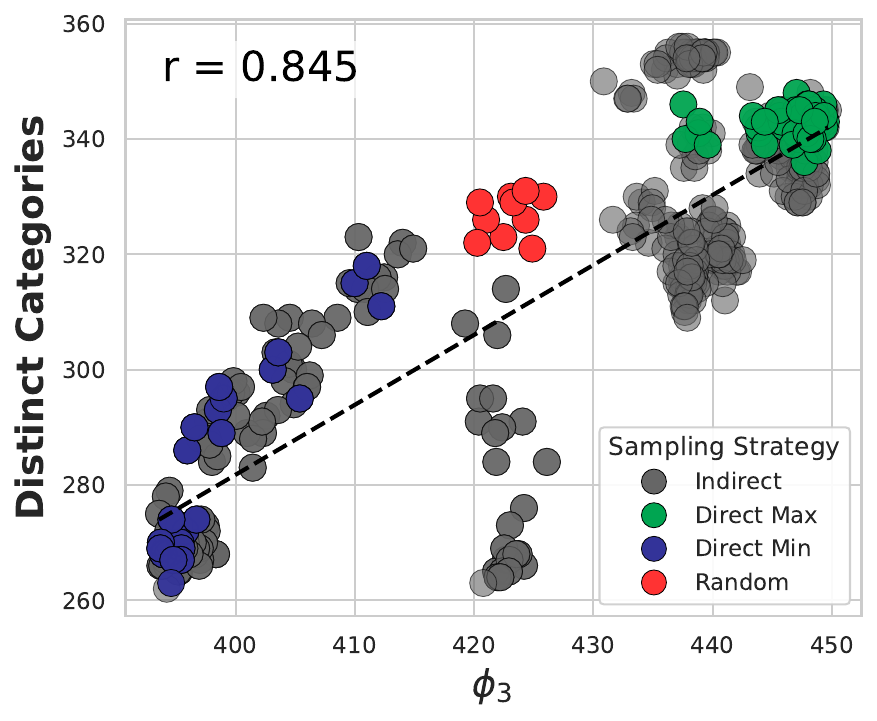}
        \caption{\mfc}
        \label{fig:bmin_budget600}
    \end{subfigure}\hfill
    \begin{subfigure}[t]{.32\textwidth}
        \centering
        \includegraphics[width=\linewidth]{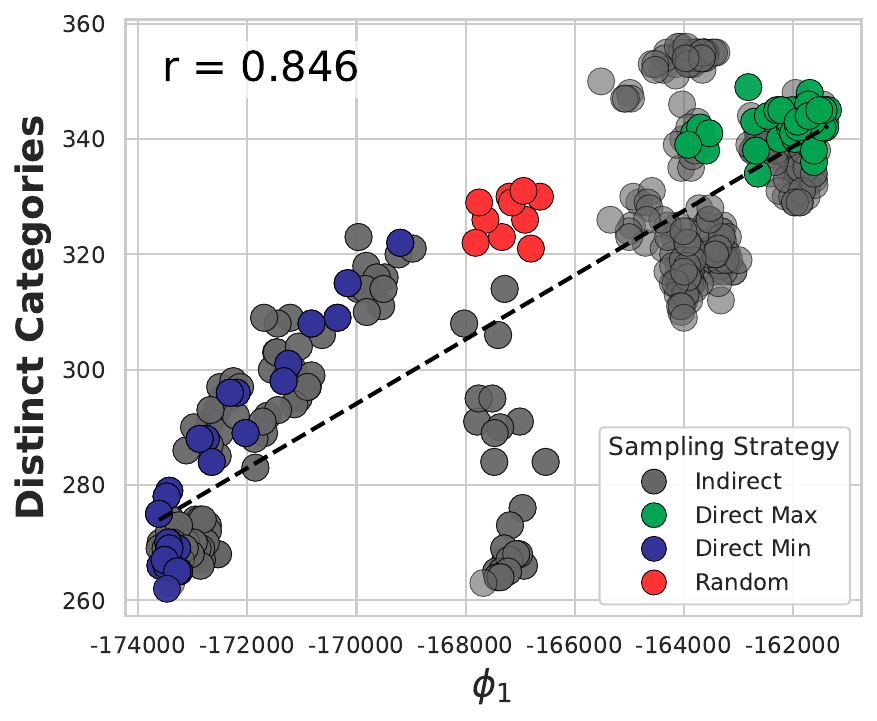}
        \caption{\mfa}
        \label{fig:plaw_budget600}
    \end{subfigure}\hfill
    \begin{subfigure}[t]{.32\textwidth}
        \centering
        \includegraphics[width=\linewidth]{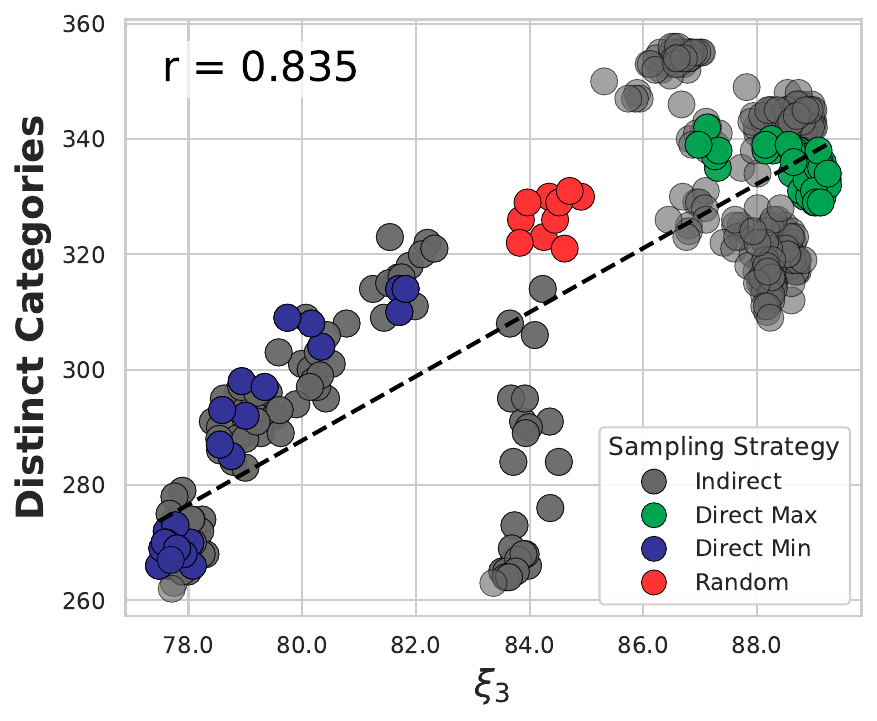}
        \caption{$\xi_3$}
        \label{fig:pow_budget600}
    \end{subfigure}
    \vspace{-.1in}
    \caption{\small \textbf{Comparison of All Appraisal Functions, Airbnb Dataset, Budget = 600.} Same setup as Figure~\ref{fig:airbnb_budget356}, with $k = 600$. Facility Location remains far ahead ($r = 0.982$); $\phi_3$, $\phi_1$, and $\xi_3$ sit at $r \in [0.83, 0.85]$, with DPP and Vendi at $r \in [0.80, 0.81]$. Vendi's non-monotonicity is more pronounced here: the Direct Max cluster sits visibly below the Indirect cluster, and the same effect is mirrored in DPP. For the definitions of $\phi_1$, $\phi_3$, and $\xi_3$, see Section~\ref{sec:weakly-matrix-monotone}.}
    \label{fig:airbnb_budget600}
\end{figure*}

\begin{figure*}[tbh]
    \centering
\begin{subfigure}[t]{.32\textwidth}
        \centering
        \includegraphics[width=\linewidth]{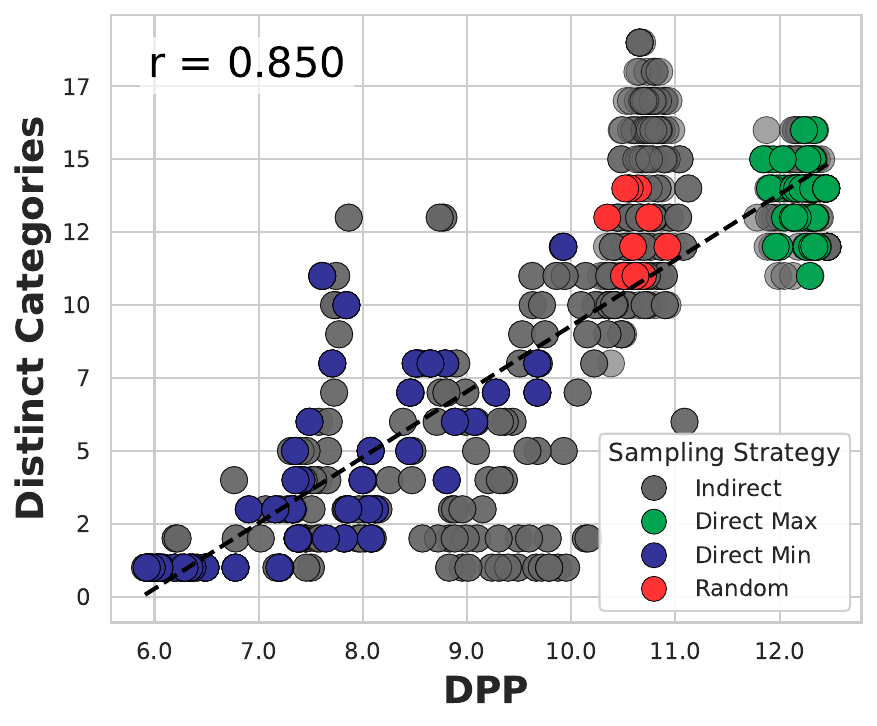}
        \caption{DPP}
        \label{fig:ldsop_20ng_budget20}
    \end{subfigure}\hfill
    \begin{subfigure}[t]{.32\textwidth}
        \centering
        \includegraphics[width=\linewidth]{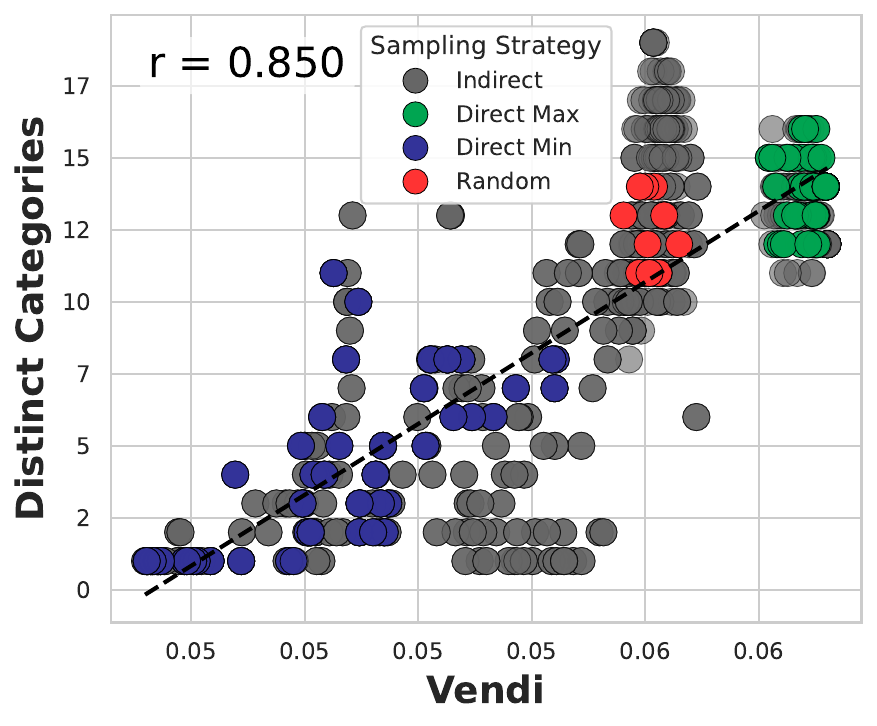}
        \caption{Vendi Score }
        \label{fig:nxlogx_20ng_budget20}
    \end{subfigure}\hfill
    \begin{subfigure}[t]{.32\textwidth}
        \centering
        \includegraphics[width=\linewidth]{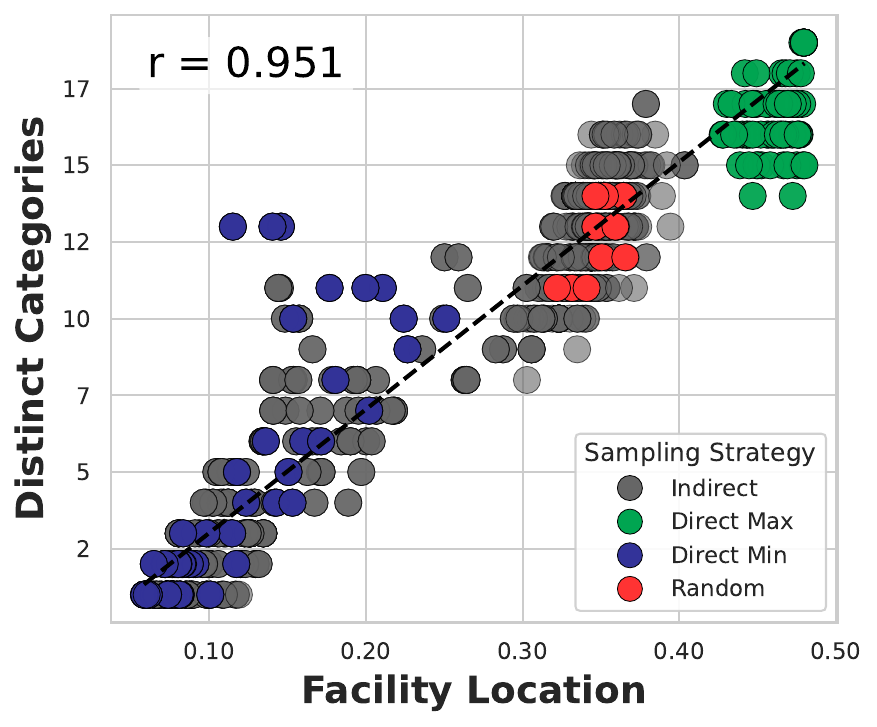}
        \caption{FL}
        \label{fig:fl_20ng_budget20}
    \end{subfigure}
\begin{subfigure}[t]{.32\textwidth}
        \centering
        \includegraphics[width=\linewidth]{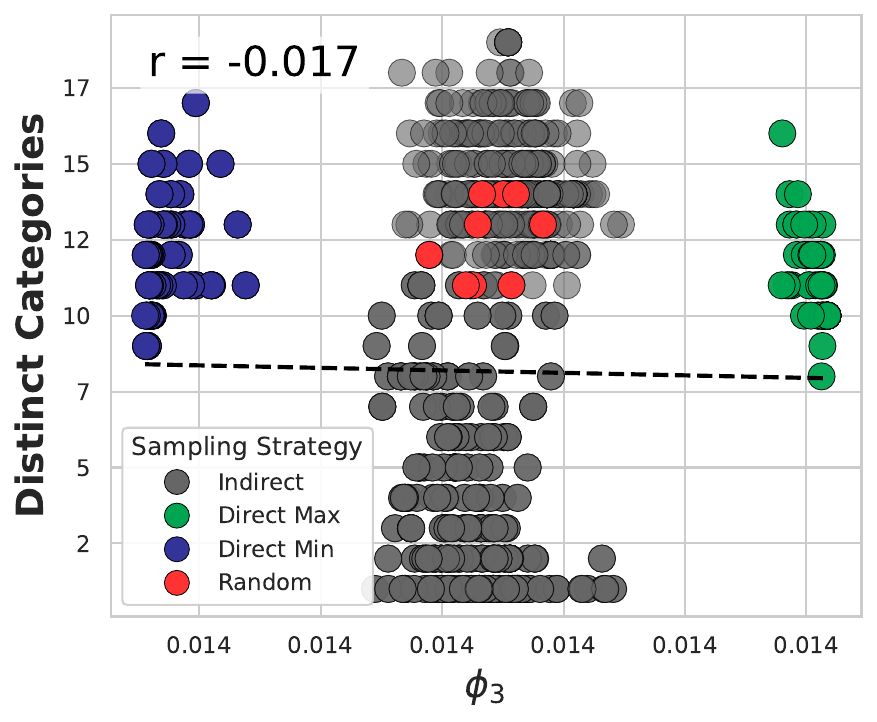}
        \caption{\mfc}
        \label{fig:bmin_20ng_budget20}
    \end{subfigure}\hfill
    \begin{subfigure}[t]{.32\textwidth}
        \centering
        \includegraphics[width=\linewidth]{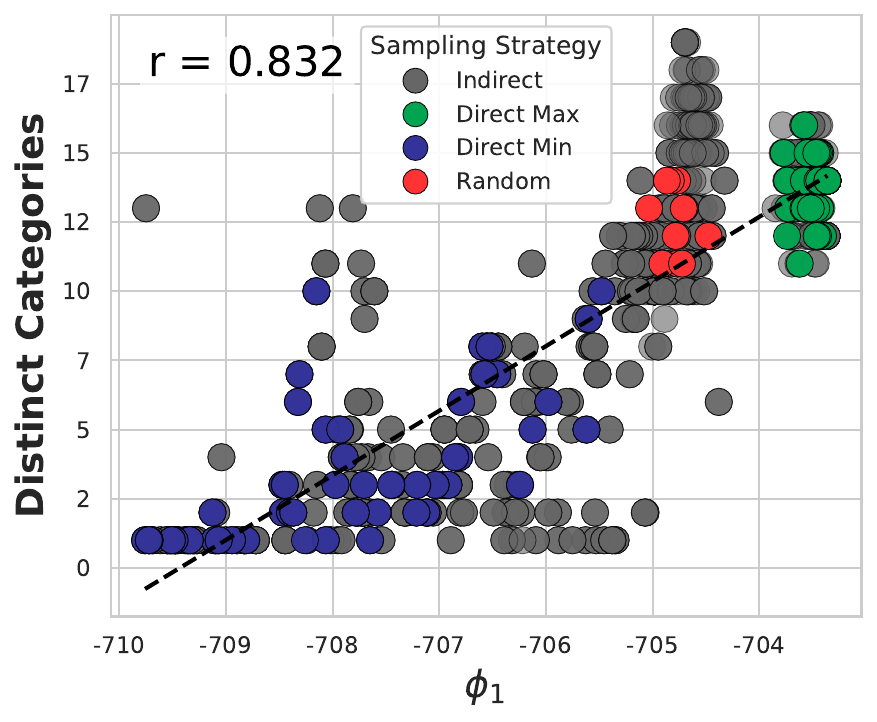}
        \caption{\mfa}
        \label{fig:plaw_20ng_budget20}
    \end{subfigure}\hfill
    \begin{subfigure}[t]{.32\textwidth}
        \centering
        \includegraphics[width=\linewidth]{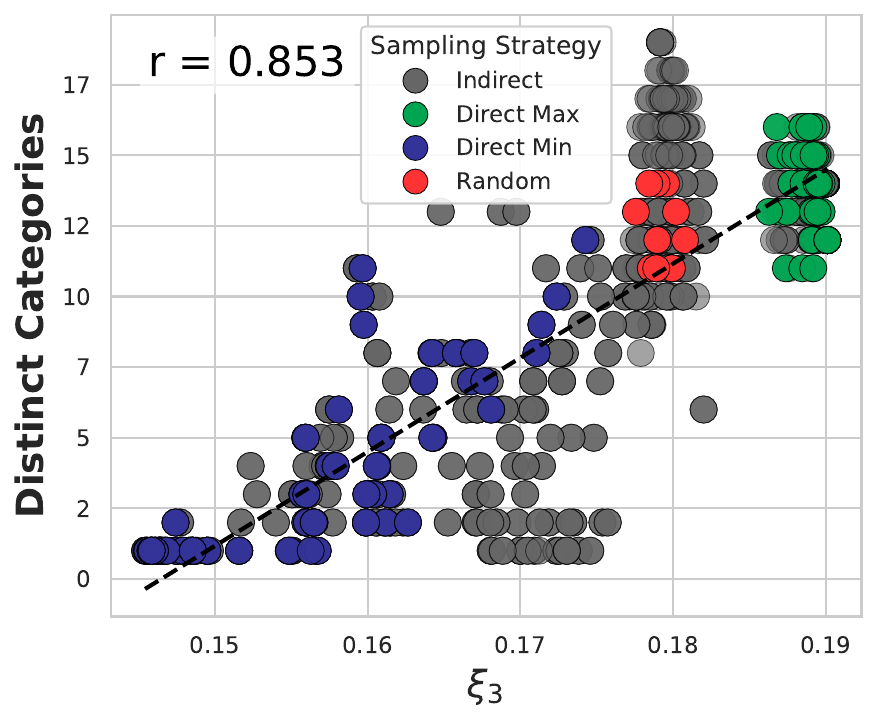}
        \caption{$\xi_3$}
        \label{fig:pow_20ng_budget20}
    \end{subfigure}
    \vspace{-.1in}
    \caption{\small \textbf{Comparison of All Appraisal Functions, 20 Newsgroups
    Dataset, Budget = 20.} Each panel plots a function's score (x-axis) against
    the number of distinct newsgroups covered (y-axis); points are colored by
    sampling strategy and the Pearson correlation $r$ is shown in the top-left.
    Facility Location is the strongest correlate of coverage ($r = 0.951$);
    $\xi_3$, $\phi_1$, DPP, and Vendi sit at $r \in [0.83, 0.85]$. The function
    $\phi_3$ collapses on this dataset ($r = -0.017$), making it unusable as a
    selector. For the definitions of $\phi_1$, $\phi_3$, and $\xi_3$, see
    Section~\ref{sec:weakly-matrix-monotone}.}
    \label{fig:20ng_budget20}
\end{figure*}

\begin{figure*}[tbh]
    \centering
\begin{subfigure}[t]{.32\textwidth}
        \centering
        \includegraphics[width=\linewidth]{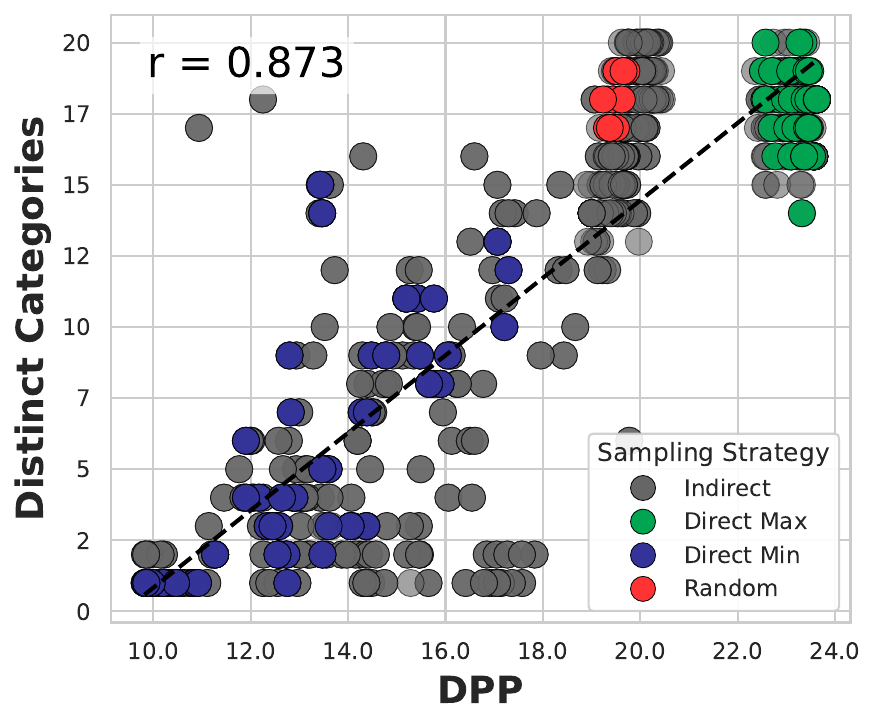}
        \caption{DPP}
        \label{fig:ldsop_20ng_budget40}
    \end{subfigure}\hfill
    \begin{subfigure}[t]{.32\textwidth}
        \centering
        \includegraphics[width=\linewidth]{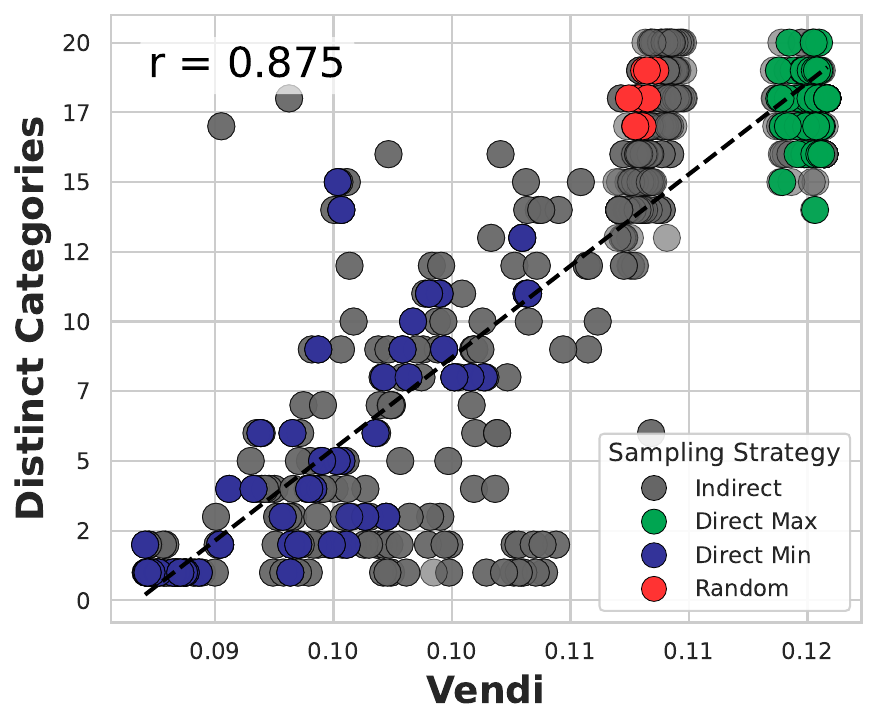}
        \caption{Vendi Score }
        \label{fig:nxlogx_20ng_budget40}
    \end{subfigure}\hfill
    \begin{subfigure}[t]{.32\textwidth}
        \centering
        \includegraphics[width=\linewidth]{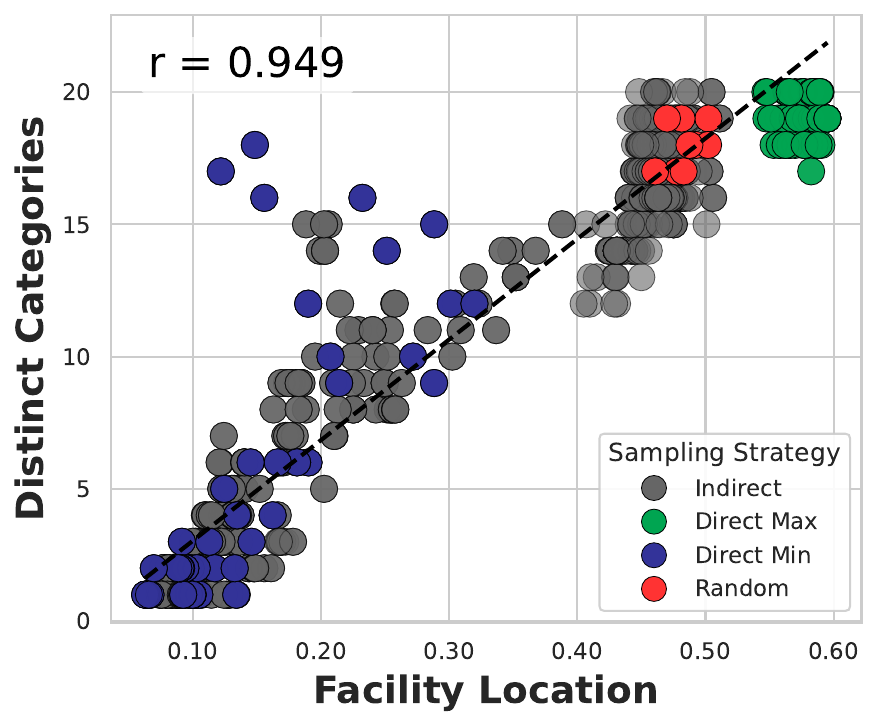}
        \caption{FL}
        \label{fig:fl_20ng_budget40}
    \end{subfigure}
\begin{subfigure}[t]{.32\textwidth}
        \centering
        \includegraphics[width=\linewidth]{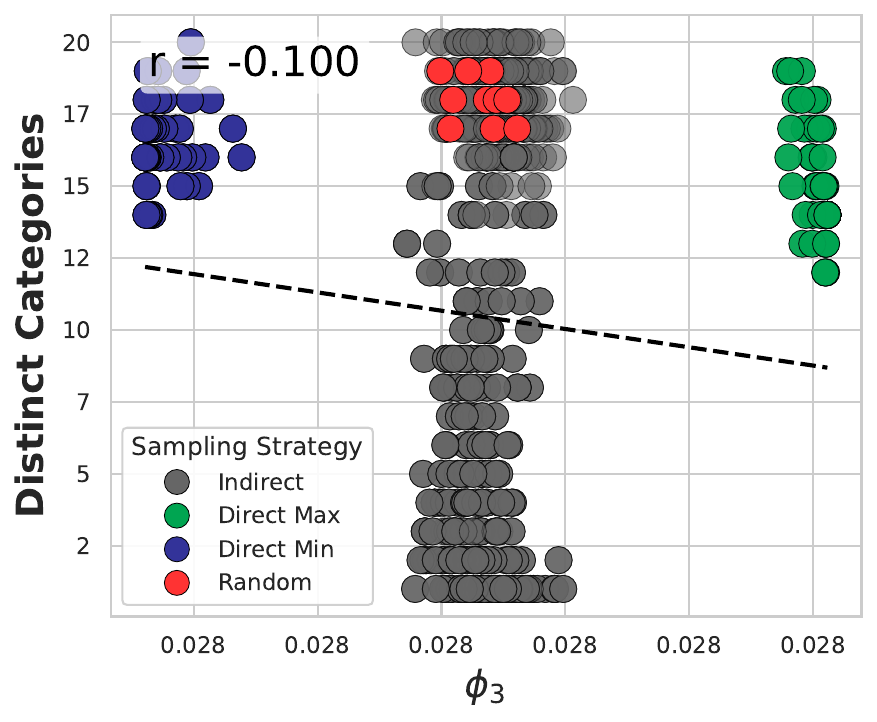}
        \caption{\mfc}
        \label{fig:bmin_20ng_budget40}
    \end{subfigure}\hfill
    \begin{subfigure}[t]{.32\textwidth}
        \centering
        \includegraphics[width=\linewidth]{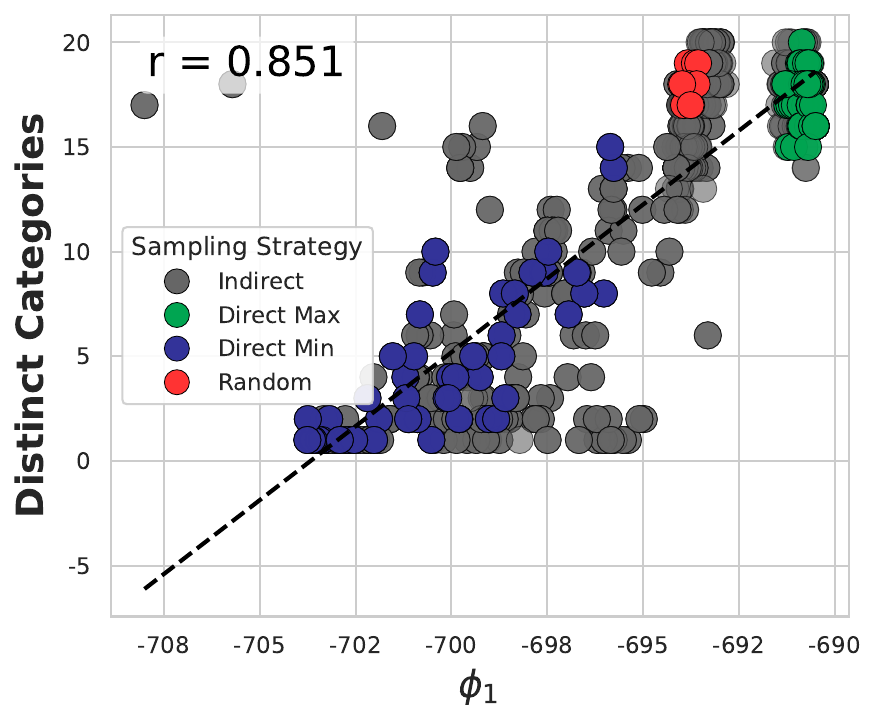}
        \caption{\mfa}
        \label{fig:plaw_20ng_budget40}
    \end{subfigure}\hfill
    \begin{subfigure}[t]{.32\textwidth}
        \centering
        \includegraphics[width=\linewidth]{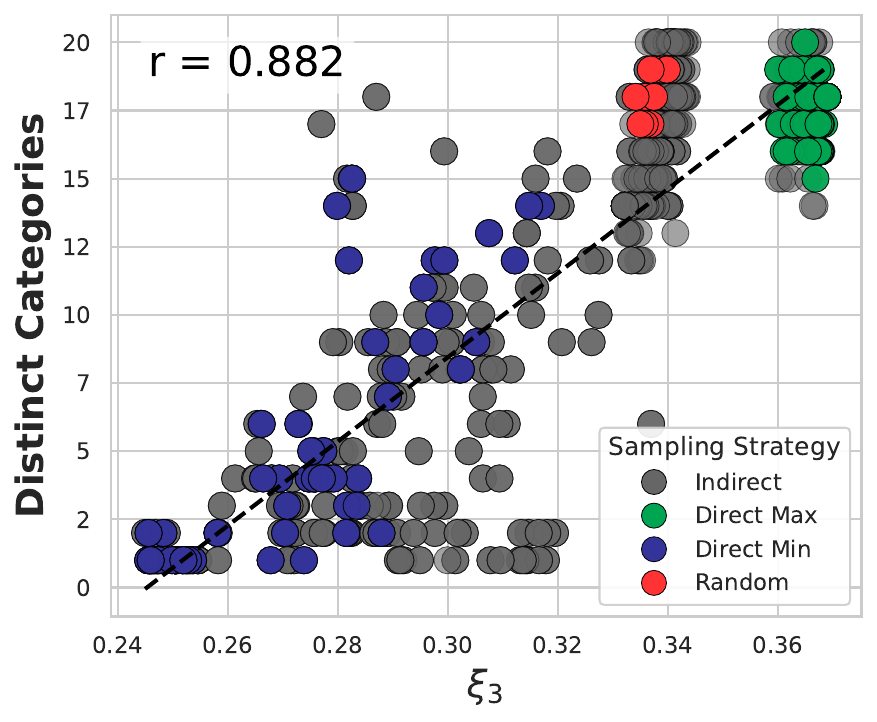}
        \caption{$\xi_3$}
        \label{fig:pow_20ng_budget40}
    \end{subfigure}
    \vspace{-.1in}
    \caption{\small \textbf{Comparison of All Appraisal Functions, 20 Newsgroups Dataset, Budget = 40.} Same setup as Figure~\ref{fig:20ng_budget20}, with $k = 40$. Facility Location remains the strongest correlate ($r = 0.949$); $\xi_3$, Vendi, DPP, and $\phi_1$ sit at $r \in [0.85, 0.88]$. Poor correlation for $\phi_3$ as observed at $k = 20$ is persistent here. For the definitions of $\phi_1$, $\phi_3$, and $\xi_3$, see Section~\ref{sec:weakly-matrix-monotone}.}
    \label{fig:20ng_budget40}
\end{figure*}

Across both datasets and all four budgets (Figures~\ref{fig:airbnb_budget356},
\ref{fig:airbnb_budget600}, \ref{fig:20ng_budget20},
and~\ref{fig:20ng_budget40}), Facility Location attains the highest correlation
with coverage, reaching $\rho \in [0.95, 0.99]$. On Airbnb, the remaining
functions sit at $\rho \in [0.80, 0.87]$ and exhibit a two-cluster pattern in
which subsets with similar function values differ by tens of unique rooms; on 20
Newsgroups, the same pattern produces differences of several newsgroups. Among
the weakly matrix monotone functions, $\mfa$ and $\mfc$ on Airbnb perform
competitively, covering 285 and 283 distinct rooms respectively, consistent with
Lemma~\ref{lem:some-useful-concave-functions-are-weakly-matrix-monotone}, which
suggests their parameters can yield a useful approximation guarantee. On 20
Newsgroups, by contrast, $\mfc$ collapses entirely ($\rho = -0.017$ at $k = 20$
and $\rho = -0.100$ at $k = 40$). On Airbnb, Vendi's highest-scoring subsets
cover fewer rooms than subsets at slightly lower Vendi scores, so maximizing
Vendi actively degrades coverage. The same non-monotone behavior appears on
ImageNet, where the highest-value Vendi subsets yield notably lower test
accuracy than subsets at intermediate Vendi scores. We explore this further with
a Gaussian kernel in Section~\ref{app:gaussian_kernel}.

\subsection{From Linear Kernel to Gaussian Kernels}
\label{app:gaussian_kernel}

So far we have used a linear kernel, $B = \data \data^\top$, where $\data \in
\mathbb{R}^{n \times m}$. Nonlinear kernels are often more expressive in
classical kernel methods~\citep{scholkopf2002learning}, and the same question is
worth asking for matrix spectral functions. Two issues make this harder than the
linear case. The secular-equation speedup of Section~\ref{sec:solv-secul-equat}
requires the linear form $B = \data \data^\top$, so direct optimization under a
nonlinear kernel cannot use it. The full Gram matrix also requires $O(n^2)$
space. The first issue is the more limiting one in practice, so we restrict this
experiment to AirBnB, where the ground set is small enough for direct
optimization to remain tractable. The weakly matrix monotone property of
Section~\ref{sec:weakly-matrix-monotone} holds for any positive semidefinite
kernel $B$, not only the linear one, so the approximation guarantees for greedy
maximization of $\mfa$ and $\mfc$ continue to apply. In this work, we
additionally experiment with Gaussian Kernels, where $k(x_i, x_j) =
\exp(-\|x_i-x_j\|^2_2 / \sigma)$; table~\ref{tab:kernel_comparison} shows that
the Gaussian kernel generally improves coverage. It is worth noting that the
kernel width of the Gaussian kernel is a salient parameter, and it is possible
to get a poor coverage with an undertuned Gaussian kernel. This suggests an
explanation for the dip in Vendi performance (higher evaluation sets achieving
much lower accuracy) on ImageNet under the linear kernel: the rank is bounded by
$\min(n, m)$, whereas the Gaussian kernel is strictly positive definite and
admits the full rank (assuming no duplicates). While we see improvements for
Vendi Score, we don't observe such improvements when used with other functions
such as \mfa (refer to \Cref{sec:weakly-matrix-monotone} for definition). A
larger-scale study of Gaussian-kernel matrix spectral functions is left to
future work.

\begin{table}[tbh]
\centering
\small
\setlength{\abovecaptionskip}{10pt}
\begin{tabular}{l|cc|cc}
\toprule
& \multicolumn{2}{c|}{Budget = 356} & \multicolumn{2}{c}{Budget = 600} \\
\cmidrule(lr){2-3} \cmidrule(lr){4-5}
& Linear & Gaussian & Linear & Gaussian \\
\midrule
Function & \shortstack{Unique\\Rooms} & \shortstack{Unique\\Rooms} & \shortstack{Unique\\Rooms} & \shortstack{Unique\\Rooms} \\
\midrule
Vendi Score & 245 & 274 & 315 & 341 \\
DPP         & 254 & 278 & 320 & 345 \\
\bottomrule
\end{tabular}
\caption{\textbf{Airbnb: linear vs.\ Gaussian kernel comparison for Vendi Score and DPP across two summary budgets.} We report the number of unique rooms covered by the selected subset for each (function, budget, kernel) combination. The Gaussian kernel uses width $\sigma = 5$. The Gaussian kernel consistently improves coverage at both budgets and for both functions (Vendi: 245 \(\to\) 274 at $k = 356$, 315 \(\to\) 341 at $k = 600$; DPP: 254 \(\to\) 278 and 320 \(\to\) 345).
Despite these improvements, these do not achieve
the performance of a facility location function on this
task which achieves 317 unique rooms with a budget of 356 using the same underlying features.
}
\label{tab:kernel_comparison}
\end{table}

\section{Additional Experimental As A Function of Compute Budget}
\label{app:additional_compute_budgets}

In this section, we show additional results 
that expand
on those given in~\Cref{fig:main-teaser}
and~\Cref{sec:data-appraisal}. In addition to
the class-balance results described
in~\Cref{sec:data-appraisal}, we show the performance
of datasets that have been created that are not constrained
to be class balanced, and are constrained only
by size. These
are given in~\Cref{fig:unconstrained-complete-pyramid}.
In the good \textcolor{FigGreen}{green}
case, we use cardinality constrained submodular
maximization on a facility location function,
in the random \textcolor{FigRed}{red} case,
we use random subsets selection of the appropriate
size. In the poor \textcolor{FigBlue}{blue} case,
we use heuristic cardinality constrained submodular
minimization using a greedy minimization procedure,
as described in~\Cref{sec:background-submodularity}.

\begin{figure}[tbh]
  \centering
\newcommand{\TopPlotWidth}{0.41\textwidth}
\makebox[\textwidth][c]{\begin{subfigure}[t]{\TopPlotWidth}
      \centering
      \includegraphics[width=\linewidth]{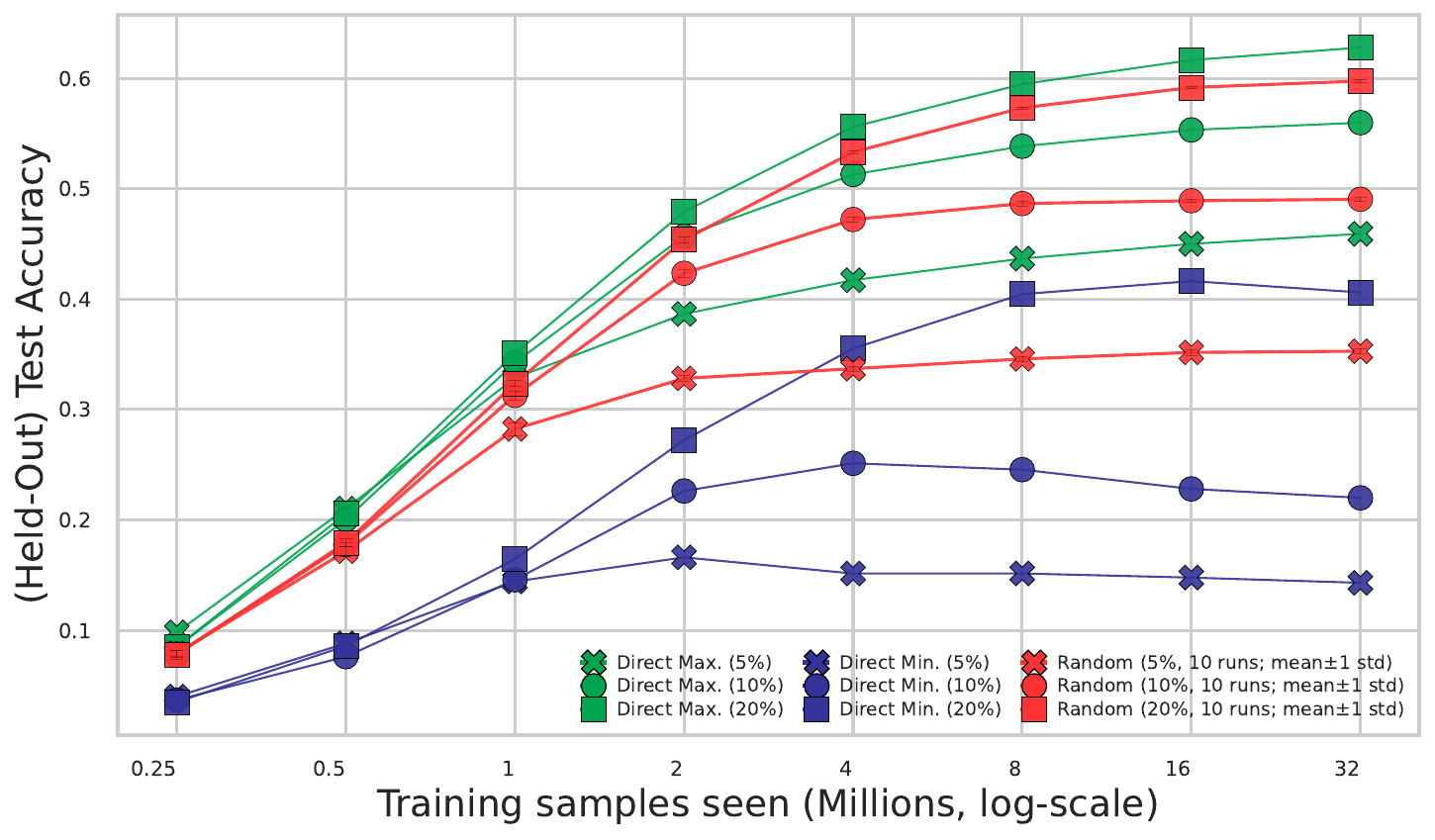}
      \caption{
        \FigureOneCaptionA This figure is the same as
        shown in~\Cref{fig:class-balanced-all-merged-teaser}.
        \looseness-1}
      \label{fig:class-balanced-all-merged-plots}
    \end{subfigure}
    \hspace{0.10\textwidth}
    \begin{subfigure}[t]{\TopPlotWidth}
      \centering
      \includegraphics[width=\linewidth]{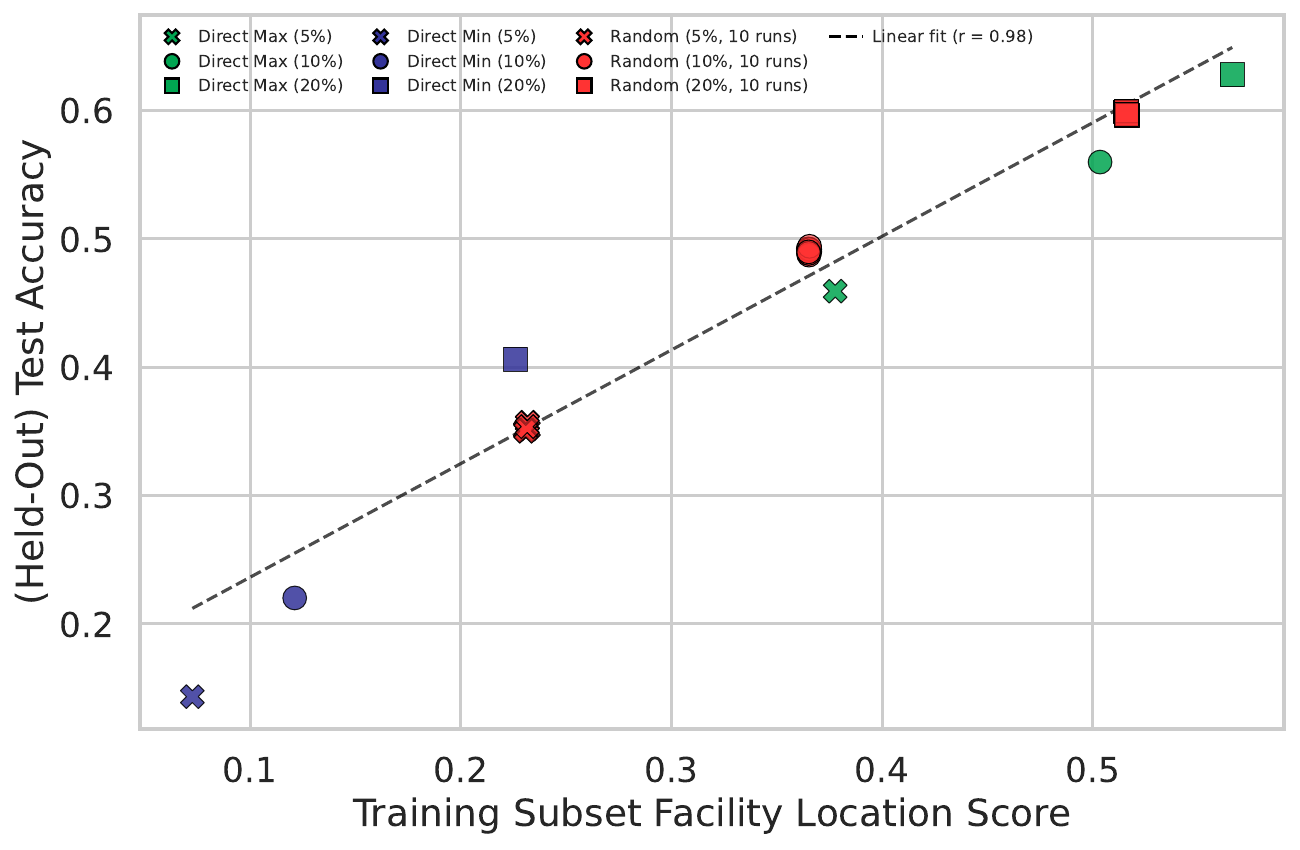}
      \caption{
        \FigureOneCaptionB Here, the surrogate data-valuation strategy
         is a facility location function valuation on each dataset. This figure
         is the same as shown in~\Cref{fig:class-balanced-accuracy-vs-fl-score-at-32M-teaser}.
        }
      \label{fig:class-balanced-accuracy-vs-fl-score-at-32m}
    \end{subfigure}
  }
  \vspace{0.8em}
\begin{subfigure}[t]{\textwidth}
    \centering
    \includegraphics[width=\linewidth]{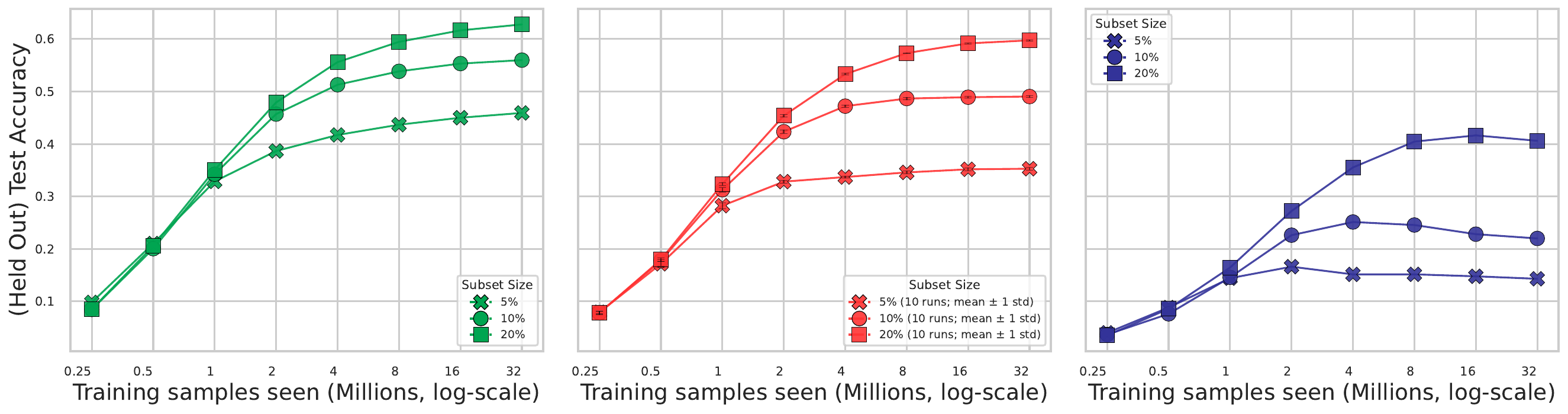}
    \caption{Side-by-side comparison of ImageNet-1K test performance
    for the highly valued \textcolor{FigGreen}{green} (left) points created using
    class-balanced partition matroid constrained submodular maximization on the facility location, \textcolor{FigRed}{red} (middle)
    created using stratified random sampling to achieve class balance,
     and \textcolor{FigBlue}{blue} (right) class-balanced heuristic partition matroid constrained
     submodular minimization again on the same facility location function. These
    three plots show the same information as in~\Cref{fig:class-balanced-all-merged-plots} but is
    expanded to three side-by-side plots for clarity.}
    \label{fig:class-balanced-side-by-side-comparison-at-32m}
  \end{subfigure}

  \caption{
    The results (top row) are the same as those presented in~\Cref{fig:main-teaser} and
    the bottom row is expanded out to three side-by-side plots to show more details more clearly.
  }
  \label{fig:class-balanced-complete-pyramid}
\end{figure}

\begin{figure}[tbh]
  \centering
\newcommand{\TopPlotWidth}{0.41\textwidth}
\makebox[\textwidth][c]{\begin{subfigure}[t]{\TopPlotWidth}
      \centering
      \includegraphics[width=\linewidth]{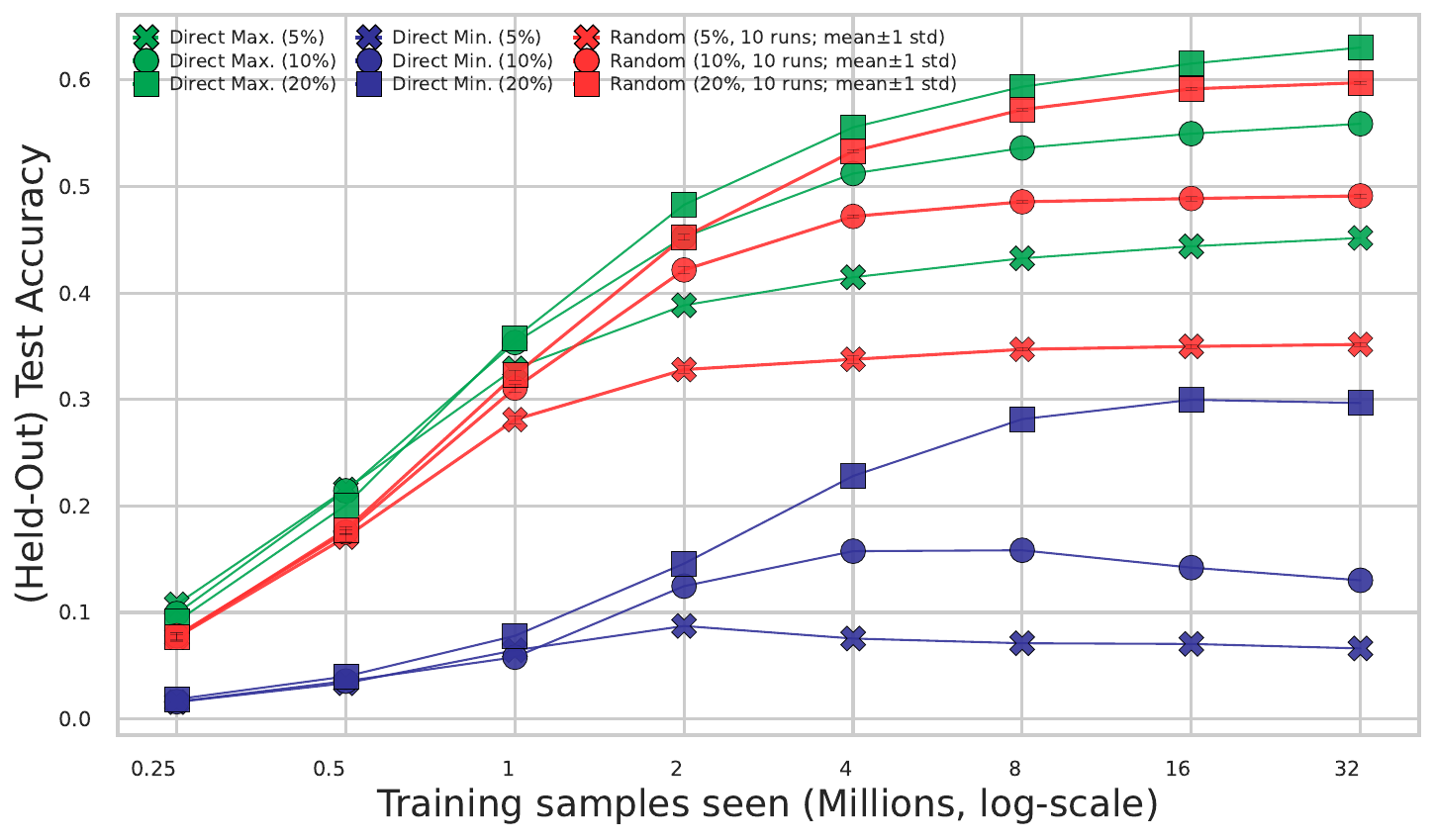}
      \caption{These results are very similar to that
      of~\Cref{fig:class-balanced-all-merged-plots} except that
      the submodular optimization was done constrained only by
      size rather than class balance (we call these "unconstrained" but they
      are really only constrained by size). We see that the heuristic
      submodular min poor \textcolor{FigBlue}{blue} sets are much worse
      since they are now allowed to become more class imbalanced, but
      both the random and the submodular maximization results have changed
      relatively little. This shows a similar pattern, which is that
      good and poor valuation is preserved across compute budget.
      \looseness-1}
      \label{fig:unconstrained-all-merged-plots}
    \end{subfigure}
    \hspace{0.10\textwidth}
    \begin{subfigure}[t]{\TopPlotWidth}
      \centering
      \includegraphics[width=\linewidth]{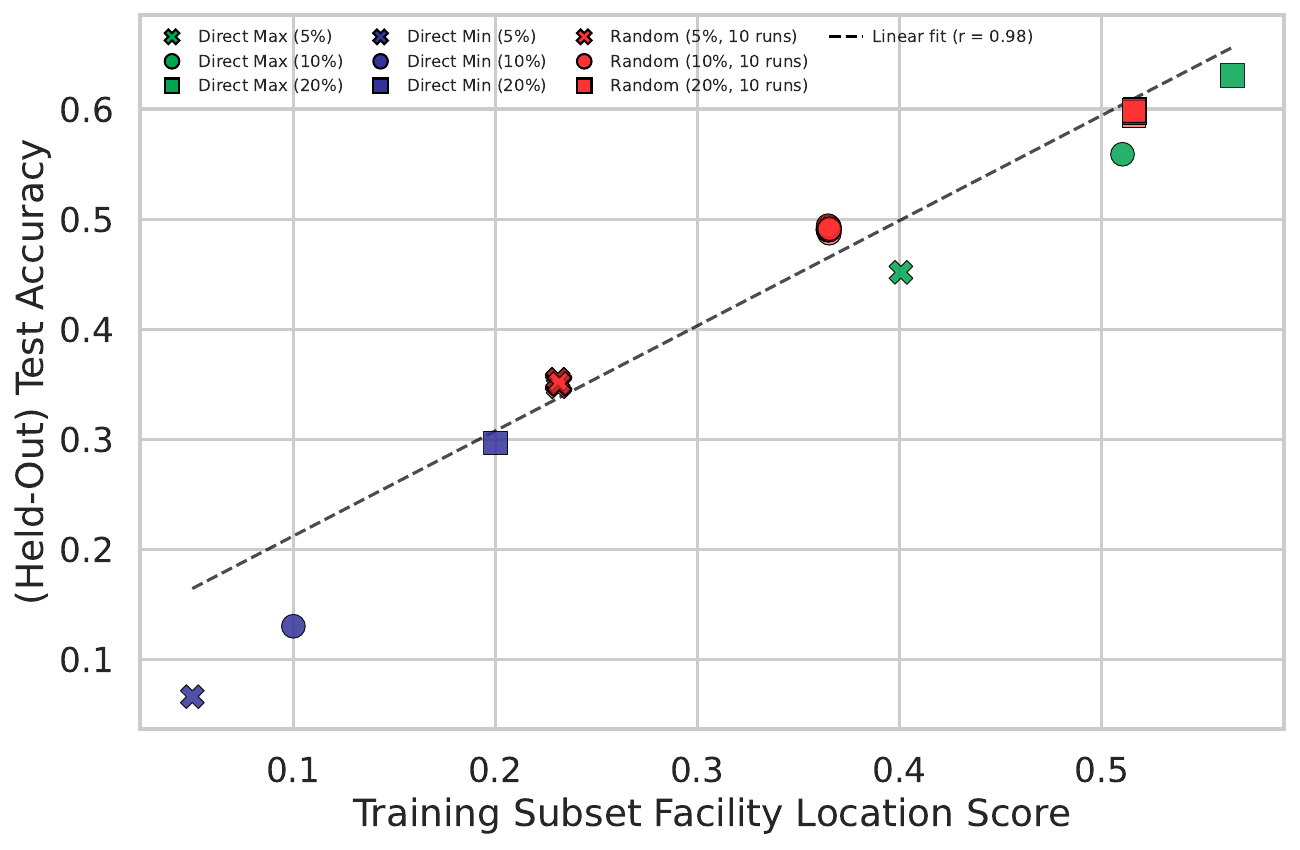}
      \caption{
        Unconstrained (i.e., size-only constrained) results corresponding
        to~\Cref{fig:class-balanced-accuracy-vs-fl-score-at-32m} showing a similar
        pattern, namely that the facility location is effective at valuing
        even not necessarily balanced datasets in terms of predicting
        their utility in training a model and testing on a held-out set.
        }
      \label{fig:unconstrained-accuracy-vs-fl-score-at-32m}
    \end{subfigure}
  }
  \vspace{0.8em}
\begin{subfigure}[t]{\textwidth}
    \centering
    \includegraphics[width=\linewidth]{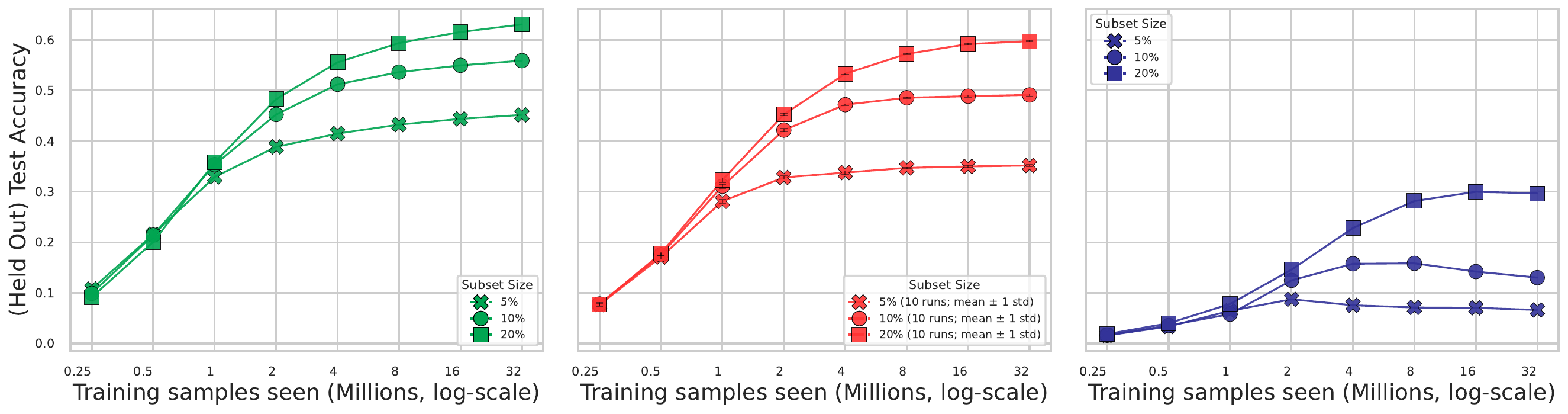}
    \caption{Expanded side-by-side comparison of~\Cref{fig:unconstrained-all-merged-plots}.}
    \label{fig:unconstrained-side-by-side-comparison-at-32m}
  \end{subfigure}

  \caption{
    This shows results similar to~\Cref{fig:class-balanced-complete-pyramid} but where the sets are not required
    to be class balanced, and are constrained only by size but are otherwise unconstrained. In the
    paper, we call this the ``unconstrained'' case. We note that a direct side-by-side comparison
    between~\Cref{fig:class-balanced-all-merged-plots} and~\Cref{fig:unconstrained-all-merged-plots}
    is shown in~\Cref{fig:class-balanced-vs-unconstrained}.
  }
  \label{fig:unconstrained-complete-pyramid}
\end{figure}

\begin{figure}[tbh]
  \centering
\begin{subfigure}[t]{\textwidth}
    \centering
    \includegraphics[width=\linewidth]{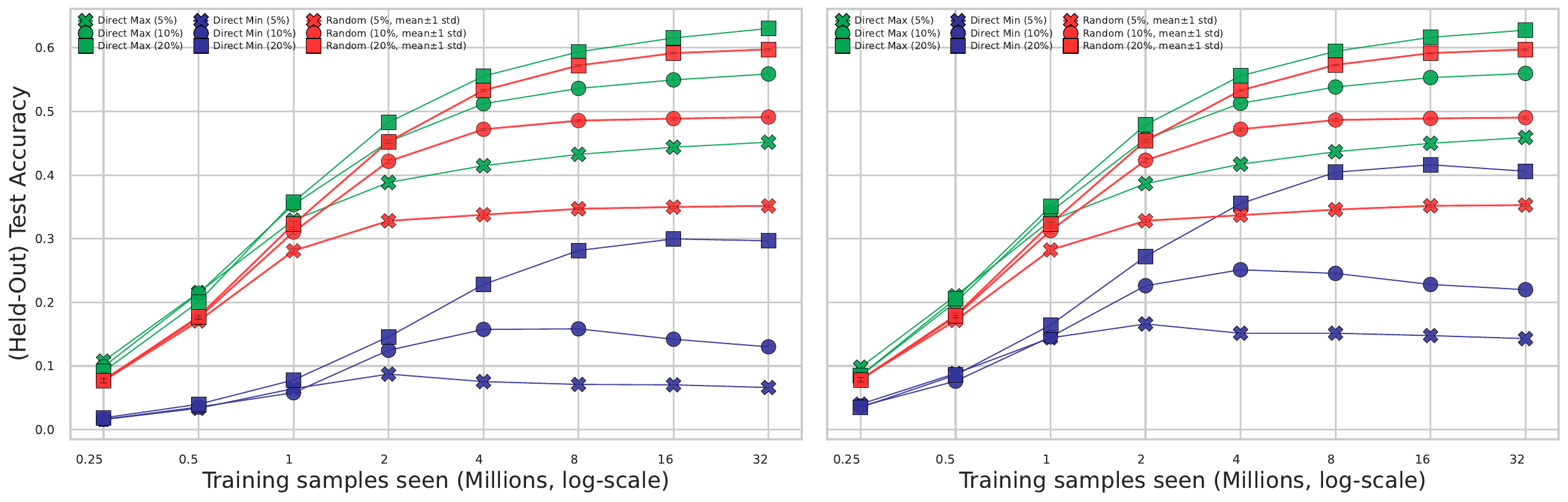}
    \caption{Side-by-side comparison of~\Cref{fig:unconstrained-all-merged-plots} (on the left)
       and~\Cref{fig:class-balanced-all-merged-plots} (on the right), with the y-axis aligned,
        to show the effect of 
    the class balance constraint on the test performance of the datasets as a function of compute budget.}
    \label{fig:side-by-side-unconstraint-vs-constrained}
  \end{subfigure}
  \vspace{0.8em}
\makebox[\textwidth][c]{\begin{subfigure}[t]{0.32\textwidth}
      \centering
      \includegraphics[width=\linewidth]{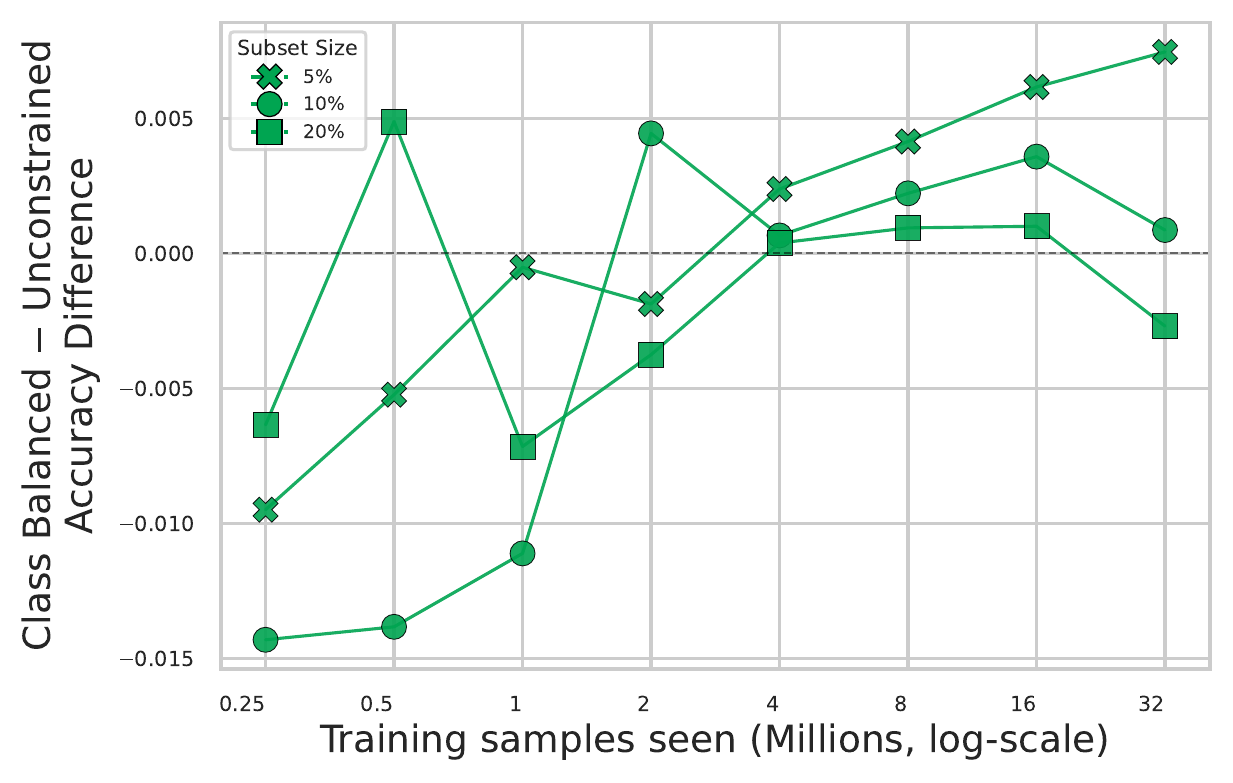}
      \caption{Difference between the high valued \textcolor{FigGreen}{green} class-balanced 
               datasets and the size constrained (but otherwise unconstrained) datasets.}
      \label{fig:diff-class-balanced-vs-unconstrained-good}
    \end{subfigure}
    \hspace{0.05\textwidth}
    \begin{subfigure}[t]{0.32\textwidth}
      \centering
      \includegraphics[width=\linewidth]{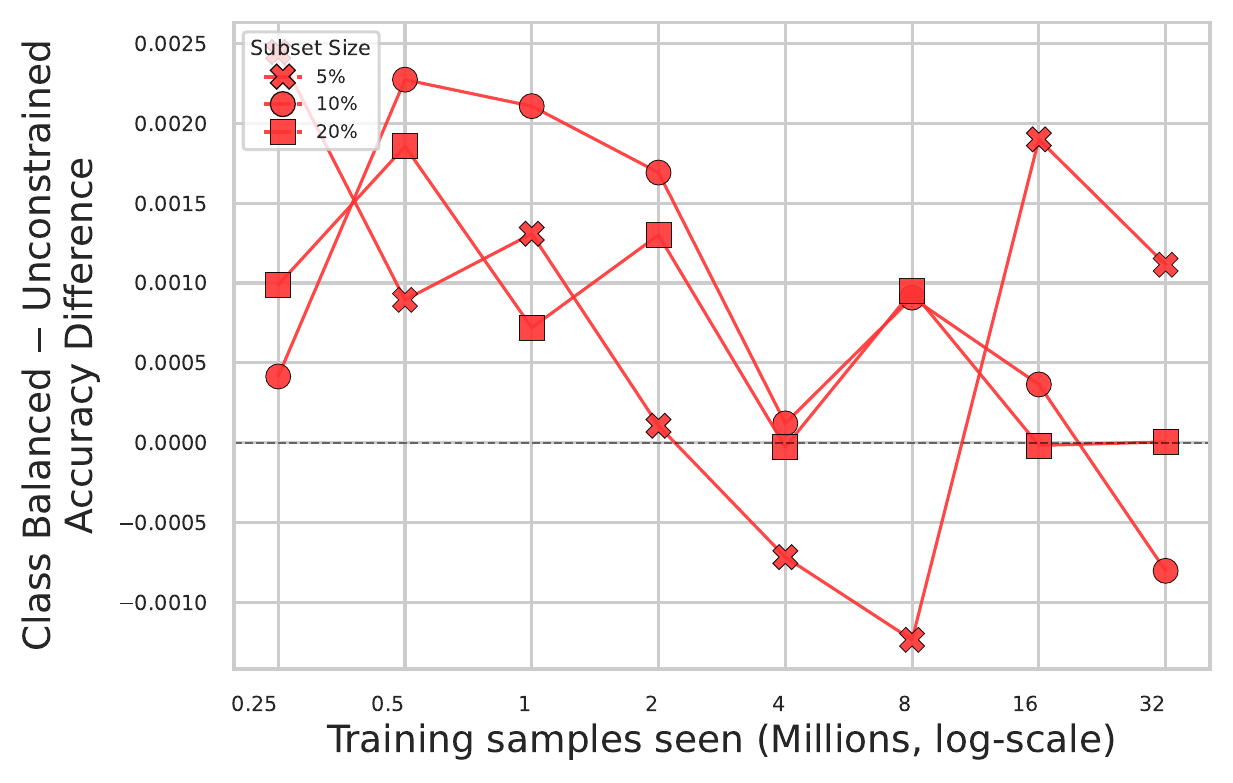}
      \caption{Difference between the stratified random \textcolor{FigRed}{red} class-balanced datasets and the 
           size constrained (but otherwise unconstrained) random datasets.}
      \label{fig:diff-class-balanced-vs-unconstrained-random}
    \end{subfigure}
    \hspace{0.05\textwidth}
    \begin{subfigure}[t]{0.32\textwidth}
      \centering
      \includegraphics[width=\linewidth]{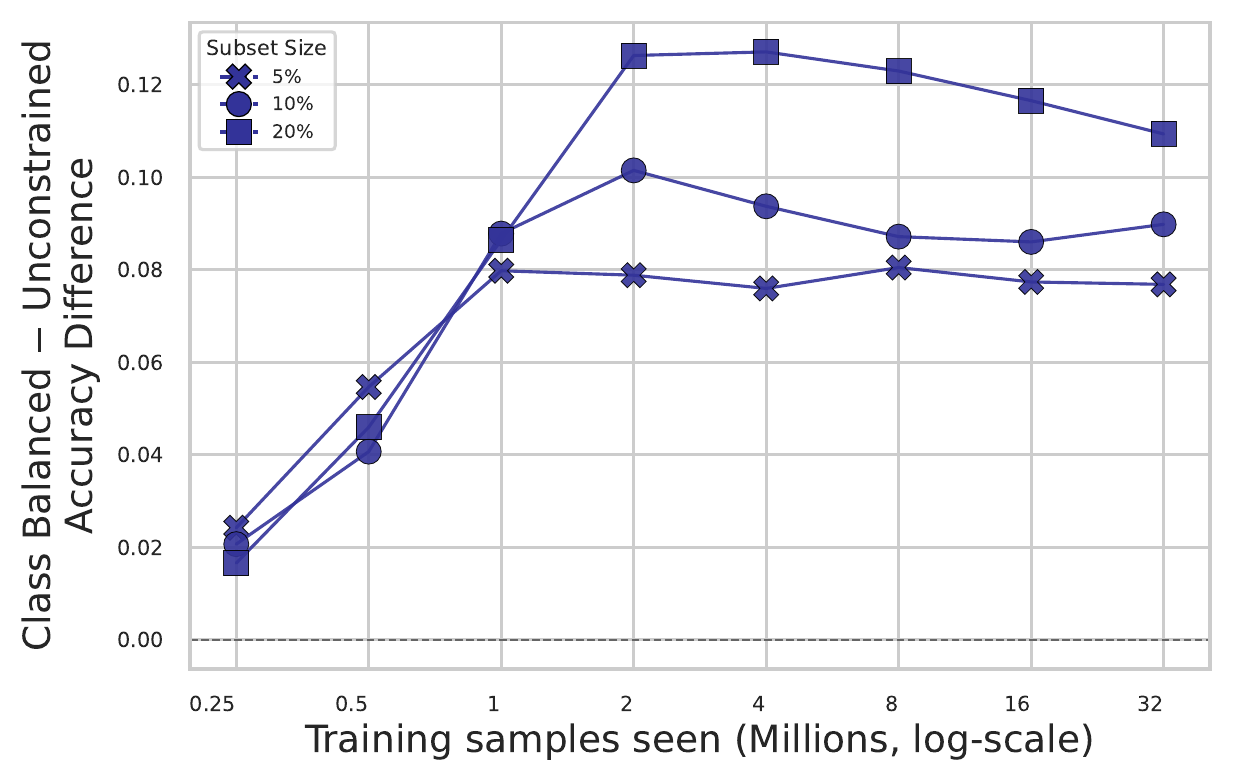}      
      \caption{Difference between the low valued \textcolor{FigBlue}{blue} class-balanced datasets and
             the size constrained (but otherwise unconstrained) datasets.}
      \label{fig:diff-class-balanced-vs-unconstrained-poor}
    \end{subfigure}
  } \caption{ On the top row, a side-by-side easy comparison of 
    unconstrained (on the left)
    vs.\ 
    class balanced (on the right), showing
    the effect of a forced balanced constraint.
    The key difference is that the poor
  \textcolor{FigBlue}{blue} sets in (b) are allowed to get worse since they are
  not forced to be class balanced. Recall that ``unconstrained'' here
  means that the sets are only constrained by size but are not constrained
 to be class balanced.
  Both the unconstrained random
  \textcolor{FigRed}{red} and unconstrained good \textcolor{FigGreen}{green}
  sets are less affected since they are likely to already naturally be much more class
  balanced --- recall, the original training dataset is class balanced from which random 
  is sampled from; also submodular max with the facility location is also likely to result
  in a naturally fairly class balanced dataset.
  The bottom row shows the difference between the class balanced
  and the size-constrained but otherwise unconstrained datasets for each
  of the three cases, broken out into individual plots for clarity since
  the scale of the differences is so different. As we can see,
  the poor \textcolor{FigBlue}{blue} sets are much more affected by the
  class balance constraint which prevents the poor datasets from being
  even worse; despite this, as mentioned above, the poor large class balanced
  dataset still performs worse than the good small balance dataset even
  at the largest (32M) compute budget.
  In all of the above cases, dataset value is determined by
  a facility location function.
  }
  \label{fig:class-balanced-vs-unconstrained}
\end{figure}

Lastly, for ease of side-by-side comparison, we
offer~\Cref{fig:class-balanced-vs-unconstrained} to show the effect of the class
balance constraint on both the valuation and the held-out test-set performance
on the subsets. This shows that while the poorly valued subsets
\textcolor{FigBlue}{blue} are much worse when they are not constrained to be
class balanced, the well-valued \textcolor{FigGreen}{green} subsets, and the
randomly selected subsets \textcolor{FigRed}{red}, have relatively similar in
performance whether they are constrained to be class balanced or not. 
This demonstrates that the 
poor \textcolor{FigBlue}{blue} class
balanced subsets in~\Cref{fig:main-teaser,fig:class-balanced-complete-pyramid}
are not as adversarially chosen as they would have been
had there not been a class balance constraint.

\section{Limitations, and Additional Conclusions and Future Work}
\label{sec:limitations}

We have not explored the $\weakly$-weakly submodular properties
of the general $\alpha$-order Vendi score (which is the Vendi
score that uses the more general Renyi entropy rather than just
the Shannon entropy). We note that matrix spectral functions
utilize a matrix function $\phi$ which is applied to each
individual eigenvalue and is then summed (in the trace).
The Renyi entropy, however, is not based on applying and then
summing one function over the eigenvalues, and so the
Renyi-entropy based Vendi score, while it is a set function,
is not a matrix spectral function as defined in this paper. This
does not mean, however, that Renyi-entropy based Vendi score
is not submodular, or possibly $\weakly$-submodular. Our current 
analysis of $\weakly$-weakly matrix monotone functions
does not apply to the Renyi-entropy case. Nonetheless,
the authors believe it is highly likely to be 
$\weakly$ DR submodular for some $\weakly$ value, and it would be useful to
analyze this case too. We stress that the secular-equation speedups
still apply since both the Renyi-entropy and Shannon-entropy require
the updated eigenvalues, something the secular equation can
provide in any case. Of course, matrix spectral submodular functions
are also much broader than just the Shannon entropy since any
function with (weakly) matrix monotone negative derivatives can be use
while still offering an approximation bound.

As mentioned in the paper, our analysis and results hold in the general
$n \times n$ positive semidefinite gram matrix case, which
can come from any PSD kernel function. Other than the small
number of experiments we have performed in the appendices, we have
not explored this further, primarily because the speed becomes prohibitive
(working in the eigenvector dual space operating on $m \times m$ matrices
means there is no potential $O(n^3)$ issue). On the other hand,
it would be useful to explore this further, as there may
be a way to use, say, a PSD kernel function on the fly while building
up the update to $B_{X+s}$ starting from an already processed $B_X$.
This would be useful future work.

It would be useful to search for other useful practical computationally feasible
(weakly) matrix monotone functions to test. As mentioned, due to
Loewner's theorem, there is a potentially infinite set to choose from,
which means there are potentially undiscovered classes of useful
practical and scalable data appraisal functions.

Ideally, we would like to try even the existing approaches on much larger and
a much wider variety of datasets. Indeed, the computational experiments we performed
were time-consuming and costly already, but useful future work would be
an empirical study on these larger resources.

While we have introduced a number of new concepts, such as that the
Vendi score is submodular, we have motivated the class of matrix
spectral functions, and defined the concept of weakly matrix monotone functions,
our results still found that the facility location function was
the best performer in predicting the value of the dataset. Firstly,
this stresses the previous paragraph since it would be useful to verify
these findings on more and larger datasets. However, the facility
location, while showing better performance, has an inherent $O(n^2)$
cost (in time and memory) and hence does not scale to extremely large
datasets. This motivates the search for other (weakly) operator
monotone functions since all matrix spectral functions in the
dual eigenspace scales, but we would like to identify one that
performs as well as the facility location function. This is ongoing
work.

As mentioned in the paper, when one has a log-det (DPP) function,
then the approach of~\cite{chen2018fast} can be used which, due to
it being specific to the log function, is faster than our approach. But
again,~\cite{chen2018fast} is only for the very specific case of the log function,
and does not apply to any other matrix function.

We note that since the Vendi score (and any matrix spectral function)
is (weakly) submodular, this means that for data appraisal, one immediately
can get conditional strategies. I.e., if you already have
data $\data_1$ and wish to appraise $\data_2$, one
can simply form $f(\data_1 \cup \data_2) - f(\data_1)$ which is
the conditional benefit of $\data_2$ given one already has $\data_1$.
This is discussed in~\Cref{sec:background-submodularity}.
This might be a valuable strategy for individuals who already are
in possession of $\data_1$ and wish to appraise $\data_2$. We have
not performed any such experiments, but we anticipate
they would be successful.

We have claimed that random sets are remarkably concentrated. We have
verified this phenomenon also on a number of much smaller datasets, 
but due to training costs, in our current experiments we have
not considered evaluating, say, thousands of random sets
to get a better sense of that distribution. Again, this is
a resource constraint issue, not due to a lack of desire.

\end{document}